\documentclass[10pt,twocolumn,letterpaper]{article}

\usepackage{3dv}
\usepackage{times}
\usepackage{epsfig}
\usepackage{graphicx}
\usepackage{amsmath}
\usepackage{amssymb}

\usepackage{booktabs} %

\usepackage{cuted}
\usepackage{capt-of}
\usepackage{float}

\usepackage{xcolor}
\usepackage{authblk}

\usepackage[pagebackref=true,breaklinks=true,letterpaper=true,colorlinks,bookmarks=false]{hyperref}

\DeclareMathOperator*{\argmin}{arg\,min}

\threedvfinalcopy %

\def\threedvPaperID{361} %

\ifthreedvfinal\fi
\begin{document}

\title{Improved Modeling of 3D Shapes with Multi-view Depth Maps}

%

\author[1]{Kamal Gupta\thanks{Equal contribution.}}
\newcommand\CoAuthorMark{\footnotemark[\arabic{footnote}]} 
\author[1]{Susmija Jabbireddy\protect\CoAuthorMark}
\author[1,2]{Ketul Shah\protect\CoAuthorMark}

\author[1]{Abhinav Shrivastava}
\author[1]{Matthias Zwicker}

\affil[1]{University of Maryland, College Park, $^{2}$John Hopkins University\vspace{0.2cm}}

\affil[ ]{\tt\small \{kampta,jsreddy,abhinav,zwicker\}@cs.umd.edu, \tt\small kshah33@jh.edu}

\renewcommand\Authands{ and }

\maketitle

\begin{strip}\centering
\includegraphics[width=0.16\textwidth]{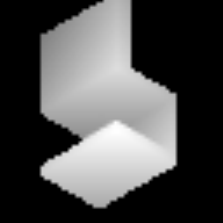}
\includegraphics[clip,trim=4cm 4cm 4cm 4cm, width=0.16\textwidth]{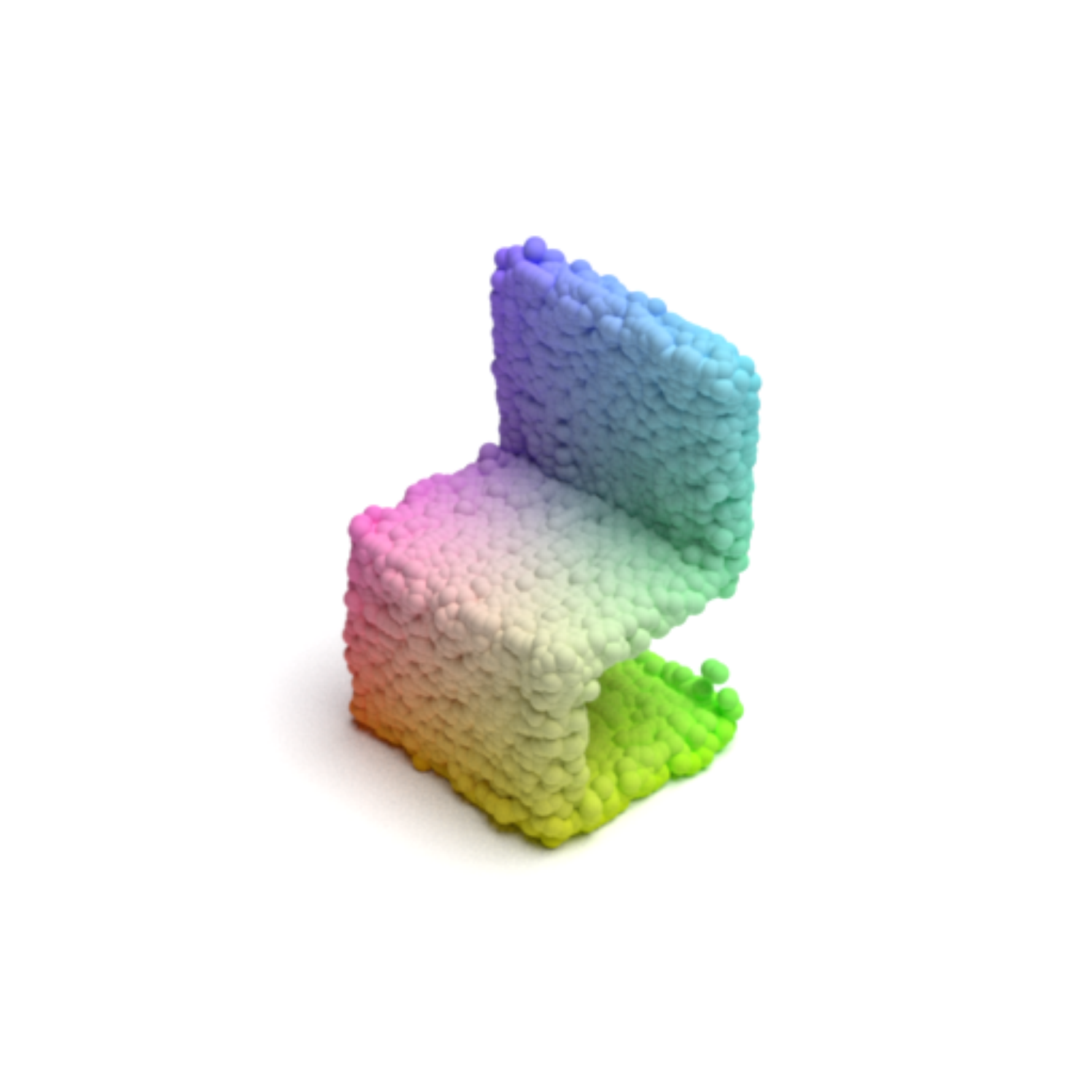}
\includegraphics[width=0.16\textwidth]{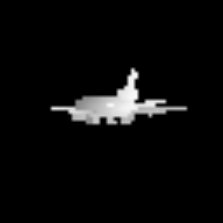}
\includegraphics[clip,trim=4cm 4cm 4cm 4cm, width=0.16\textwidth]{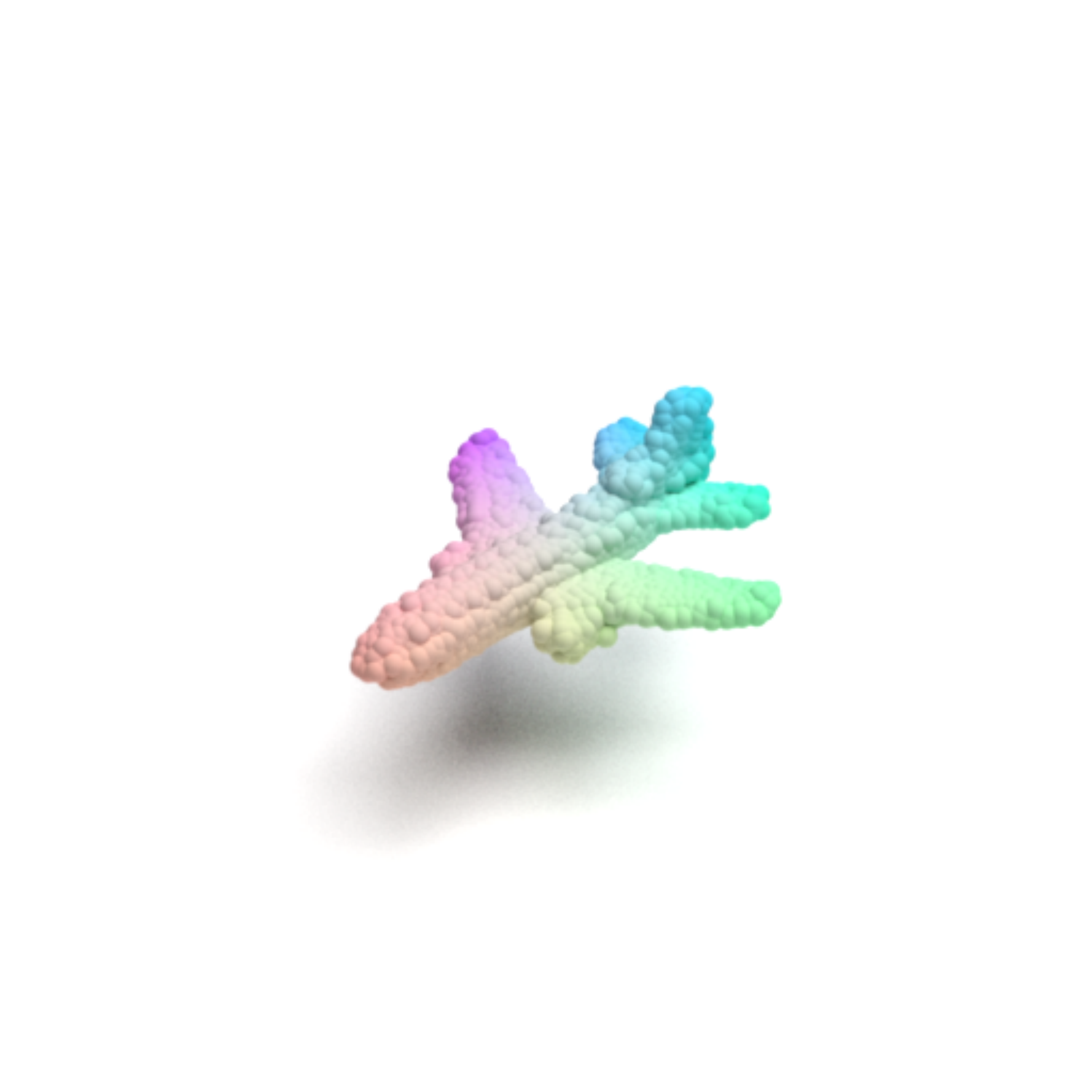}
\includegraphics[width=0.16\textwidth]{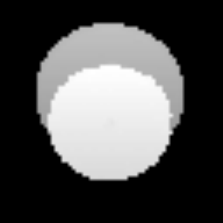}
\includegraphics[clip,trim=4cm 4cm 4cm 4cm, width=0.16\textwidth]{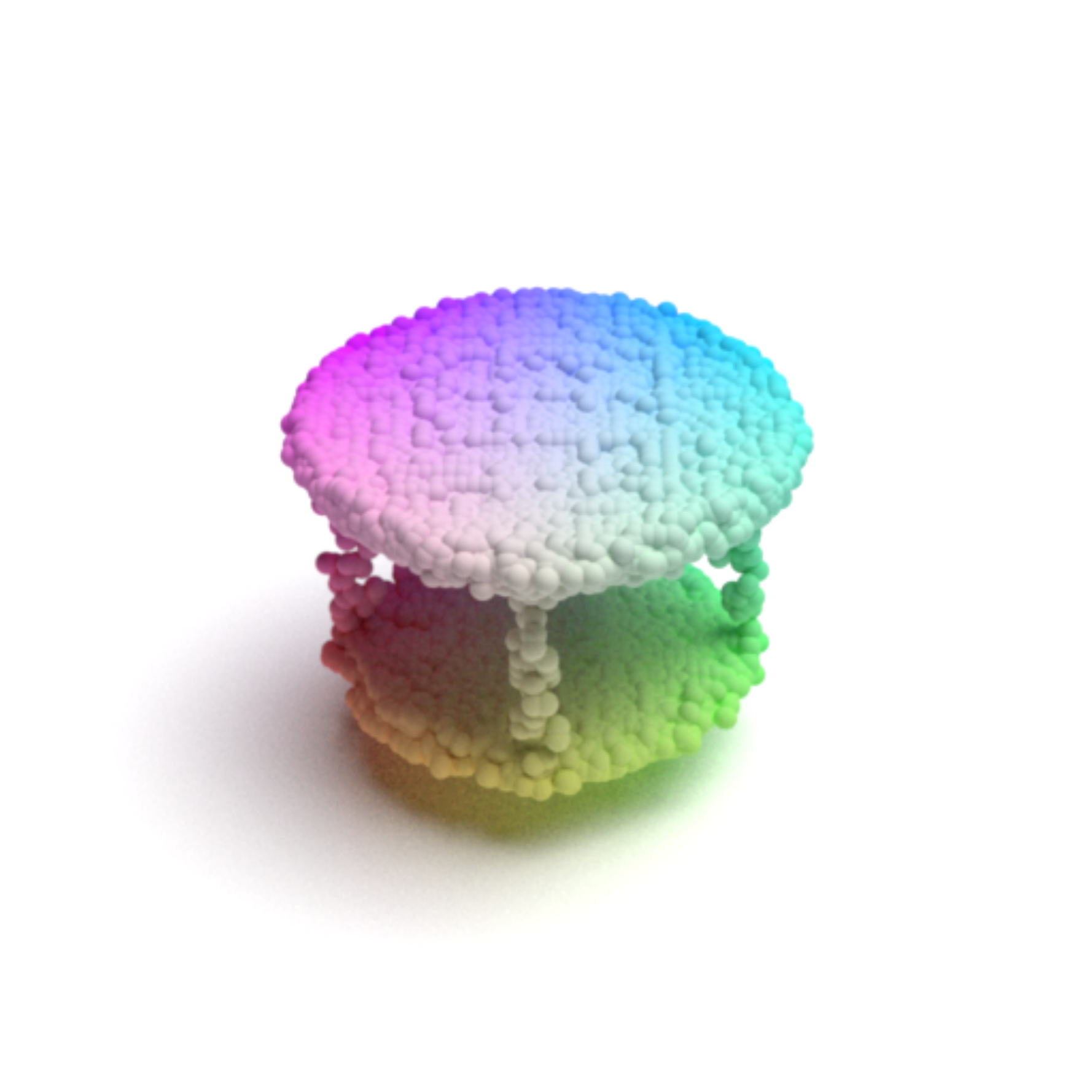}

\vspace*{0.5cm}%
\hspace*{-0.5cm}%
\includegraphics[clip,trim=4cm 4cm 4cm 4cm, width=0.16\textwidth]{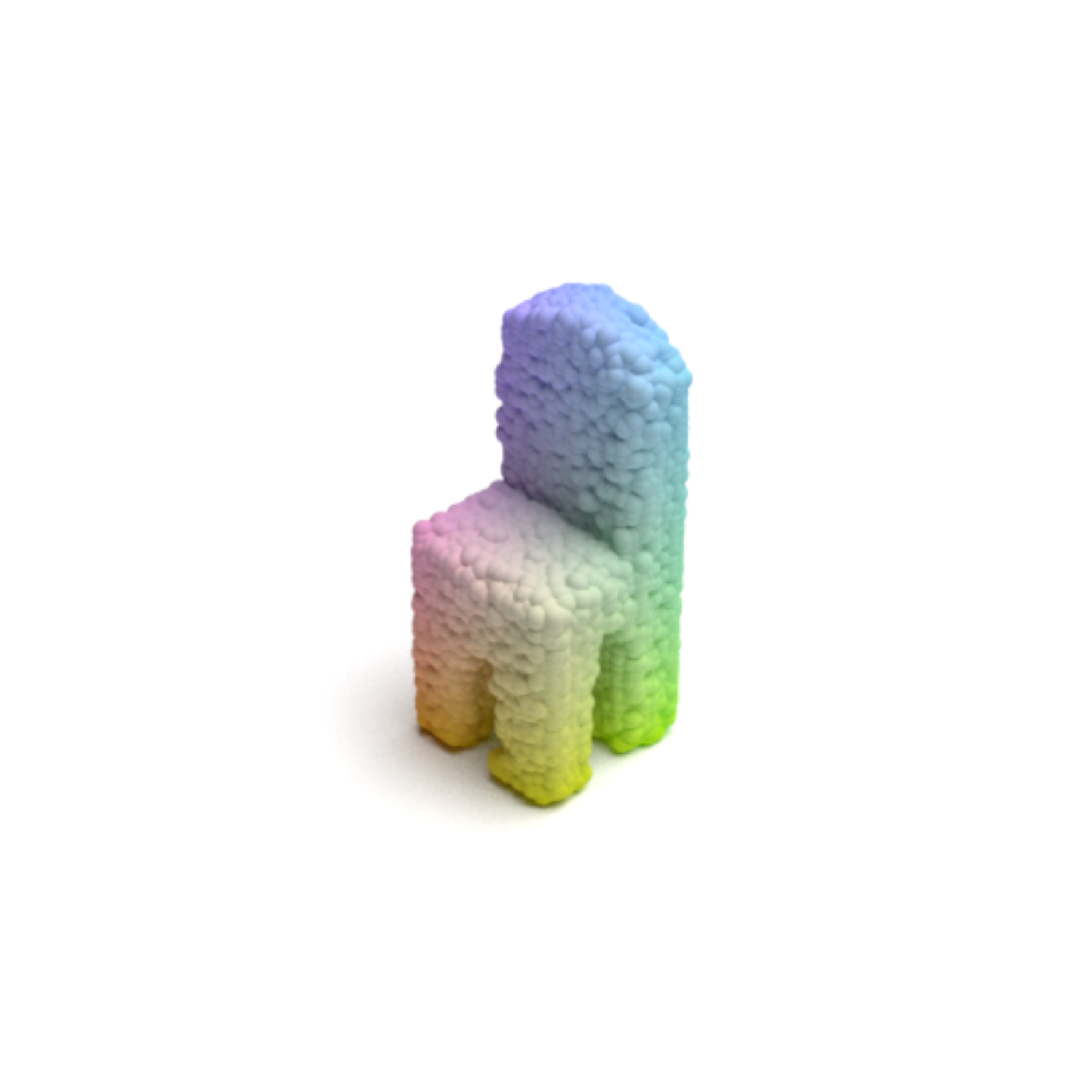}
\includegraphics[clip,trim=4cm 4cm 4cm 4cm, width=0.16\textwidth]{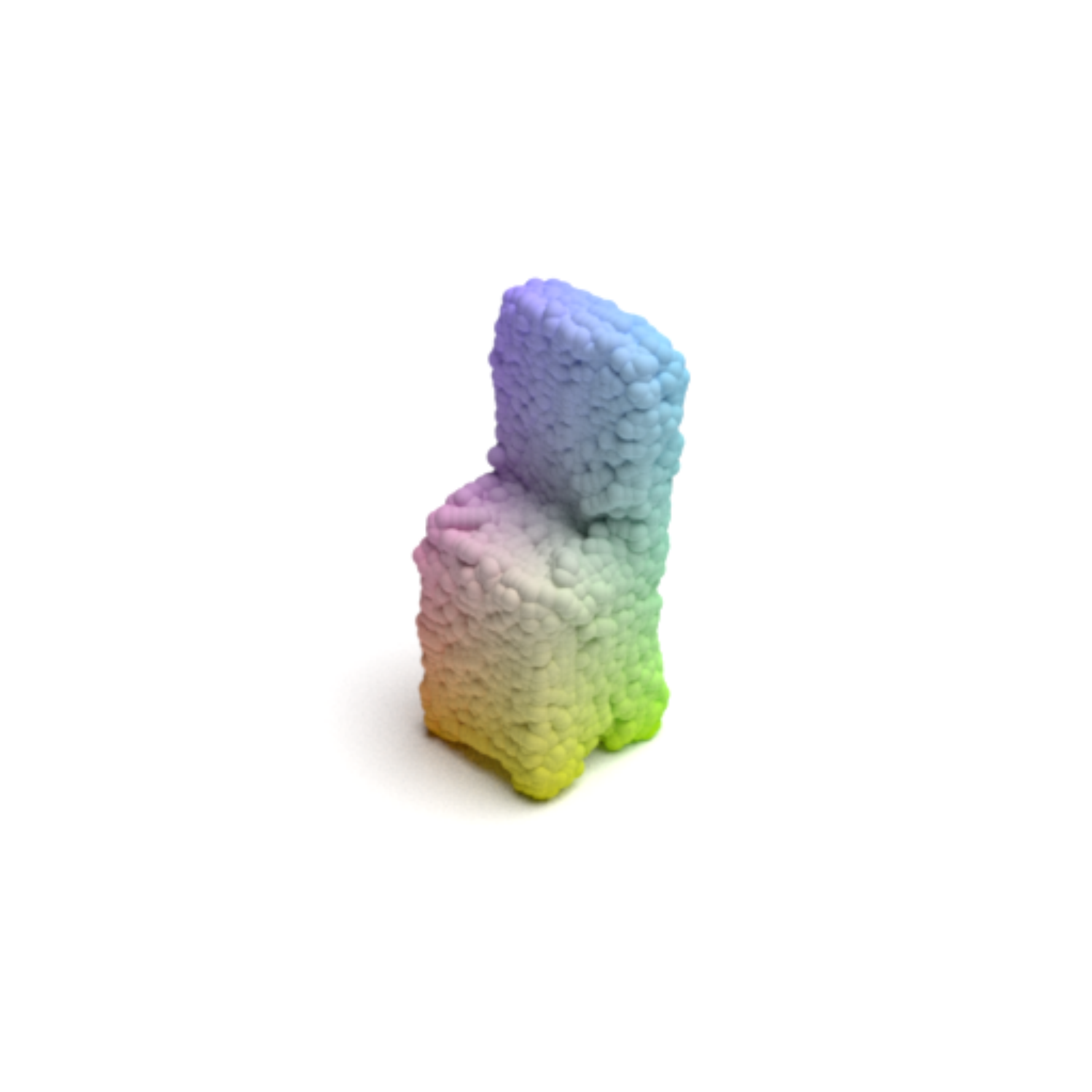}
\includegraphics[clip,trim=4cm 4cm 4cm 4cm, width=0.16\textwidth]{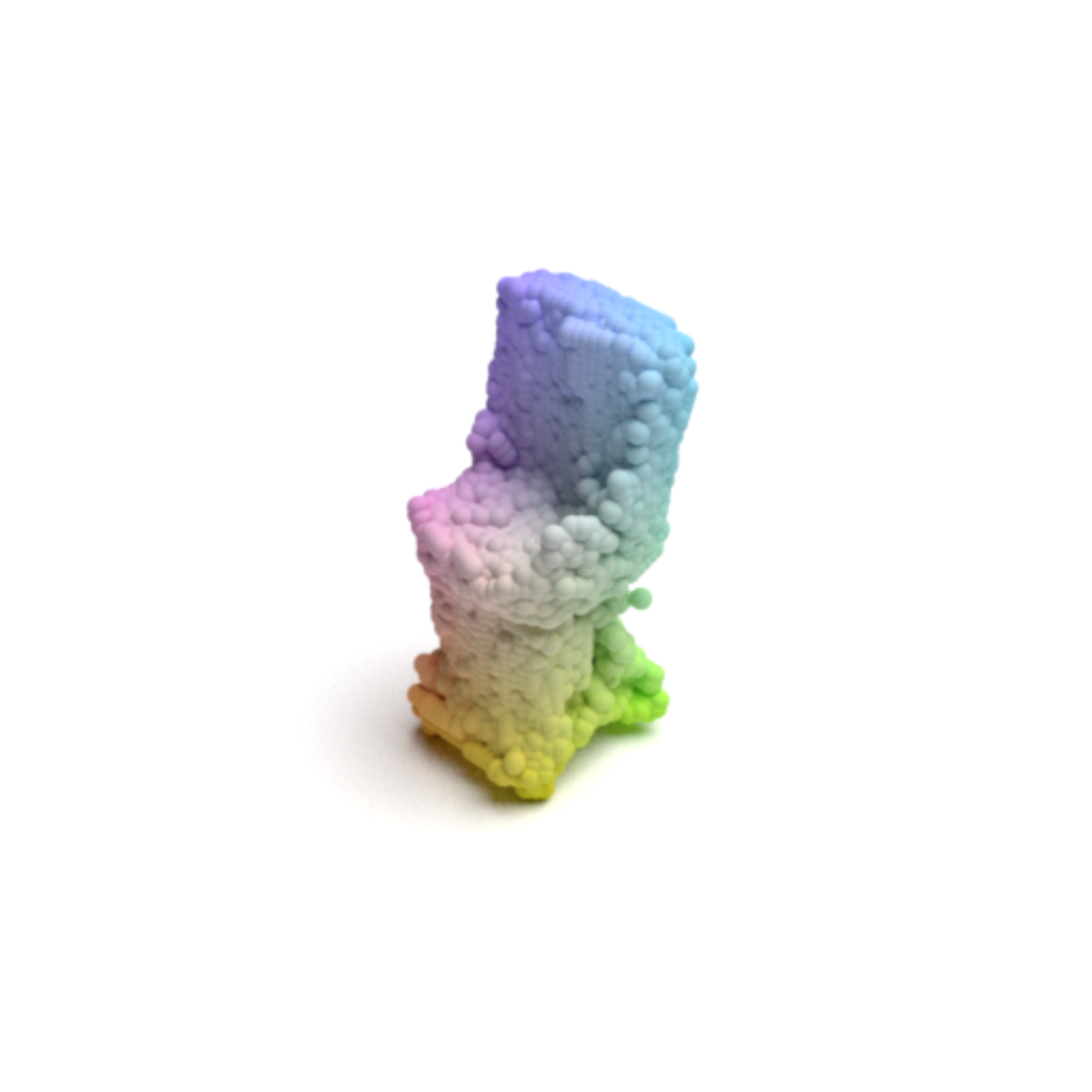}
\includegraphics[clip,trim=4cm 4cm 4cm 4cm, width=0.16\textwidth]{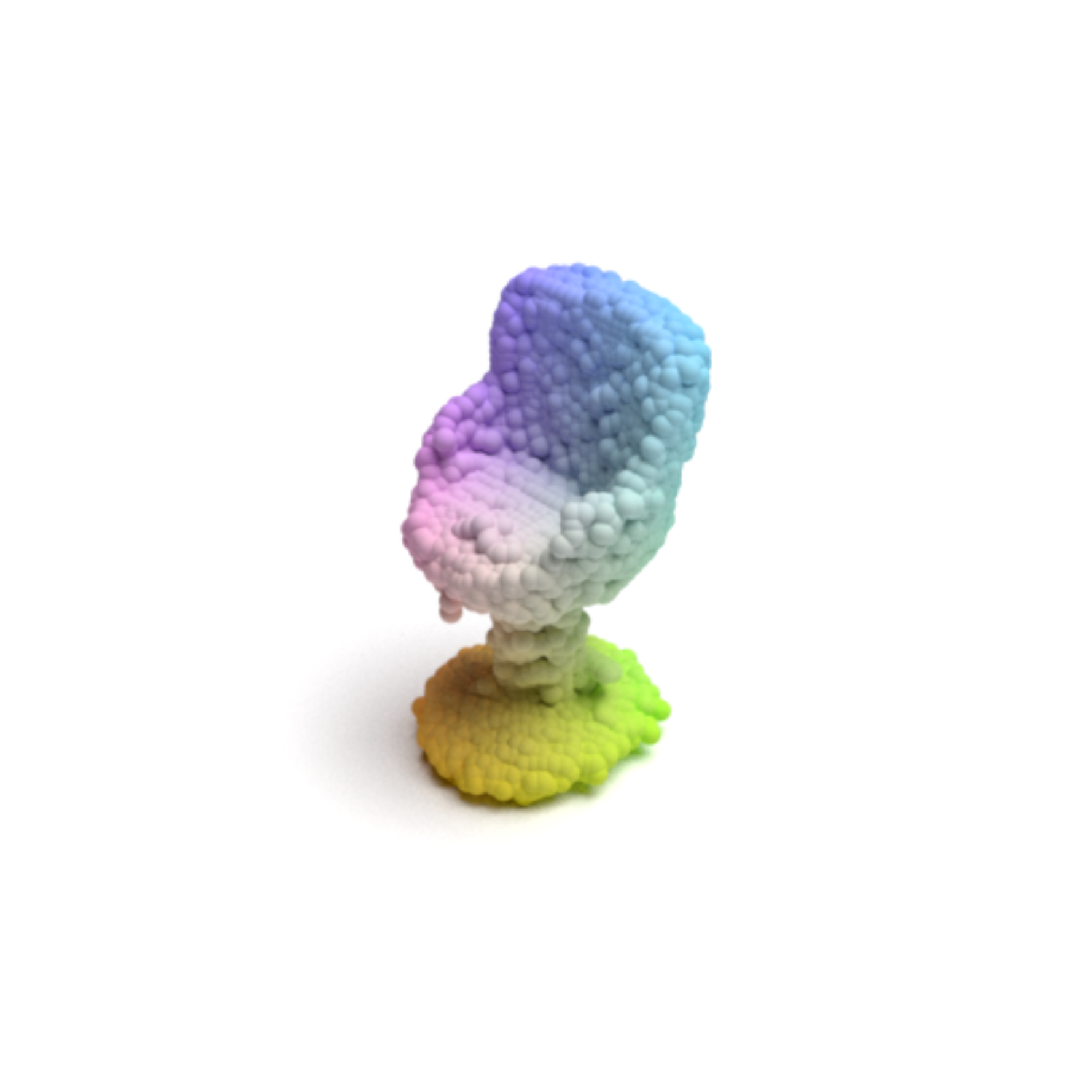}
\includegraphics[clip,trim=4cm 4cm 4cm 4cm, width=0.16\textwidth]{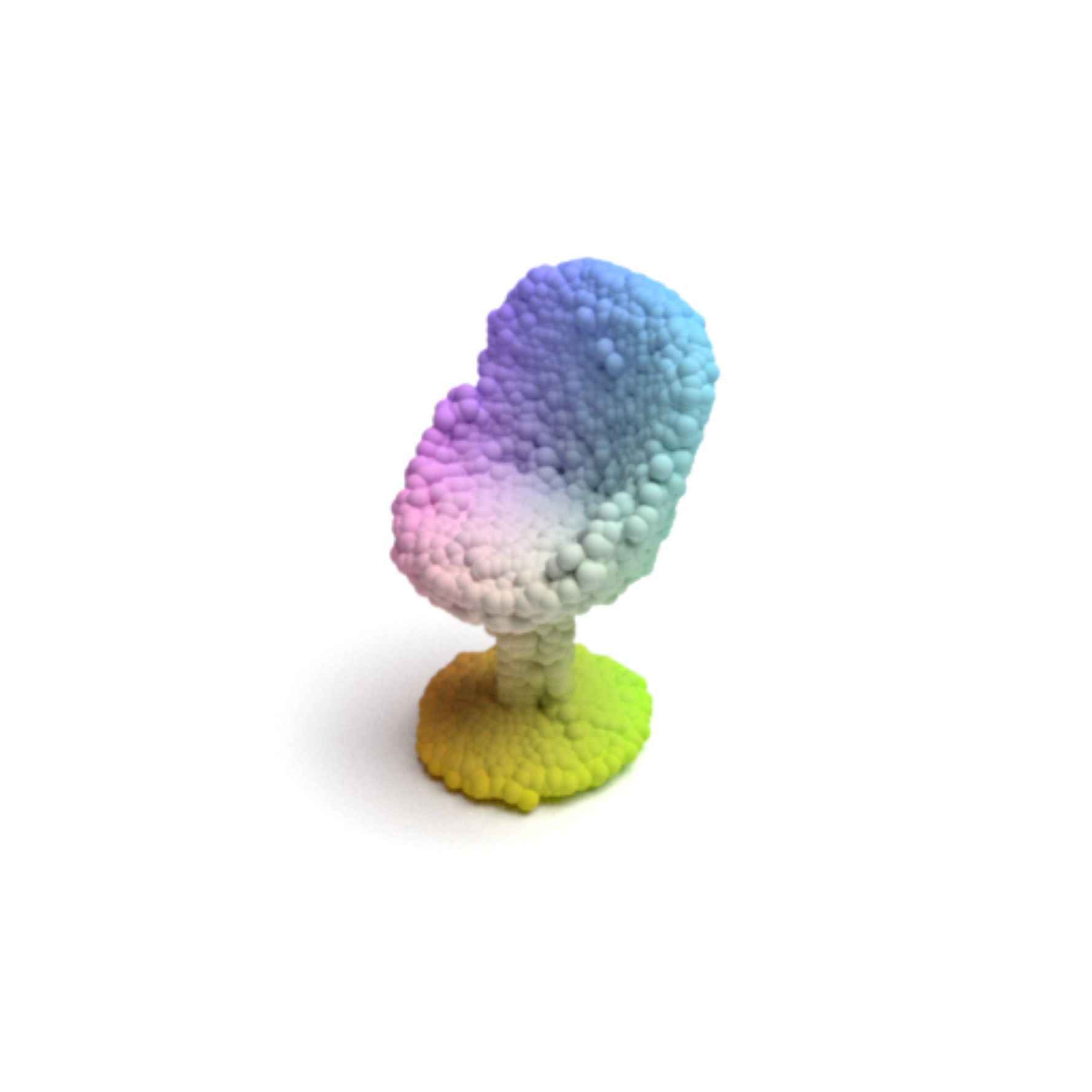}
\includegraphics[clip,trim=4cm 4cm 4cm 4cm, width=0.16\textwidth]{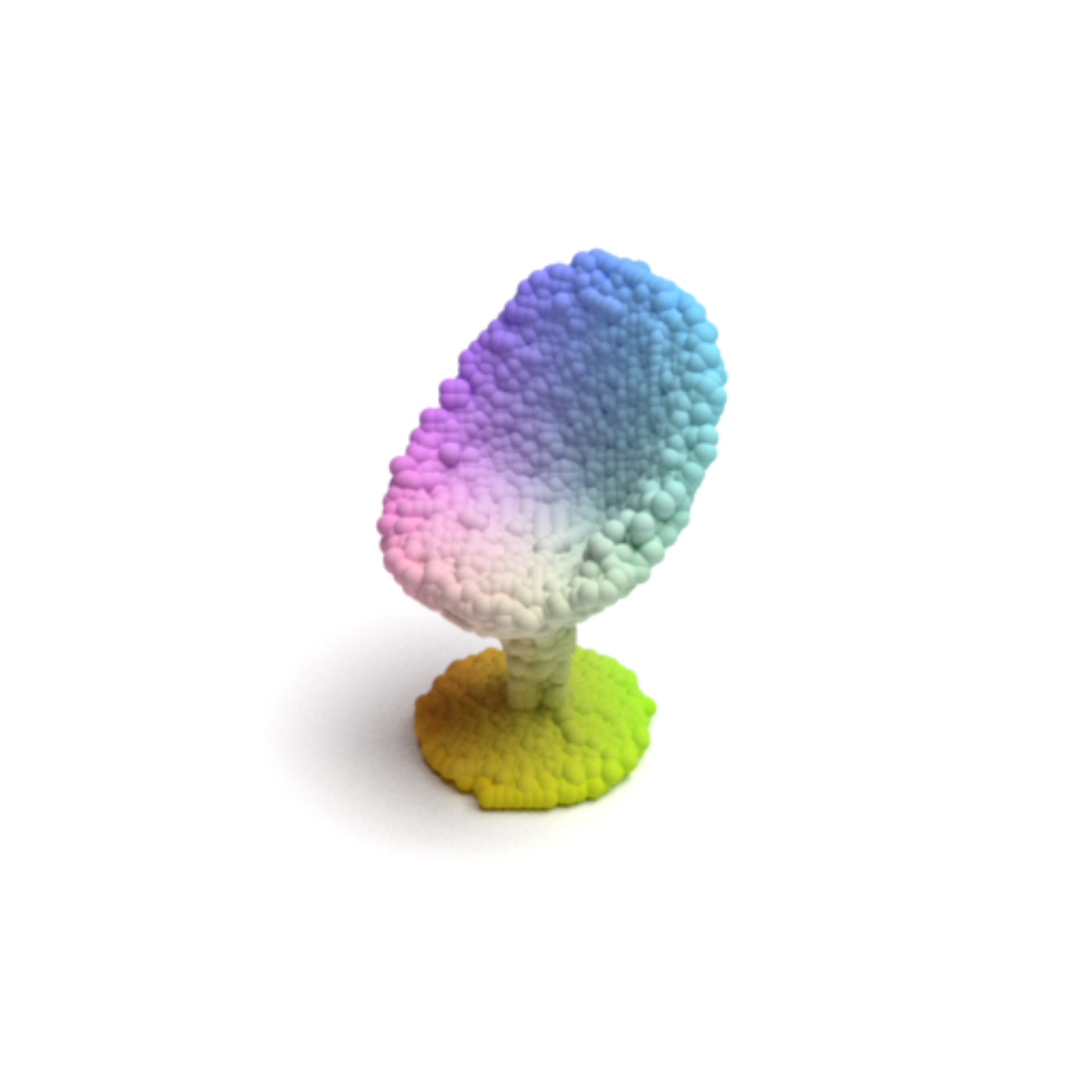}

\captionof{figure}{We demonstrate a simple encoder-decoder architecture to model 3D shapes with multi-view depth maps. Our model can be used to reconstruct complex 3D objects from a single depth map as shown in the top row. Second row shows the 3D shapes generated by interpolating latent representations of two 3D shapes, using our viewpoint generator to produce depth maps. Since our model outputs depth maps, we project the depth maps to point cloud and use ~\cite{yang2019pointflow}'s renderer for visualization. Best viewed in color.
\label{fig:teaser}}
\end{strip}

\begin{abstract}
\vspace{-0.05in}
We present a simple yet effective general-purpose framework for modeling 3D shapes by leveraging recent advances in 2D image generation using CNNs. Using just a single depth image of the object, we can output a dense multi-view depth map representation of 3D objects. Our simple encoder-decoder framework, comprised of a novel identity encoder and class-conditional viewpoint generator, generates 3D consistent depth maps. Our experimental results demonstrate the two-fold advantage of our approach. First, we can directly borrow architectures that work well in the 2D image domain to 3D. Second, we can effectively generate high-resolution 3D shapes with low computational memory. Our quantitative evaluations show that our method is superior to existing depth map methods for reconstructing and synthesizing 3D objects and is competitive with other representations, such as point clouds, voxel grids, and implicit functions.
\end{abstract}

\section{Introduction}
\label{sec:introduction}

Humans excel at perceiving the 3D structure of a large variety of objects in very diverse conditions. 
A combination of cues from signals such as color, texture, memory, and, most importantly, disparity from multi-view images enables 3D perception in humans. 
Biological evidence ~\cite{welchman2016human} suggests that the ventral cortex in our brain stores representations or features of various object configurations used while performing various tasks related to visual perception. We can naturally disentangle the identity of the object, and apply it on top of prior learned or memorized shape of objects. Some of the recent advances in neural network architectures primarily based on StyleGAN~\cite{karras2018style} generator ~\cite{karras2020analyzing, abdal2019image2stylegan, fetty2020latent, nie2020semi} attempt to accomplish the same by learning an initial embedding of say, faces, and adding noise vector to it in subsequent convolution layers to generate a new face. 
Improved neural network architectures along with large scale datasets and compute have helped us rapidly advance in the domain of representing and synthesizing images. However, due to the curse of dimensionality, we are still far from achieving such progress with 3D shapes, at scale. Two fundamental problems prevent us from bridging the gap between 2D and 3D synthesis - (1) finding an appropriate 3D representation and, (2) the availability of large-scale 3D datasets. 
In this paper, we postulate multi-view depth maps as a viable 3D representation for efficiently storing and generating high-quality 3D shapes. We show that architectures developed to work with 2D images can be easily adapted to disentangle viewpoint from shape or identity information for 3D objects.

The use of multi-view representation for 3D objects has existed for a long time~\cite{coxeter1964projective}, however, it has been overshadowed in favor of representations such as meshes~\cite{cheng2019meshgan, wang2018pixel2mesh, hanocka2019meshcnn}, point clouds~\cite{yang2019pointflow, pointnet, qi2017pointnet++}, voxel grids~\cite{3dgan, wu20153d, riegler2017octnet}, and more recently, implicit functions~\cite{park2019deepsdf, Chen_2019_CVPR, NIPS2019_9039,jiang2020sdfdiff, NIPS2019_8340}. Deep learning methods, specifically CNNs, have shown excellent capabilities for modeling complex data distributions, such as images. Adopting CNNs to express continuous 3D surfaces with complex topologies only seems natural. 
However, non-image based 3D representations limit the application of CNNs due to several reasons.
For example, voxel-based representations such as dense occupancy grid~\cite{3dgan} grow cubically in memory and processing requirements and are hard to scale to achieve higher resolutions. Both point cloud-based and mesh-based representations can provide high-quality 3D shapes with less memory, however, learning on these kinds of representations is a challenging task since they require a new way of defining convolution and pooling operations.
Meshes additionally can efficiently represent only fixed topology like faces and do not generalize to simple objects with varying topologies such as chairs and tables.
Multi-view depth maps offer a promising alternative. They are memory efficient and can be directly used to store and visualize the 3D object from various viewpoints without explicitly storing the entire 3D object.
Learning with CNNs on depth maps is relatively trivial. 
And last but not the least, with the availability of depth sensors on new generation smartphones, the amount of RGB-D data available for learning is only going to explode in the future.

In this work, we propose a framework for learning to reconstruct and synthesize 3D shapes using multi-view depth maps. We improve upon previously proposed multi-view approaches~\cite{3DVAE} in the following manners: (1) we show a simple way to adopt image-based CNN architectures to extract the identity of an object from one or more depth maps or silhouettes using an averaging heuristic, and reconstruct the entire 3D shape from a single viewpoint, (2) our framework consumes very little memory and can learn from as few as two viewpoints of an object during training (3) using a single class conditional model, we show results competitive with methods that learn only using class-specific data in tasks such as single-view reconstruction.
\section{Related Work}
\label{sec:related}

Learning good representations of 3D shapes from complex domains such as cars, chairs, humans, clothes, etc. is a classical problem in 3D Vision.
Unlike 2D images, 3D shapes do not have a standard representation. 
Traditional approaches to modeling shape information have focused on identifying primitives that combine in a meaningful way to form existing shapes.
\cite{chaudhuri2011probabilistic} developed one of the early generative model for representing shapes as assembly of its parts.
\cite{kollerprob} proposed a component-based generative model that learns the probabilistic relationships between properties of shape components and relates them to learned underlying causes of structural variability (latent variables). \cite{huang2015analysis} uses part-based templates to construct a probabilistic deformation model for generating shapes.

Recent efforts for 3D reconstruction and generation are primarily based on neural networks and can be broadly classified based on the 3D representation used: voxel-based, mesh-based, point-based, implicit function-based, or image-based methods. We discuss key recent works for each category below:

\noindent\textbf{Voxel. }
Voxel-based representations allow for easily extending advances in 2D convolutions to 3D convolutions. Several works~\cite{choy20163d, chang2015shapenet} have attempted to learn the 3D shapes by reconstructing the voxel representation of the shape.
3DGAN~\cite{3dgan} was the first GAN-based work to synthesize 3D shapes by using a voxel grid to represent the shapes.
Their work is a straight forward extension of DCGAN \cite{radford2015unsupervised} to voxel volumes with the use of 3D convolutions. However, the use of 3D convolutions is very memory intensive and hard to scale to higher resolution voxel grids, making it difficult to represent shapes with fine details. 
Octree-based methods like ~\cite{tatarchenko2017octree},~\cite{riegler2017octnet} make it possible to learn shapes up to a resolution of $512^3$ by relieving the compute and memory limitations of dense voxel methods. 

\noindent\textbf{Mesh. }
Recent works~\cite{cheng2019meshgan, bouritsas2019neural, ranjan2018generating} have exploited mesh representation of 3D shapes along with Graph Convolution Networks~\cite{henaff2015deep} for modeling human face and body.
AtlasNet~\cite{groueix2018papier} and 3D-CODED~\cite{groueix20183d} parameterize the surface of the 3D shape as a set of simple primitives and learn a mapping from the set of primitive shapes to the 3D surface. Such surface-parametric approaches rely on carefully chosen shape primitives and have shown promising results in reconstruction tasks.

\noindent\textbf{Point Cloud. }
Point cloud is another way of representing 3D objects and closely matches the raw data from sensors like LiDARs. Early works~\cite{pointnet, qi2017pointnet++} show that distinctive 3D shapes can be learned using point cloud-based representation. 
~\cite{achlioptas2017latent_pc} learns the shape embeddings of 3D objects using an auto-encoder framework and samples new objects by training a generative model on the learned embeddings.
PointFlow~\cite{yang2019pointflow} parameterizes 3D shapes as a two-level hierarchy of distributions, where the first level learns the distribution of shapes, and the second level learns the distribution of points within the shape using continuous normalizing flows. 
Though point cloud-based networks show promising results, the main limitation of learning with point clouds is that they lack topological information.

\noindent\textbf{Implicit function. }
Recently, there has been an increased interest in using implicit functions to represent 3D shapes. Implicit functions assign a value to each 3D point. The set of points representing a specific value represents the shape of the object.
Occupancy Networks~\cite{mescheder2019occupancy}, IM-Net~\cite{Chen_2019_CVPR}, and DeepSDF~\cite{park2019deepsdf} learns the surface of an object represented as signed distance function (SDF) as a continuous decision boundary of a neural network.
Implicit approaches are typically low memory and have shown to perform well at various tasks, however, the inference is usually very slow since each point in 3D still needs to be tested for presence or absence of the object.

\noindent\textbf{Image. }
Image-based representations have demonstrated the potential for 3D shape understanding and reconstruction. 
Early works~\cite{su15mvcnn} show that the view-based descriptors learned using CNN can provide better representations for 3D shapes compared to the descriptors learned in 3D space.
Owing to the efficient view-based representations of 3D shapes, several works have explored reconstructing 3D shapes from single or multi-view images. 
\cite{dosovitskiy2017learning} attempted to generate images given the viewpoint and properties of the 3D shapes, albeit in a deterministic manner. 
\cite{3DVAE} proposed a VAE-based approach to model 3D shapes using multi-view depth maps and silhouettes. 

In Section~\ref{sec:approach}, we discuss some of the limitations of the existing image-based methods and how we address them in our proposed framework.
In Section~\ref{sec:evaluation}, we evaluate our framework both quantitatively and qualitatively against approaches using different representations for 3D objects. Finally, we summarize and discuss future work in Section~\ref{sec:conculsion}.

\section{Our Approach}
\label{sec:approach}

\begin{figure*}[!t]
  \centering
  \includegraphics[width=0.95\linewidth]{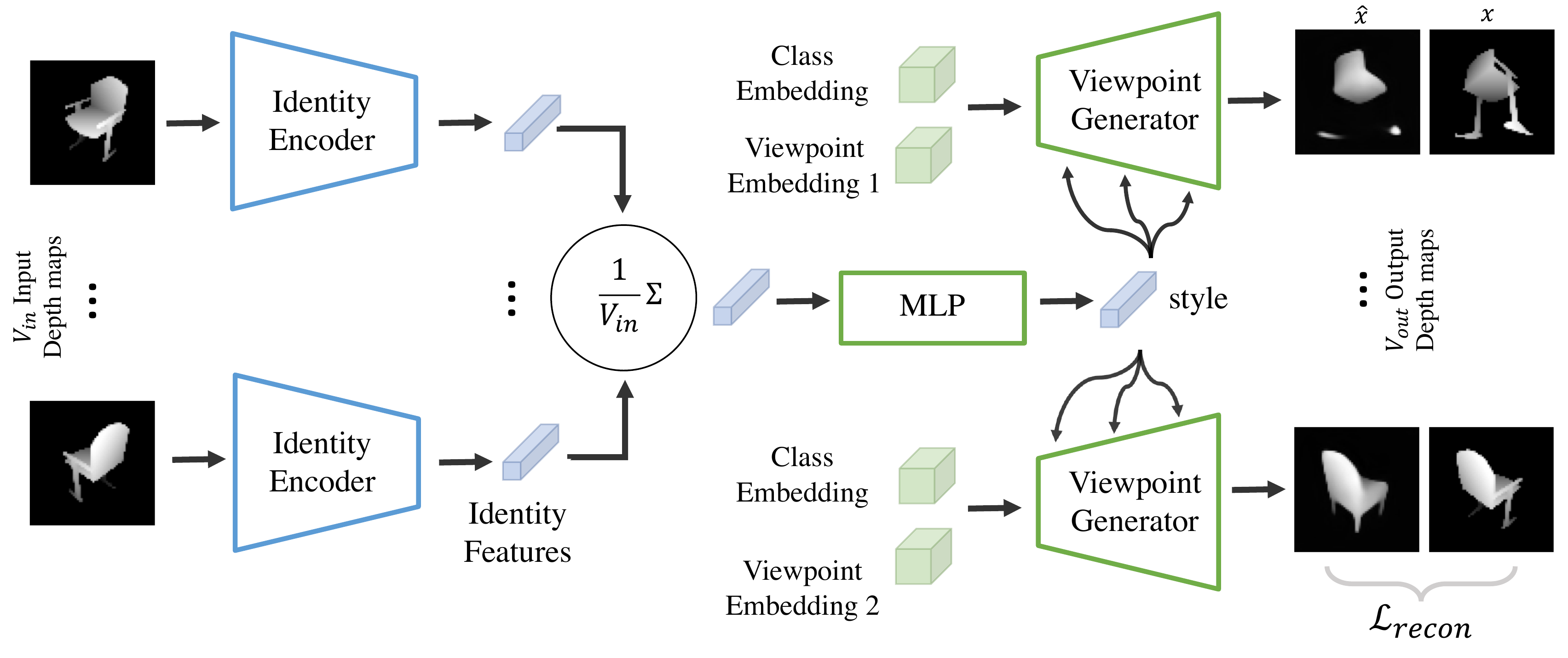}
  \caption{\textbf{Architecture}: Our Identity Encoder Network takes one or more depth maps of 3D object as input and encodes each of them into latent vectors. We use expected value of these latent vectors as the identity vector of the object. The decoder or viewpoint generator network uses this identity vector, category and viewpoint as input to generate depthmap of the object. It consists of an MLP that maps the identity vector to a style vector. This style vector is added to each block in the Viewpoint Generator Network using adaptive instance normalization.}
  \label{fig:architecture}
\end{figure*}

Existing neural network architectures for learning distributions over image data are oblivious to the 3D structure inscribed within images. 
A major challenge for the architectures working with depth maps is to generate depth maps from multiple viewpoints that are consistent in the 3D space. Soltani et al.~\cite{3DVAE} resolve the multi-view consistency by treating depth maps and silhouettes from different viewpoints as various channels of the final output. 
Overall their model generates a $40 \times H \times W$ dimensional output representing 20 depth maps and 20 silhouettes, each of size $H \times W$.
We hypothesize that representing a 3D shape as a 40 channel image poses two major challenges. 
First, a given spatial coordinate (pixel) in the 40 output channels does not correspond to the same location in 3D space, as the output channels correspond to different viewpoints.
This is in stark contrast with typical three channel RGB or four channel RGBD images, where a pixel over all channels corresponds to a single point in 3D space (given by the intersection of camera ray with the scene). 
This also makes the training more challenging since each convolution filter aggregates features of a small neighborhood to predict features at a particular pixel location. 
However, a 40 channel pixel representing 20 locations in 3D would require a much large neighborhood for accurate predictions. 
Second, having to predict all 40 channels simultaneously, makes it very memory intensive to work with large batch sizes.
Smaller batch sizes often lead to unstable gradients which leads to the requirement of a small learning rate and longer training duration~\cite{l_2018dont,goyal2017accurate}. For example, ~\cite{3DVAE} uses a maximum batch size of 8 and a learning rate of $5 \times 10^{-6}$.

In this section, we discuss our approach for modeling 3D shapes using depth maps rendered from multiple viewpoints surrounding the object. We propose several improvements to overcome the shortcomings in existing depth maps-based approaches. We also show that the proposed model can trivially adapt to modern neural network architectures developed for learning image representation and generation.

\subsection{Framework}
Our overall framework is surprisingly simple, yet effective, and depicted in Figure~\ref{fig:architecture}. It consists of two components: (1) an identity encoder, that extracts viewpoint independent, shape specific representation of a 3D object, and (2) a viewpoint conditioned decoder, that generates the depth map corresponding to the encoded shape as seen from the given viewpoint.
Unlike non-image based representations for 3D shapes, depth maps allows us to plugin and benefit from various neural network techniques developed for image-like data, such as convolutions, self-attention, or normalization layers. We hope that the simplicity of our method will bring more focus on image-based representations of 3D scenes and objects, like 2.5D, in the community. Next, we discuss each major component in detail.

\medskip
\noindent\textbf{Discrete Depth Maps.}
To keep our framework compatible with literature, we render depth maps and silhouettes from 20 fixed camera viewpoints.
As opposed to representing the depth map and silhouette as a two-channel image, we apply an 8-bit uniform quantization to the depth map. 
This expresses the depth in discrete values between 0 and 255, with 0 representing background and 1-255 representing the foreground.
Though the maximum precision for depth values is reduced, the quality of shapes is significantly improved compared to continuous values.
Similar approach of discretizing continuous signals has shown to be effective in ~\cite{van2016pixel,oord2016wavenet,gupta2020layout}. Please refer to Appendix for additional details.

\medskip
\noindent\textbf{Identity Encoder.}  
Our encoder takes as input a single depth map $\mathrm{R}^{C \times H \times W}$ and computes a view independent $L$-dimensional embedding of the depth map. During training, we obtain the disentangled shape identity from the viewpoint information using a simple averaging heuristic. We take two or more randomly sampled viewpoints of the same object and compute their embedding. We then take the expected value of this embedding and pass it on to the decoder. 
Since the decoding task can require the network to generate completely different viewpoints, the encoder is forced to learn identity information of the shape. The encoder starts with $H \times W$ resolution and applies a series of encoder blocks, each block reducing the resolution by half, recursively till we reach $4 \times 4$ resolution. Each encoder block consists of 3 convolutional layers and a downsampling layer. In our experiments, we also tried a viewpoint conditional variation of the encoder. More details of our identity encoder architecture can be found in Appendix.

\medskip
\noindent\textbf{Viewpoint Generator.}
Our framework comprises of a viewpoint generator that takes shape identity code generated by the identity encoder and a viewpoint to generate depth map of the shape as viewed from that viewpoint. The architecture of our generator is similar to the one in StyleGAN v2~\cite{karras2018style, karras2020analyzing}. Instead of starting with latent, the generator starts with a constant viewpoint embedding in shape of $256 \times 4 \times 4$ feature maps. The averaged latent vector obtained from the identity encoder is passed through a Multi-layered Perceptron (MLP) to generate a style vector, which is added to each layer of the viewpoint generator using AdaIN (Adaptive Instance Normalization) layers. For more details on the generator architecture, please refer to Figure~\ref{fig:architecture} and the appendix. 

\medskip
\noindent\textbf{Conditional generation.}
We extend our framework to be conditioned on the category of 3D objects. Along with the viewpoint embedding, we add an additional category embedding in shape of $256 \times 4 \times 4$ feature maps (same size as viewpoint embedding). This embedding is also learned during the training. Figure~\ref{fig:tsne} shows a 2d visualization of this embedding plotted using TSNE~\cite{maaten2008visualizing}.

\subsection{Loss function and other training details}
\label{subsec:training}
For each batch of $B$ objects during training, we sample $V_{in}$ depth maps per object.
The depth maps are passed through the encoder to generate viewpoint invariant latent embeddings for each of the $B$ objects.
The generator then generates $V_{out}$ number of depth maps for randomly selected viewpoints. 
We keep $V_{in} = V_{out} \leq V_{max}=20$ constant throughout the training process. 
In ablation studies, we experiment with different values of 1, 2, 3 and 4 for $V_{in}$ and $V_{out}$. 
Since our depth maps are discrete, at the end of the generator, we apply a softmax to get a probability distribution over 256 possible depth values. 
Instead of using a standard cross-entropy loss, we use a label smoothing loss~\cite{szegedy2016rethinking}. 
For each ground truth depth value, we create a pseudo-ground truth distribution with a weight of $\epsilon=0.2$ distributed uniformly at all locations except for the ground truth index which has a weight of $1-\epsilon = 0.8$. 
We, finally minimize KL-Divergence loss between the softmax predictions $\hat{x}$ and label-smoothed ground truth $x$, $\mathcal{L}(x, \hat{x}) = \mathbb{KL}\left[ x \parallel \hat{x} \right]$. 
We use Adam optimizer~\cite{kingma2014adam} with a learning rate of $0.004$, and a batch size of 32, in all our experiments.

\subsection{Generative modeling with Implicit MLE}
\label{sec:imle}

Given the trained identity encoder discussed above, we further extend our method to generate new 3D shapes by learning a latent space over the shape priors from the previous step using Implicit Maximum Likelihood Estimation (IMLE)~\cite{li2019implicit}. We use IMLE since it addresses the three main challenges of training a Generative Adversarial Network -- mode collapse, vanishing gradients, and training instability. It also claims to generate superior quality images compared to GANs and VAEs when trained for images~\cite{hoshen2019non}.

Much similar to GAN, an IMLE uses an implicit model $T$ to transform a noise vector $e$ to the data distribution. Instead of learning from a discriminator, the parameters of the implicit model are learned by ensuring that in a random batch of training data samples, every sample is close to at least one of the transformed noise vectors. Empirically, we observe that training an IMLE on the depth maps directly results in blurry depth maps samples. So instead, we train a simple fully connected IMLE model $T$ on latent vectors obtained from our identity encoder. 

Our IMLE adaptation works as follows.
Given the shape identities, at the start of each epoch, we first sample $M$ noise vectors $\{e_1, e_2, \dots, e_M\}$. The value of $M$ is chosen to be twice the size of training data. A simple 2 hidden-layer fully connected feed-forward network $T$ maps the noise vectors to $T(e_j)$ which has the same dimensions as shape identity vectors. Then, we sample a mini-batch of $K$ training points. For each shape identity $s_i$ in this mini-batch, nearest-neighbour search finds the closest transformed noise vector $T(e_i)$.

\[
e_i = \argmin_{e_j} \|T(e_j) - s_i\|_2^2
\]

The weights of network T are then learned by minimizing the distance between the nearest-neighbor correspondences
\[
T = \argmin_{\hat{T}} \sum_{i=1}^K \|s_i - \hat{T}(e_i)\|_2^2
\]

Once the model is trained, we can simply sample a vector $e$ from normal distribution, transform it to map to the distribution of shape vectors with a learned transformation $T(e)$. Finally, the viewpoint generator can then generate depth maps of this newly generated identity vector as viewed from different viewpoints, hence generating a new 3D shape. 

\section{Experiments}
\label{sec:evaluation}

We evaluate our model both quantitatively and qualitatively on various tasks to compare it against the state-of-the-art approaches.  We consider approaches that use various 3D representations, such as point clouds, voxel grids, implicit functions, and depth maps. We discuss four experiments to test the ability of our framework to learn the distribution of 3D data: (1) auto-encoding reconstructing seen and unseen shapes, (2) single view 3D reconstruction, (3) analyze the smoothness of the embeddings, and (4) sample new shapes using the learned model. 
Since most approaches train an independent model for each category, we also train independent unconditional models for 6 of the categories of ShapeNet (airplane, car, chair, lamp, sofa, and table), for quantitative comparisons. However, for qualitative analyses, we use the class conditional model.

\subsection{Dataset}
\label{subsec:dataset}

We use ShapeNetCore v2 dataset \cite{chang2015shapenet, wu20153d} which consists of $52472$ 3D aligned models from $55$ categories for all our experiments. We further use the provided train, validation, and test splits of $36814$, $5306$ and $10276$ models. To render depth maps, we follow the same procedure as~\cite{3DVAE} and place 20 virtual cameras at 20 vertices of a regular dodecahedron enclosing the object. All cameras are assumed to be located at a fixed distance of 2.5m from the origin and point towards the origin. The focal length of the camera used is 50mm and field of view is $40^\circ$. We use Blender~\cite{blender}, an open source 3D model creation and rendering suite to render the depth maps and silhouettes.

\subsection{Baselines}
We select a variety of representative baseline approaches for modeling 3D shapes.

\noindent
\textbf{3D-EPN.}~\cite{dai2017shape} proposed a voxel-based approach for shape completion of 3D shapes. Their model consists of a sequence of 3D convolution layers to predict missing voxels.
 
\noindent
\textbf{AtlasNet.}~\cite{groueix2018papier} parameterizes meshes as surface elements, and tries to auto-encode the mesh or infer the mesh using a single view of the object.

\noindent
\textbf{DeepSDF.}~\cite{park2019deepsdf} learns continuous Signed Distance Function (SDF) where the zero-level-set of the learned function represents the shape's surface. A shape embedding, with a Gaussian prior, is learned in an auto-decoder setting. At inference time, the latent shape code is estimated using MAP from the full or partial observations of 3D shape.

\noindent
\textbf{Soltani et al.}~\cite{3DVAE} is perhaps closest to our work since they also work with multi-view depth maps and follow a similar encoder-decoder approach. 

\noindent
\textbf{PointFlow.}~\cite{yang2019pointflow} takes a probabilistic approach and learns a two-level distribution to model 3D data, each learned using continuous normalizing flow. The first level learns the distribution of shapes and second learns the distribution of points for a given shape. 

\begin{table}[t]
    \centering
    \caption{Overview of baselines}
    \resizebox{\linewidth}{!}{
        \begin{tabular}{@{}llccc@{}}
            \toprule
            Method & Representation & Model size (GB) & Inference time (s) \\
            \midrule
            3D-EPN~\cite{dai2017shape}           & Voxel           &  0.42          & -  \\
            AtlasNet-25~\cite{groueix2018papier} & Mesh            &  0.17          & 0.32 \\
            DeepSDF~\cite{park2019deepsdf}       & SDF             &  0.01 & 9.72 \\
            Soltani et al.~\cite{3DVAE}          & Multiview Depth &  0.39          & 0.03  \\
            PointFlow ~\cite{yang2019pointflow} & Point cloud & \textbf{0.005} & 18.87 \\
            Ours                                 & Multiview Depth &  0.41          & \textbf{0.01}  \\
            \bottomrule
        \end{tabular}
    }
\vspace{-0.1in}
\label{table:baselines}
\end{table}

\subsection{Metrics}
We use the following to compare a pair of point clouds:

\noindent
\textbf{Chamfer Distance (CD)}: is the sum of distances between points from a set to its nearest neighbors in the other set.

\noindent
\textbf{Earth Mover's Distance (EMD)}: is used for measuring optimal transport distance between two discrete distributions.

Both CD and EMD don't work well if large holes exist in either source or target models. 
EMD is relatively slower to compute and we have used a small sample of 500 points from each point cloud in all our evaluations. For CD, we use $30000$ points and compute the distance in both directions in order to make it symmetric. The reported CD is multiplied by $10^3$.

We follow the protocols of ~\cite{park2019deepsdf} to evaluate our model's capability in several ways. To compare against other approaches quantitatively, we project the depth maps generated by model to 3D point clouds. Specifically, we measure (1) how well our model can faithfully reconstruct unseen 3D shapes with a single latent vector while preserving the geometric details, and (2) if our model can extrapolate complete 3D structure from a single viewpoint information. We also analyze (1) the latent space learned by the model both in terms of smoothness, and (2) semantics of the category embeddings learned by the model. Finally we show the generative capabilities of the model, using an IMLE approach as discussed in Section~\ref{sec:imle}.

\begin{figure*}
\centering

\hspace*{-0.5cm}%

\includegraphics[clip,trim=4cm 4cm 4cm 4cm, width=0.095\textwidth]{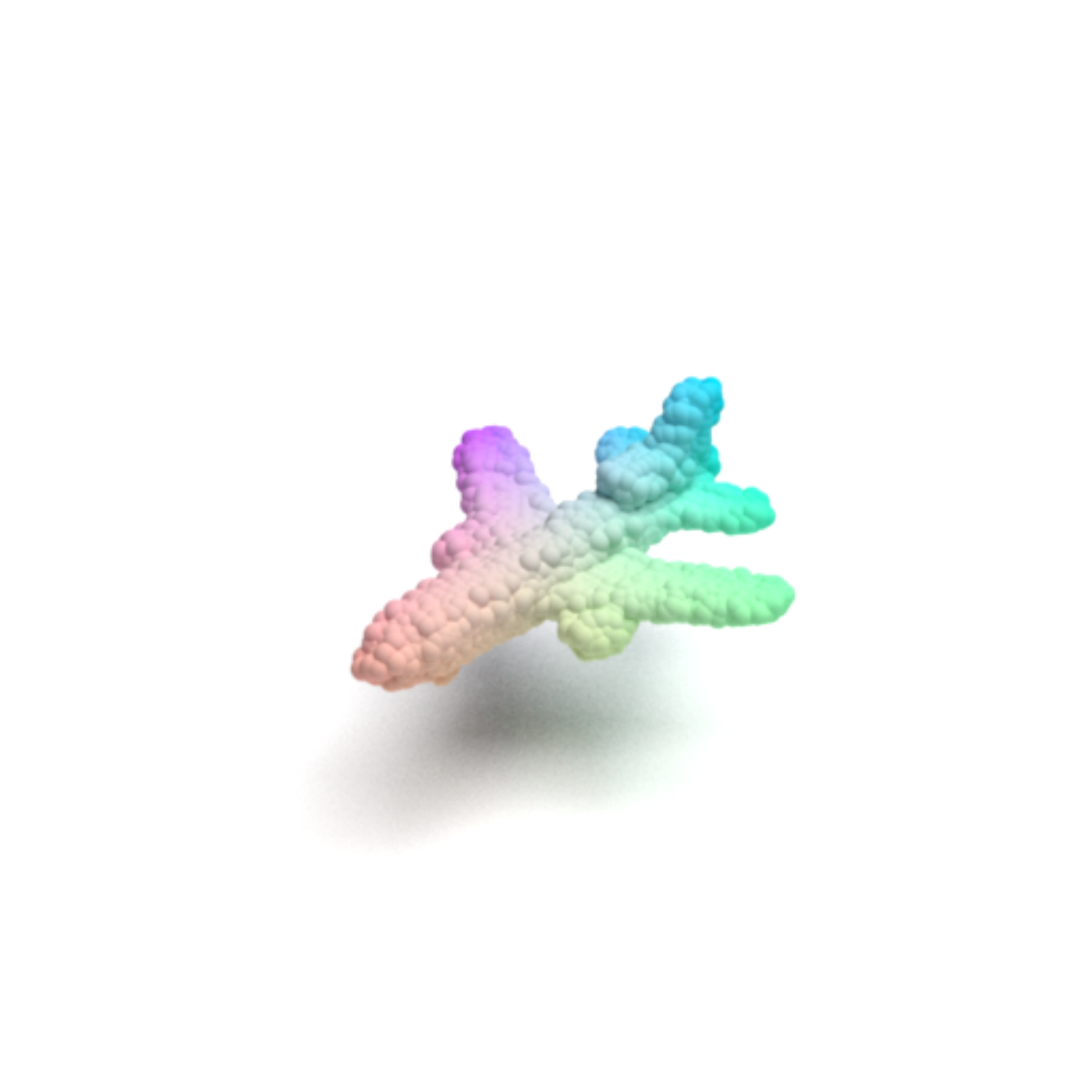}
\includegraphics[clip,trim=4cm 4cm 4cm 4cm, width=0.095\textwidth]{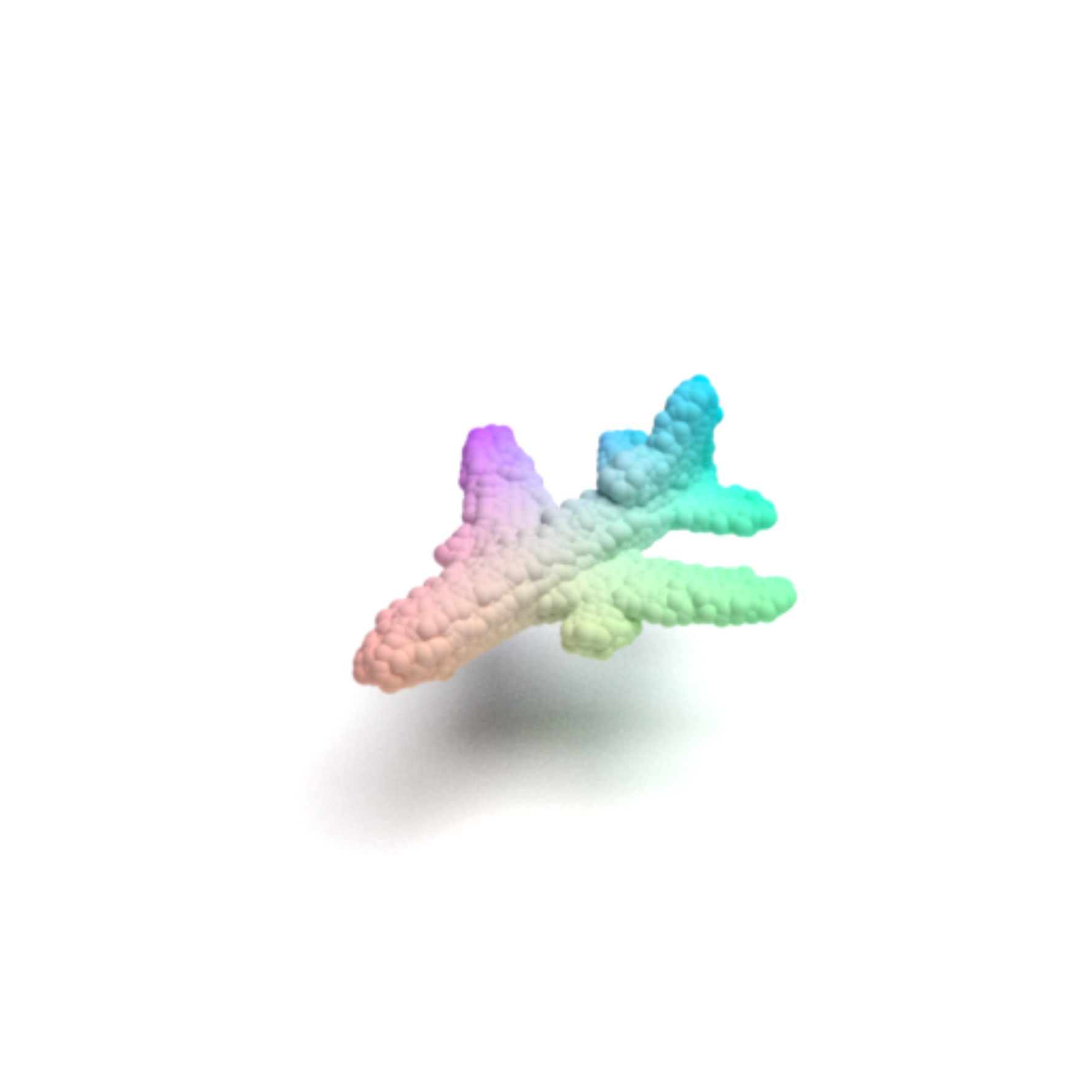}
\includegraphics[clip,trim=4cm 4cm 4cm 4cm, width=0.095\textwidth]{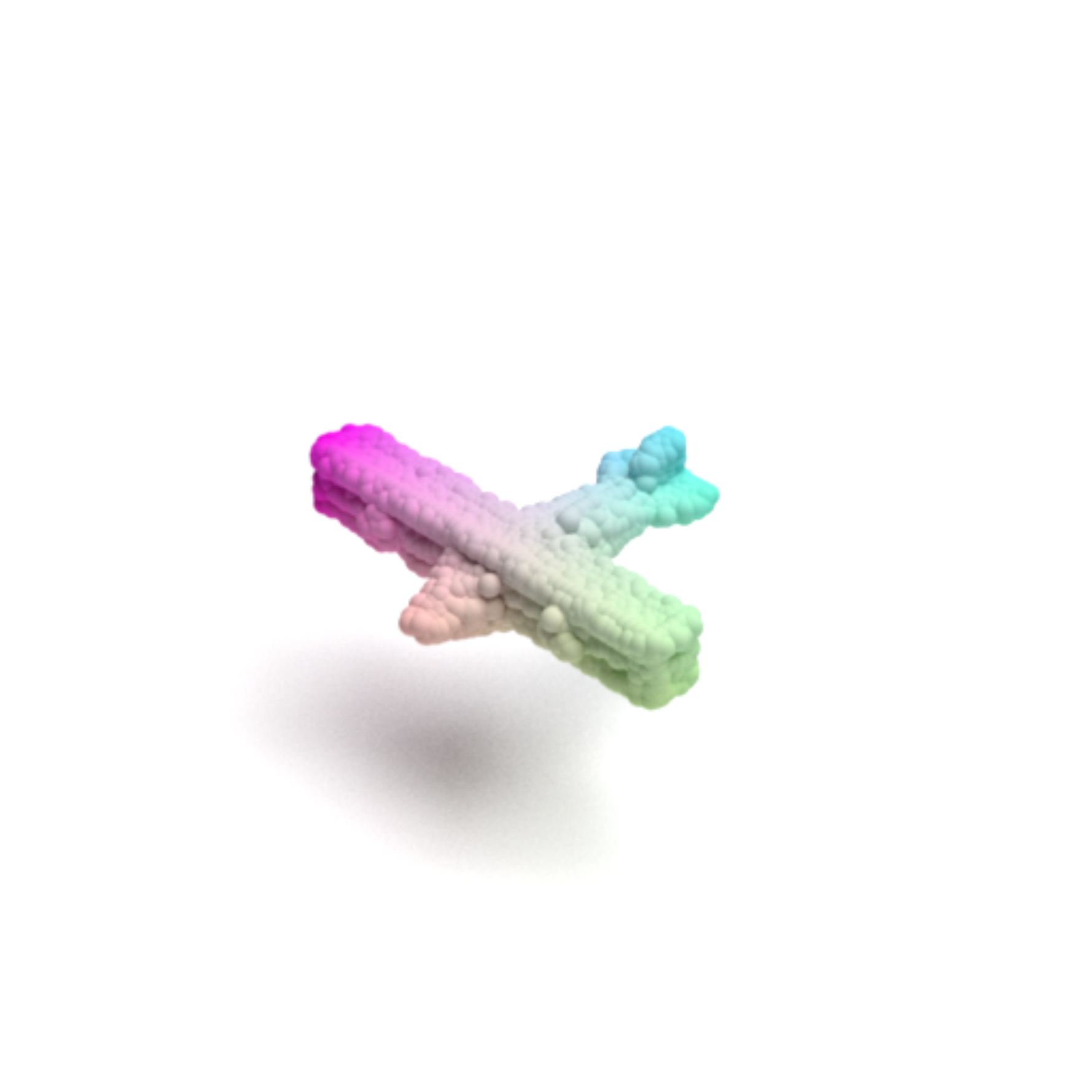}
\includegraphics[clip,trim=4cm 4cm 4cm 4cm, width=0.095\textwidth]{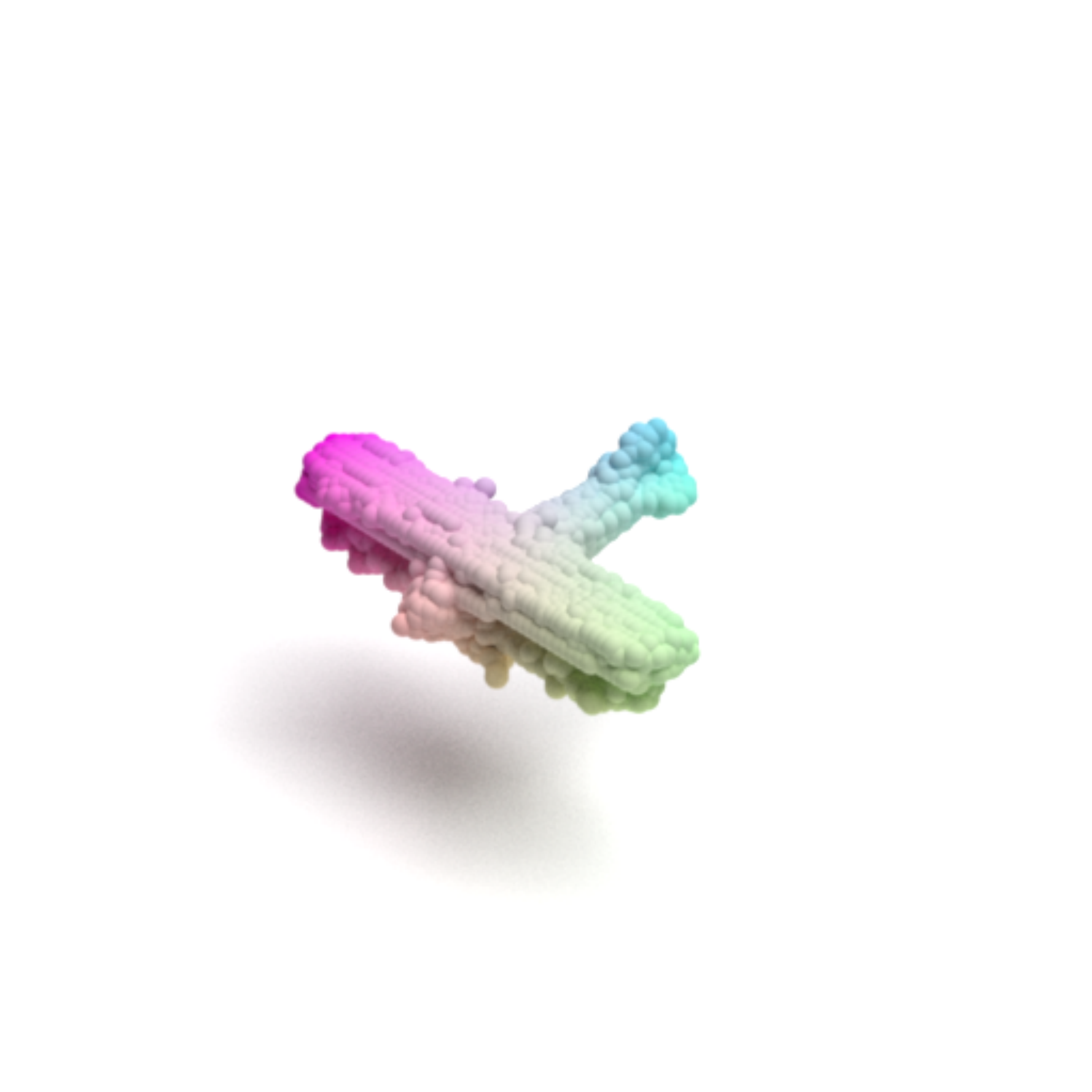}
\includegraphics[clip,trim=4cm 4cm 4cm 4cm, width=0.095\textwidth]{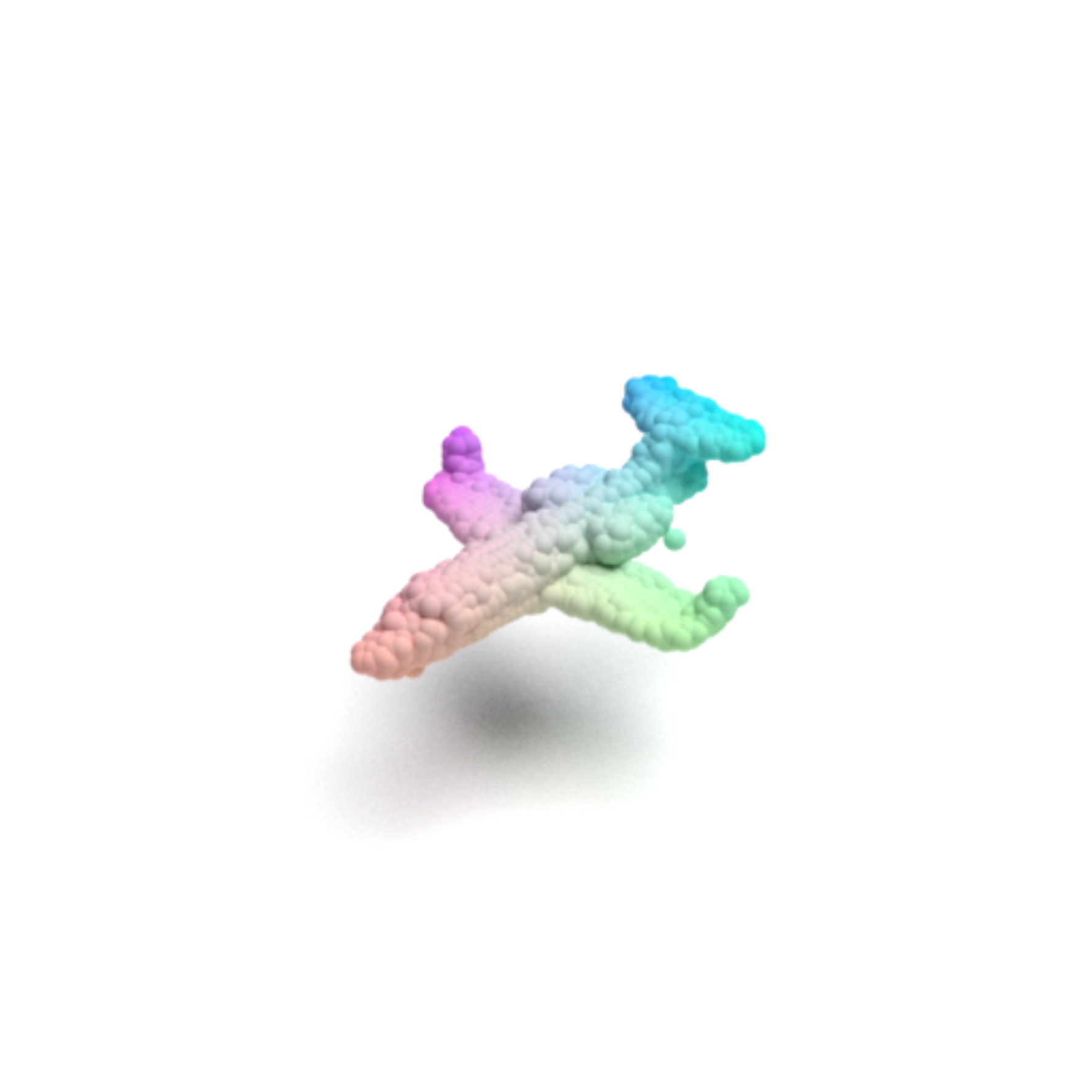}
\includegraphics[clip,trim=4cm 4cm 4cm 4cm, width=0.095\textwidth]{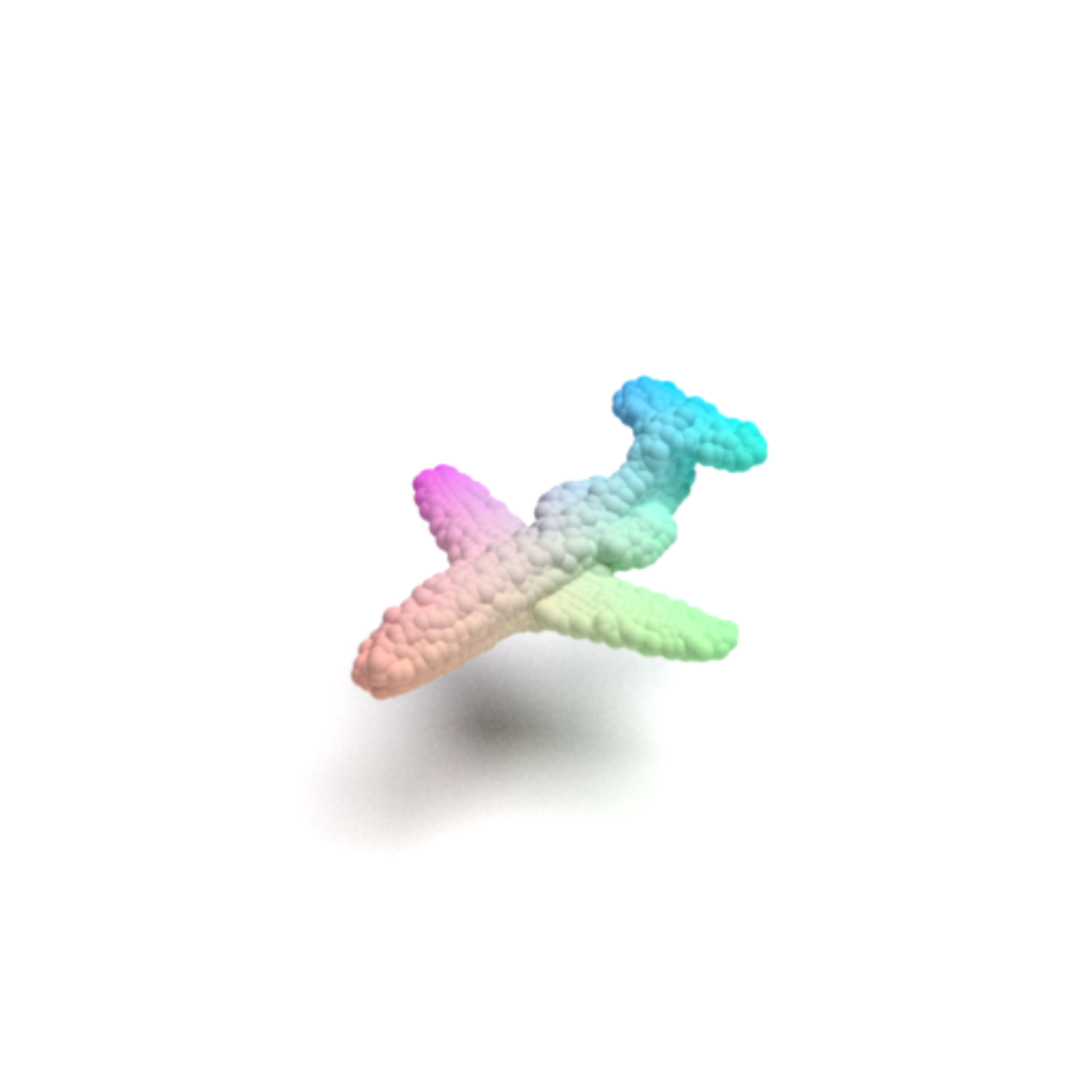}
\includegraphics[clip,trim=4cm 4cm 4cm 4cm, width=0.095\textwidth]{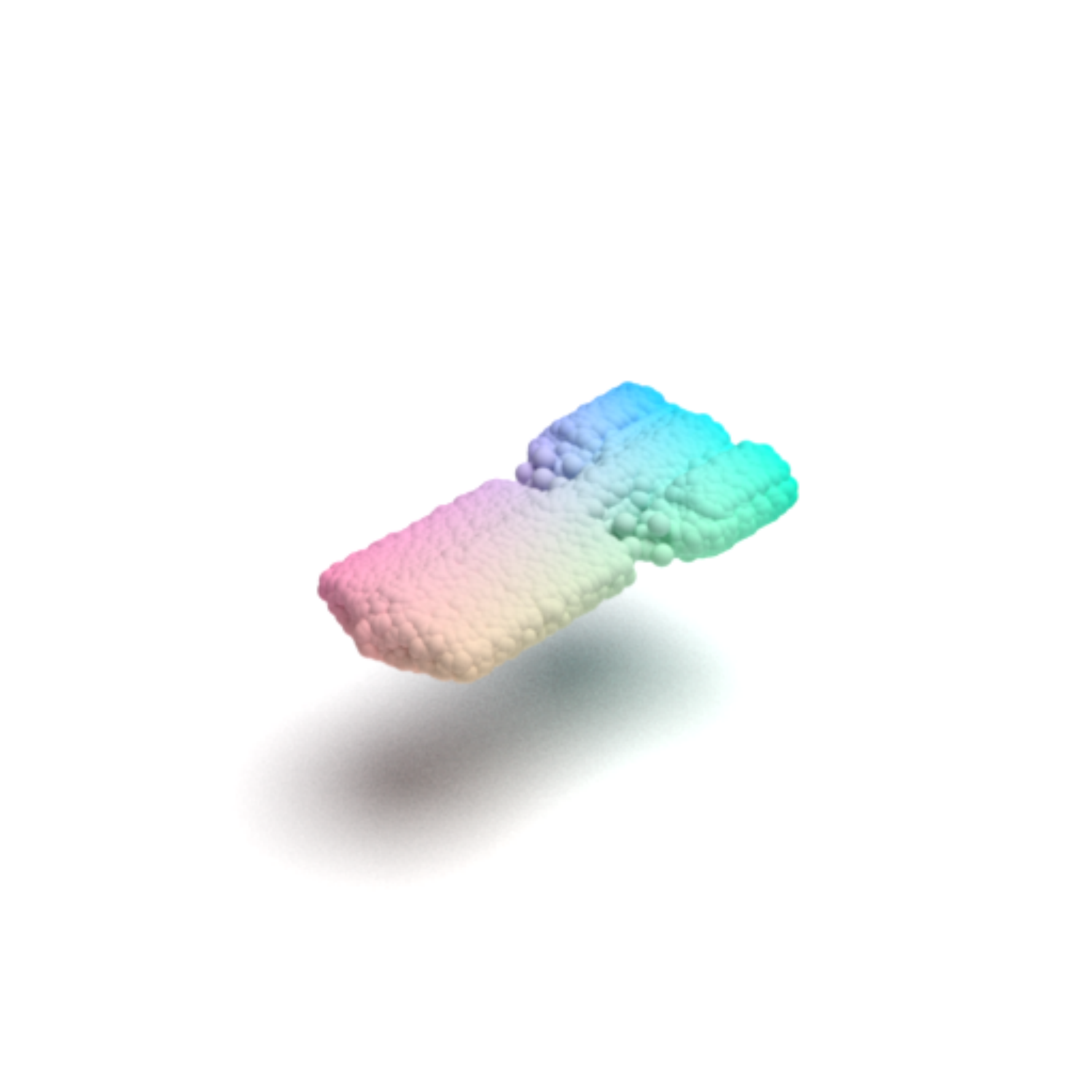}
\includegraphics[clip,trim=4cm 4cm 4cm 4cm, width=0.095\textwidth]{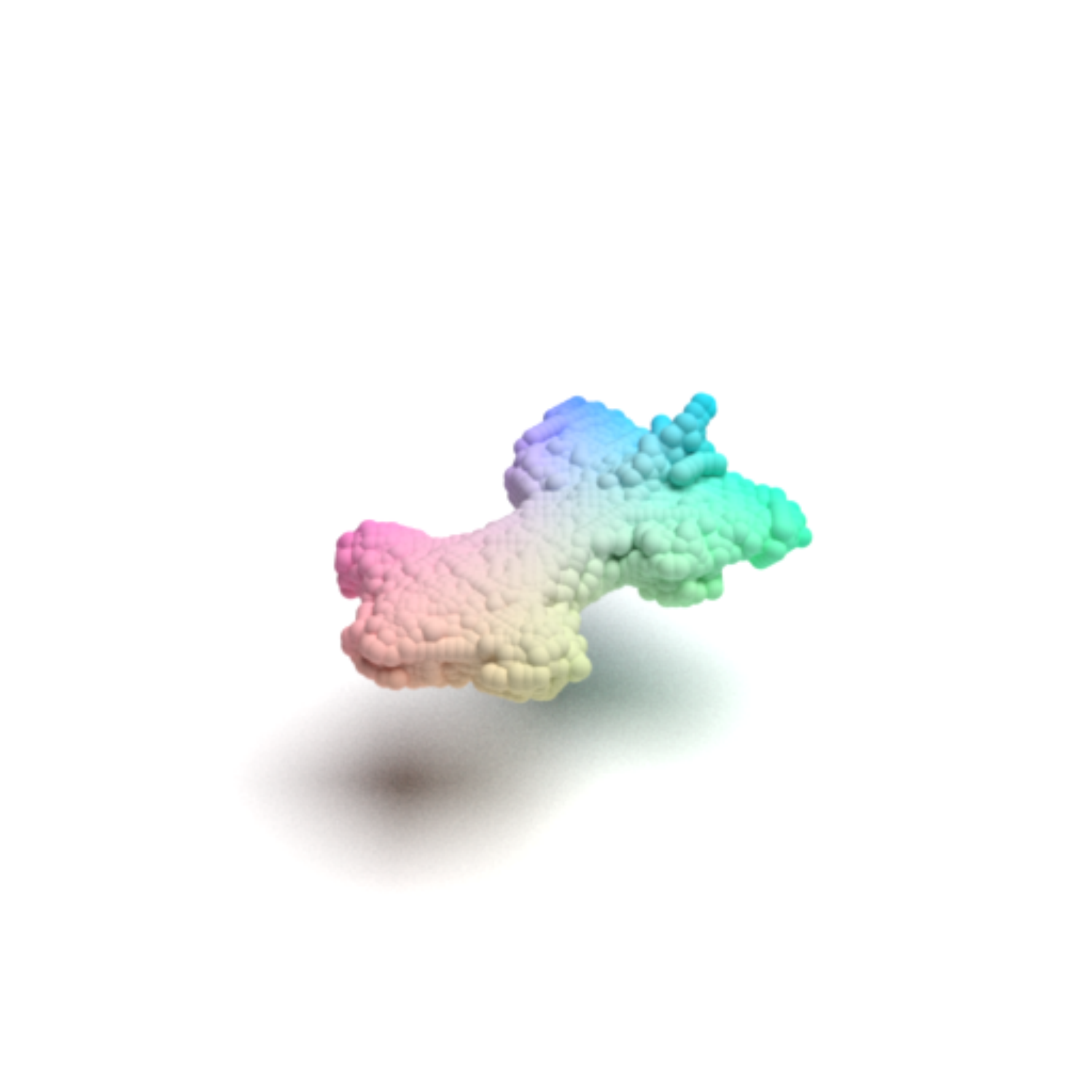}
\includegraphics[clip,trim=4cm 4cm 4cm 4cm, width=0.095\textwidth]{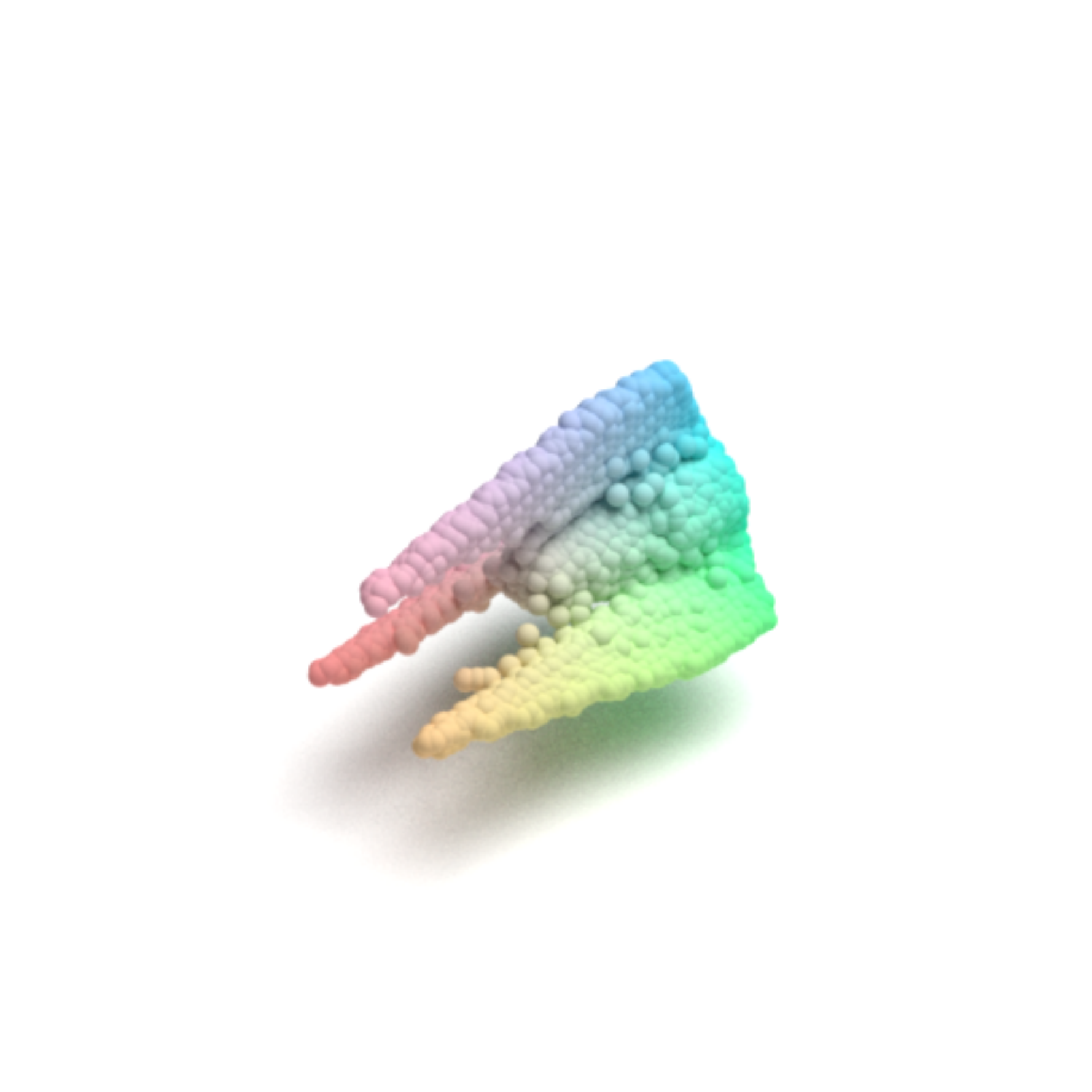}
\includegraphics[clip,trim=4cm 4cm 4cm 4cm, width=0.095\textwidth]{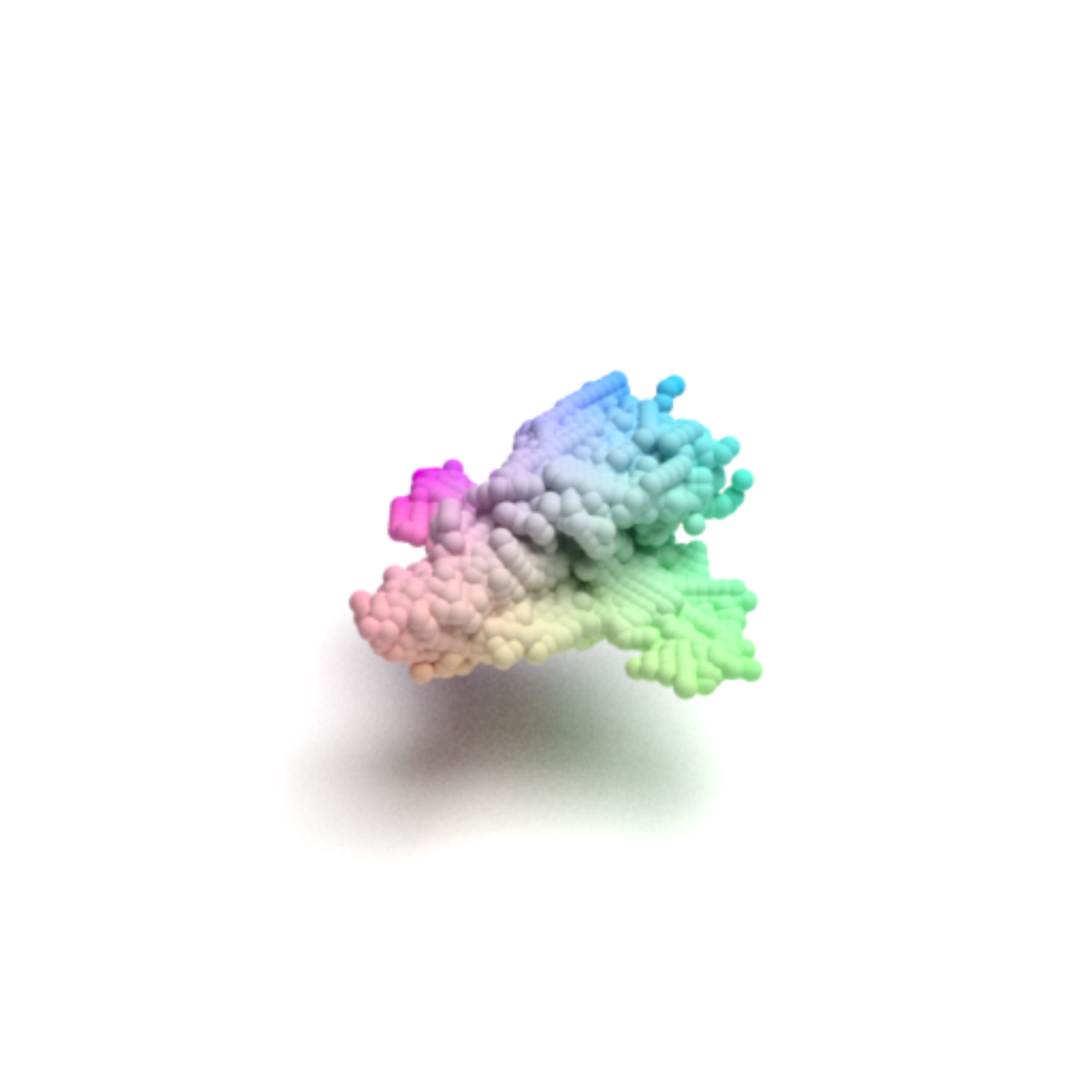}

\includegraphics[clip,trim=3cm 3cm 3cm 3cm, width=0.095\textwidth]{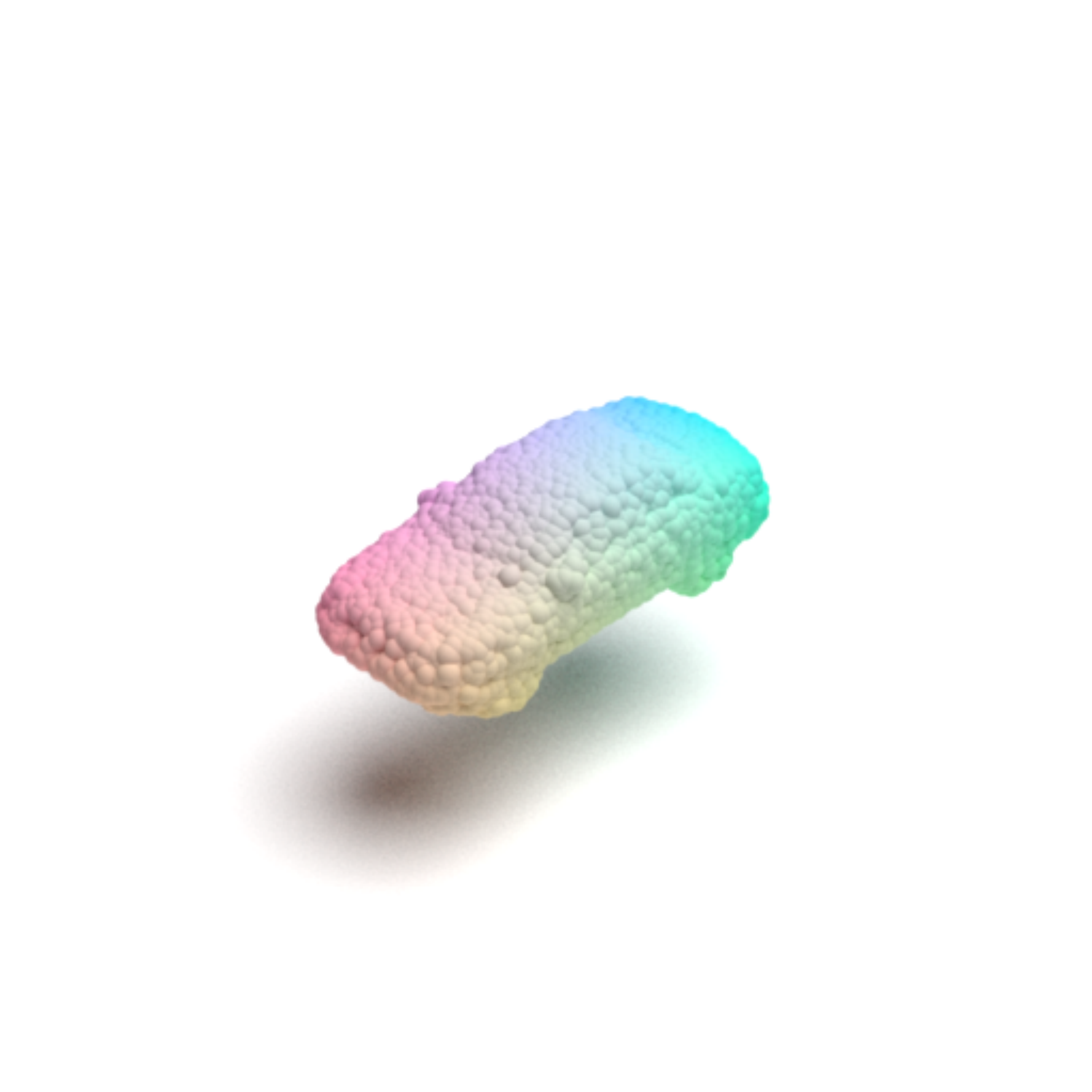}
\includegraphics[clip,trim=3cm 3cm 3cm 3cm, width=0.095\textwidth]{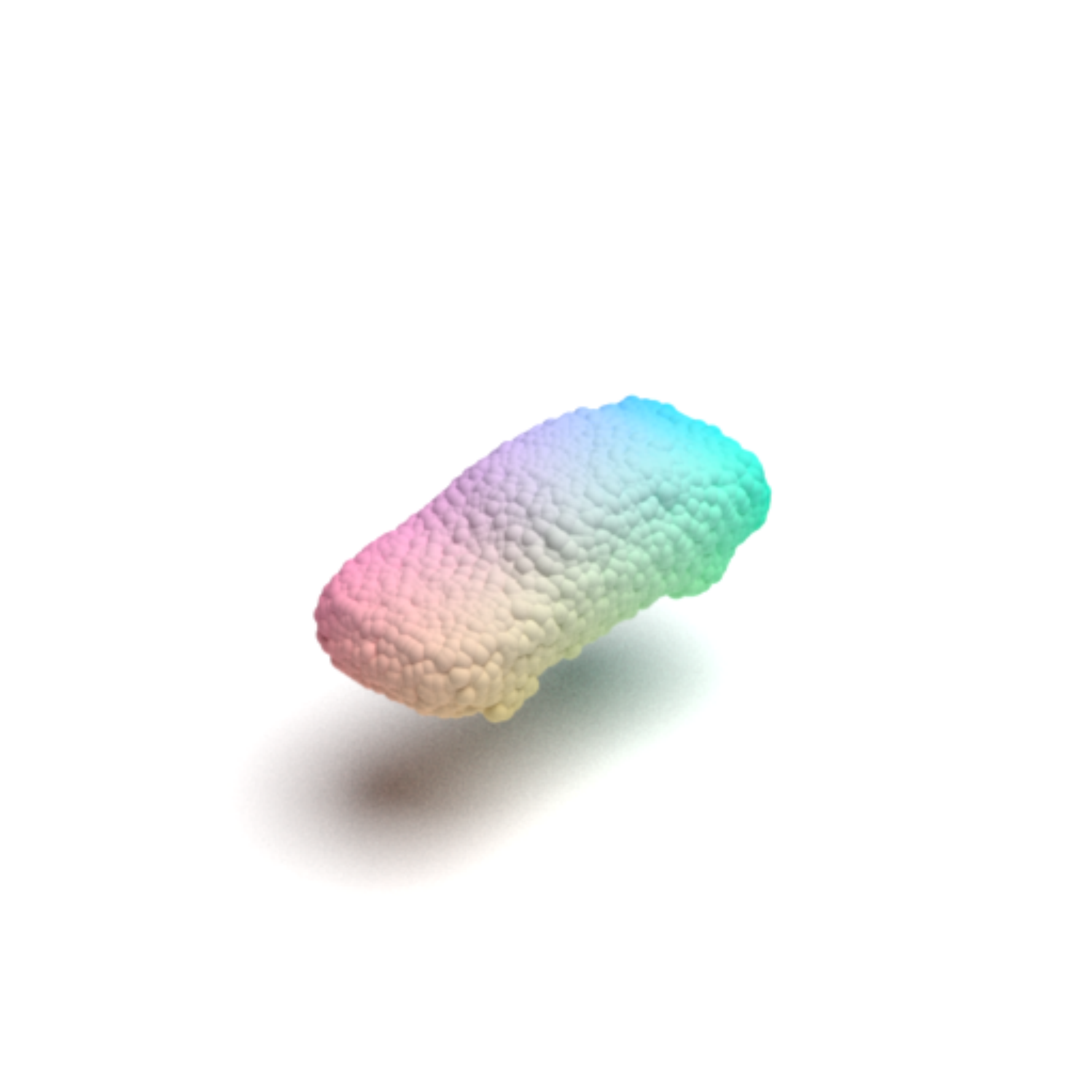}
\includegraphics[clip,trim=3cm 3cm 3cm 3cm, width=0.095\textwidth]{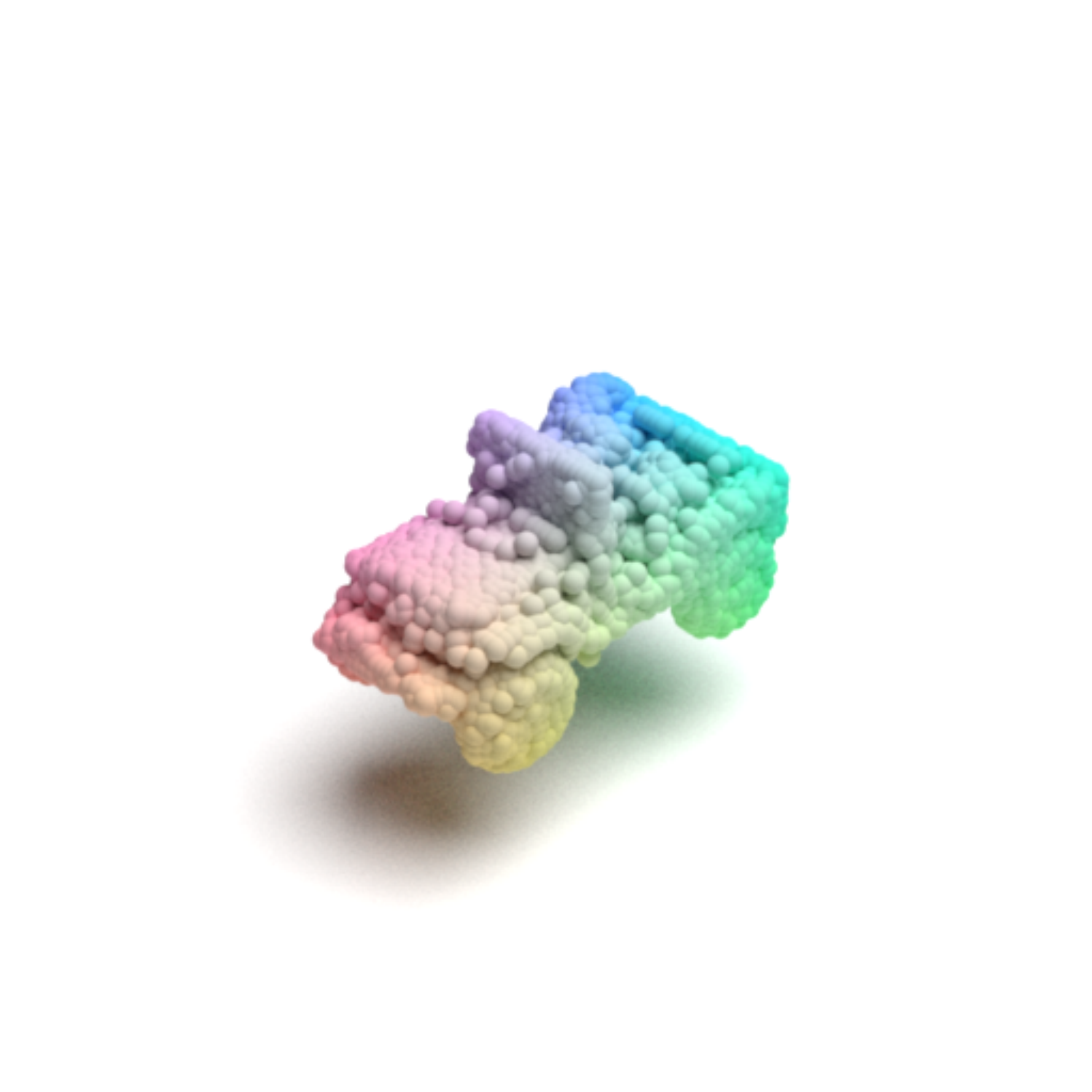}
\includegraphics[clip,trim=3cm 3cm 3cm 3cm, width=0.095\textwidth]{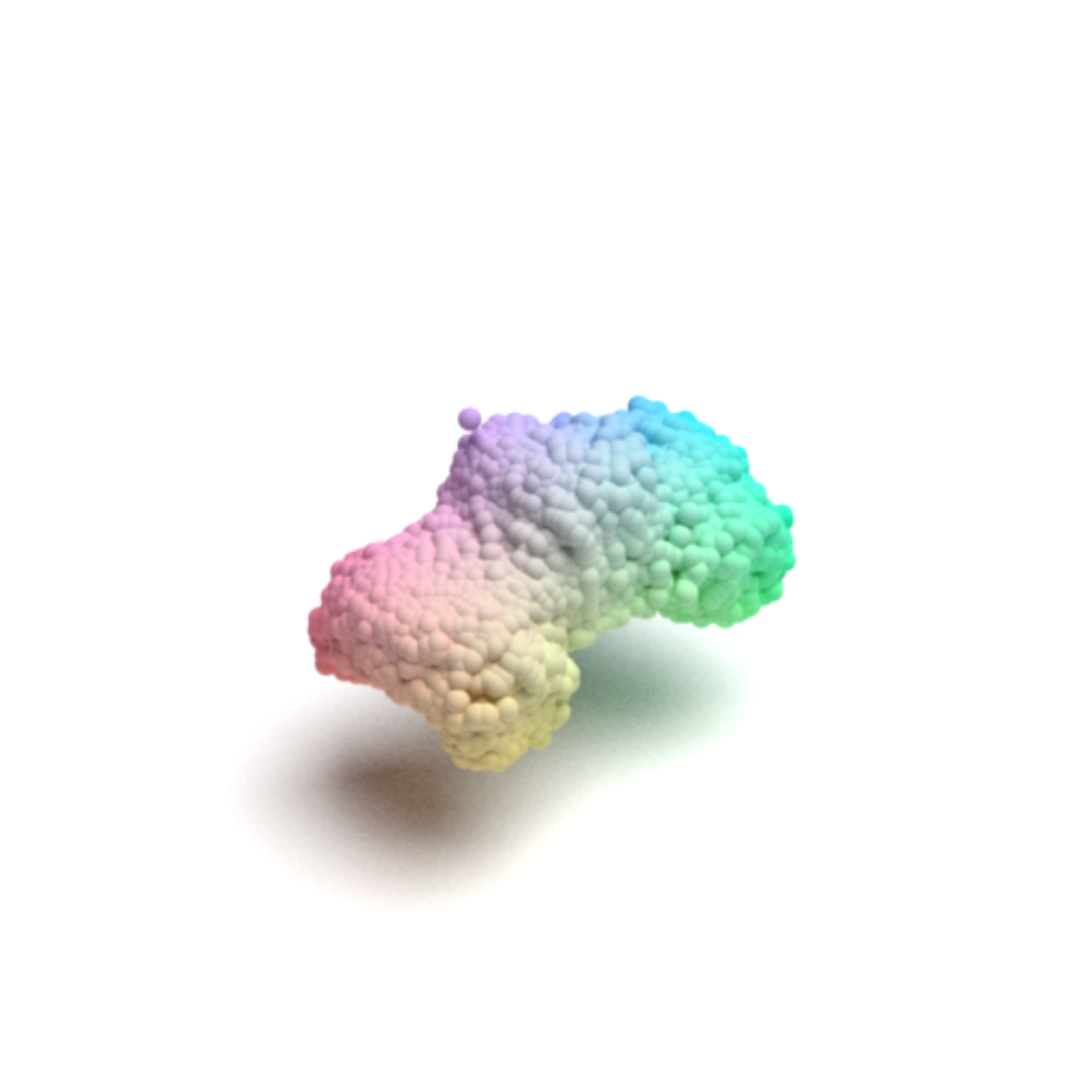}
\includegraphics[clip,trim=3cm 3cm 3cm 3cm, width=0.095\textwidth]{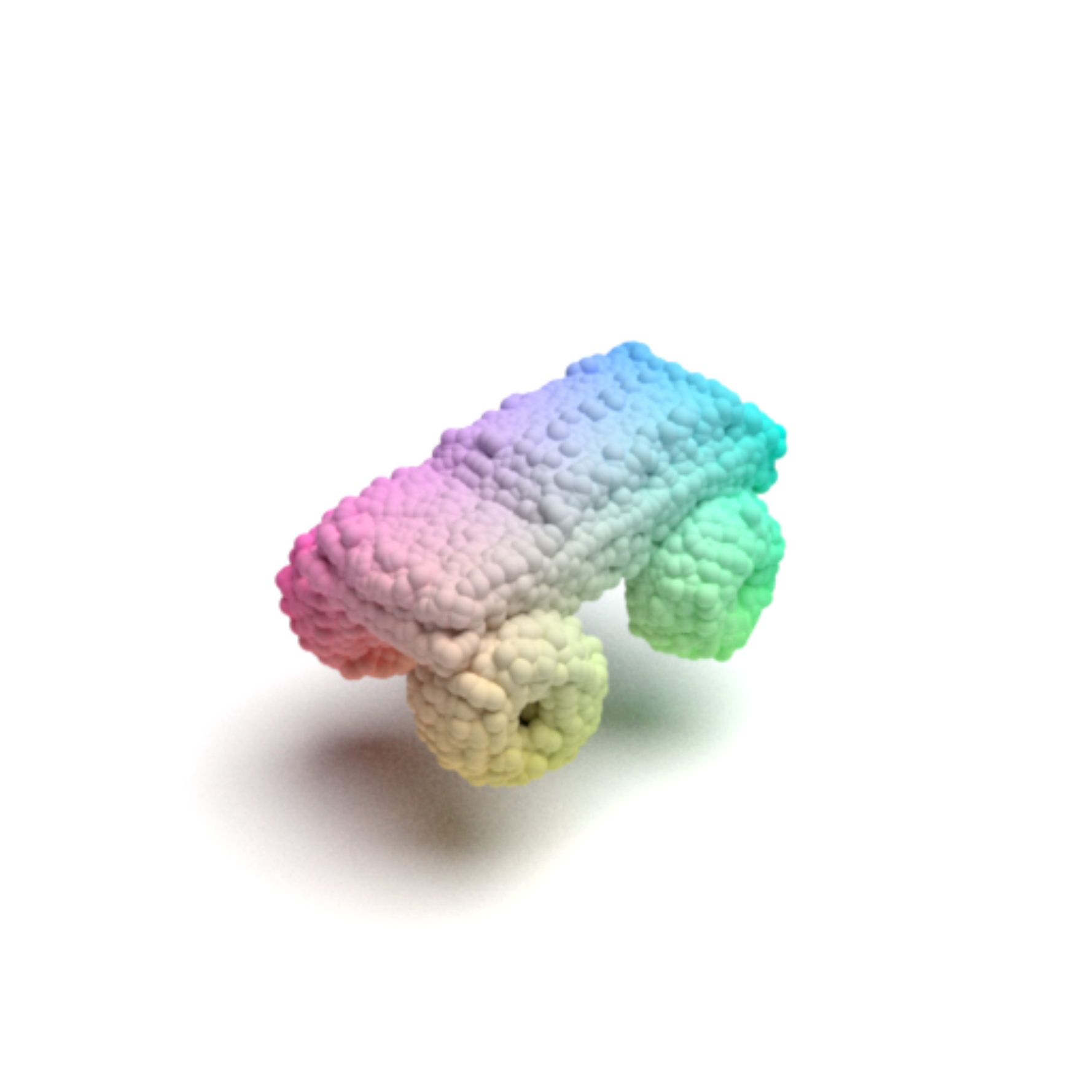}
\includegraphics[clip,trim=3cm 3cm 3cm 3cm, width=0.095\textwidth]{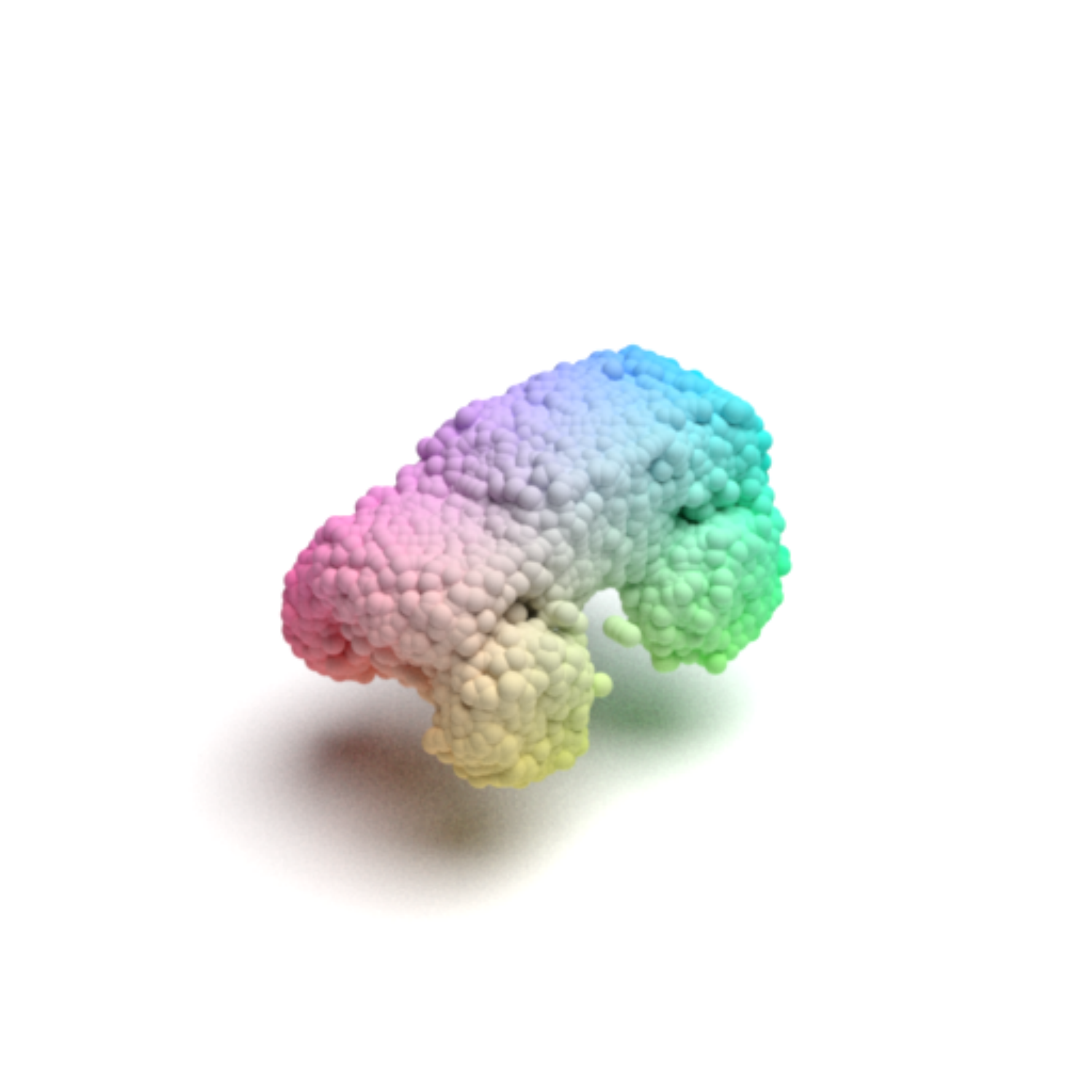}
\includegraphics[clip,trim=3cm 3cm 3cm 3cm, width=0.095\textwidth]{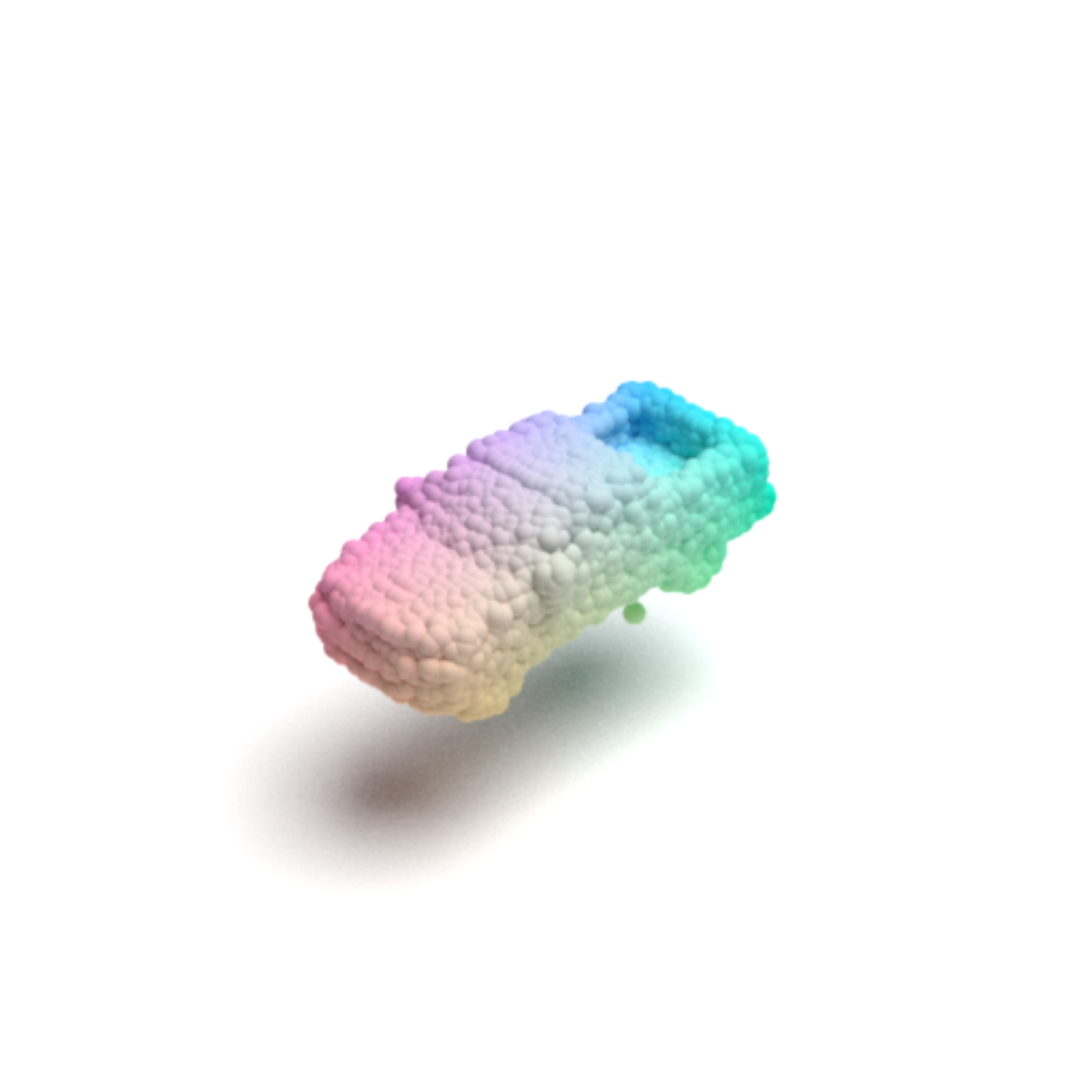}
\includegraphics[clip,trim=3cm 3cm 3cm 3cm, width=0.095\textwidth]{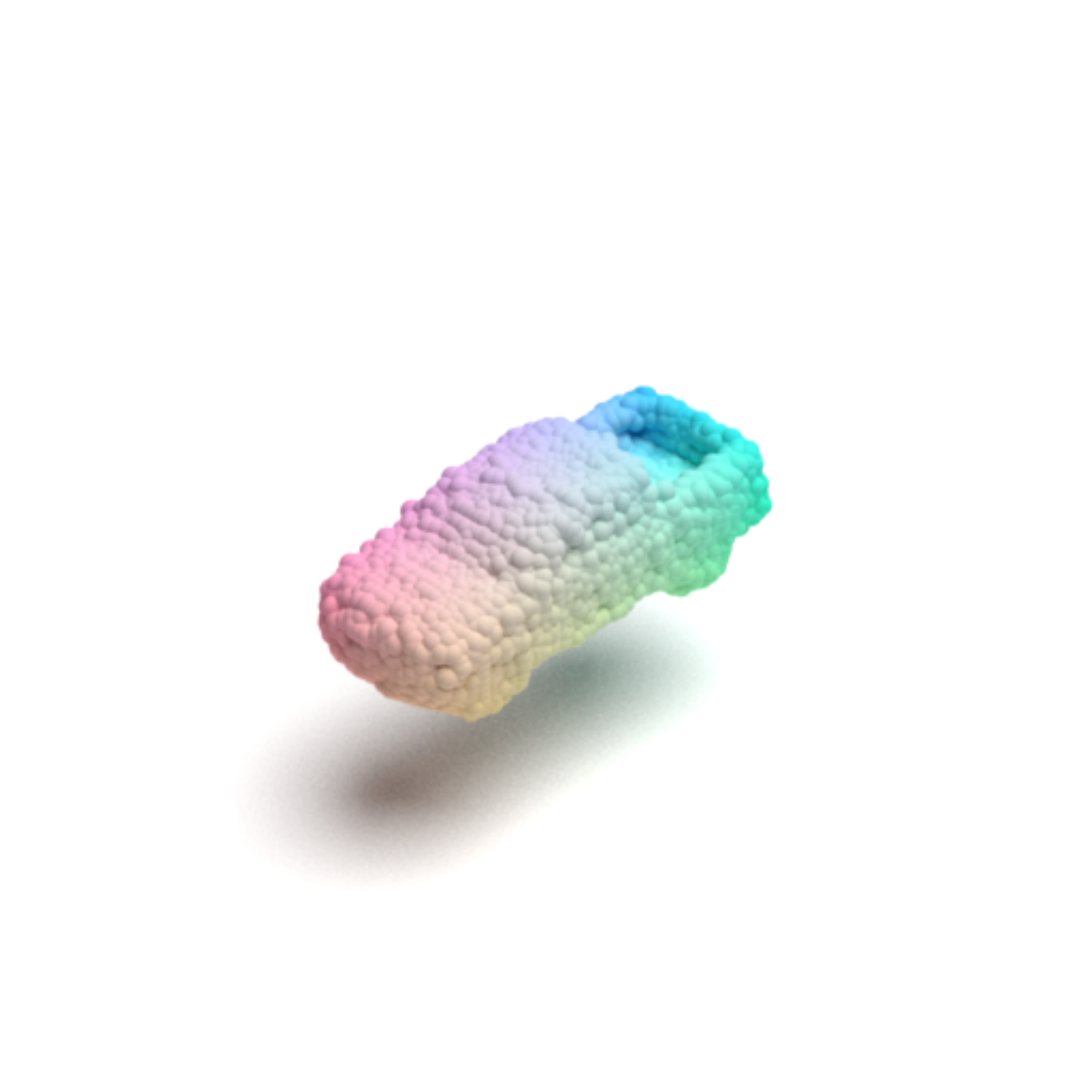}
\includegraphics[clip,trim=3cm 3cm 3cm 3cm, width=0.095\textwidth]{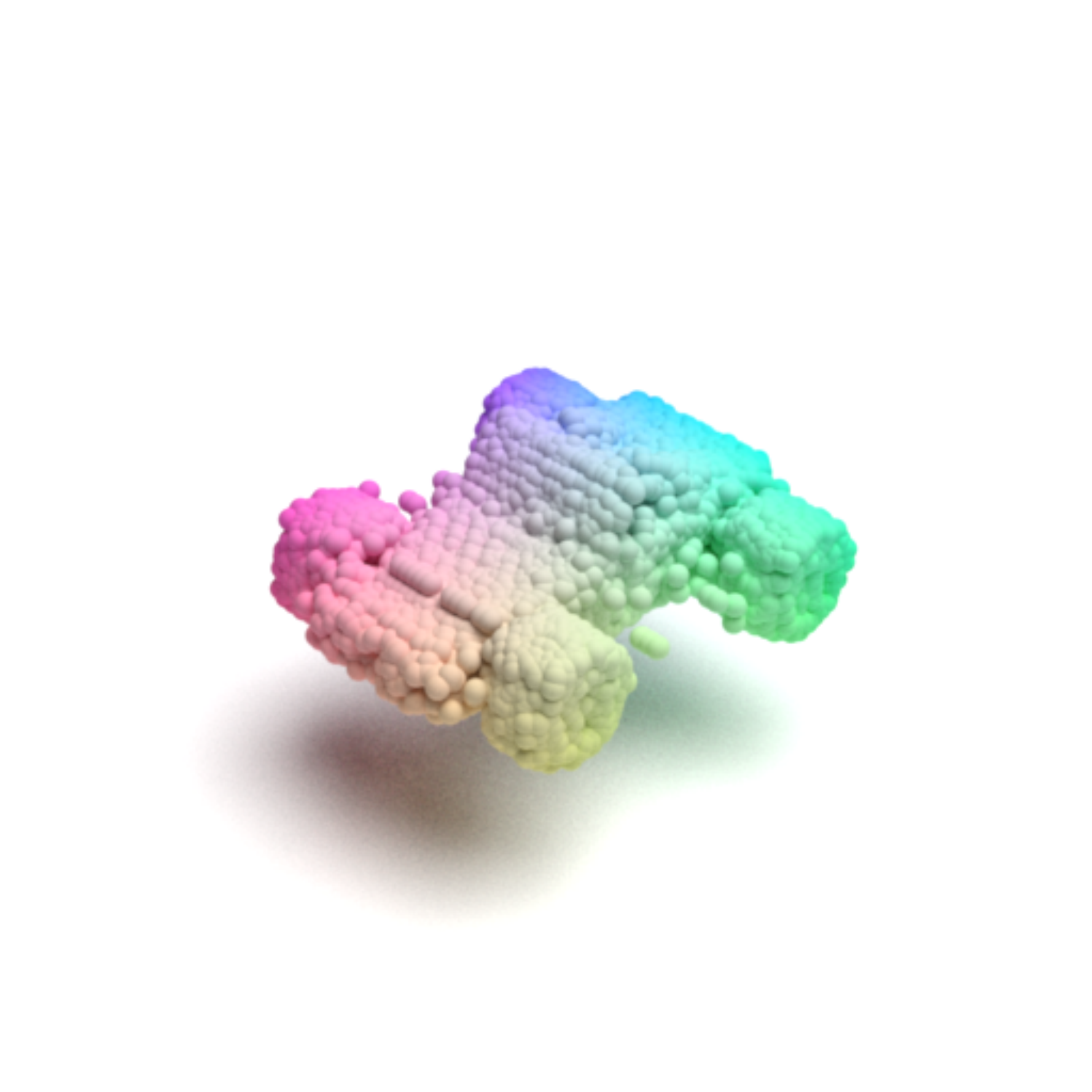}
\includegraphics[clip,trim=3cm 3cm 3cm 3cm, width=0.095\textwidth]{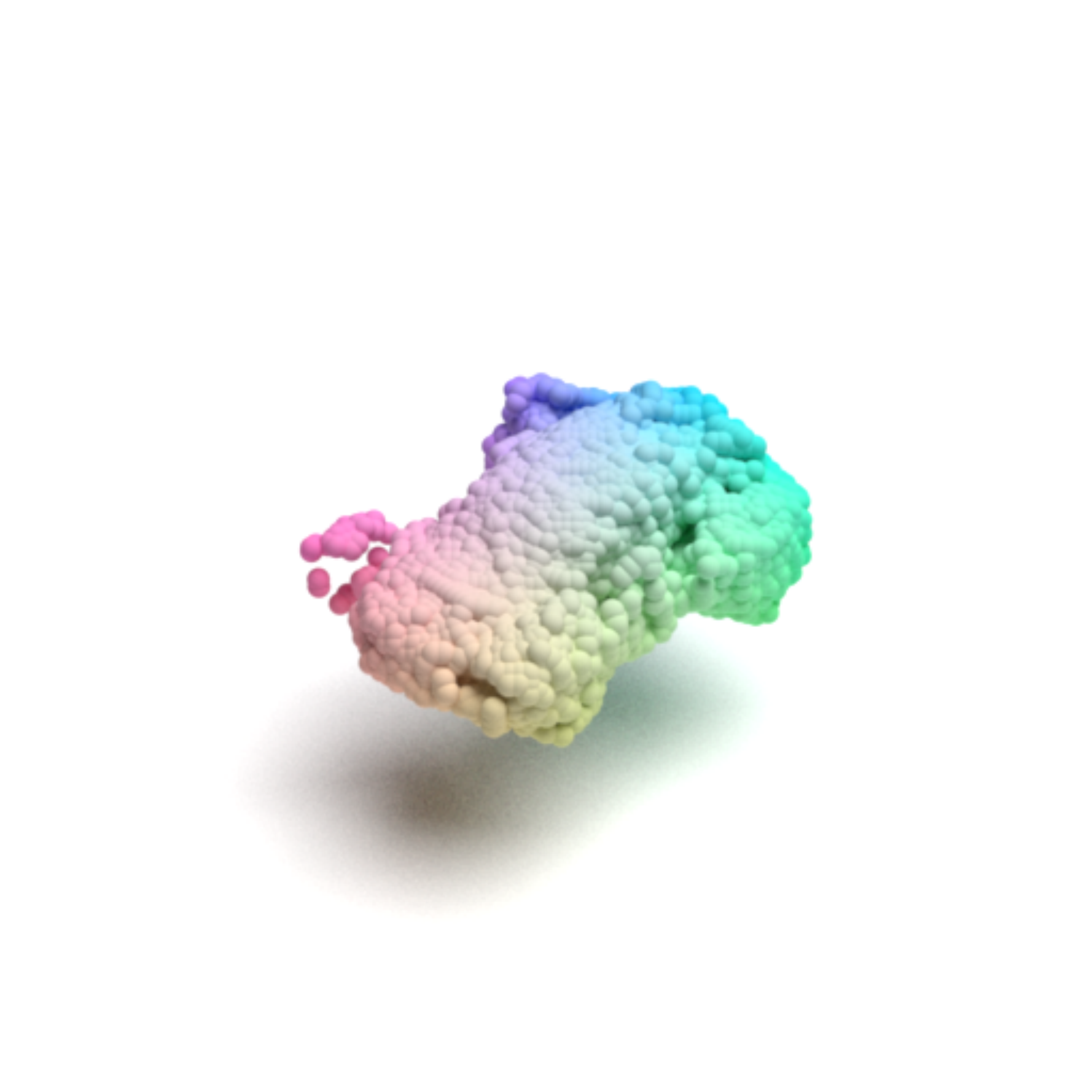}

\includegraphics[clip,trim=3cm 3cm 3cm 3cm, width=0.095\textwidth]{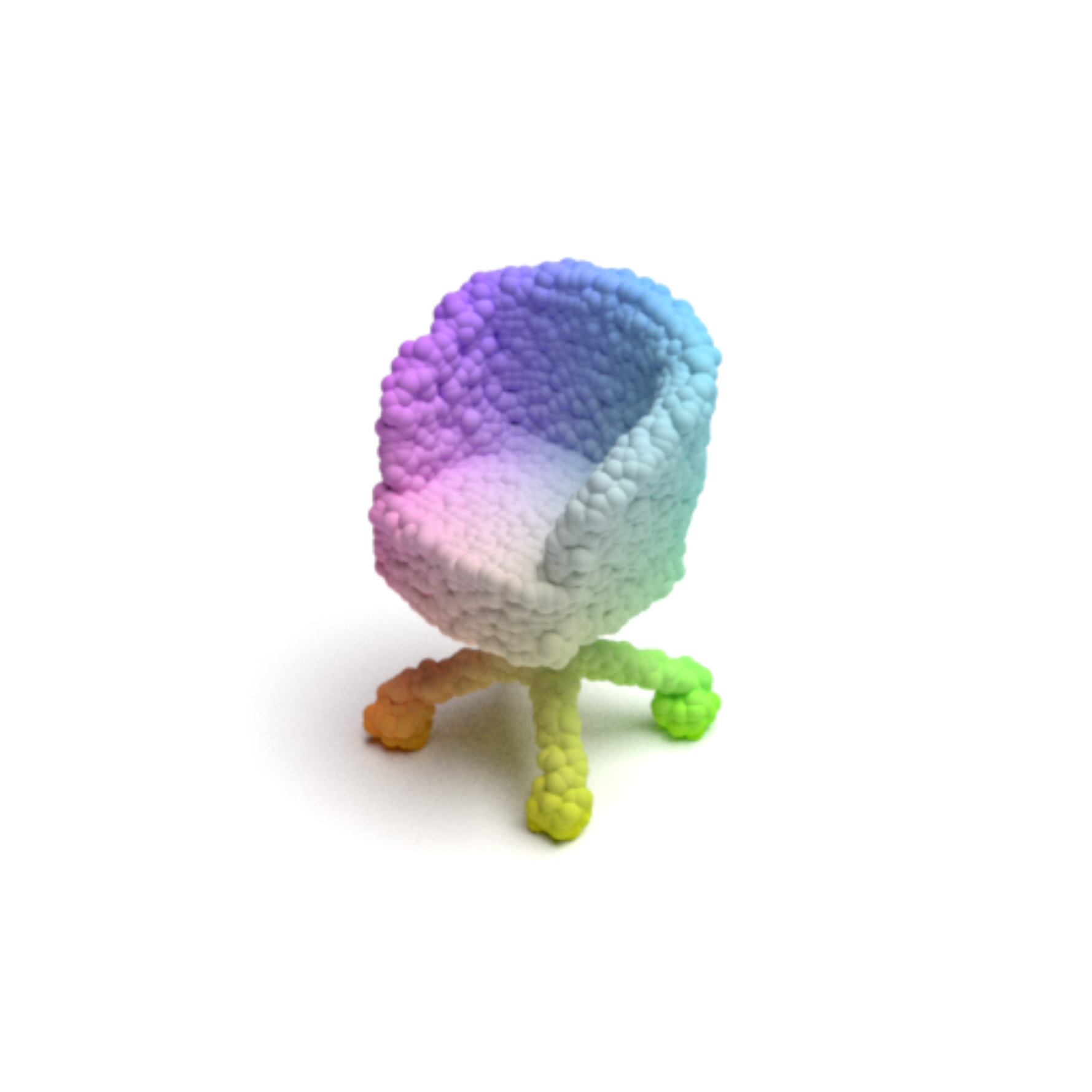}
\includegraphics[clip,trim=3cm 3cm 3cm 3cm, width=0.095\textwidth]{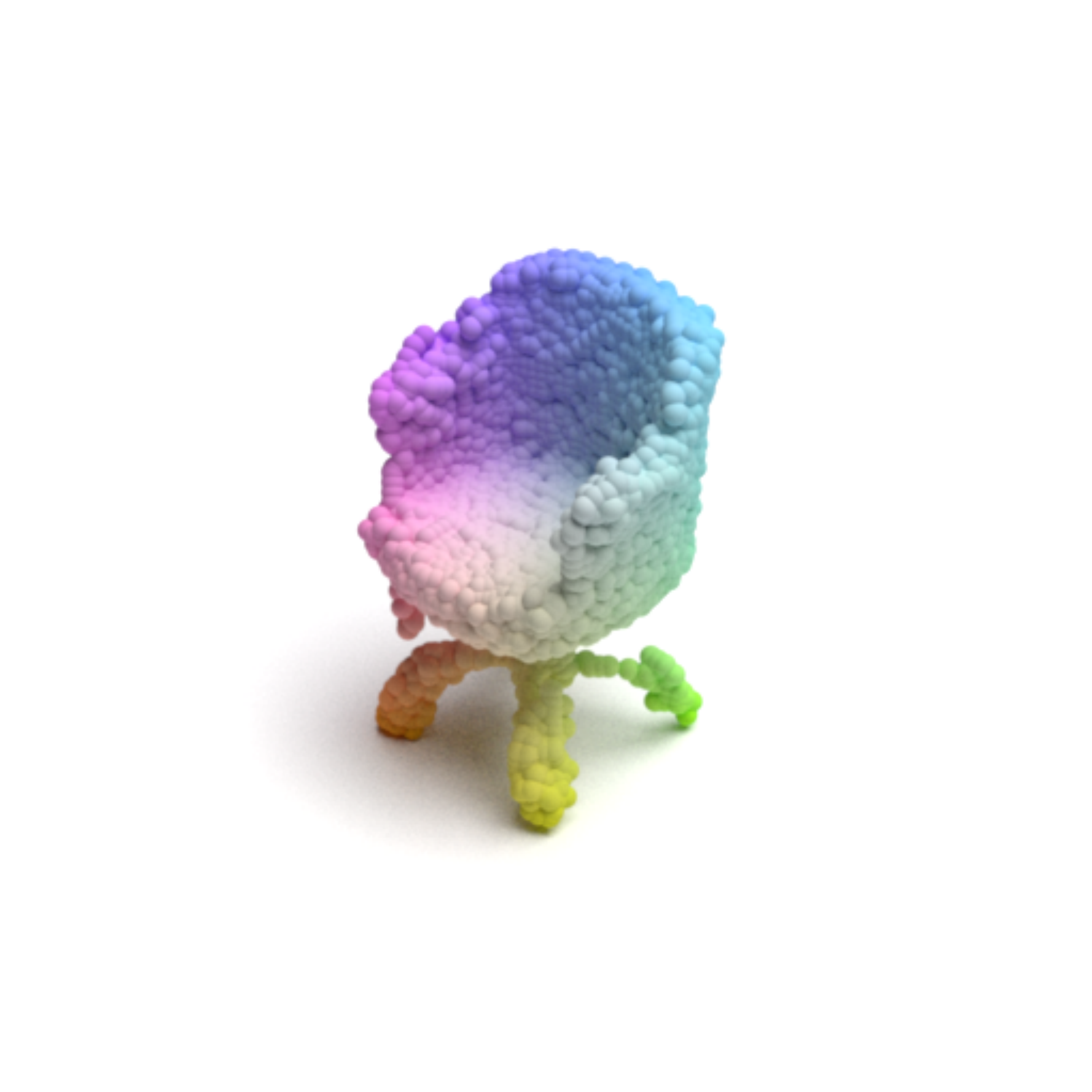}
\includegraphics[clip,trim=3cm 3cm 3cm 3cm, width=0.095\textwidth]{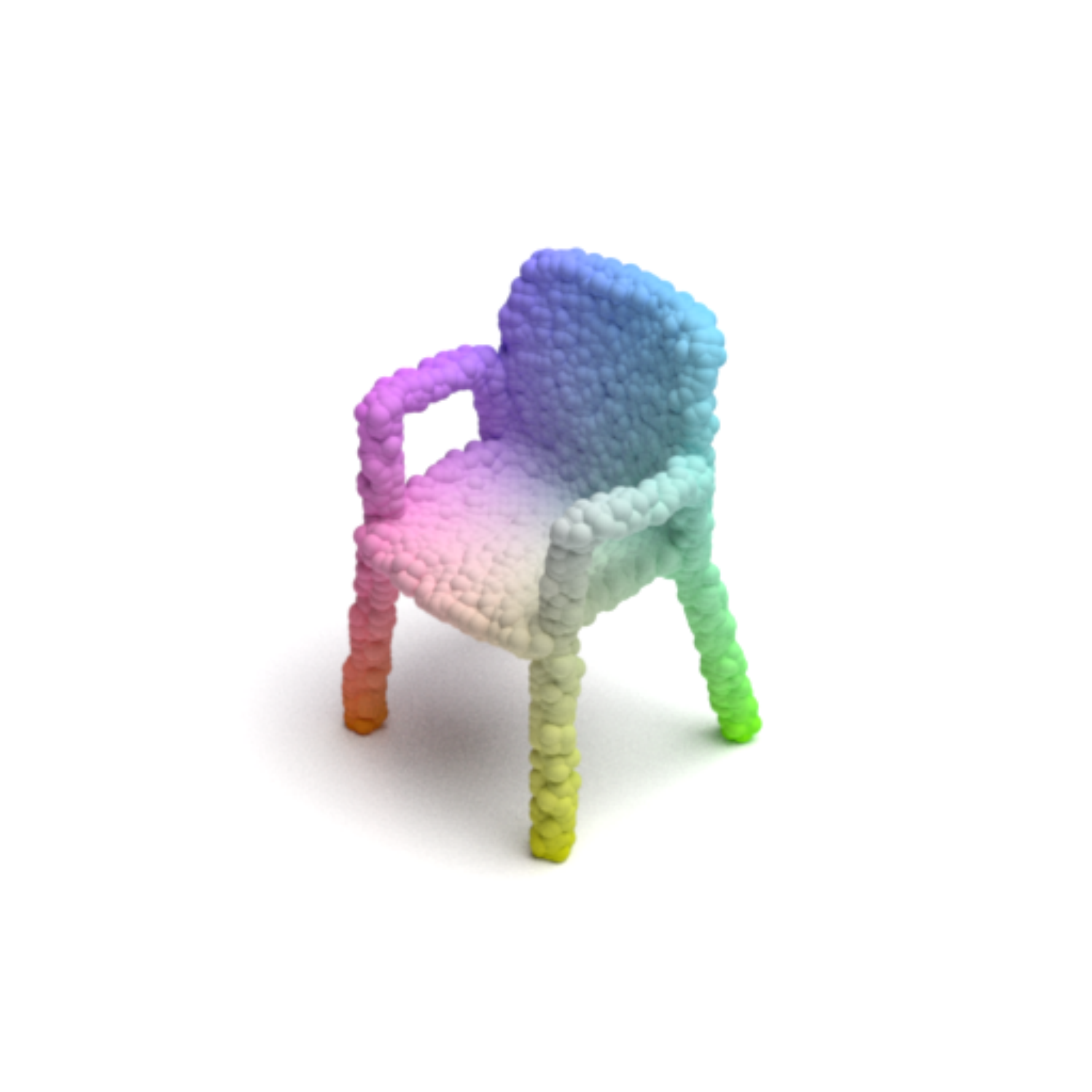}
\includegraphics[clip,trim=3cm 3cm 3cm 3cm, width=0.095\textwidth]{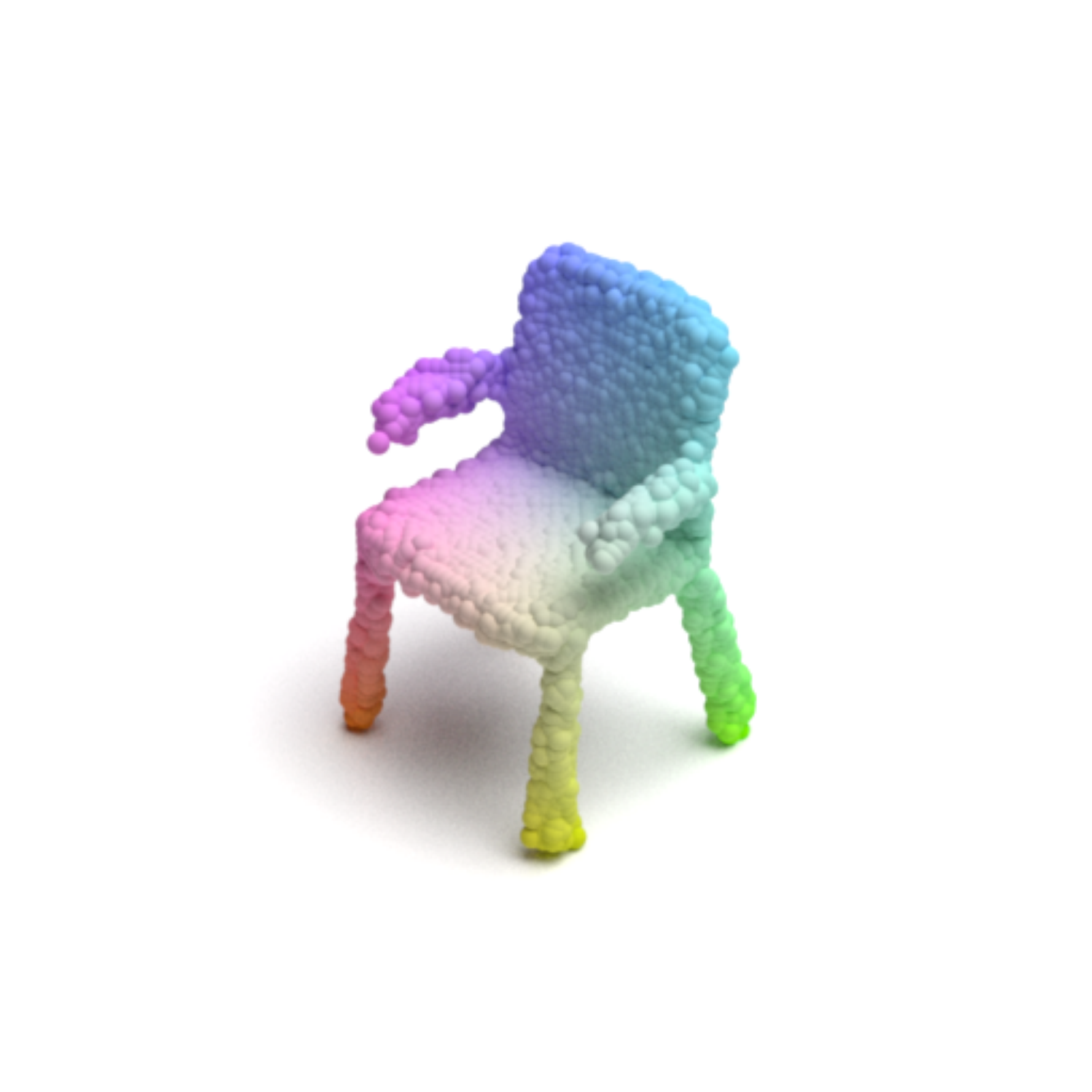}
\includegraphics[clip,trim=3cm 3cm 3cm 3cm, width=0.095\textwidth]{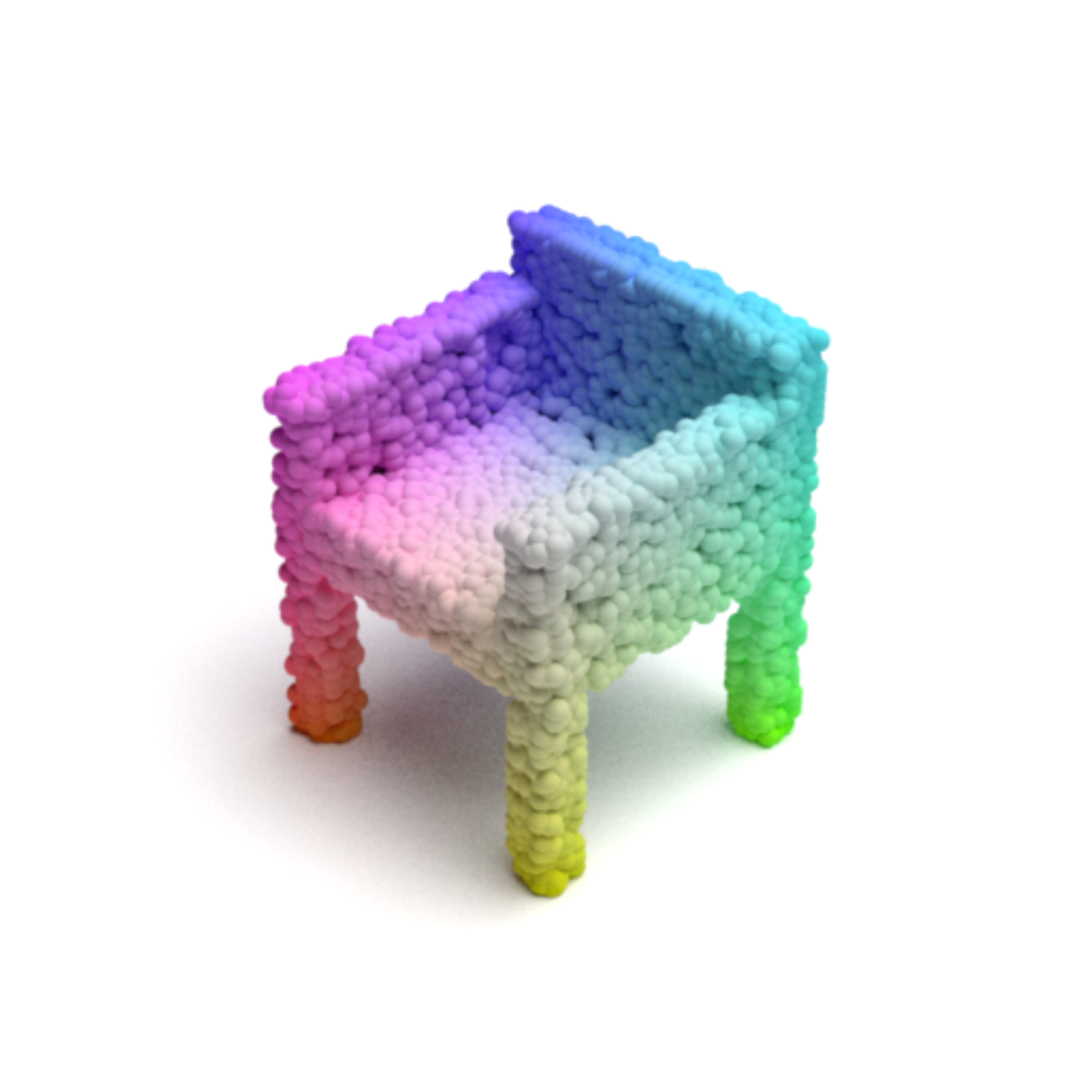}
\includegraphics[clip,trim=3cm 3cm 3cm 3cm, width=0.095\textwidth]{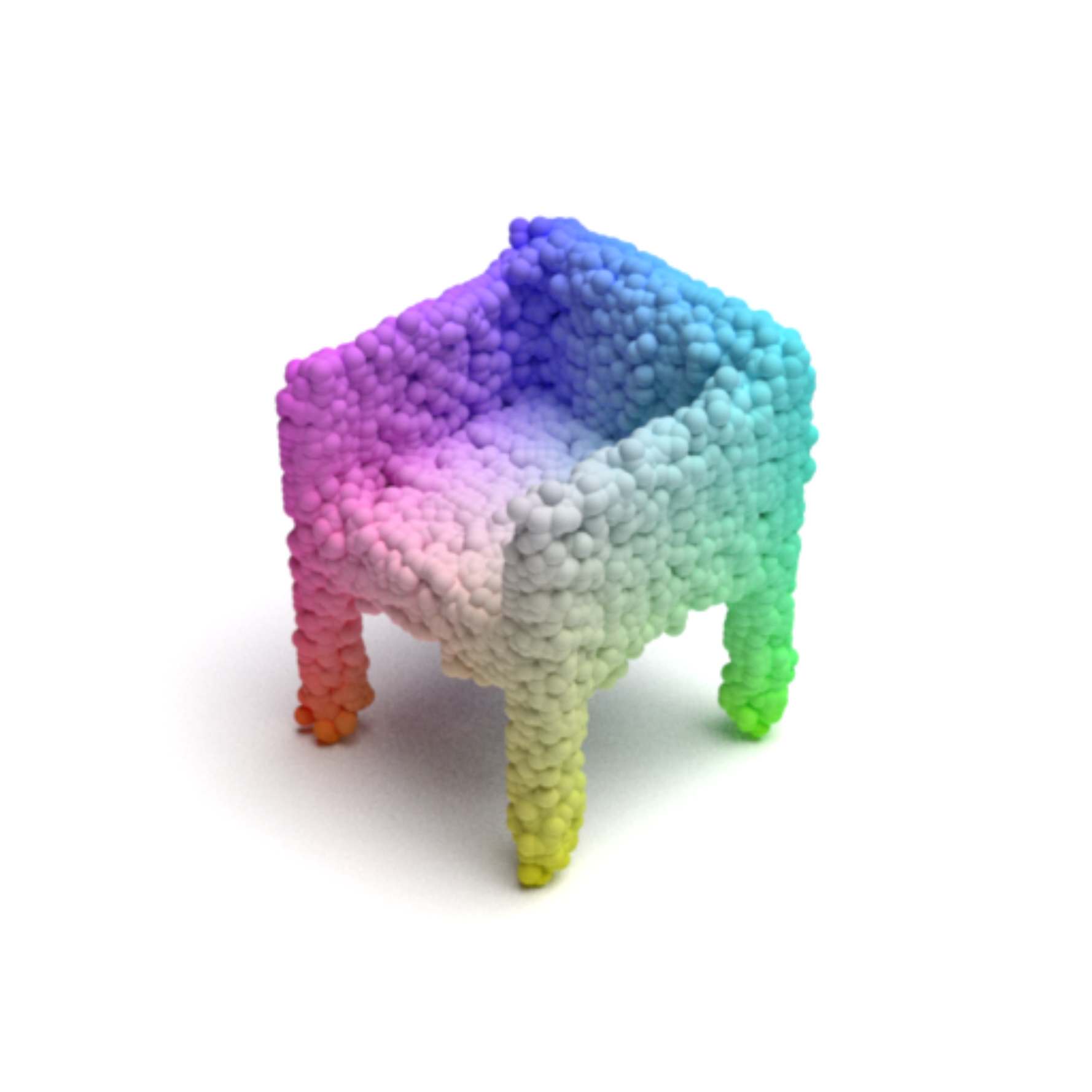}
\includegraphics[clip,trim=3cm 3cm 3cm 3cm, width=0.095\textwidth]{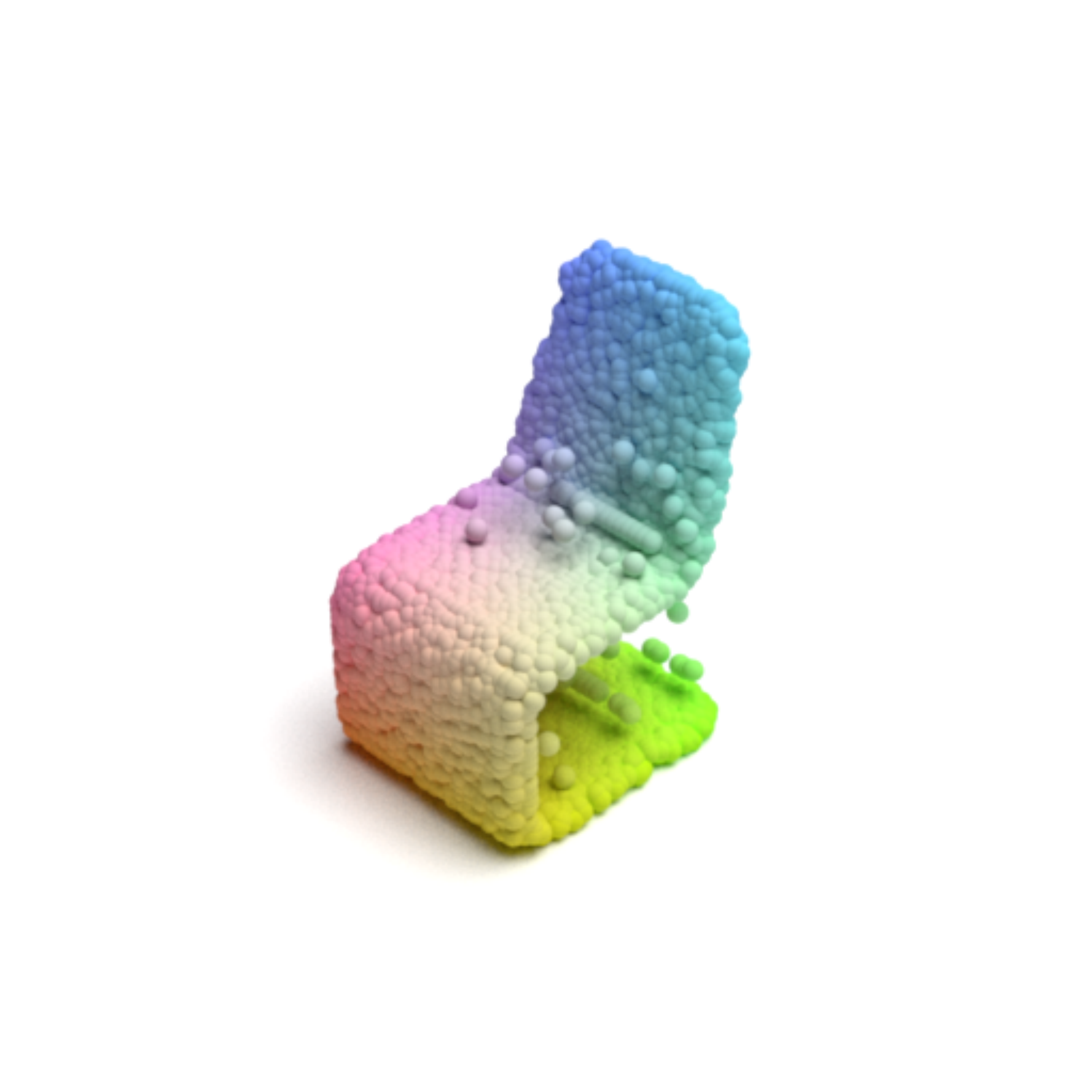}
\includegraphics[clip,trim=3cm 3cm 3cm 3cm, width=0.095\textwidth]{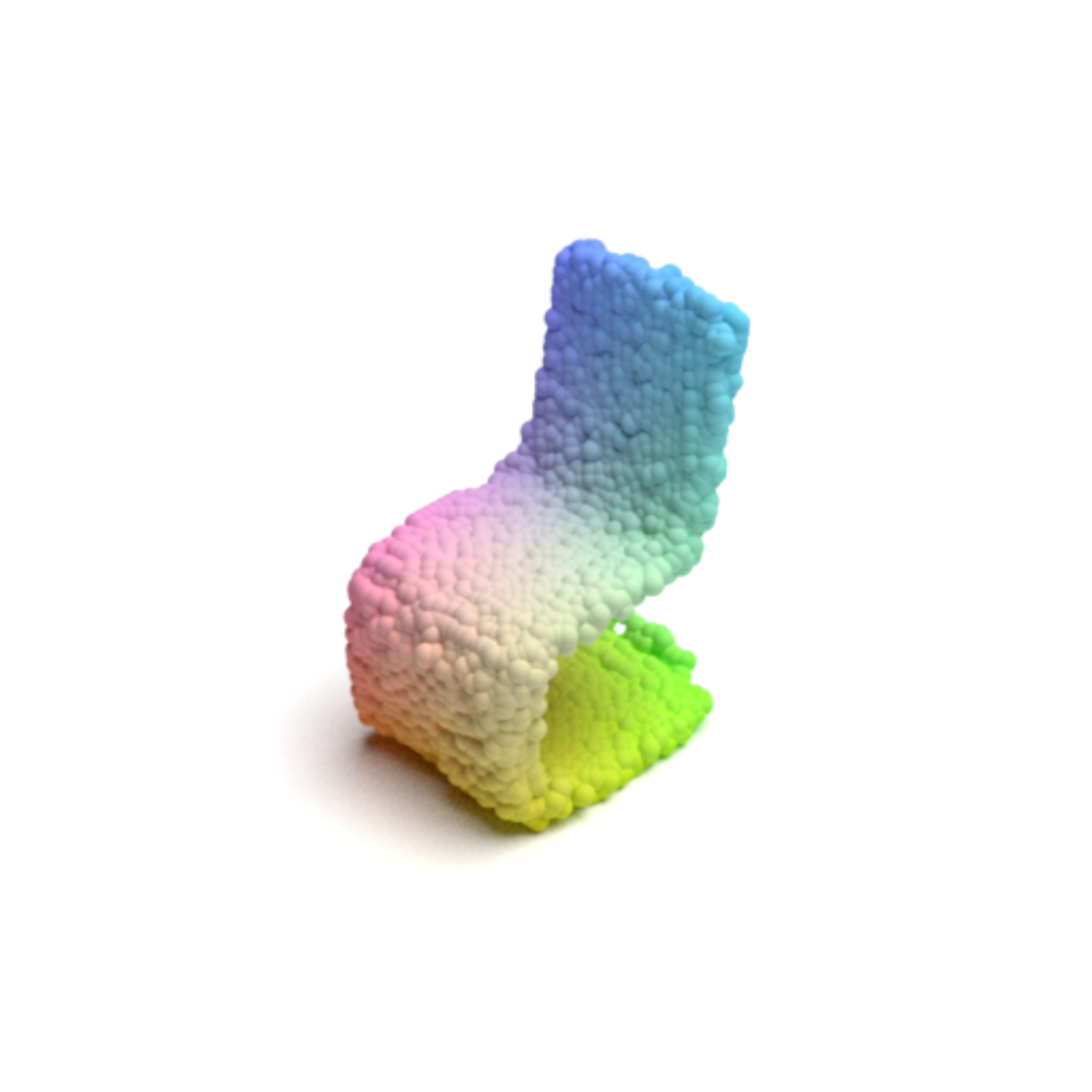}
\includegraphics[clip,trim=3cm 3cm 3cm 3cm, width=0.095\textwidth]{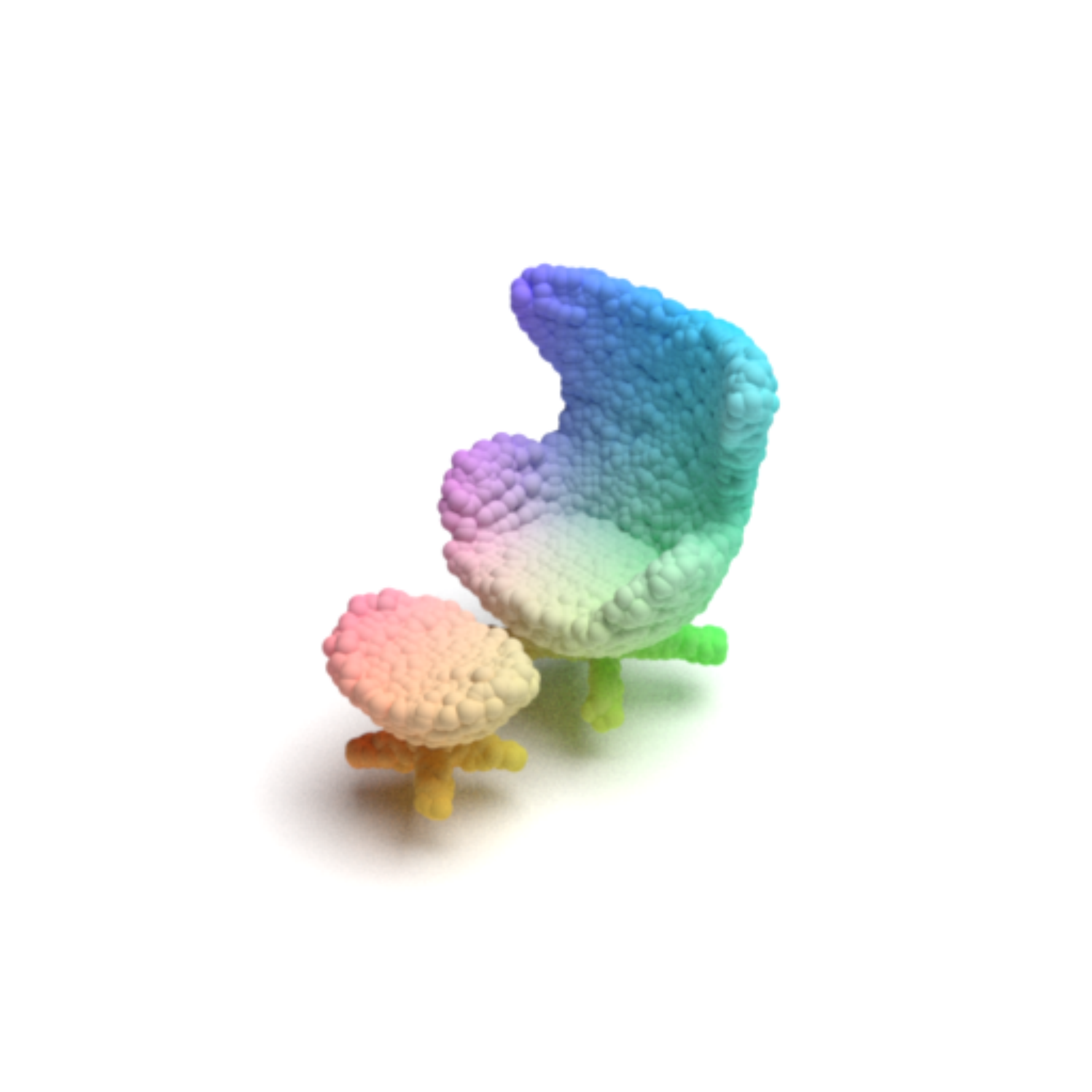}
\includegraphics[clip,trim=3cm 3cm 3cm 3cm, width=0.095\textwidth]{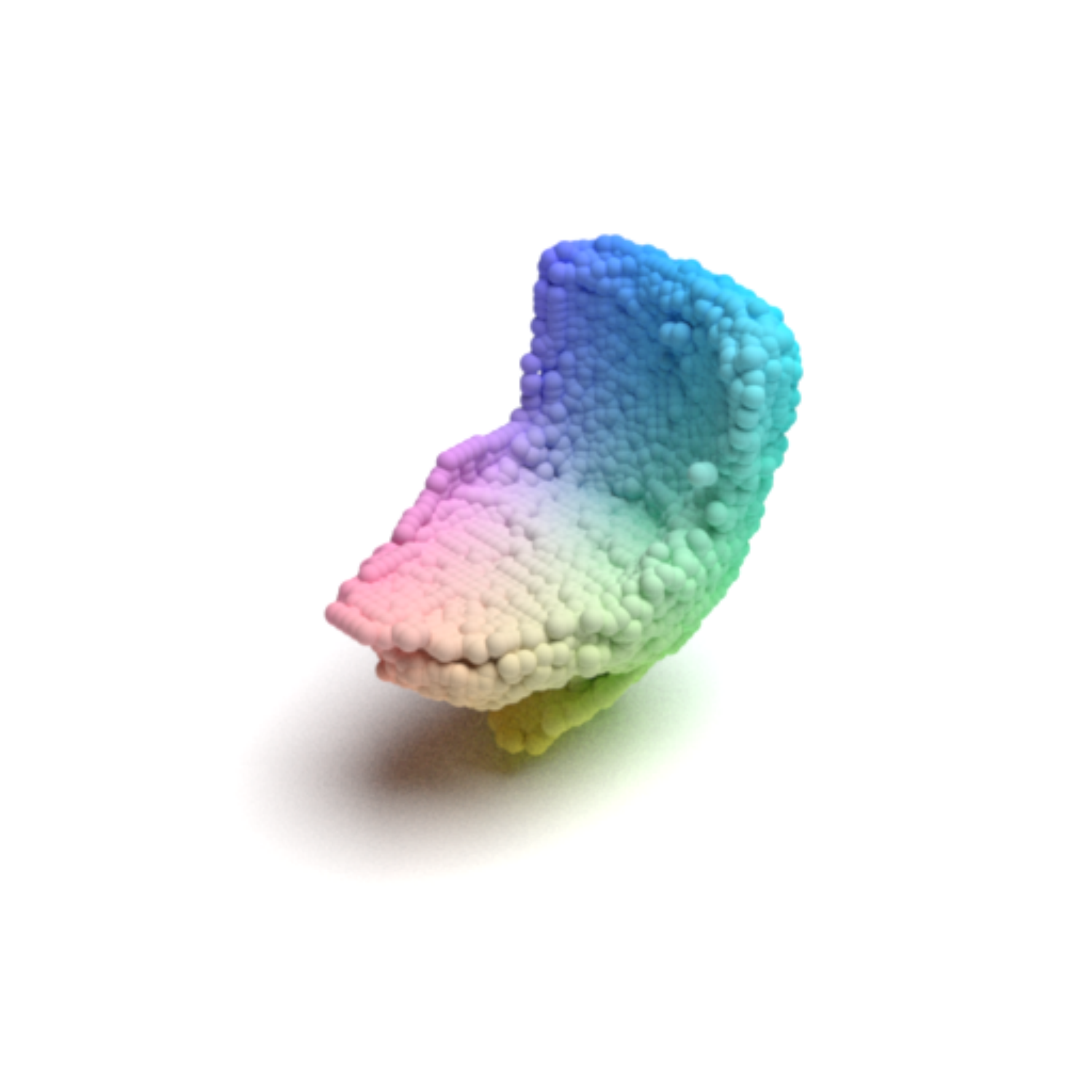}

\includegraphics[clip,trim=4cm 4cm 4cm 4cm, width=0.095\textwidth]{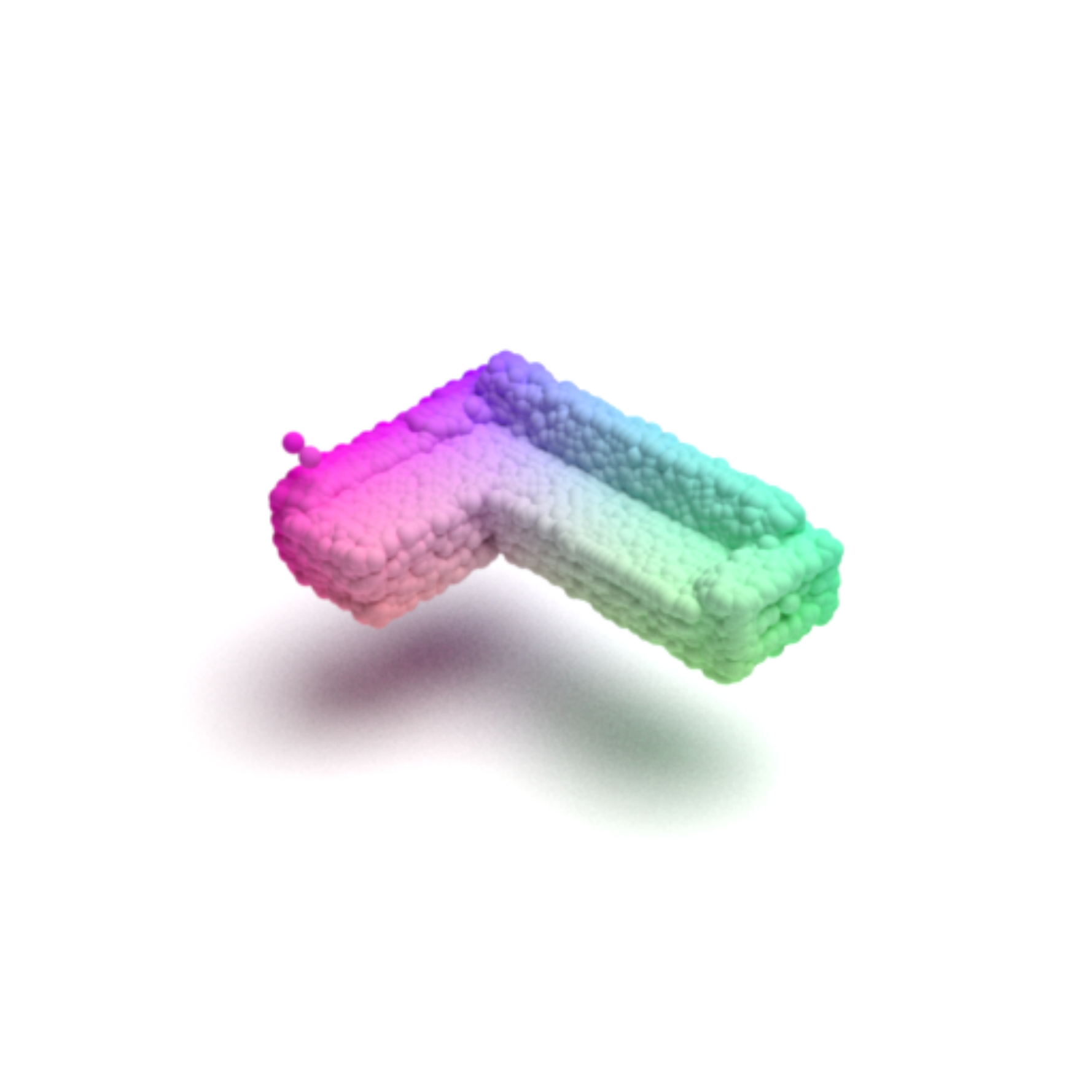}
\includegraphics[clip,trim=4cm 4cm 4cm 4cm, width=0.095\textwidth]{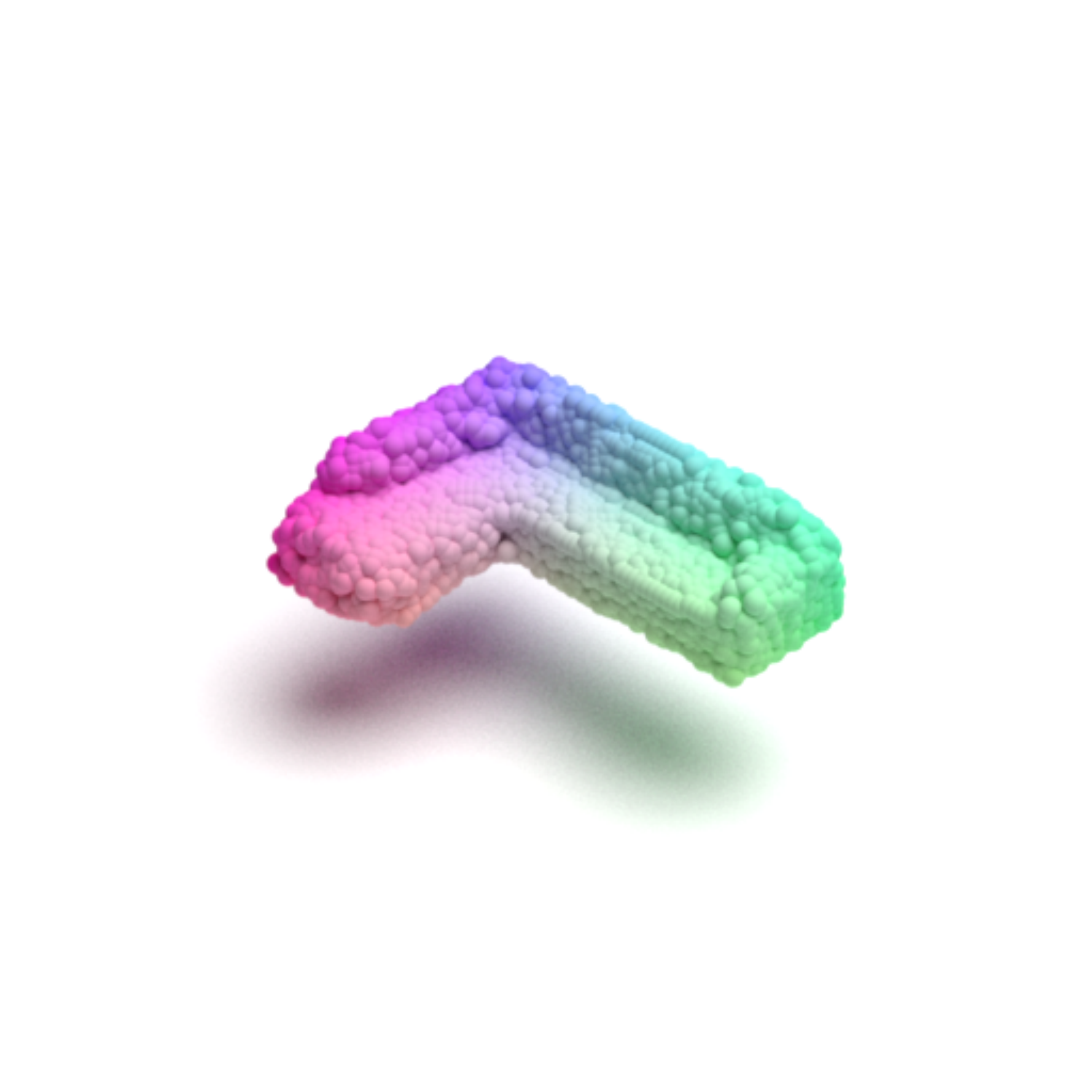}
\includegraphics[clip,trim=4cm 4cm 4cm 4cm, width=0.095\textwidth]{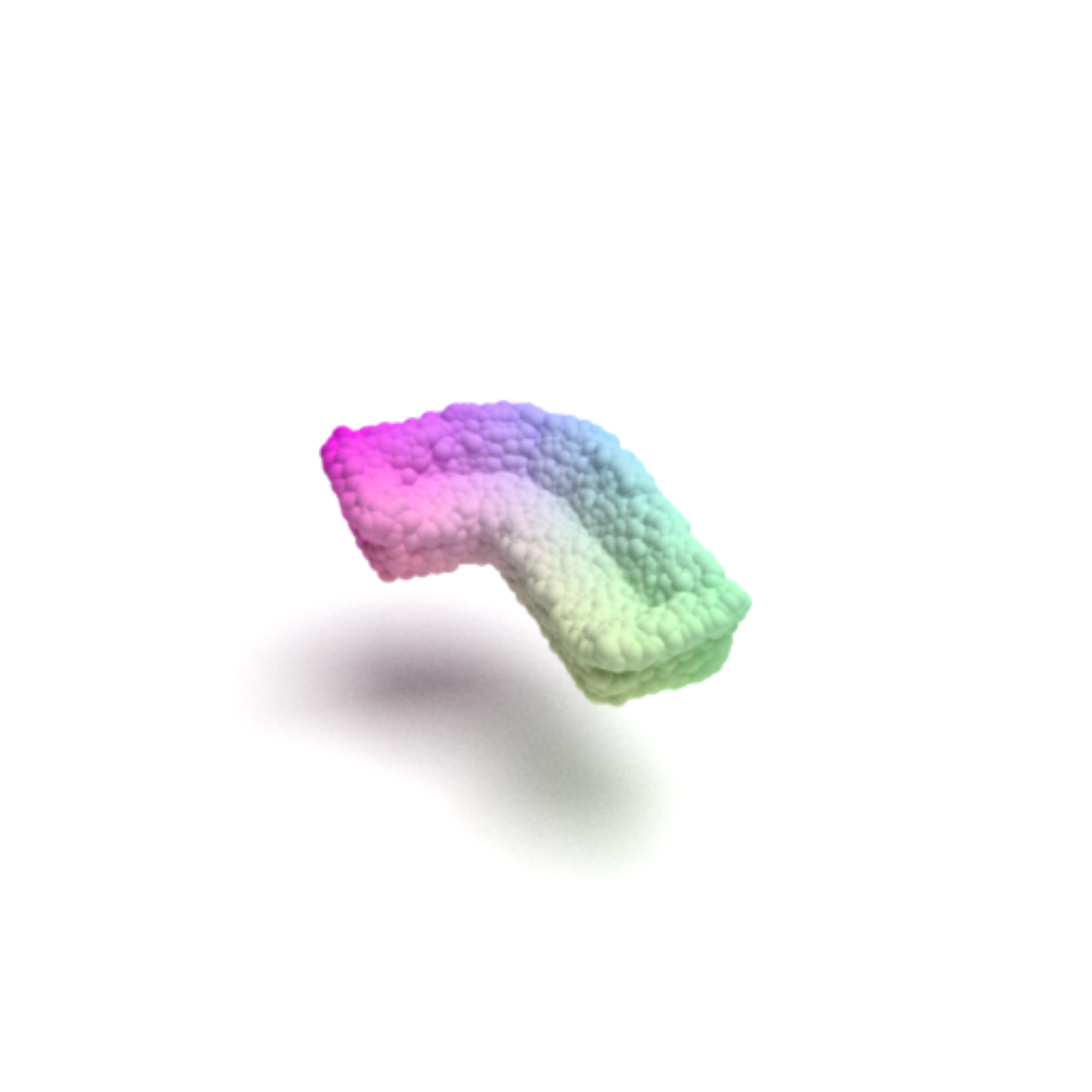}
\includegraphics[clip,trim=4cm 4cm 4cm 4cm, width=0.095\textwidth]{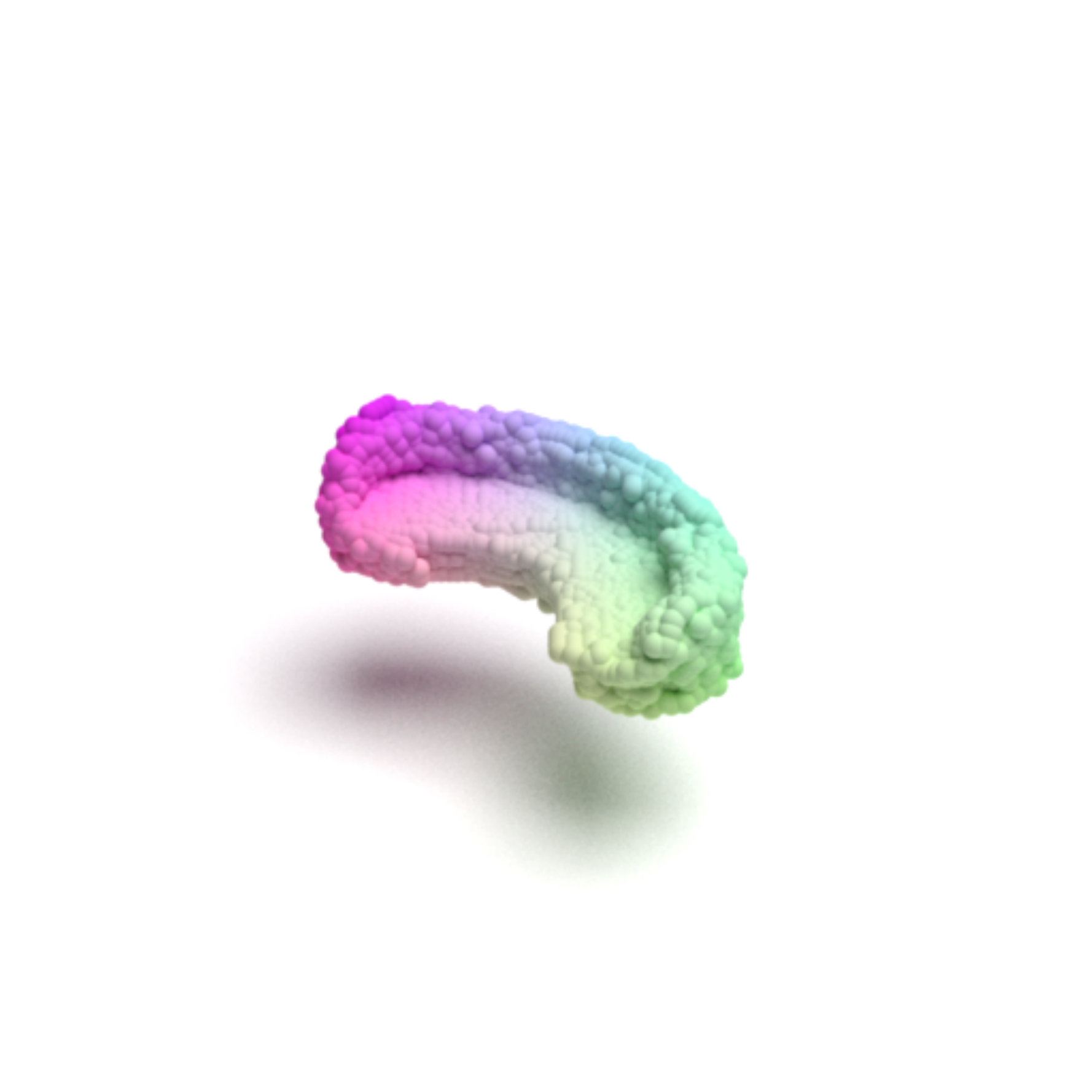}
\includegraphics[clip,trim=4cm 4cm 4cm 4cm, width=0.095\textwidth]{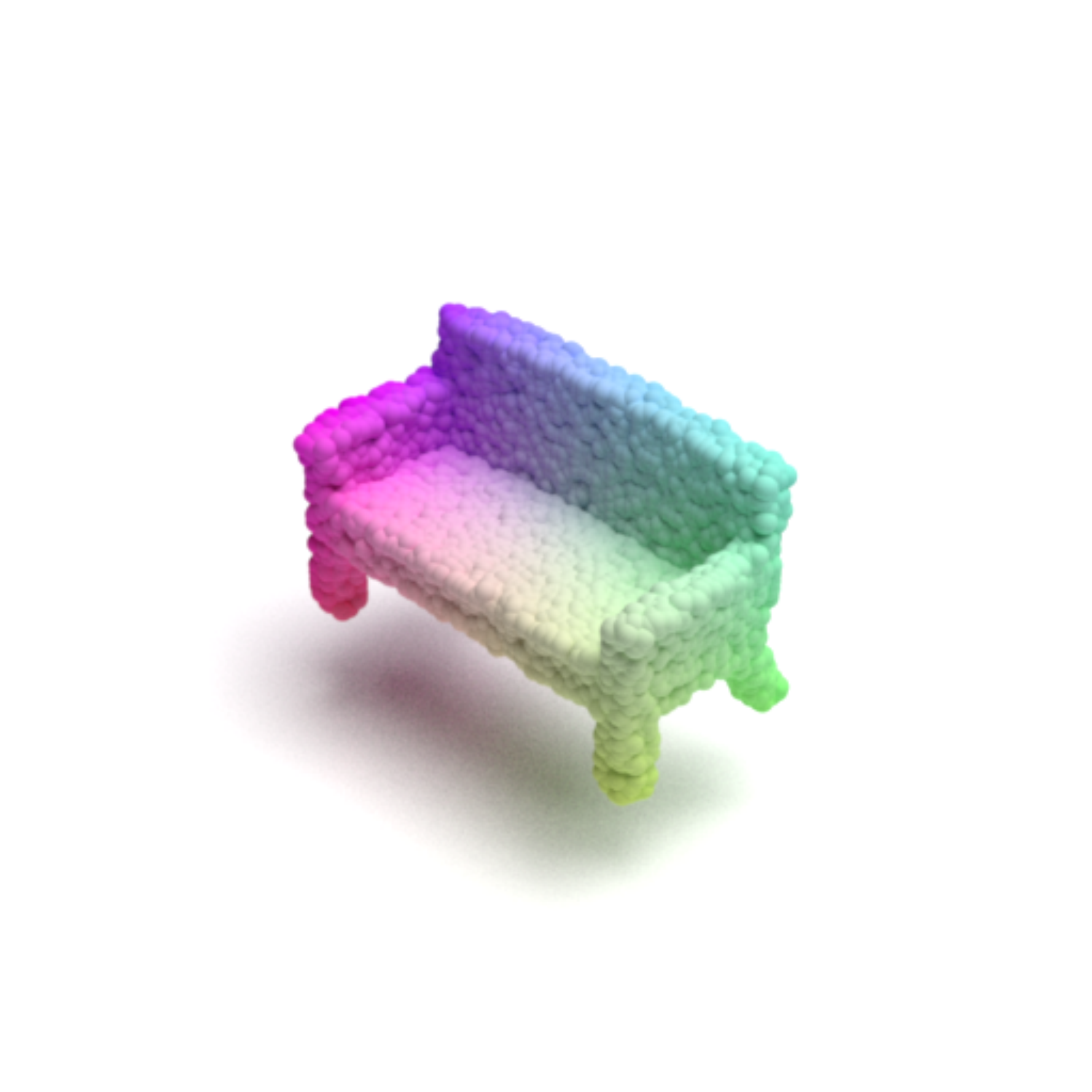}
\includegraphics[clip,trim=4cm 4cm 4cm 4cm, width=0.095\textwidth]{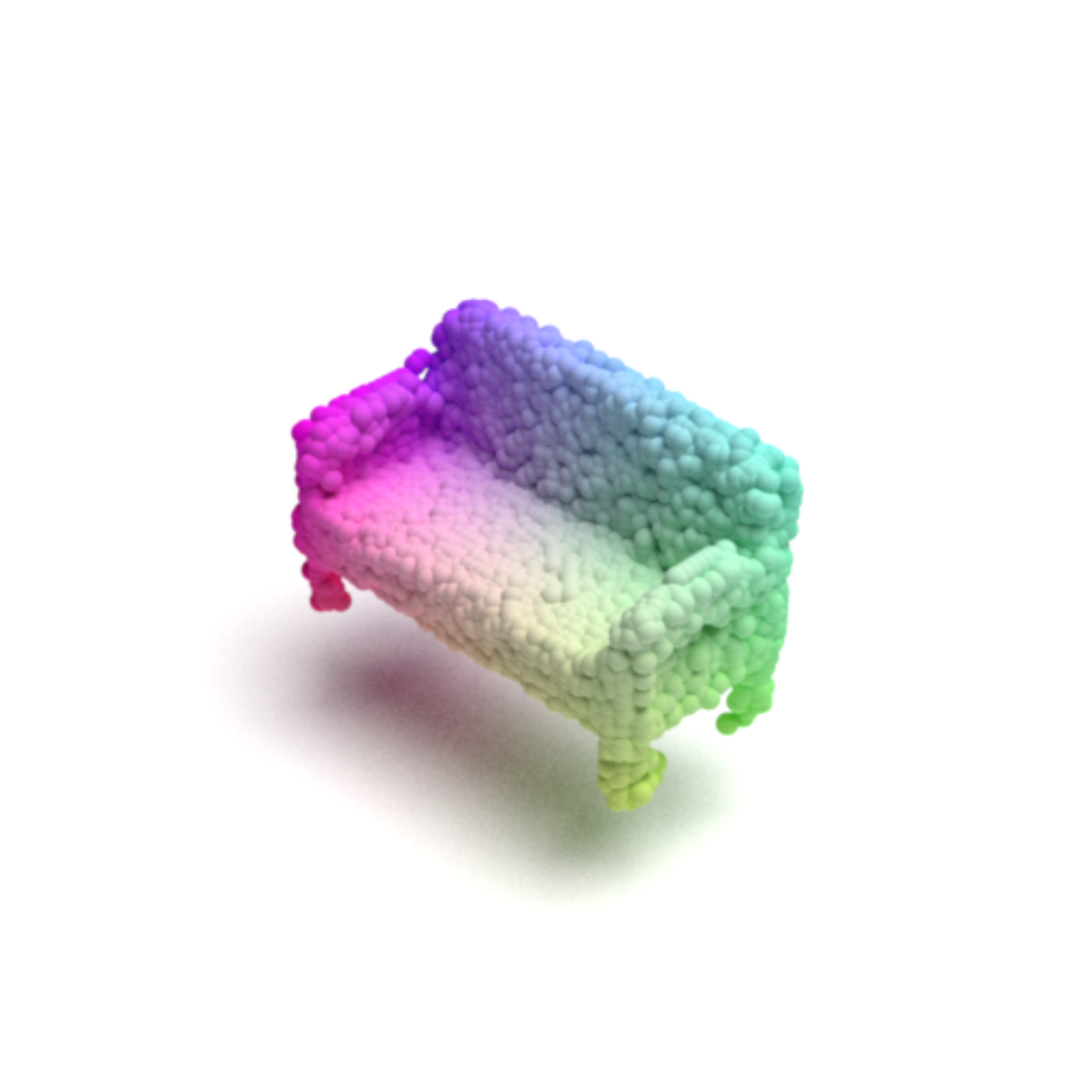}
\includegraphics[clip,trim=4cm 4cm 4cm 4cm, width=0.095\textwidth]{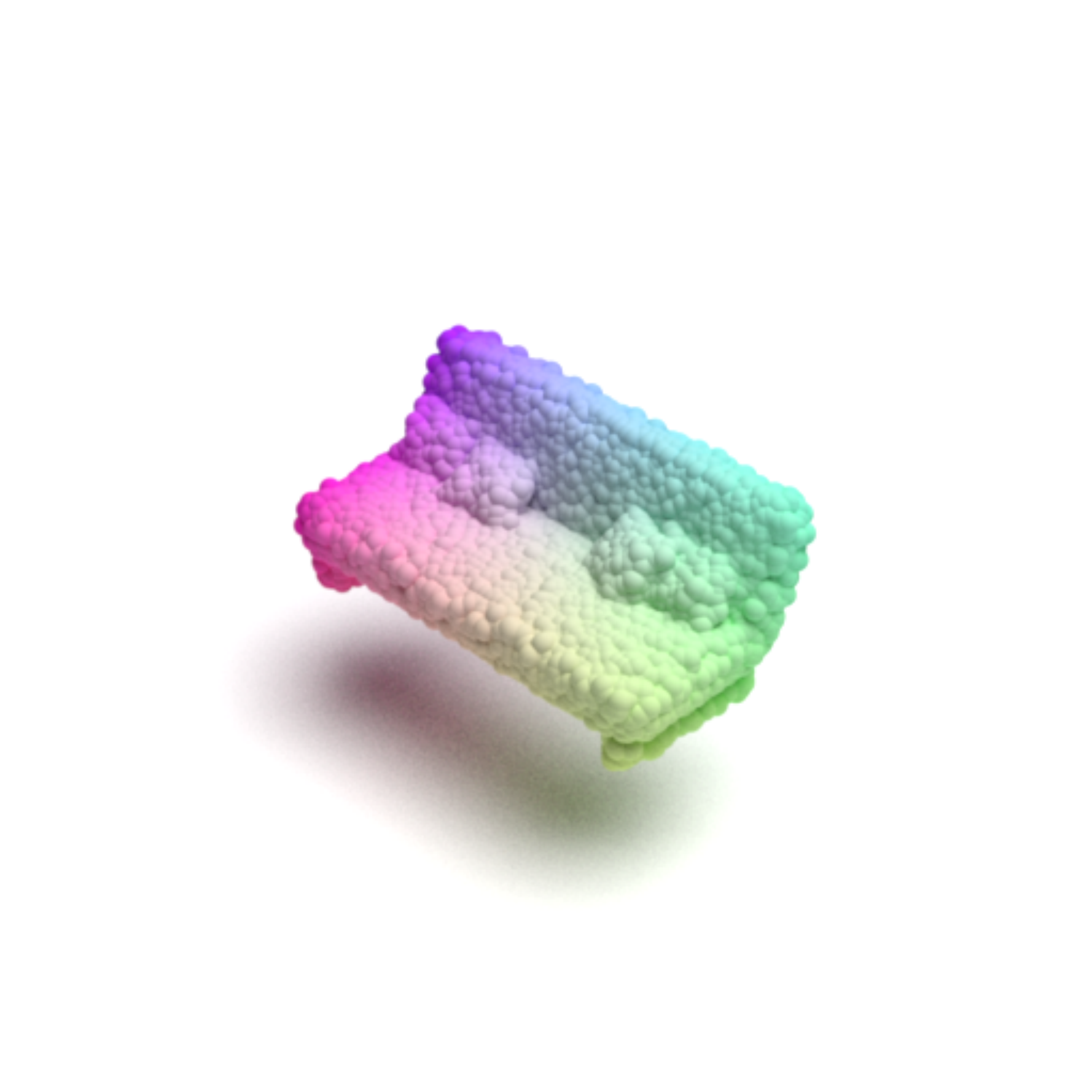}
\includegraphics[clip,trim=4cm 4cm 4cm 4cm, width=0.095\textwidth]{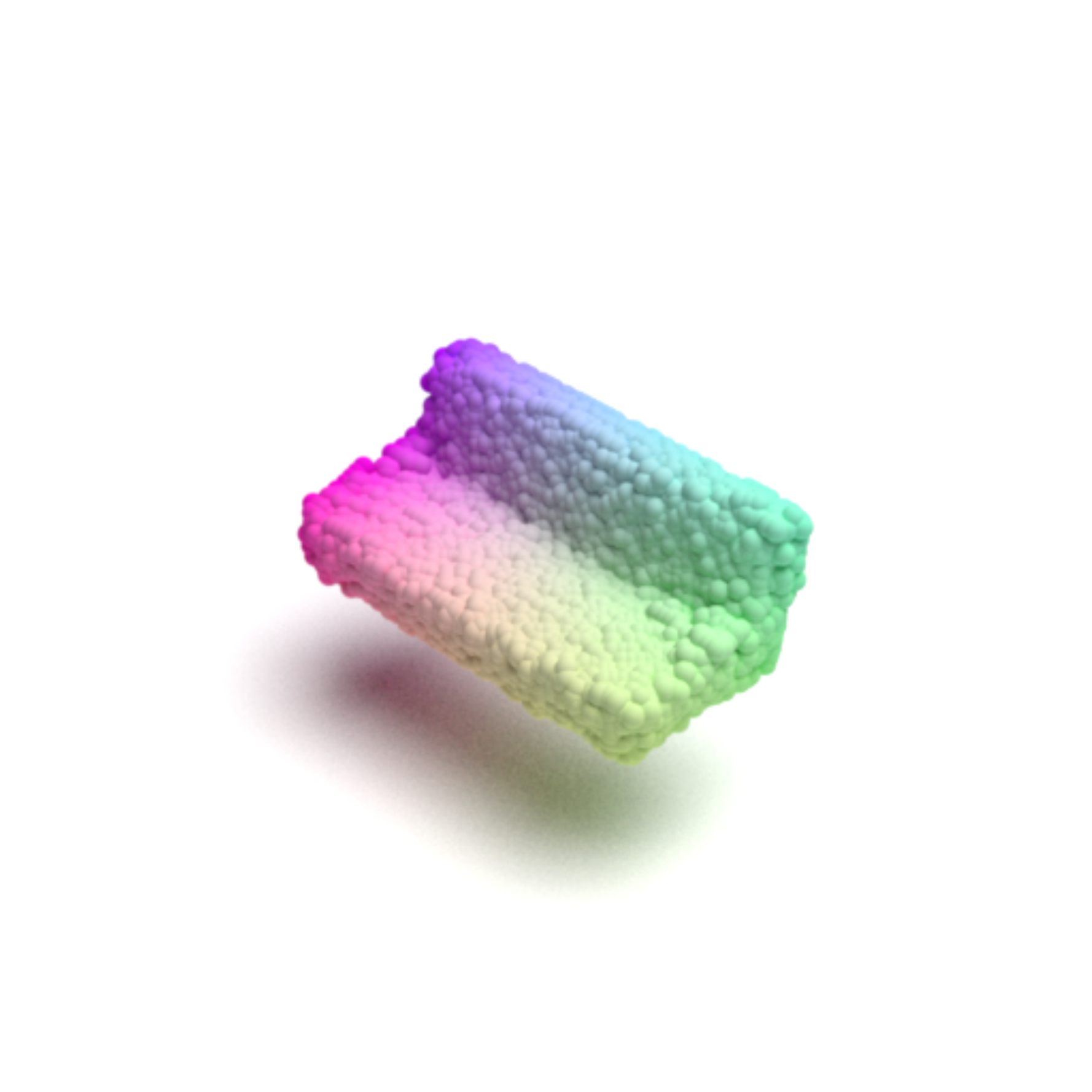}
\includegraphics[clip,trim=4cm 4cm 4cm 4cm, width=0.095\textwidth]{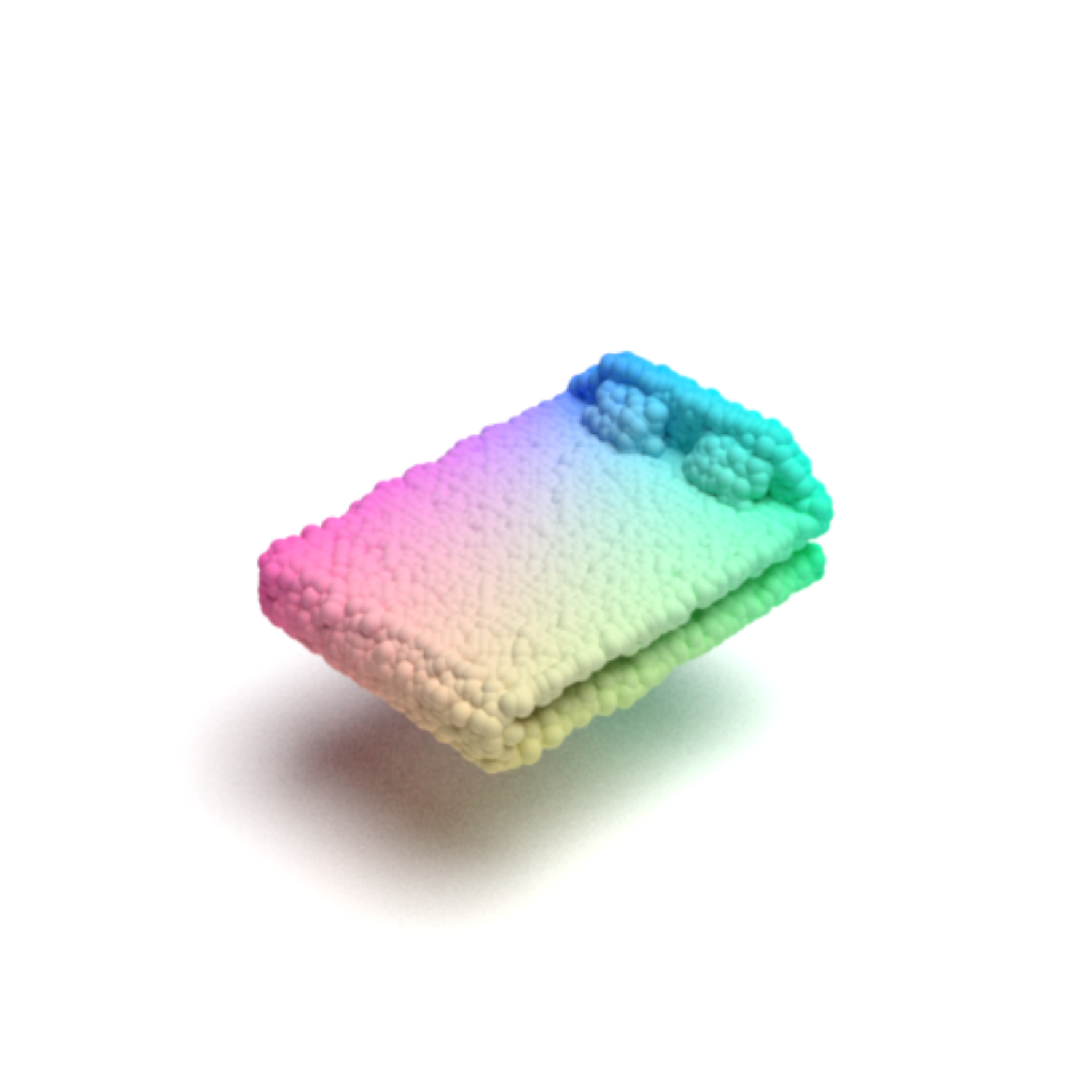}
\includegraphics[clip,trim=4cm 4cm 4cm 4cm, width=0.095\textwidth]{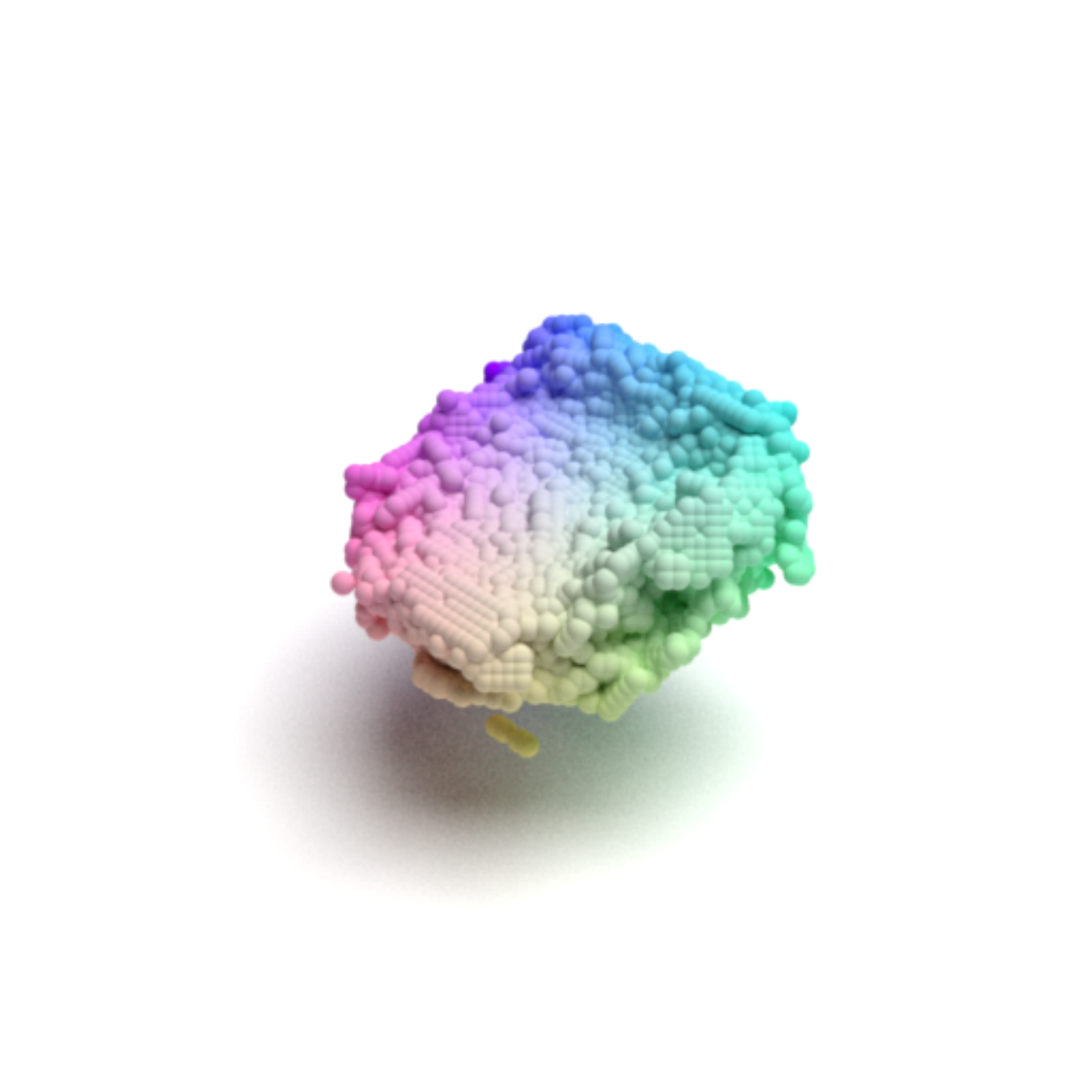}

\captionof{figure}{\textbf{Reconstruction on test objects.} We evaluate our model's ability to encode geometry of 3D shapes with its latent representation. Each row represents 3D shapes from different categories. Odd columns are the input objects and even columns are the reconstructions. We render depth maps of input objects from different viewpoints. Our encoder encodes each of the depth map independently and outputs a latent code. We average the latent code over all the viewpoints of the object. Finally we use our generator to use the same latent code and generate multiple viewpoints from the latent code. Final output depth maps are projected back to 3D. Last two columns shows some of the failure cases likely because of lack of similar samples in training data.}
\label{fig:recon}
\end{figure*}

\subsection{Results}
\paragraph{Auto-encoding.}
We first evaluate the model's ability to represent a given 3D shape with a fixed latent vector size. We follow~\cite{park2019deepsdf} and compute mean and median CD, and mean EMD between ground truth point clouds and reconstructed point clouds obtained from depth maps. Note that our model optimization is done in 2D depth map space and not for either of these metrics but we are able to perform competitively against other voxel and mesh-based methods. Table~\ref{table:reconstruction} shows point cloud and SDF based approaches perform better at these reconstruction metrics. The trade-off is reduced performance in single view reconstruction as discussed next. 
Figure~\ref{fig:recon} shows the reconstruction results on the test set.

\begin{table}[t]
    \centering
    \caption{Reconstruction on test data. We measure the reconstruction performance of different techniques on the test dataset.~\cite{park2019deepsdf} and ~\cite{yang2019pointflow} outperforms the reconstruction in majority of the cases.}
    \resizebox{\linewidth}{!}{
        \begin{tabular}{@{}lcccccc@{}}
            \toprule
            CD, mean & chair & sofa & table & lamp & plane & car\\
            \midrule
            AtlasNet-Sph~\cite{groueix2018papier}  & 0.75 & 0.45 & 0.73 & 2.38 & 0.19 & - \\
            AtlanNet-25~\cite{groueix2018papier}   & 0.37 & 0.41 & \textbf{0.33} & 1.18 & 0.22 & - \\
            DeepSDF~\cite{park2019deepsdf}         & \textbf{0.20} & \textbf{0.13} & 0.55 & \textbf{0.83} & 0.14 & - \\
            Soltani et al.~\cite{3DVAE} & 1.32 & 0.88    & -    & 3.20   & 1.82   & - \\
            PointFlow~\cite{yang2019pointflow} & 0.75    & -    & -    & -    & \textbf{0.07}    & 0.40 \\
            Ours                                   & 0.69 & 0.57 & 1.33 & 2.19 & 0.44 & \textbf{0.36} \\
            \bottomrule
            \toprule
            CD, median & chair & sofa & table & lamp & plane & car\\
            \midrule
            AtlasNet-Sph\cite{groueix2018papier}  & 0.51 & 0.33 & 0.39 & 2.18 & 0.08 & - \\
            AtlanNet-25\cite{groueix2018papier}   & 0.28 & 0.31 & 0.20 & 0.99 & 0.07 & - \\
            DeepSDF~\cite{park2019deepsdf}           & \textbf{0.07} & \textbf{0.09} & \textbf{0.07} & \textbf{0.22} & \textbf{0.04} & - \\
            Soltani et al.~\cite{3DVAE} & 1.28    & 0.76    & -    & 1.99   & 1.71   & - \\
            PointFlow~\cite{yang2019pointflow} & 0.55    & -    & -    & -    & 0.05    & 0.36 \\
            Ours.         & 0.34 & 0.28 & 0.38 & 0.89 & 0.23 & \textbf{0.19} \\
            \bottomrule
            \toprule
            EMD, mean & chair & sofa & table & lamp & plane & car\\
            \midrule
            AtlasNet-Sph\cite{groueix2018papier}  & 0.071 & 0.050 & 0.060 & 0.085 & 0.038 & -     \\
            AtlanNet-25\cite{groueix2018papier}   & 0.064 & 0.063 & 0.073 & 0.062 & 0.041 & -     \\
            DeepSDF~\cite{park2019deepsdf}           & \textbf{0.049} & \textbf{0.047} & 0.050 & \textbf{0.059} & \textbf{0.033} & -     \\
            Soltani et al.~\cite{3DVAE} & 0.139 & 0.072     & 0.075     & 0.096     & 0.063    & -     \\
            PointFlow~\cite{yang2019pointflow} & 0.078    & -    & -    & -    & 0.039    & 0.066 \\
            Ours          & 0.077 & 0.078 & \textbf{0.049} & 0.099 & 0.056 & \textbf{0.046} \\
            \bottomrule
        \end{tabular}
    }
\vspace{-0.2in}
\label{table:reconstruction}
\end{table}

\begin{figure*}
\centering

\includegraphics[width=0.09\textwidth]{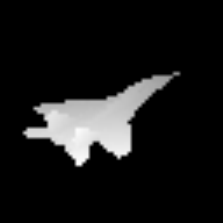}
\includegraphics[width=0.09\textwidth, clip,trim=3cm 3cm 3cm 3cm]{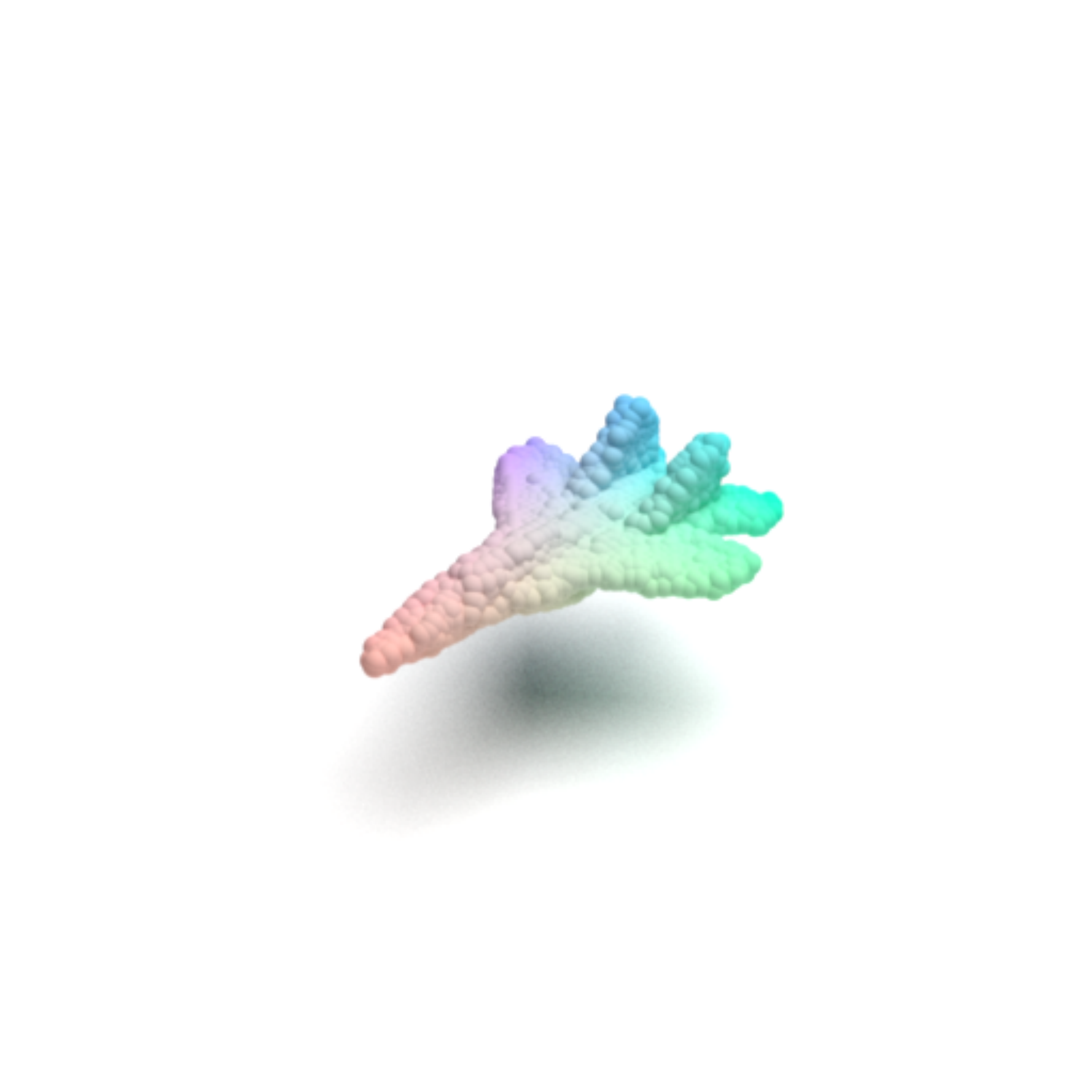}
\includegraphics[width=0.09\textwidth]{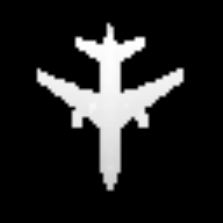}
\includegraphics[width=0.09\textwidth, clip,trim=3cm 3cm 3cm 3cm]{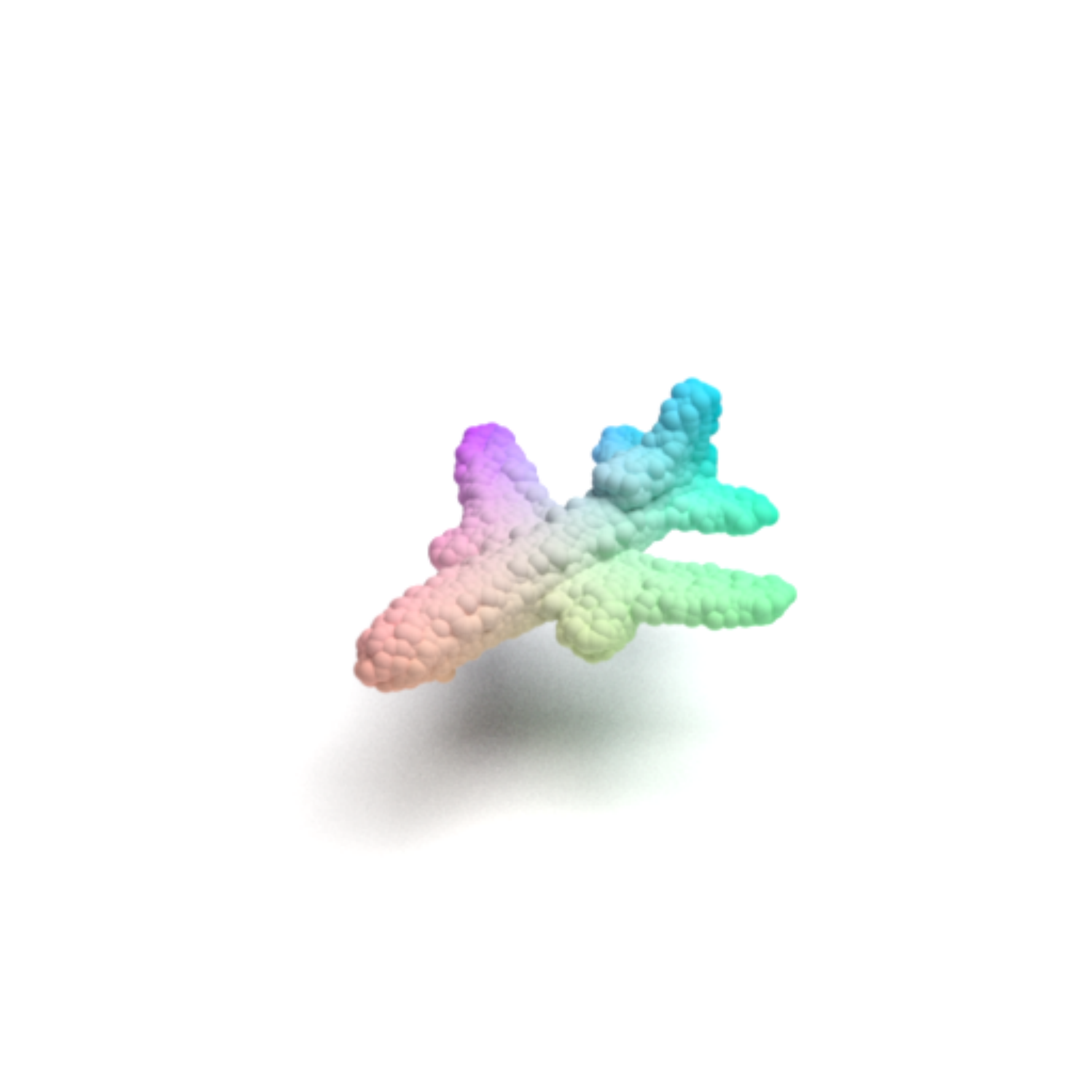}
\includegraphics[width=0.09\textwidth]{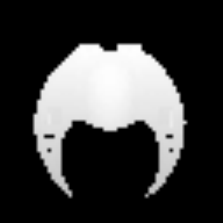}
\includegraphics[width=0.09\textwidth,clip,trim=3cm 3cm 3cm 3cm]{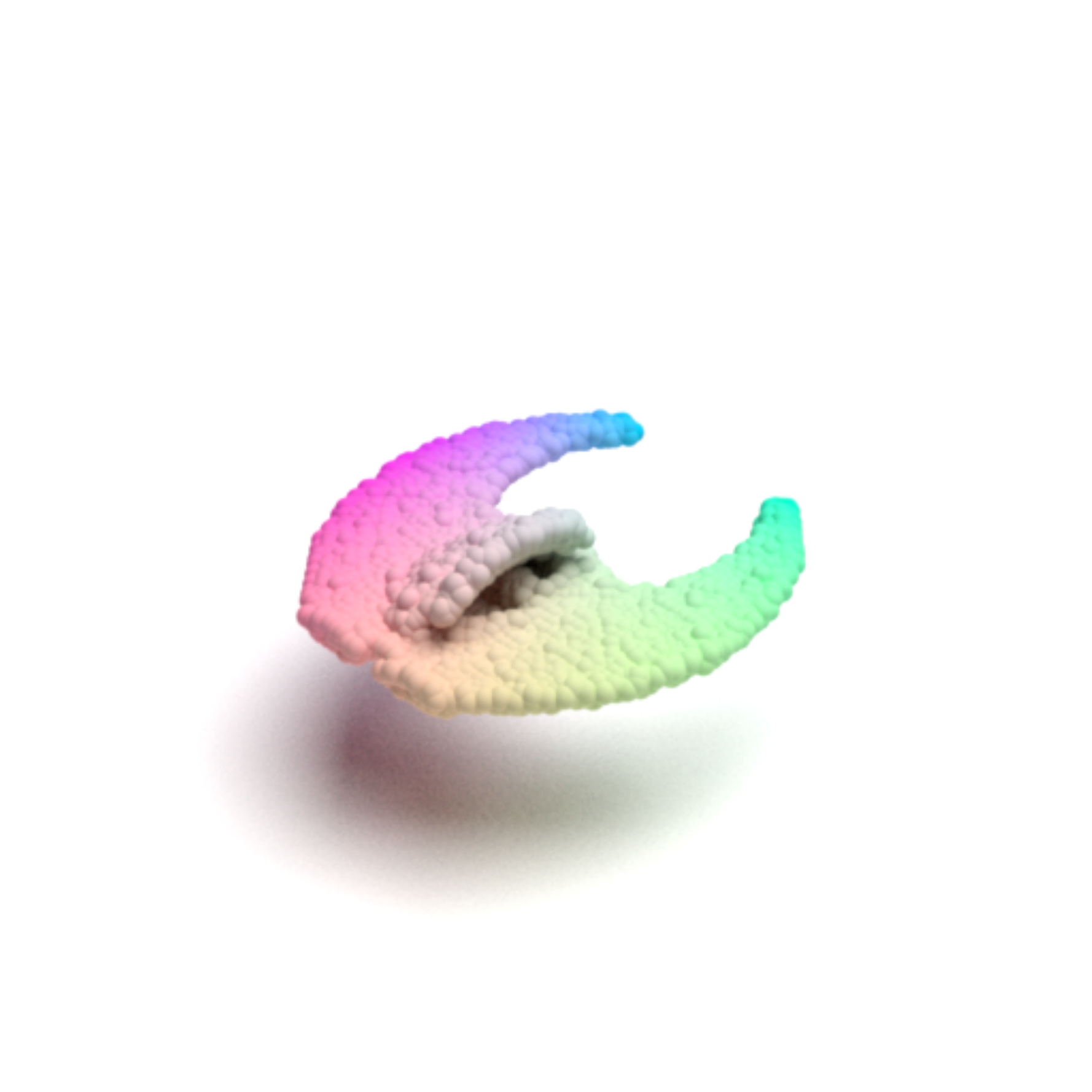}
\includegraphics[width=0.09\textwidth]{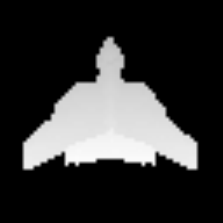}
\includegraphics[width=0.09\textwidth,clip,trim=3cm 3cm 3cm 3cm]{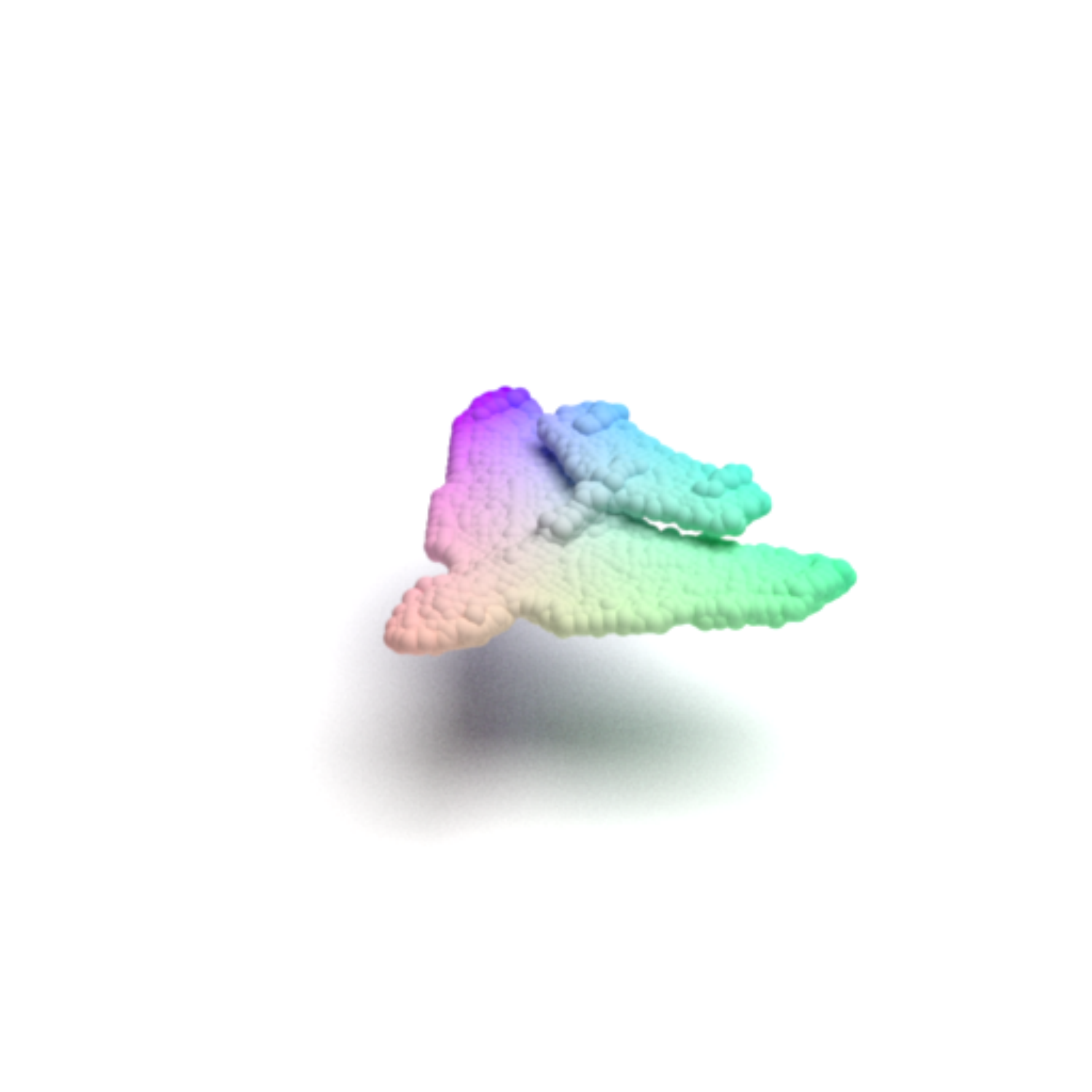}
\includegraphics[width=0.09\textwidth]{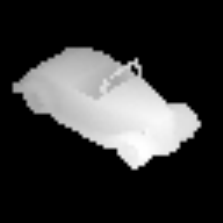}
\includegraphics[width=0.09\textwidth,clip,trim=3cm 3cm 3cm 3cm]{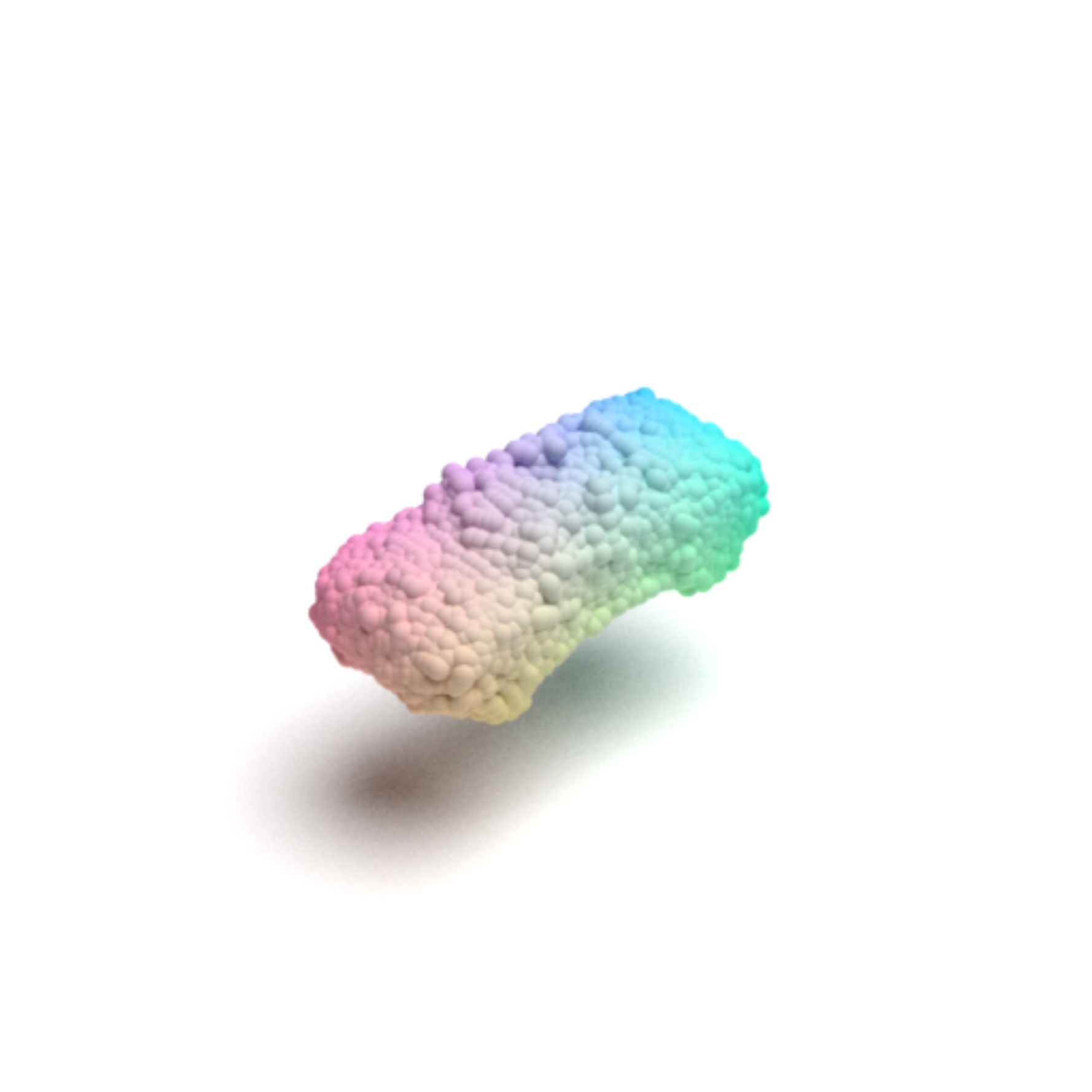}

\includegraphics[width=0.09\textwidth]{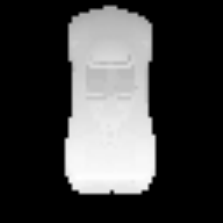}
\includegraphics[width=0.09\textwidth,clip,trim=3cm 3cm 3cm 3cm]{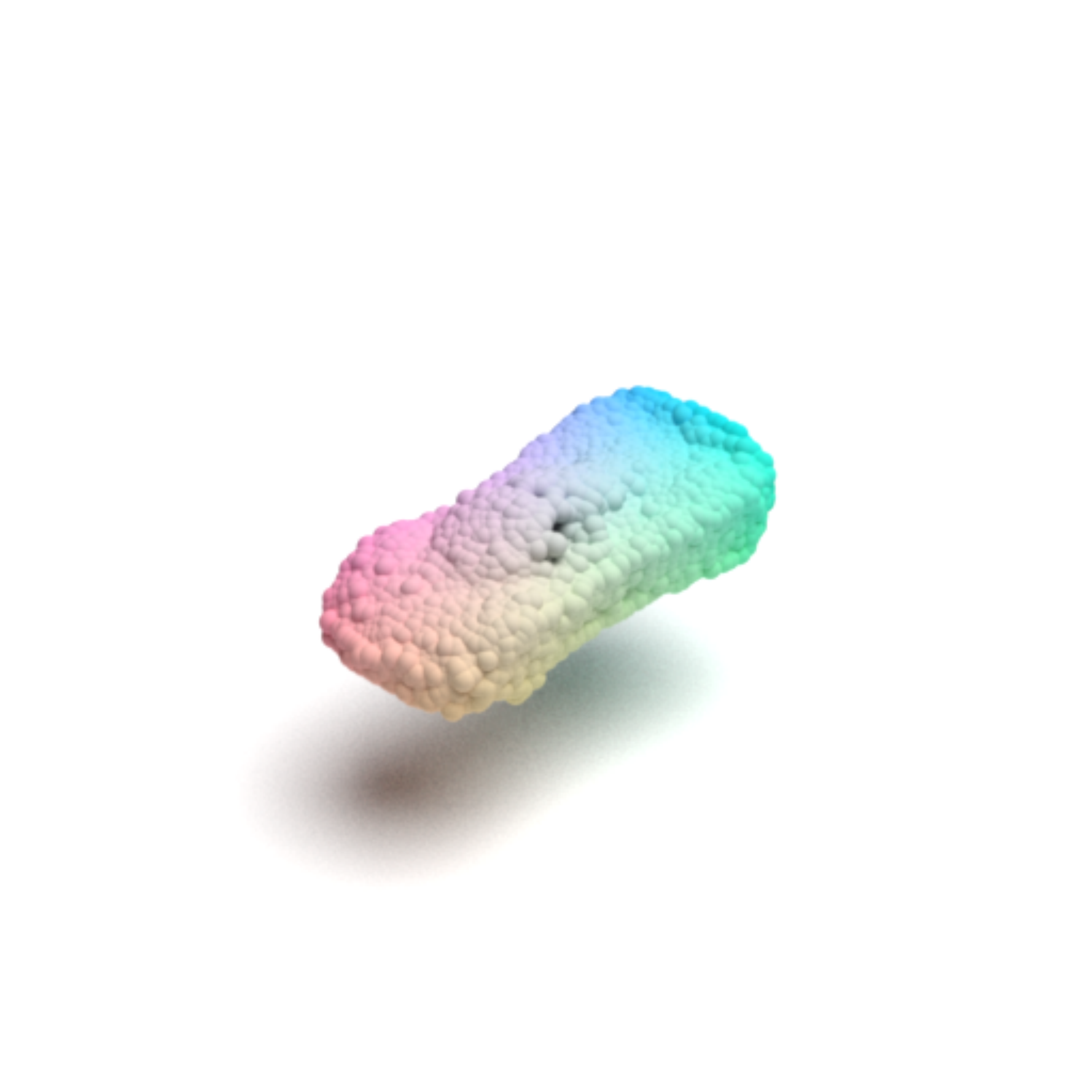}
\includegraphics[width=0.09\textwidth]{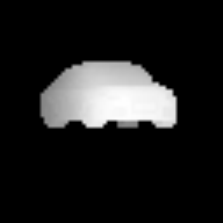}
\includegraphics[width=0.09\textwidth,clip,trim=3cm 3cm 3cm 3cm]{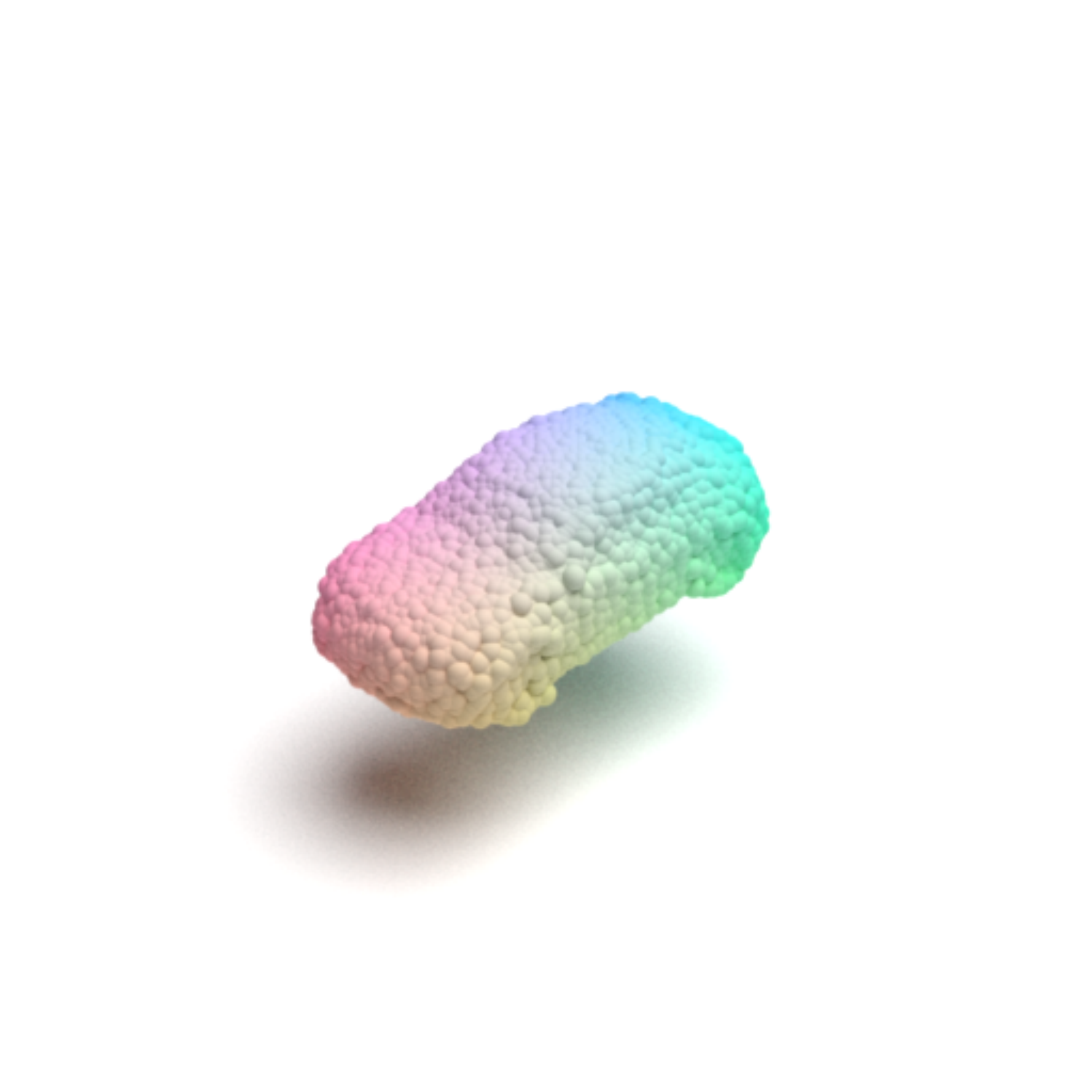}
\includegraphics[width=0.09\textwidth]{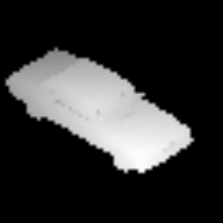}
\includegraphics[width=0.09\textwidth,clip,trim=3cm 3cm 3cm 3cm]{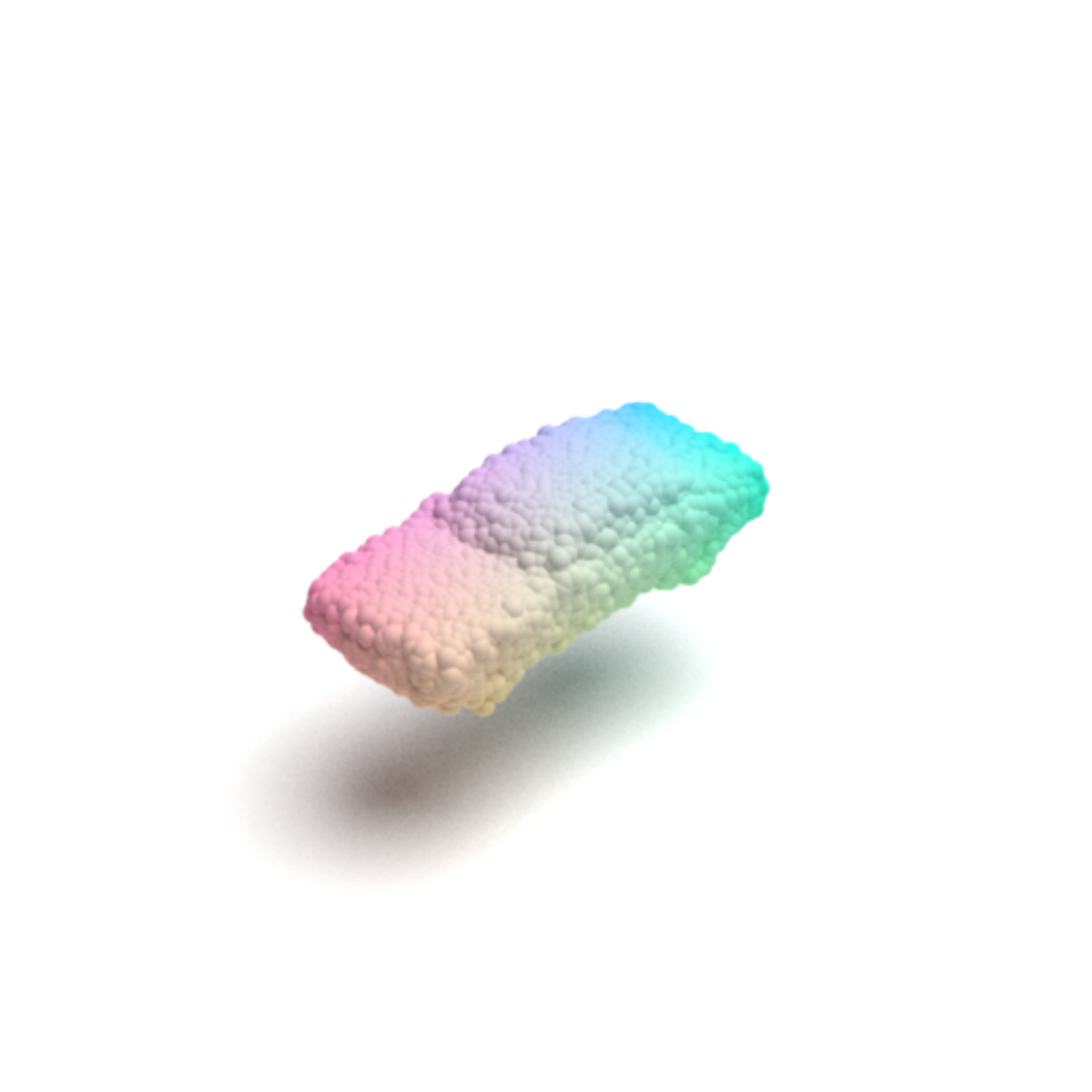}
\includegraphics[width=0.09\textwidth]{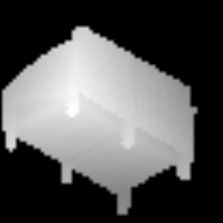}
\includegraphics[width=0.09\textwidth,clip,trim=3cm 3cm 3cm 3cm]{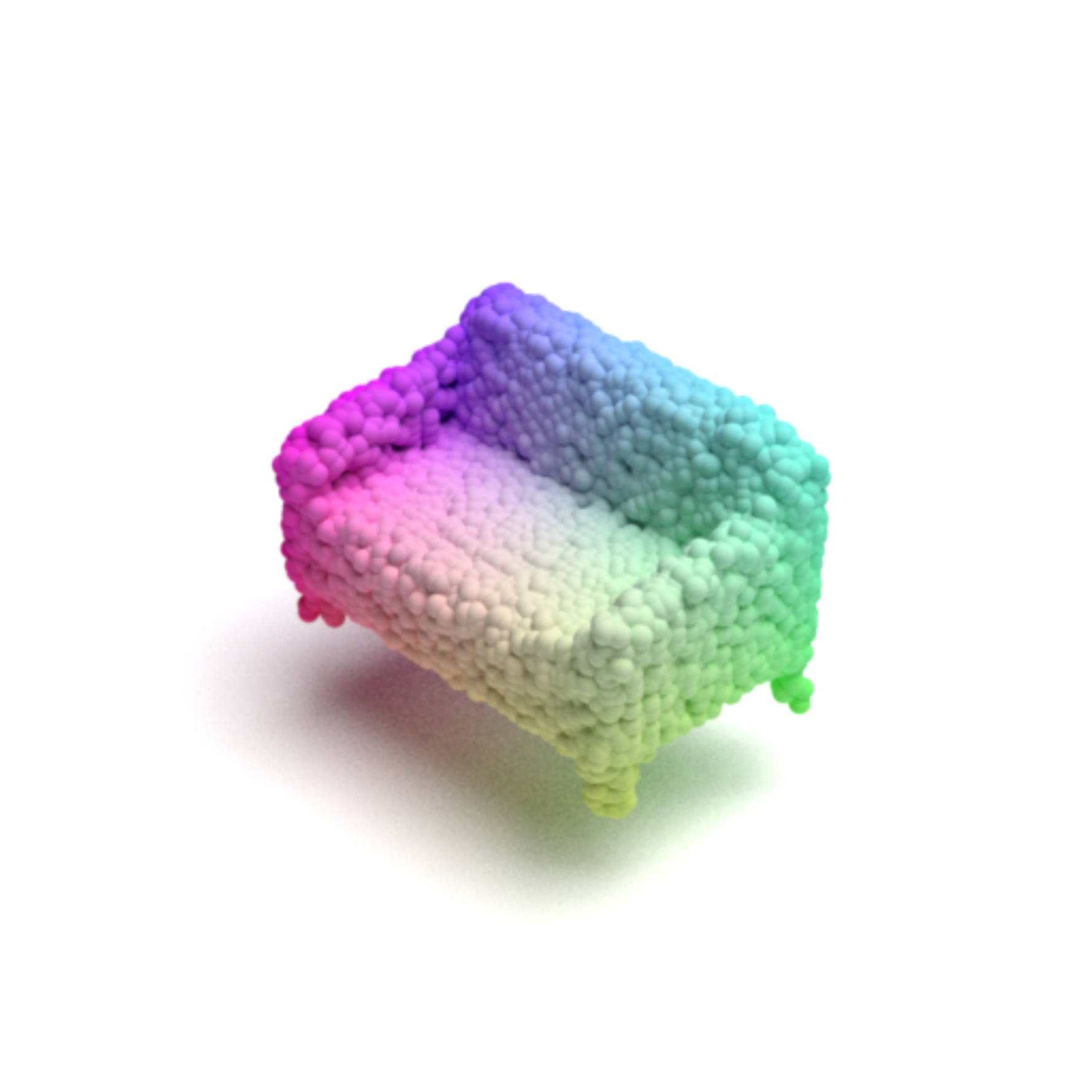}
\includegraphics[width=0.09\textwidth]{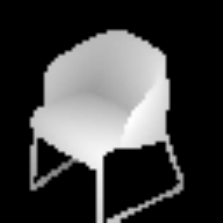}
\includegraphics[width=0.09\textwidth,clip,trim=3cm 3cm 3cm 3cm]{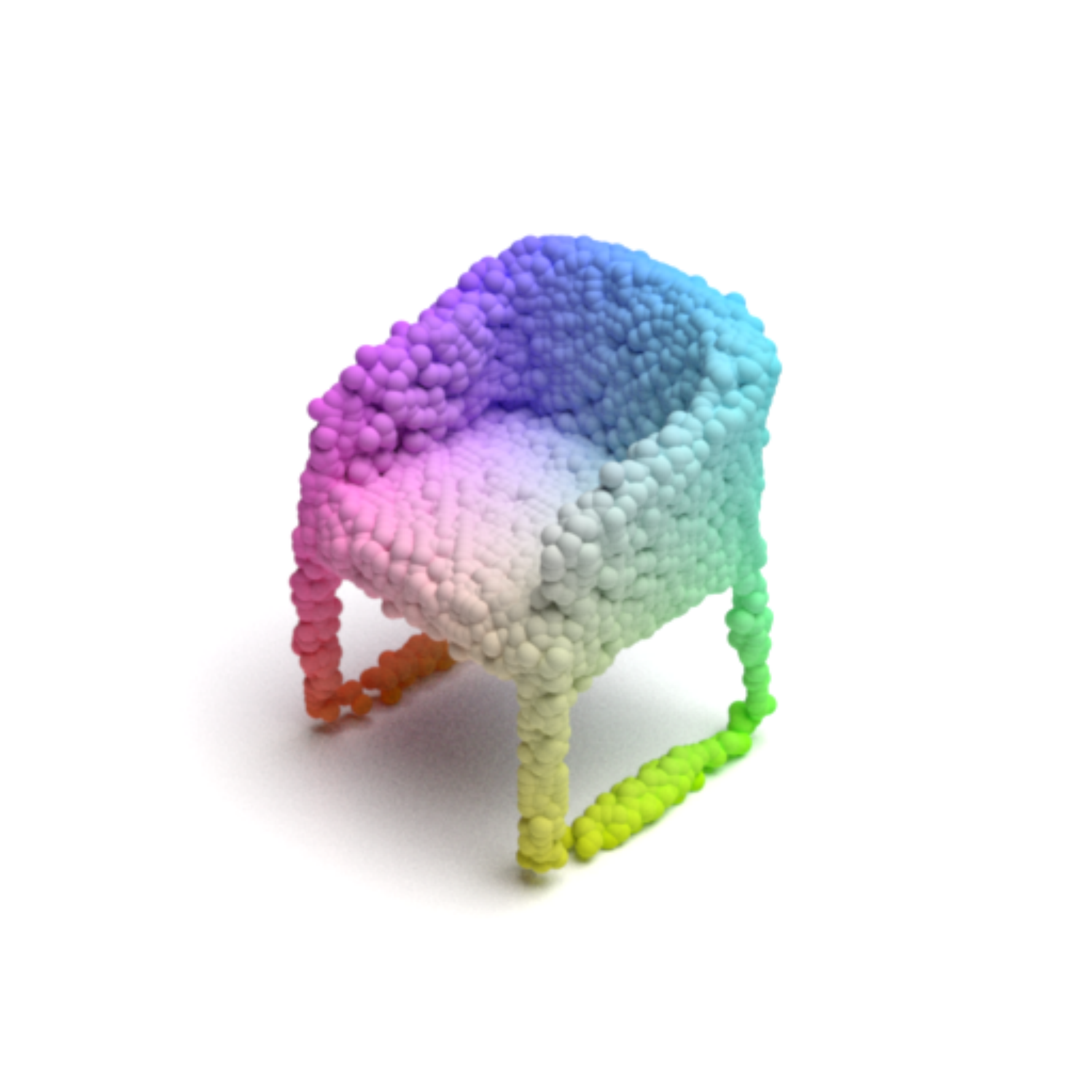}

\includegraphics[width=0.09\textwidth]{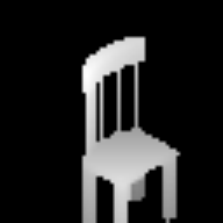}
\includegraphics[width=0.09\textwidth,clip,trim=3cm 3cm 3cm 3cm]{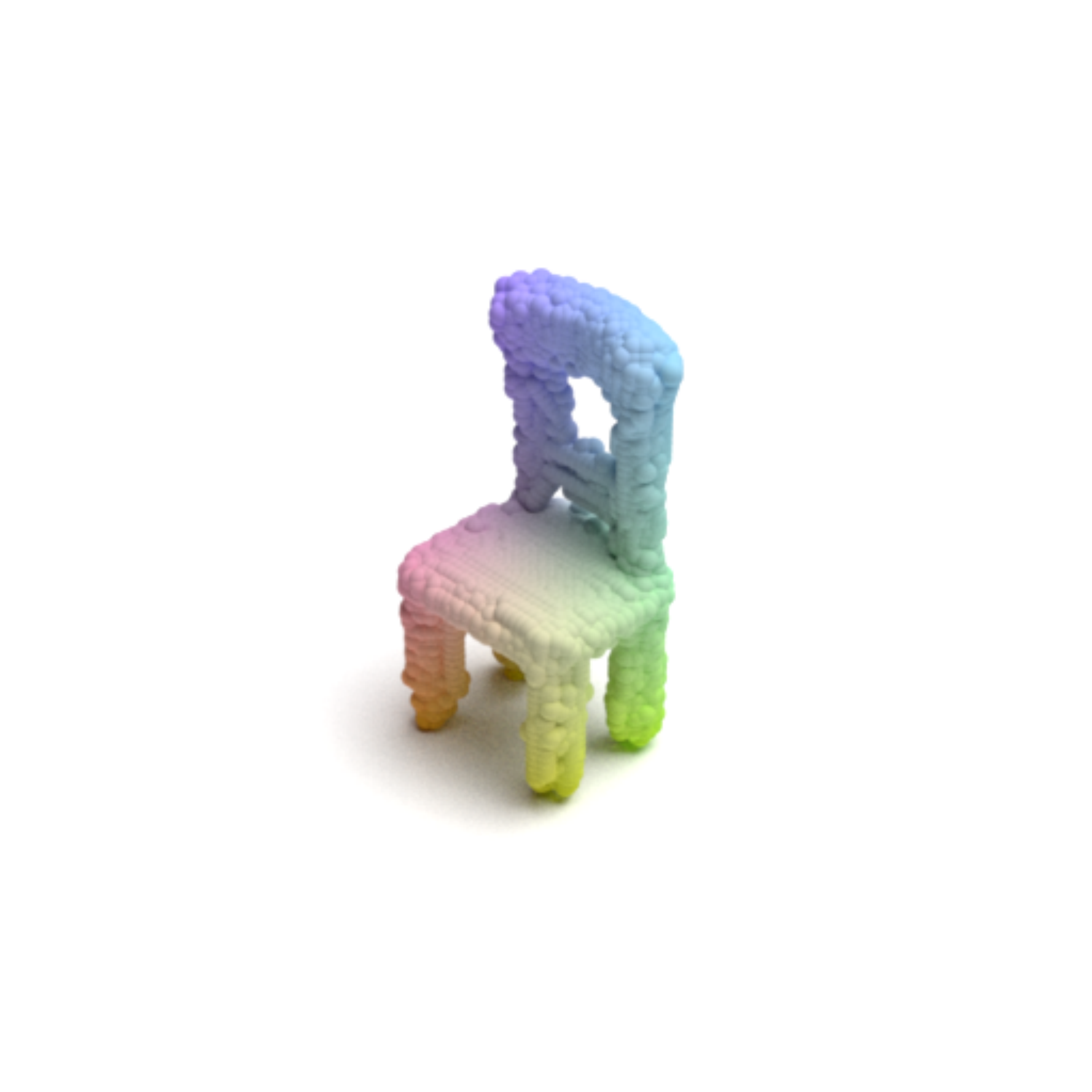}
\includegraphics[width=0.09\textwidth]{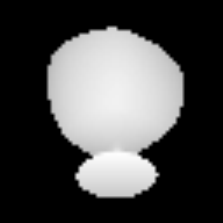}
\includegraphics[width=0.09\textwidth,clip,trim=3cm 3cm 3cm 3cm]{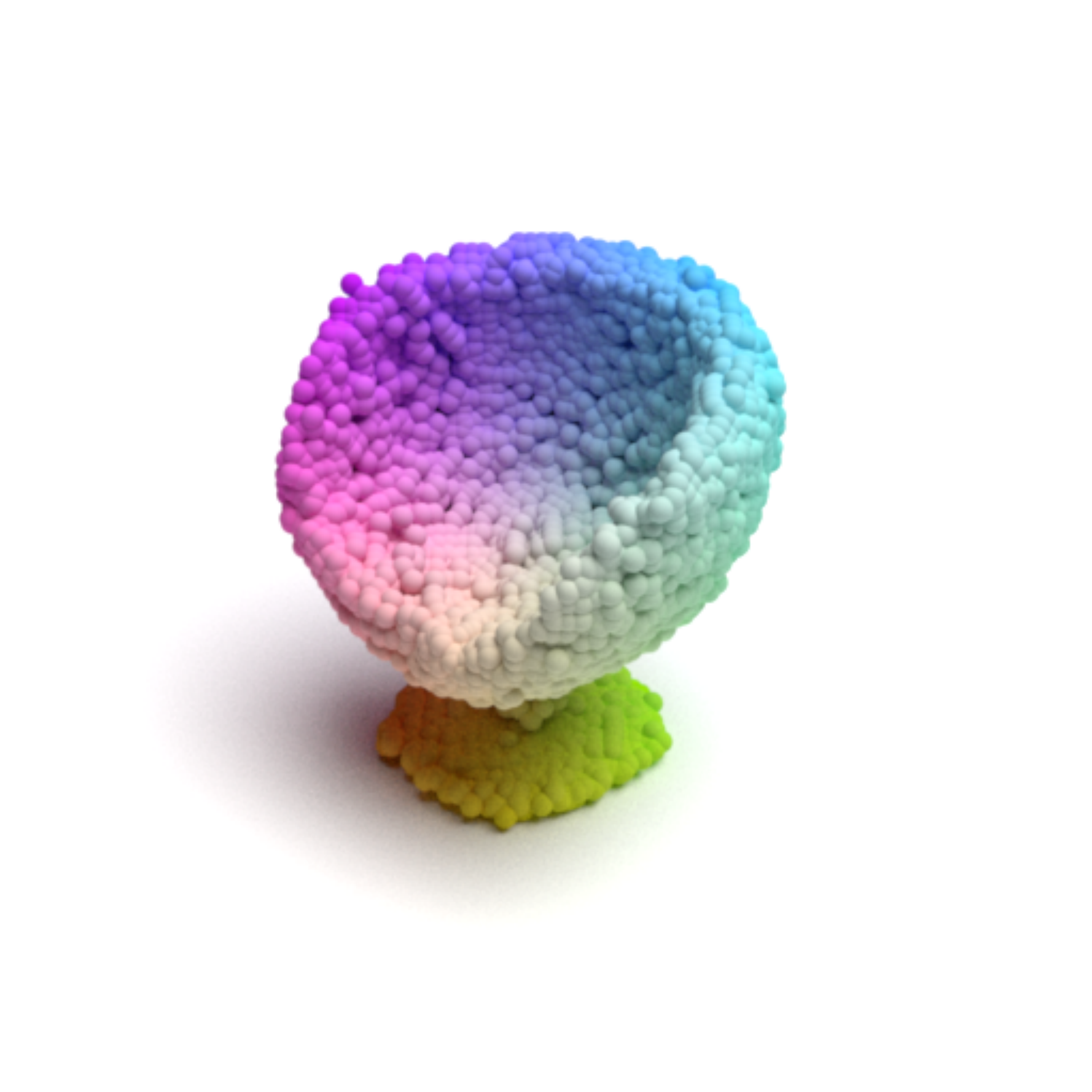}
\includegraphics[width=0.09\textwidth]{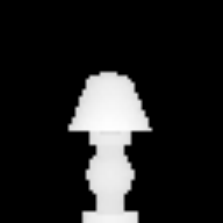}
\includegraphics[width=0.09\textwidth,clip,trim=3cm 3cm 3cm 3cm]{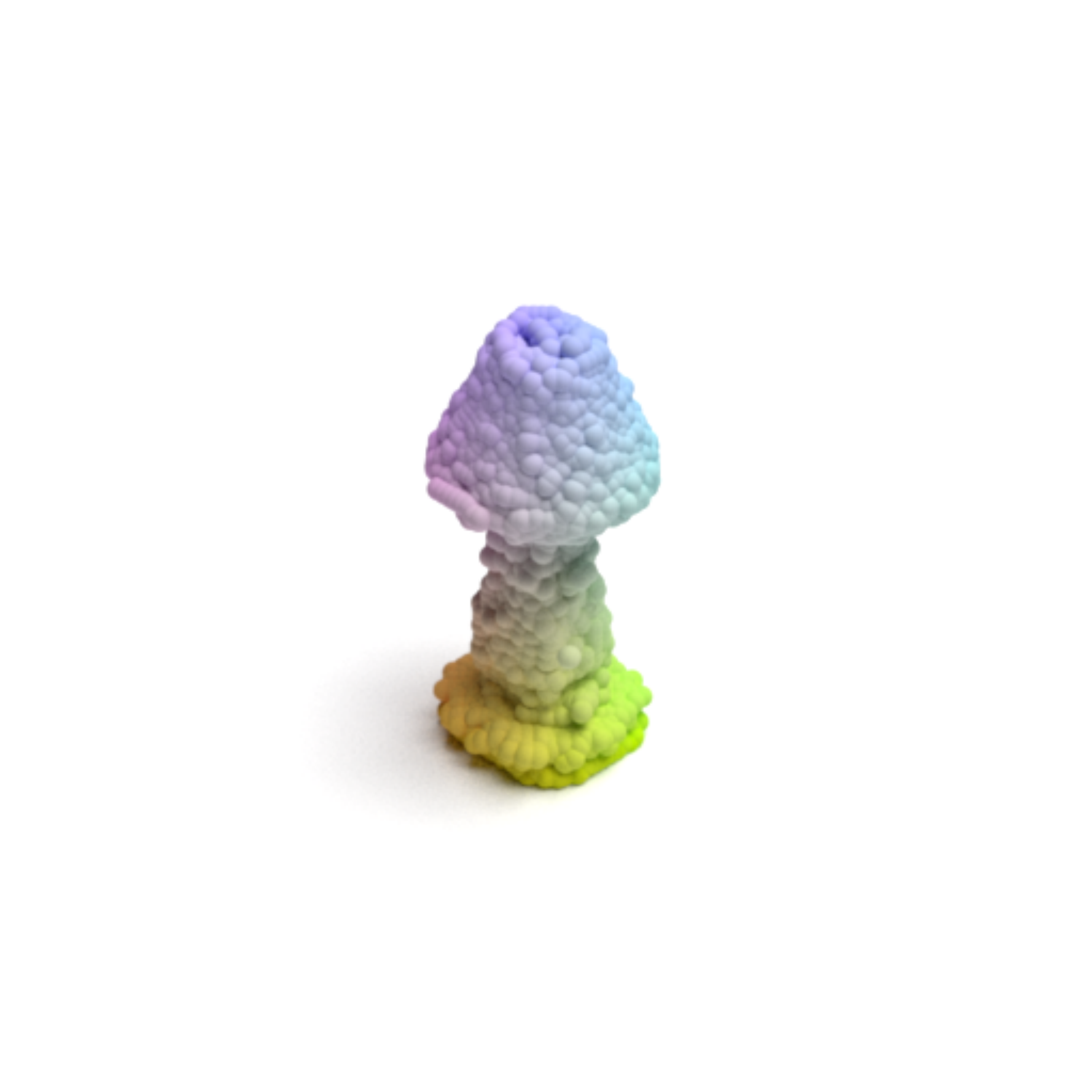}
\includegraphics[width=0.09\textwidth]{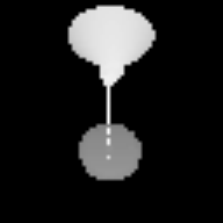}
\includegraphics[width=0.09\textwidth,clip,trim=3cm 3cm 3cm 3cm]{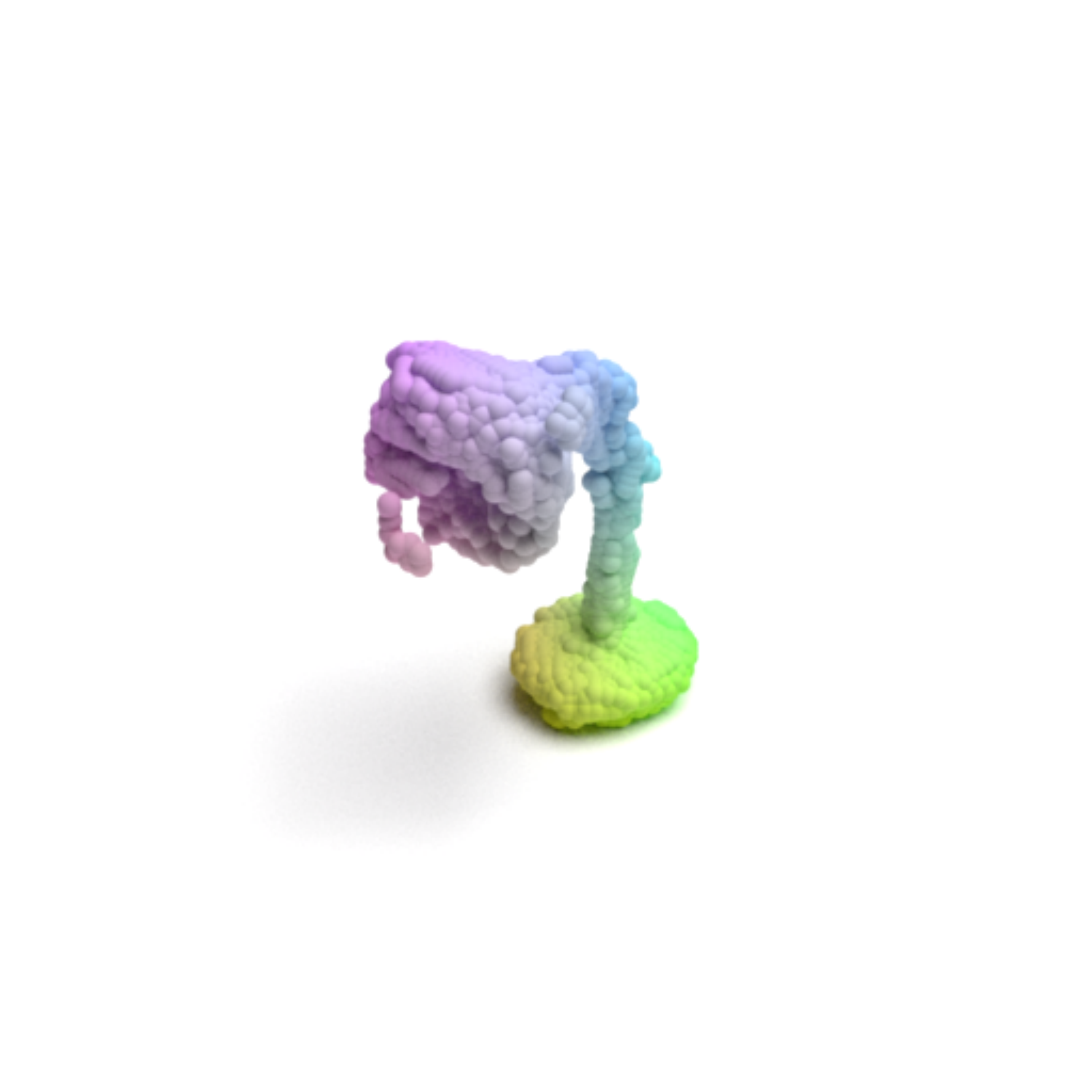}
\includegraphics[width=0.09\textwidth]{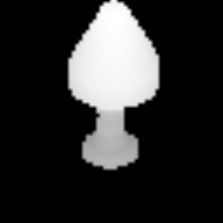}
\includegraphics[width=0.09\textwidth,clip,trim=3cm 3cm 3cm 3cm]{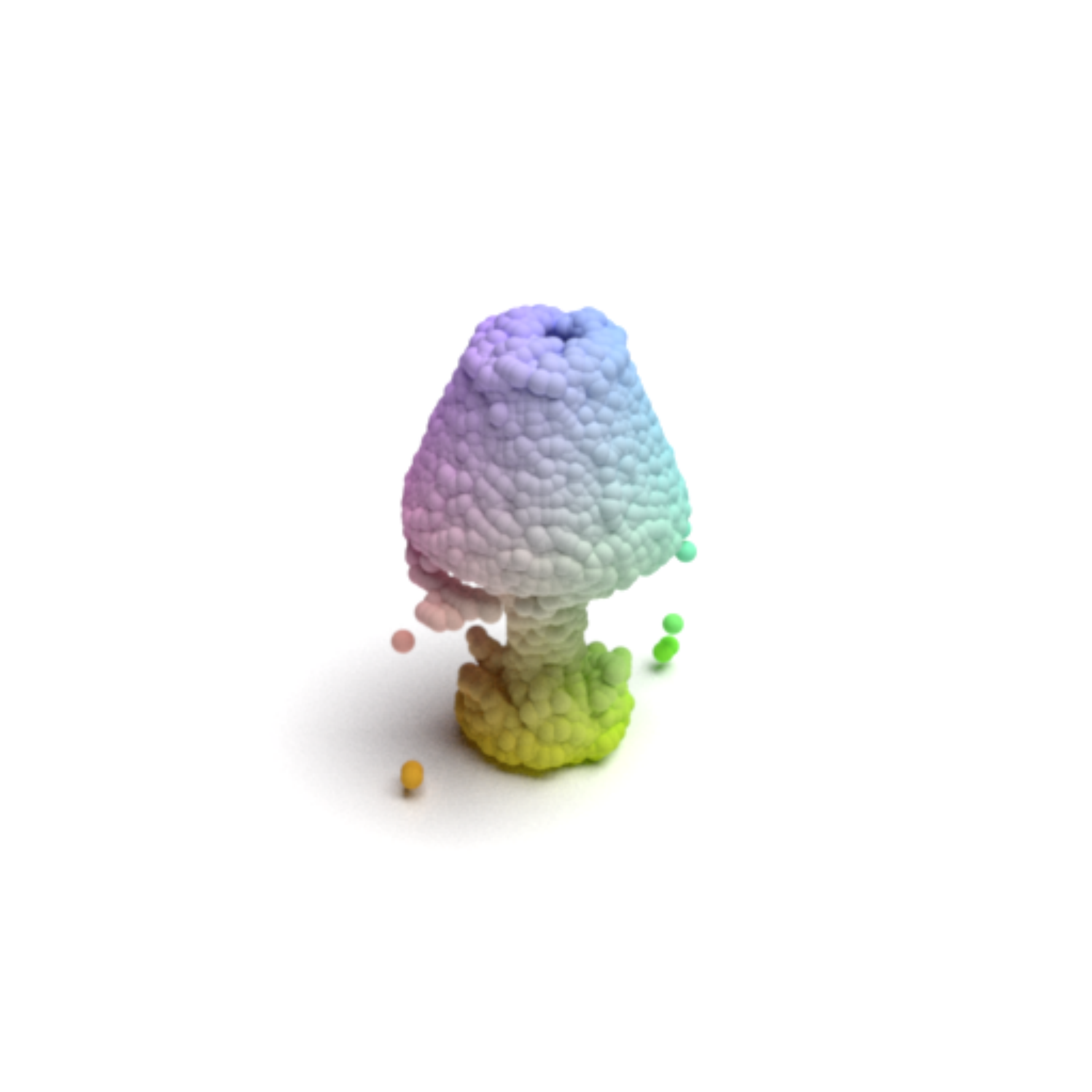}

\includegraphics[width=0.09\textwidth]{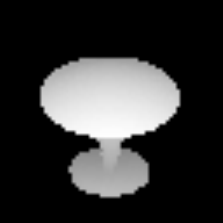}
\includegraphics[width=0.09\textwidth,clip,trim=3cm 3cm 3cm 3cm]{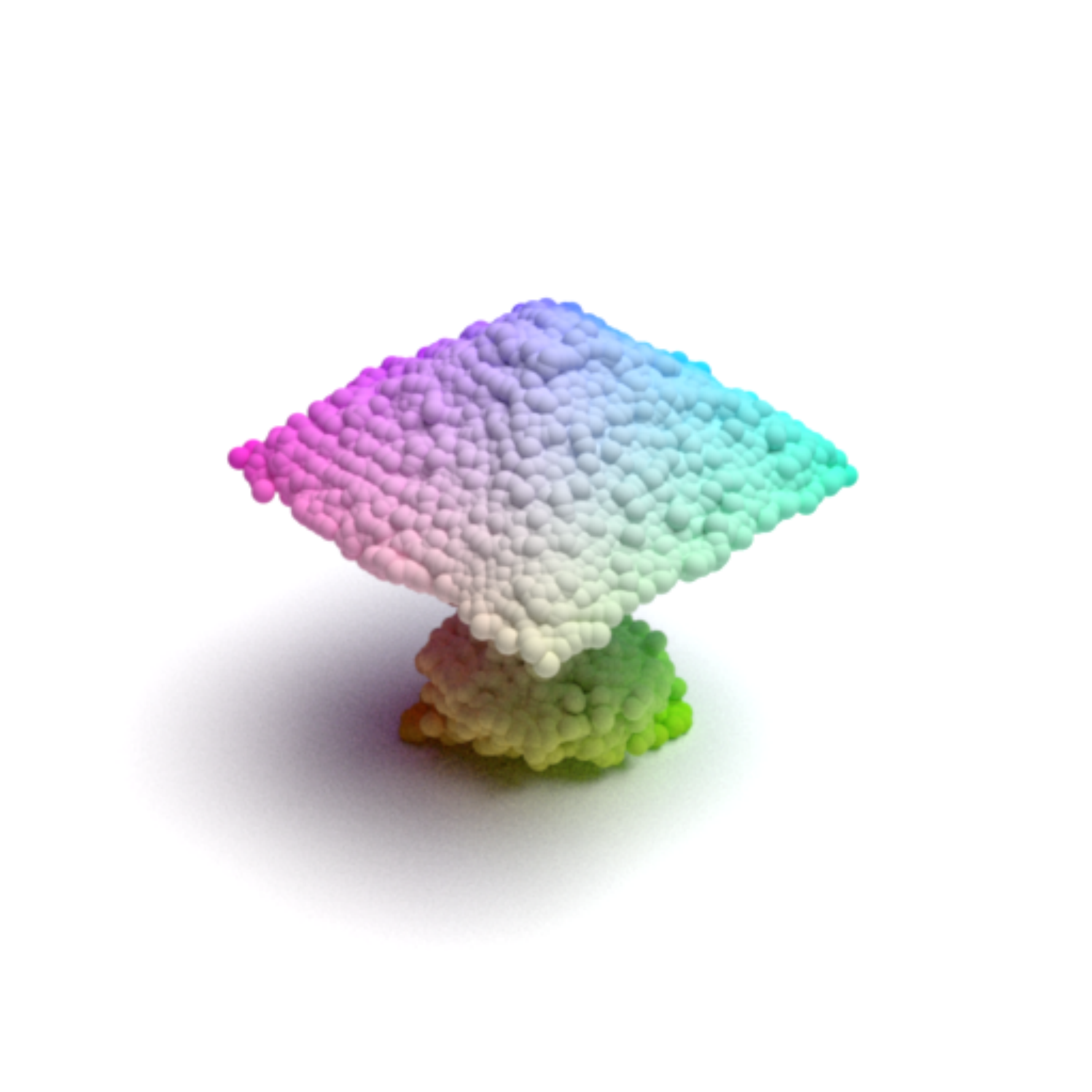}
\includegraphics[width=0.09\textwidth]{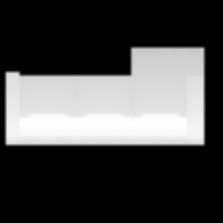}
\includegraphics[width=0.09\textwidth,clip,trim=3cm 3cm 3cm 3cm]{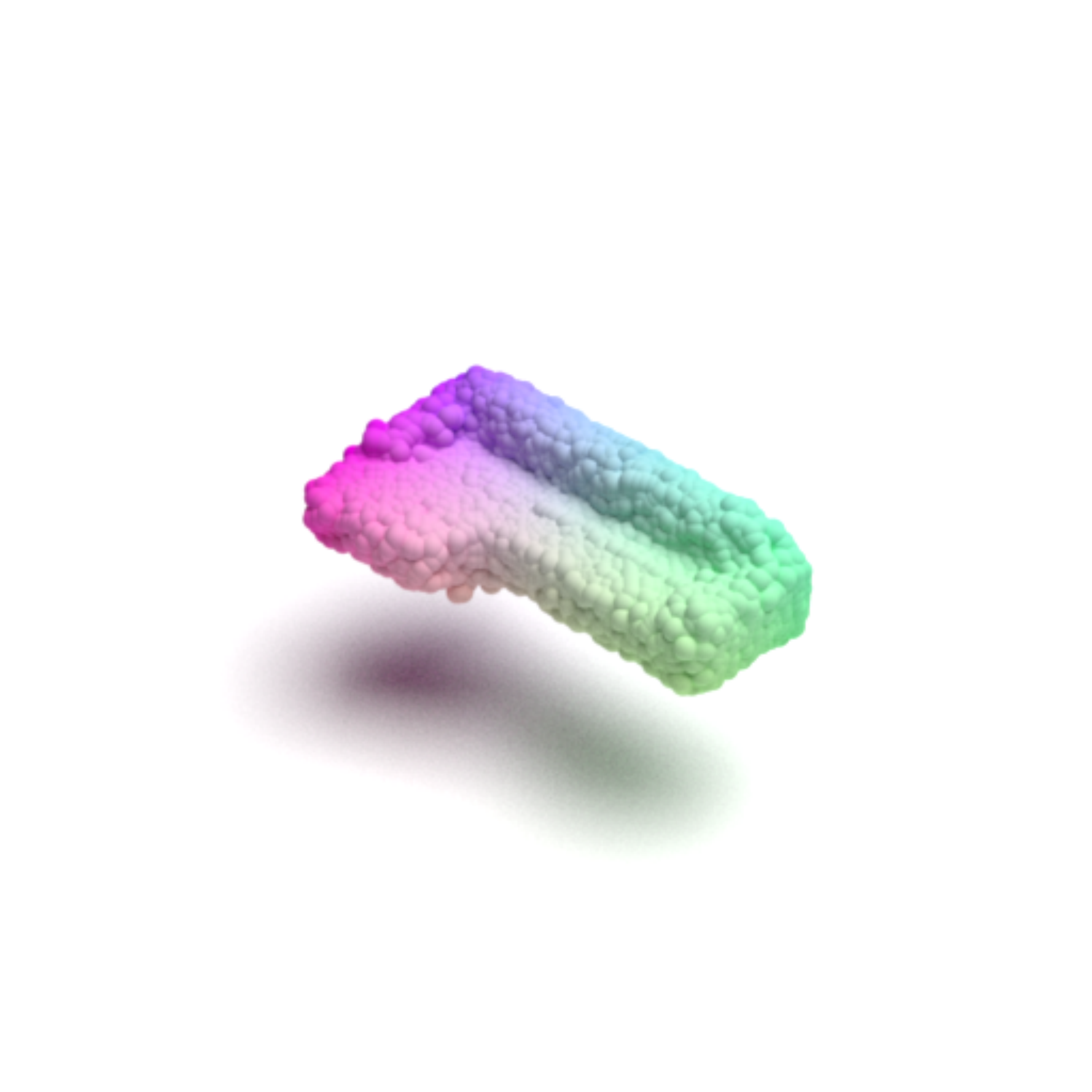}
\includegraphics[width=0.09\textwidth]{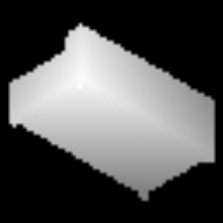}
\includegraphics[width=0.09\textwidth,clip,trim=3cm 3cm 3cm 3cm]{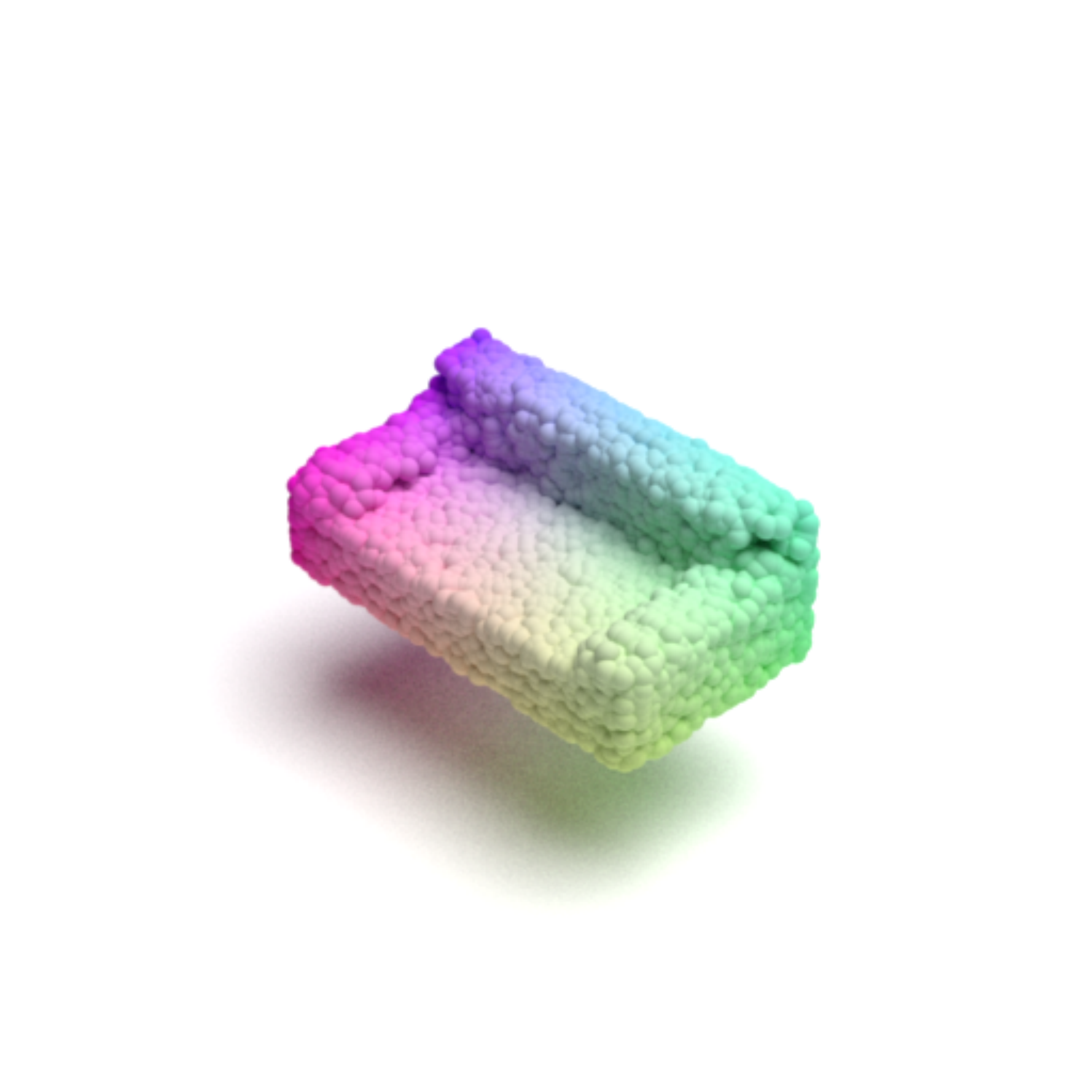}
\includegraphics[width=0.09\textwidth]{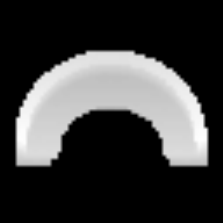}
\includegraphics[width=0.09\textwidth,clip,trim=3cm 3cm 3cm 3cm]{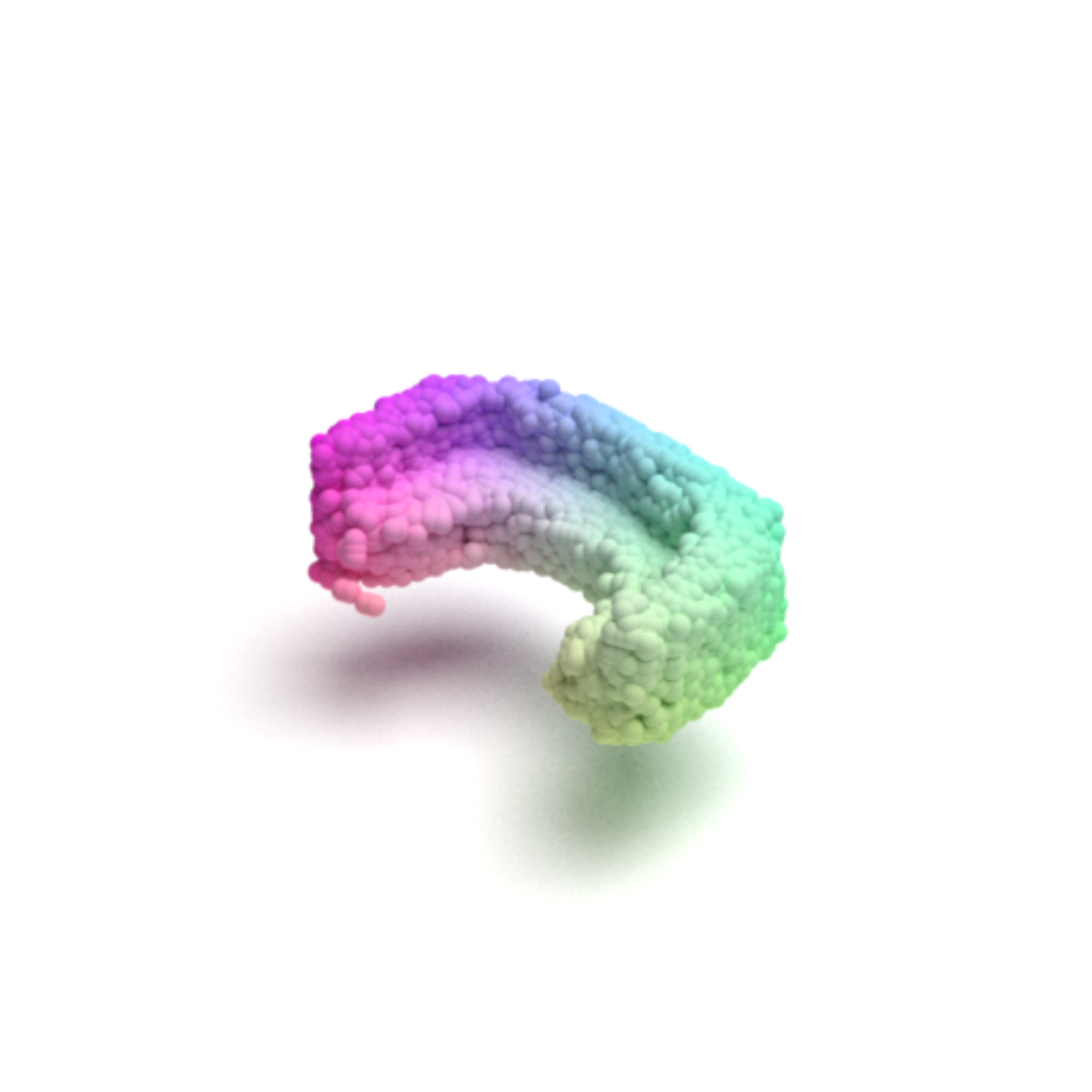}
\includegraphics[width=0.09\textwidth]{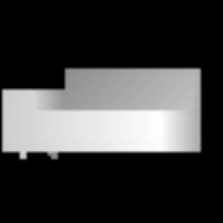}
\includegraphics[width=0.09\textwidth,clip,trim=3cm 3cm 3cm 3cm]{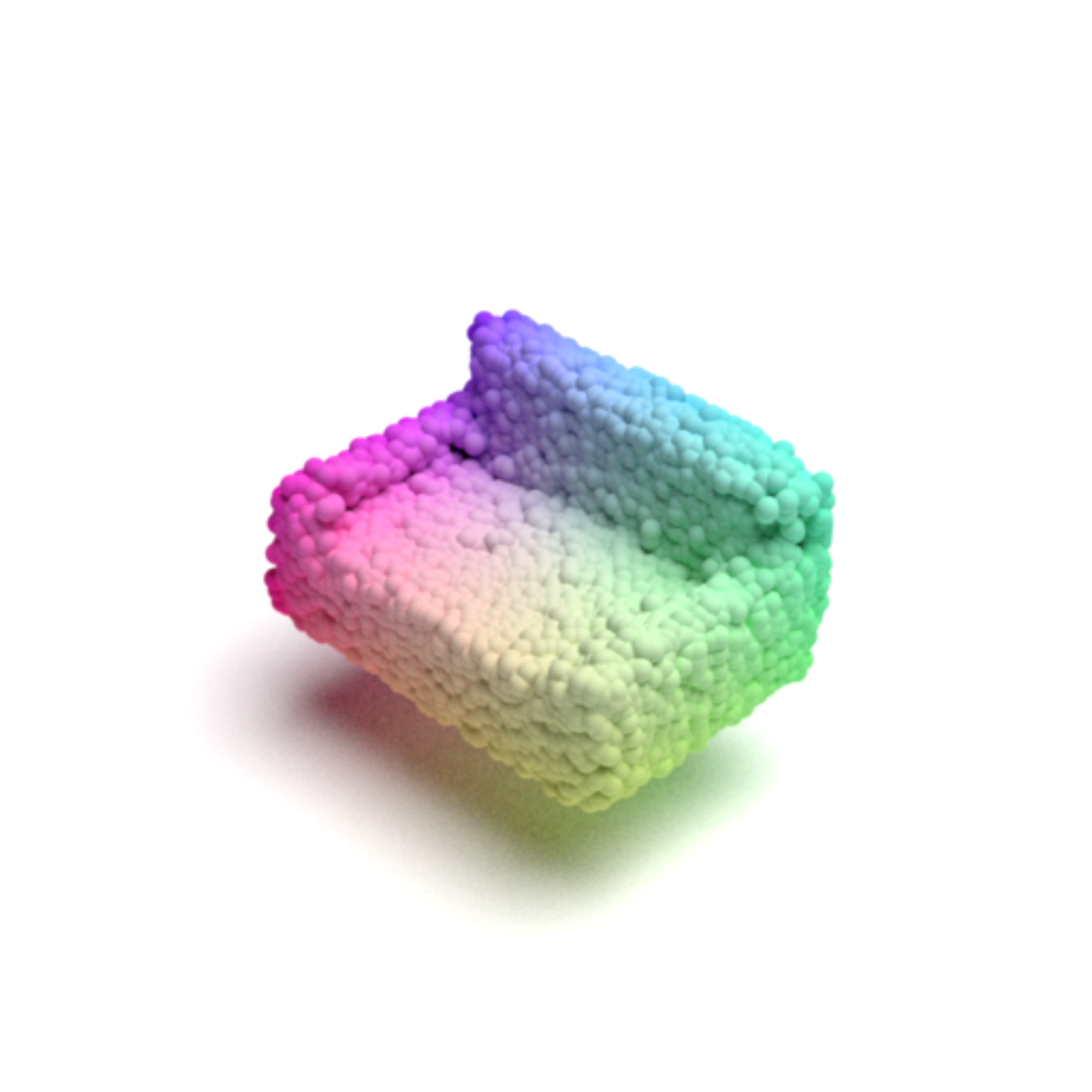}

\includegraphics[width=0.09\textwidth]{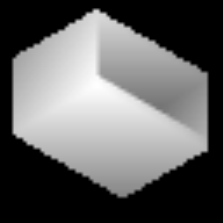}
\includegraphics[width=0.09\textwidth,clip,trim=3cm 3cm 3cm 3cm]{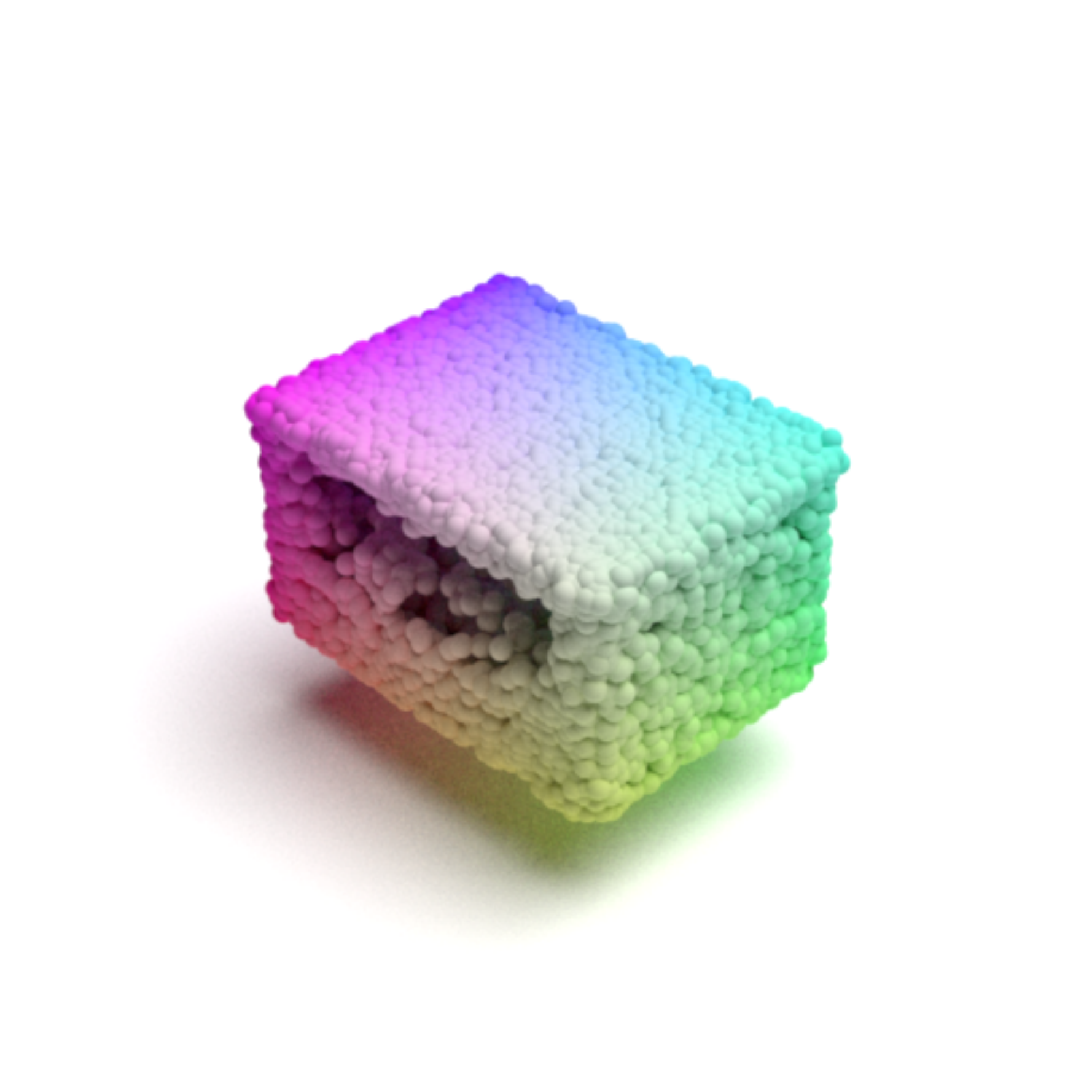}
\includegraphics[width=0.09\textwidth]{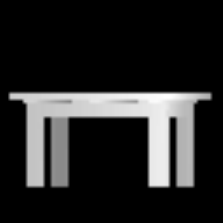}
\includegraphics[width=0.09\textwidth,clip,trim=3cm 3cm 3cm 3cm]{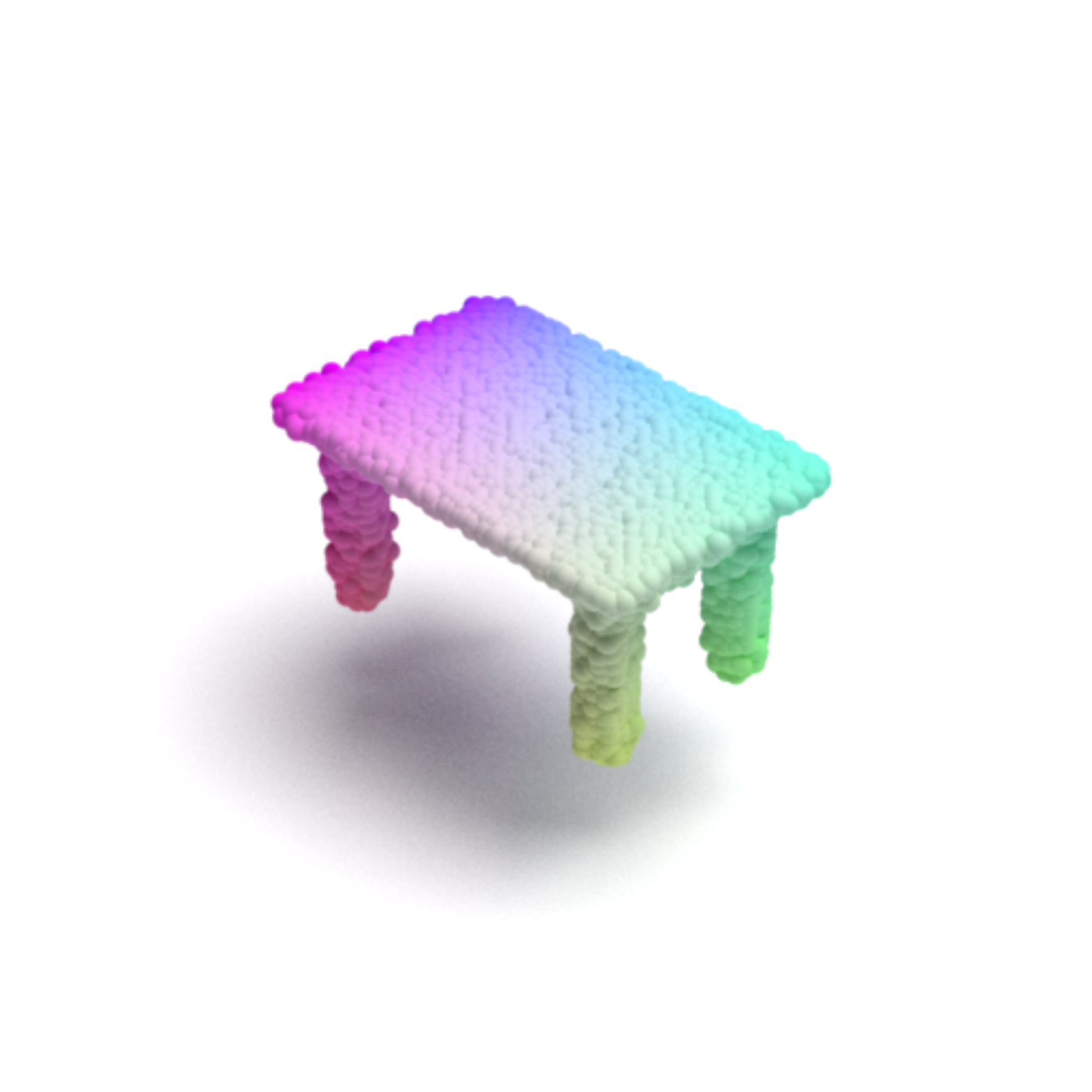}
\includegraphics[width=0.09\textwidth]{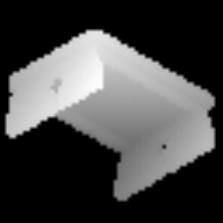}
\includegraphics[width=0.09\textwidth,clip,trim=3cm 3cm 3cm 3cm]{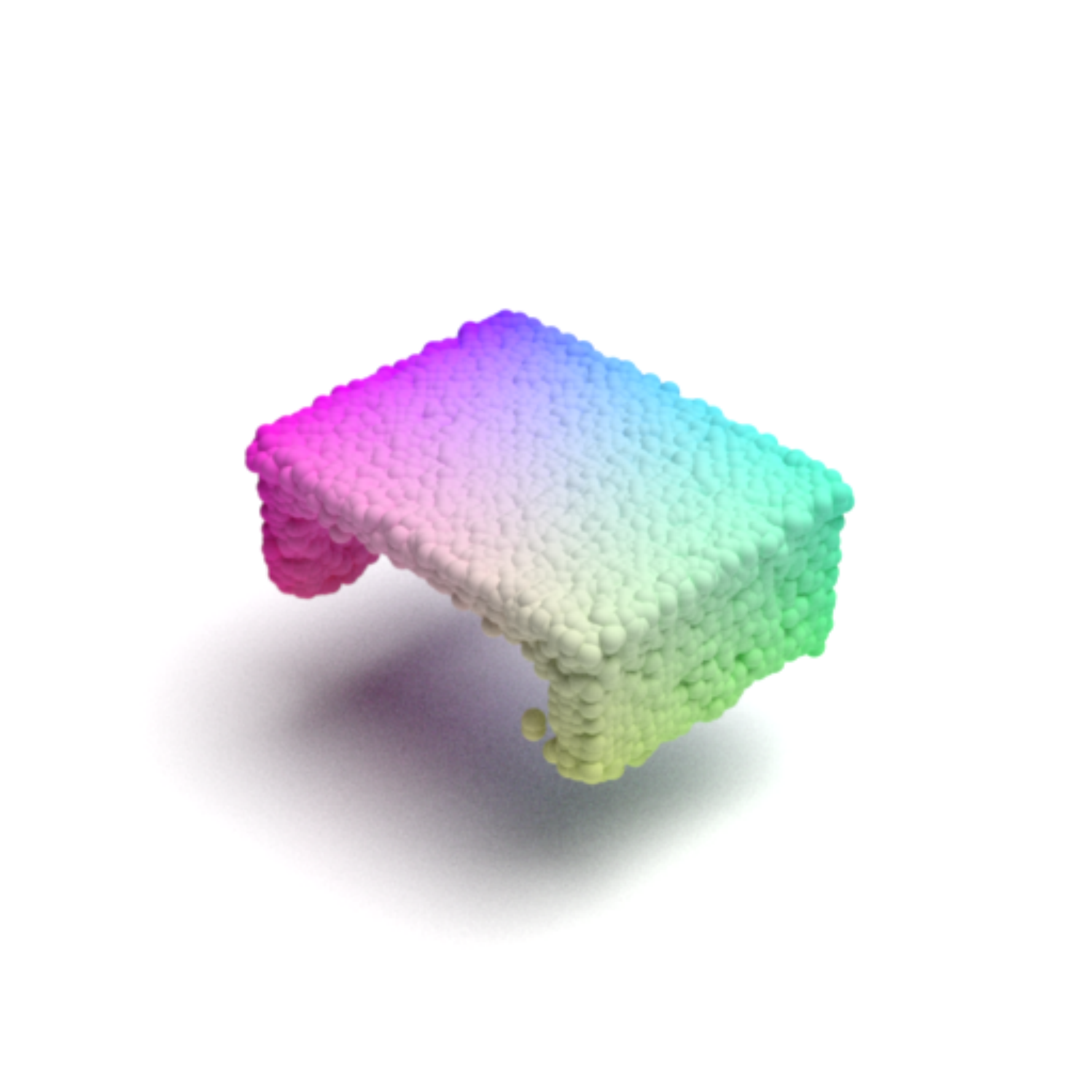}
\includegraphics[width=0.09\textwidth]{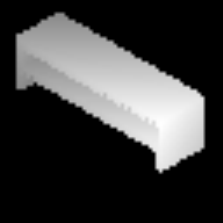}
\includegraphics[width=0.09\textwidth,clip,trim=3cm 3cm 3cm 3cm]{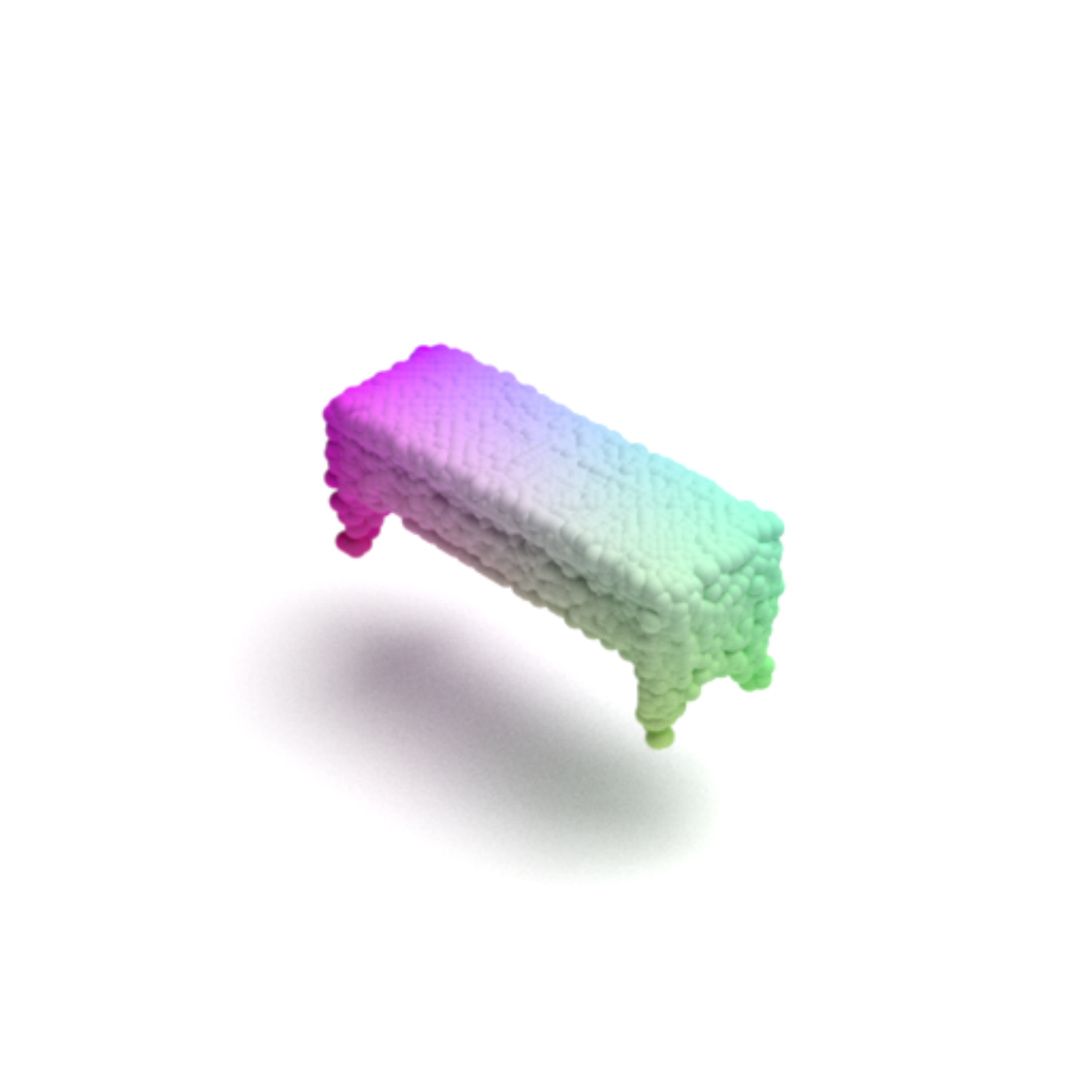}
\includegraphics[width=0.09\textwidth]{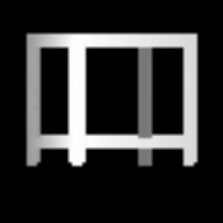}
\includegraphics[width=0.09\textwidth,clip,trim=3cm 3cm 3cm 3cm]{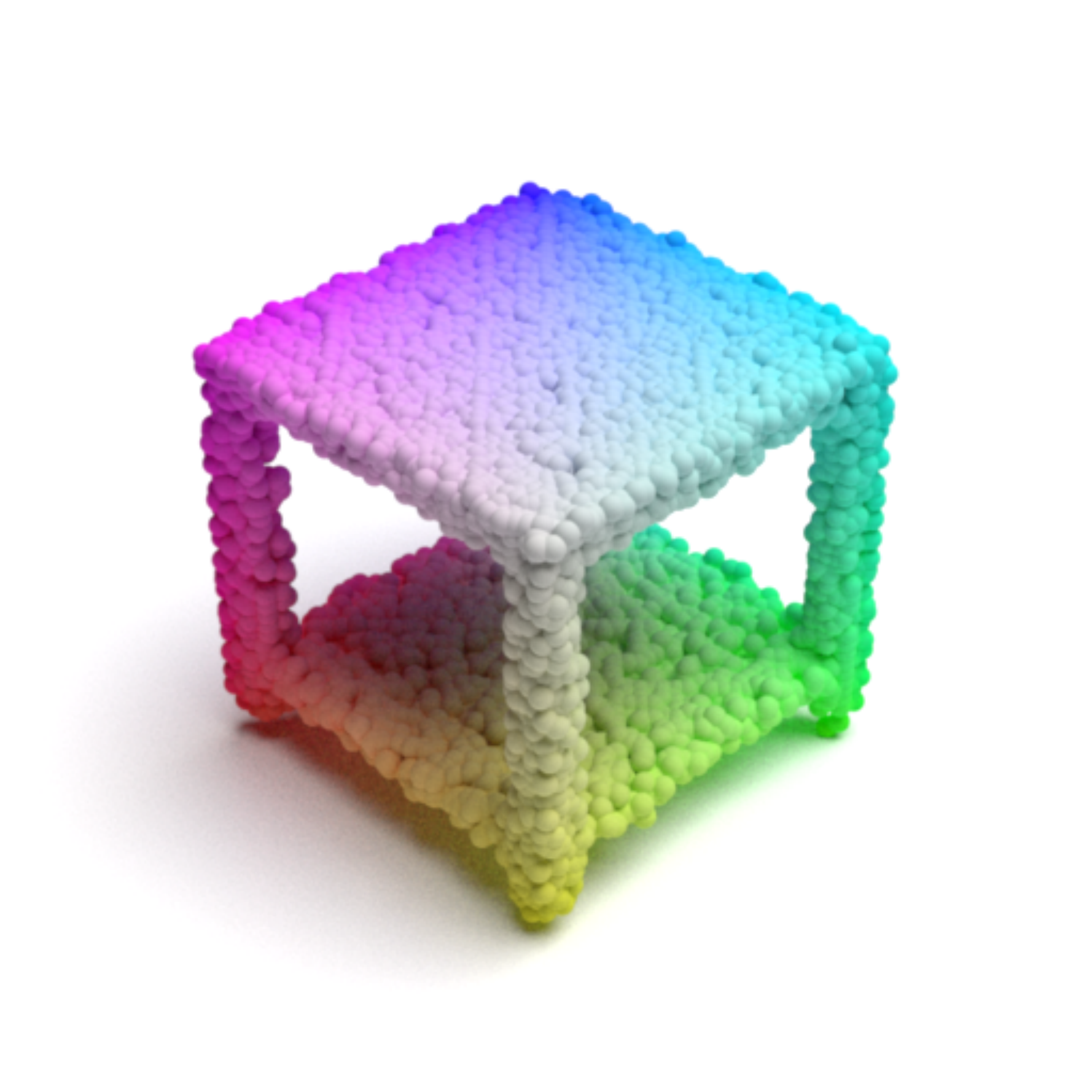}

\captionof{figure}{\textbf{Single View Reconstruction.} Given a single input depth map, we show reconstructions of complete 3D object using our approach. Odd columns are the input depth maps, even columns are the full 3D reconstructions viewed from a canonical viewpoint.}
\vspace*{-0.2cm}%
\label{fig:svr}
\end{figure*}

\begin{table}[!h]
    \centering
    \caption{Shape reconstruction from a single viewpoint}
    \resizebox{\linewidth}{!}{
    \begin{tabular}{@{}lcccccc@{}}
        \toprule
        CD, mean & chair & sofa & table & lamp & plane & car\\
        \midrule
        3D-EPN~\cite{dai2017shape}     & 2.83  & 2.18  & -    & -    & 2.19   & - \\
        DeepSDF~\cite{park2019deepsdf} & 2.11  & 1.59  & -    & -    & 1.16   & - \\
        Soltani et al.~\cite{3DVAE}.   & 1.89 & 3.07 & \textbf{1.99}    & 7.18    & 2.71  & - \\
        Ours                           & \textbf{1.05}  & \textbf{0.85}  & 2.26 & \textbf{2.72} & \textbf{0.63}   & \textbf{0.59} \\
        \bottomrule
        \toprule
        CD, median & chair & sofa & table & lamp & plane & car\\
        \midrule
        3D-EPN~\cite{dai2017shape}        & 2.25 & 2.03 & -    & -    & 1.63 & - \\
        DeepSDF~\cite{park2019deepsdf}       & 1.28 & 0.82 & -    & -    & 0.37 & - \\
        Soltani et al.~\cite{3DVAE}& 1.58 & 1.29 & 1.41    & 3.16    & 2.27    & - \\
        Ours          & \textbf{0.49} & \textbf{0.42} & \textbf{0.62} & \textbf{1.00} & \textbf{0.26} & \textbf{0.25} \\
        \bottomrule
        \toprule
        EMD, mean & chair & sofa & table & lamp & plane & car\\
        \midrule
        3D-EPN~\cite{dai2017shape}        & 0.084 & 0.071 & -     & -     & 0.063 & - \\
        DeepSDF~\cite{park2019deepsdf}       & \textbf{0.071} & \textbf{0.059} & -     & -     & \textbf{0.049} & - \\
        Soltani et al.~\cite{3DVAE}& 0.150 & 0.091     & 0.086     & 0.143     & 0.083 & - \\
        Ours          & 0.091 & 0.090 & \textbf{0.067} & \textbf{0.109} & 0.064 & \textbf{0.056} \\
        \bottomrule
    \end{tabular}
    }
\label{table:svr}
\end{table}

\medskip
\noindent
\textbf{Single View 3D Reconstruction.}
Next, we consider the problem of recovering the 3D object from a given depth map from a single viewpoint. Our model can trivially perform this task since encoder encodes view independent representation of the object and decoder can use the encoded latent vector to generate depth maps from any viewpoint. For quantitative evaluation, we take the single viewpoint depth map of objects from test split and compute CD and EMD loss between the full ground truth and predicted 3D point clouds. 
Table~\ref{table:svr} shows that our method outperforms all other methods in Singe View Reconstruction task with CD as distance metric. Our method is close second (and better than other multi-view depth baseline) when we use EMD.
Figure~\ref{fig:svr} shows the results of single view constructions using our model.

\begin{figure*}
\centering

\includegraphics[clip,trim=3cm 3cm 3cm 3cm, width=0.095\textwidth]{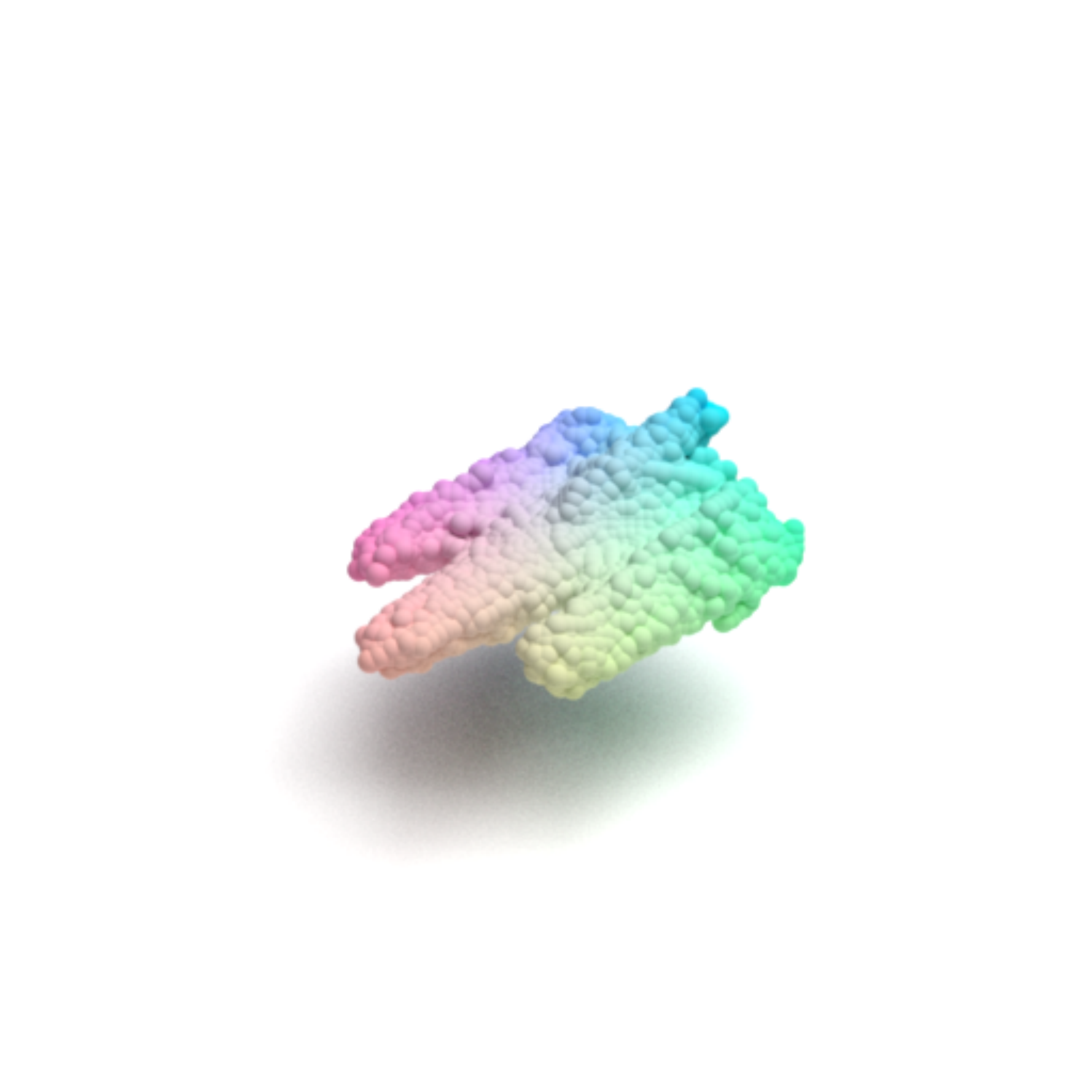}
\includegraphics[clip,trim=3cm 3cm 3cm 3cm, width=0.095\textwidth]{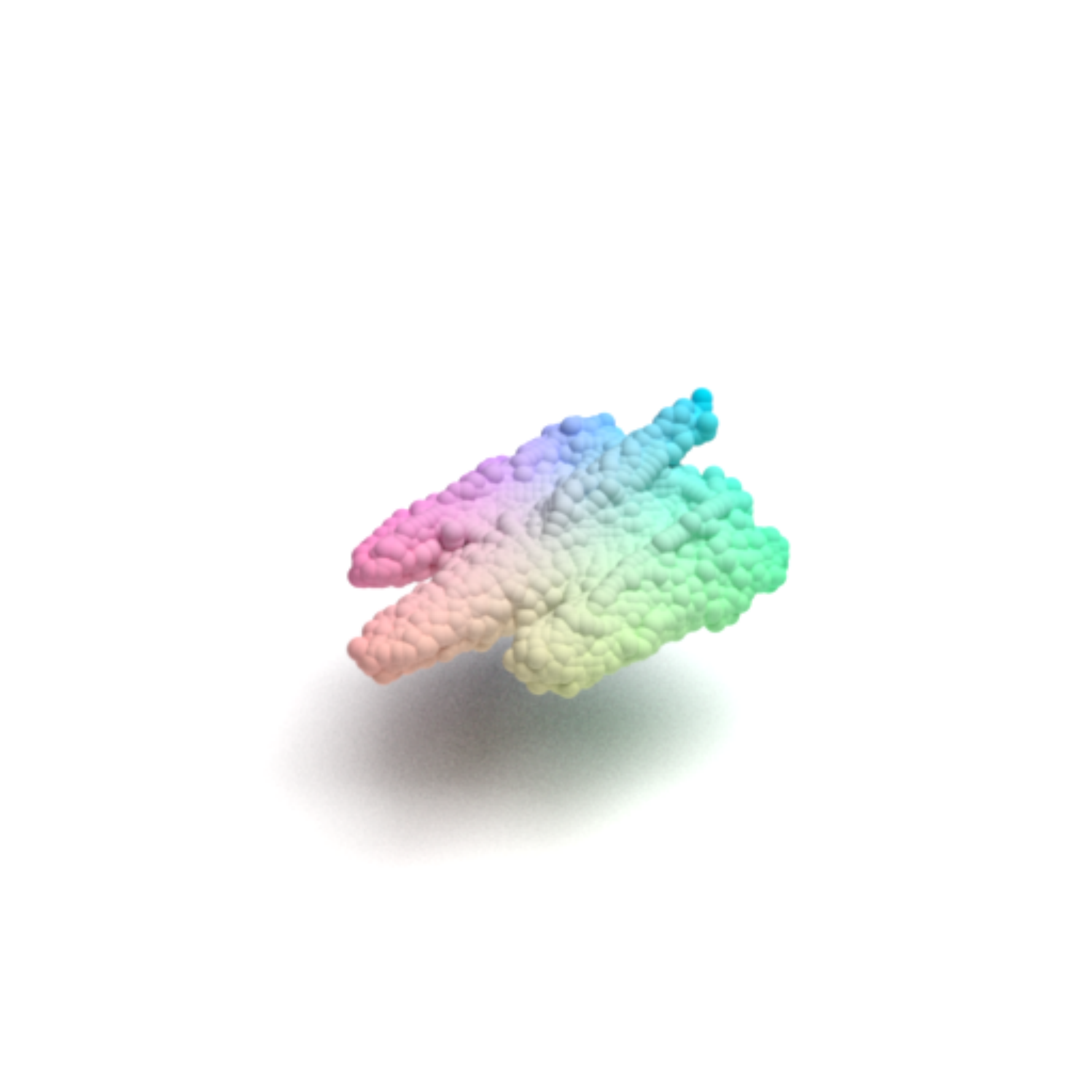}
\includegraphics[clip,trim=3cm 3cm 3cm 3cm, width=0.095\textwidth]{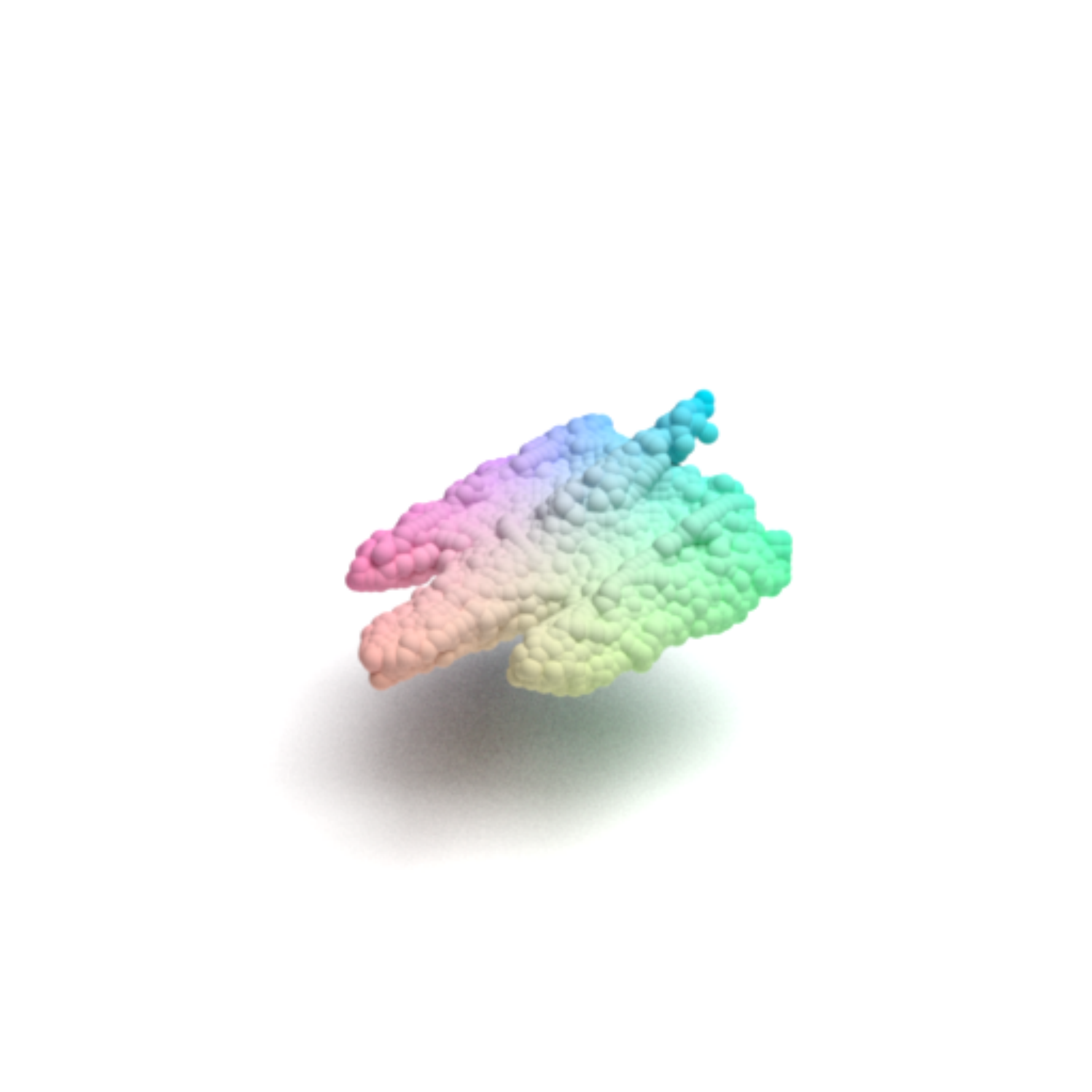}
\includegraphics[clip,trim=3cm 3cm 3cm 3cm, width=0.095\textwidth]{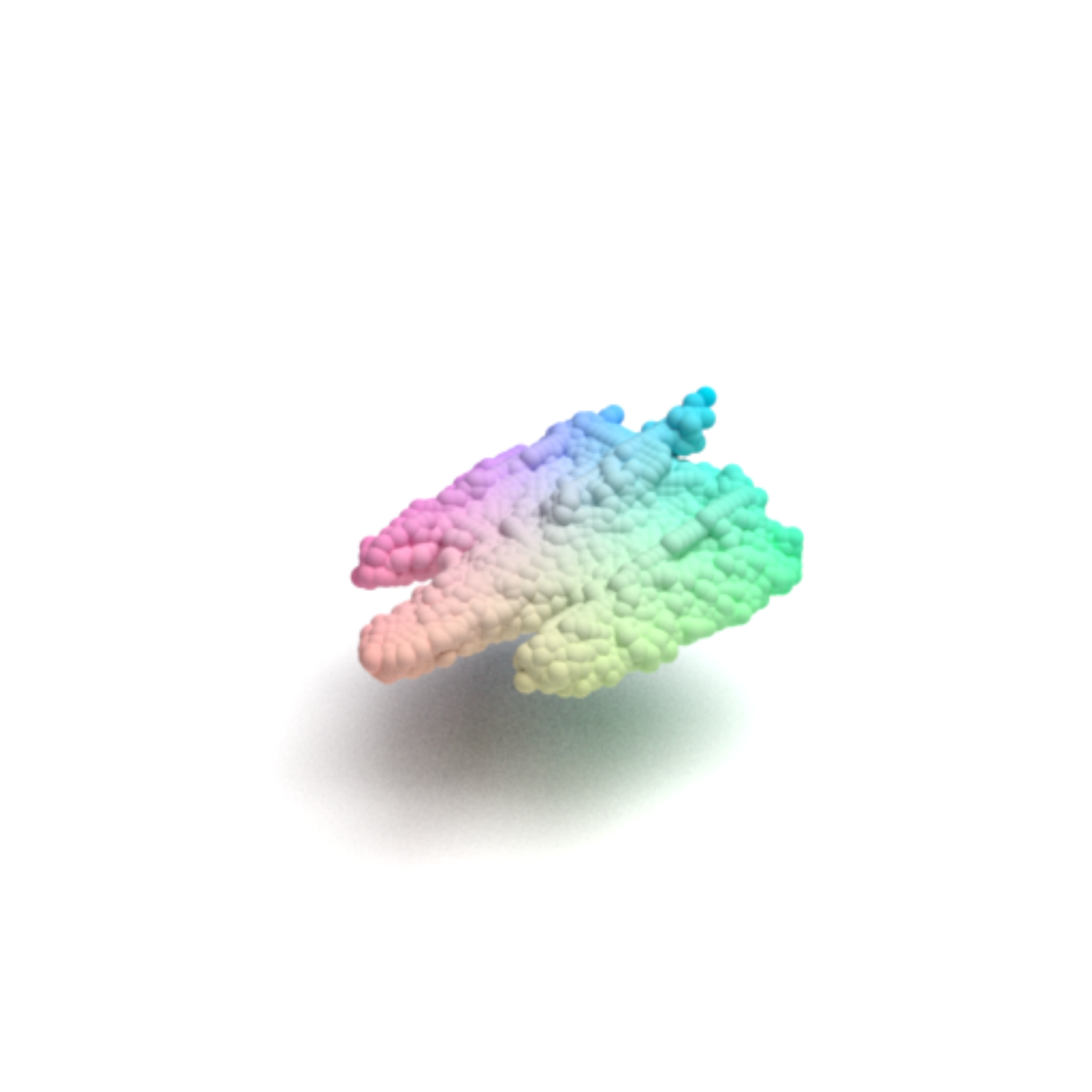}
\includegraphics[clip,trim=3cm 3cm 3cm 3cm, width=0.095\textwidth]{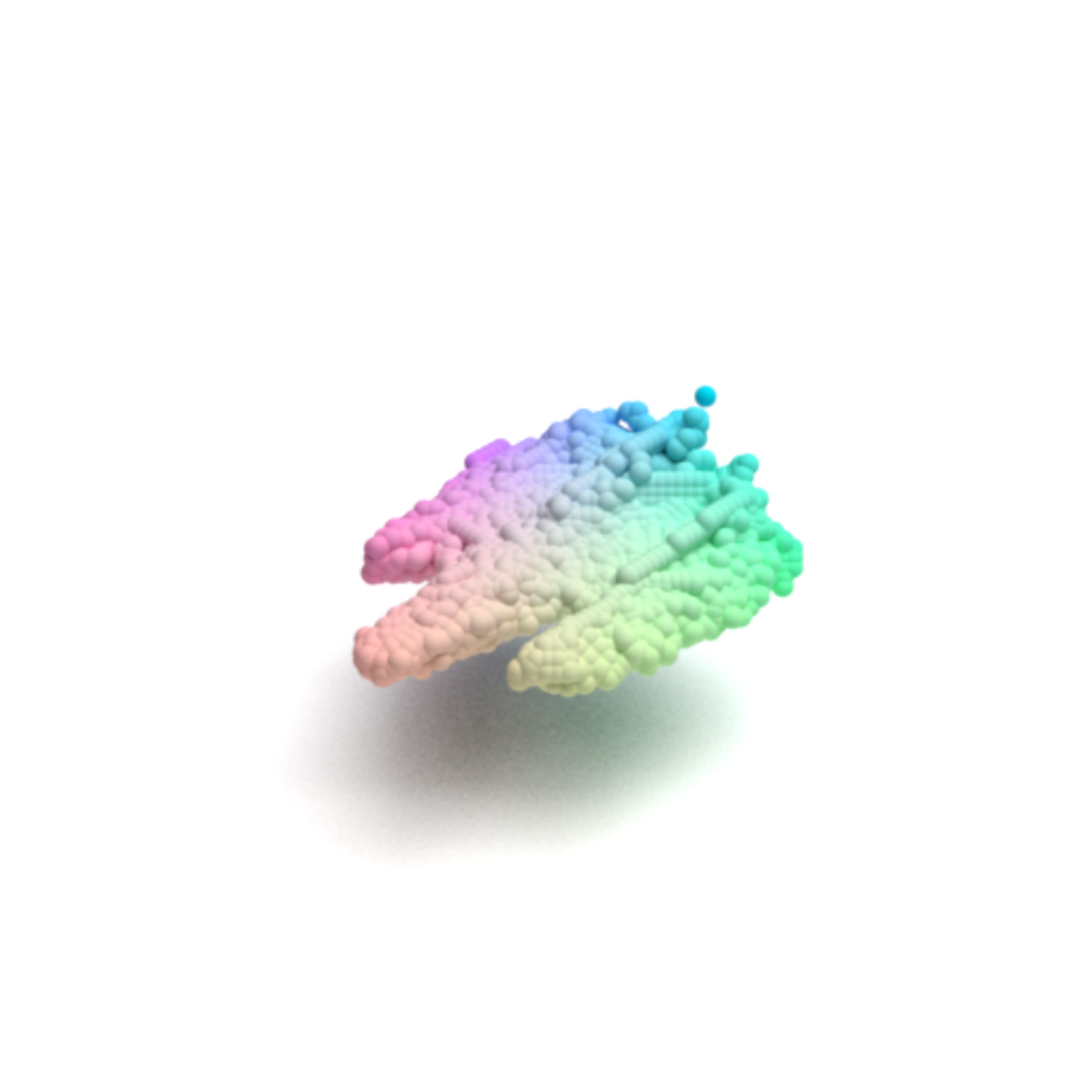}
\includegraphics[clip,trim=3cm 3cm 3cm 3cm, width=0.095\textwidth]{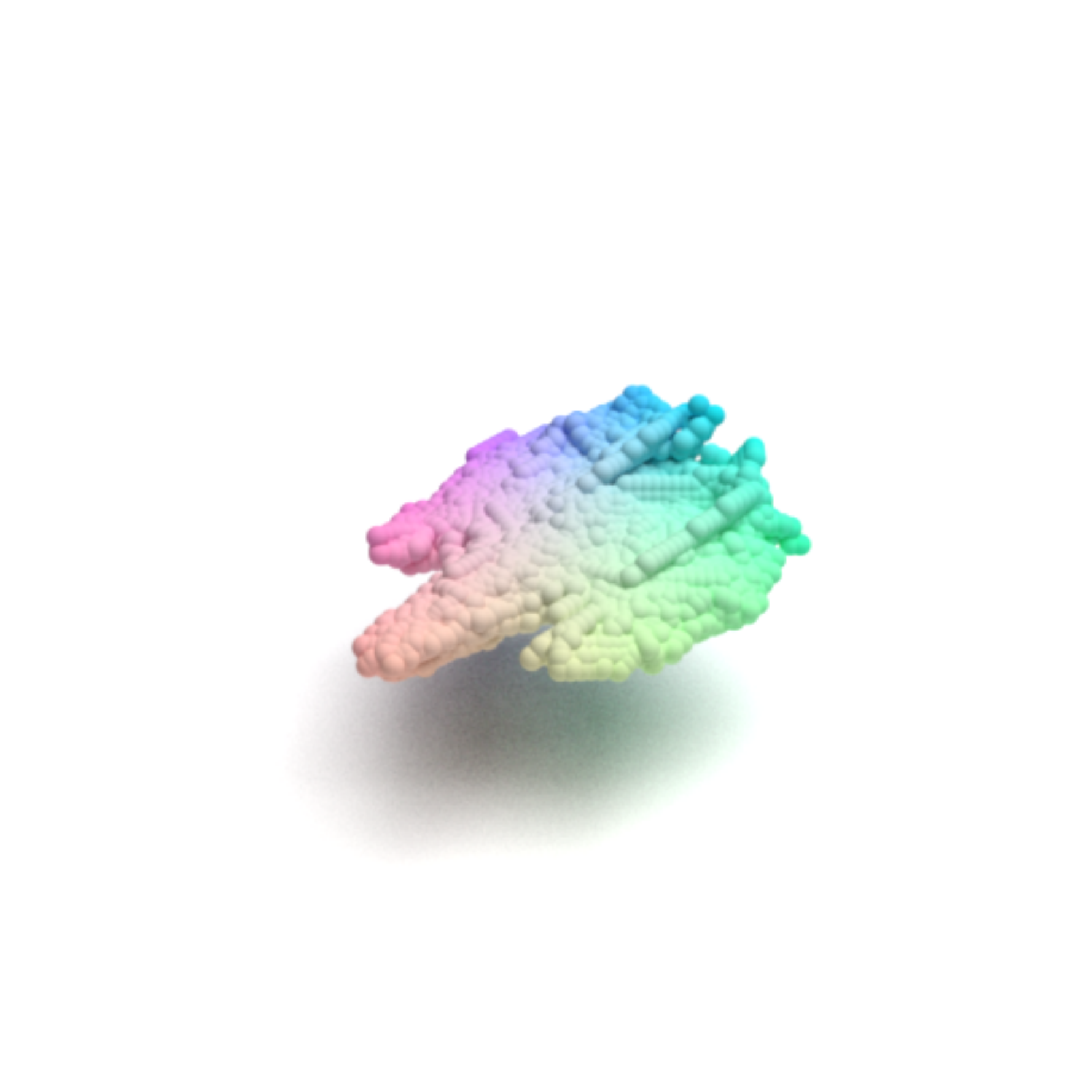}
\includegraphics[clip,trim=3cm 3cm 3cm 3cm, width=0.095\textwidth]{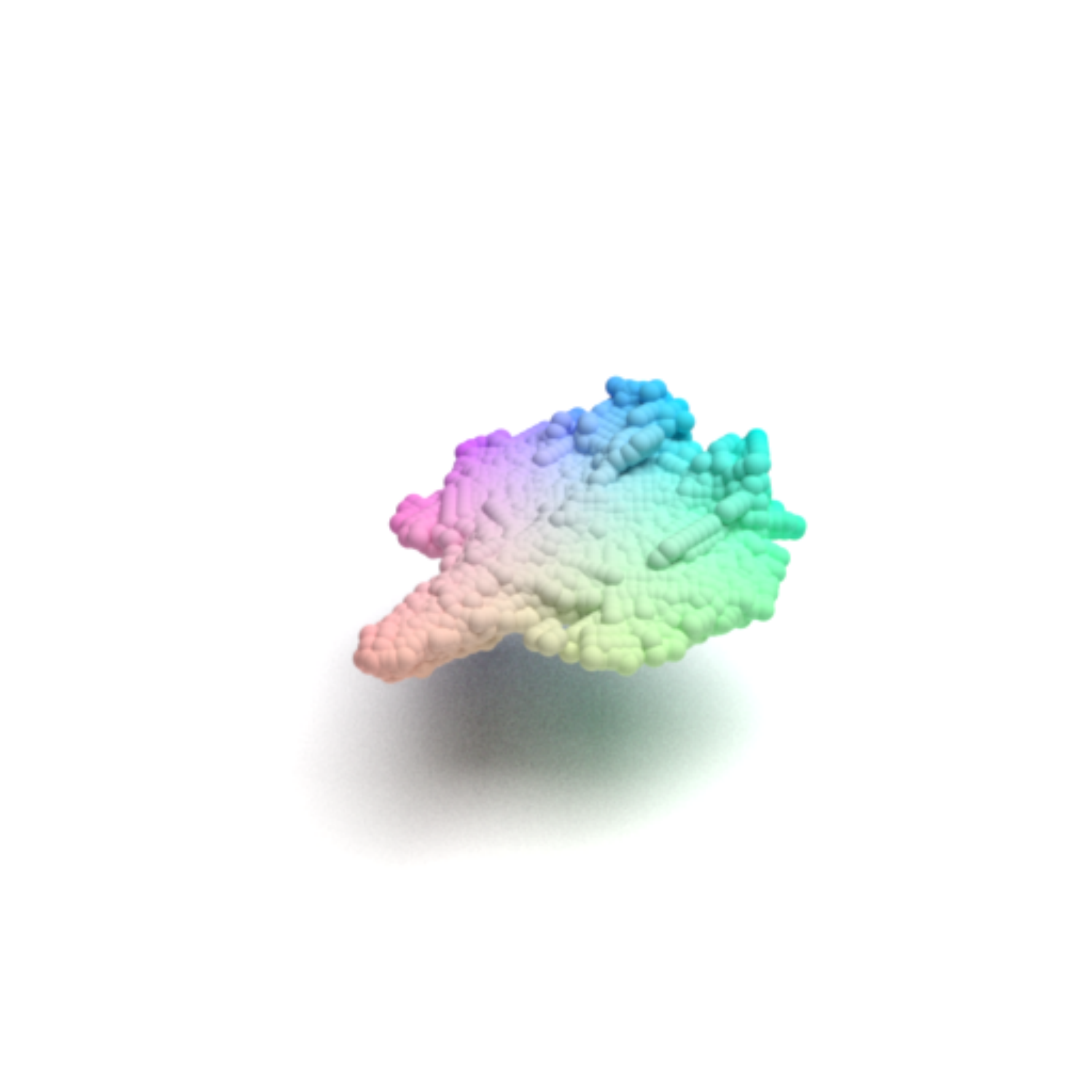}
\includegraphics[clip,trim=3cm 3cm 3cm 3cm, width=0.095\textwidth]{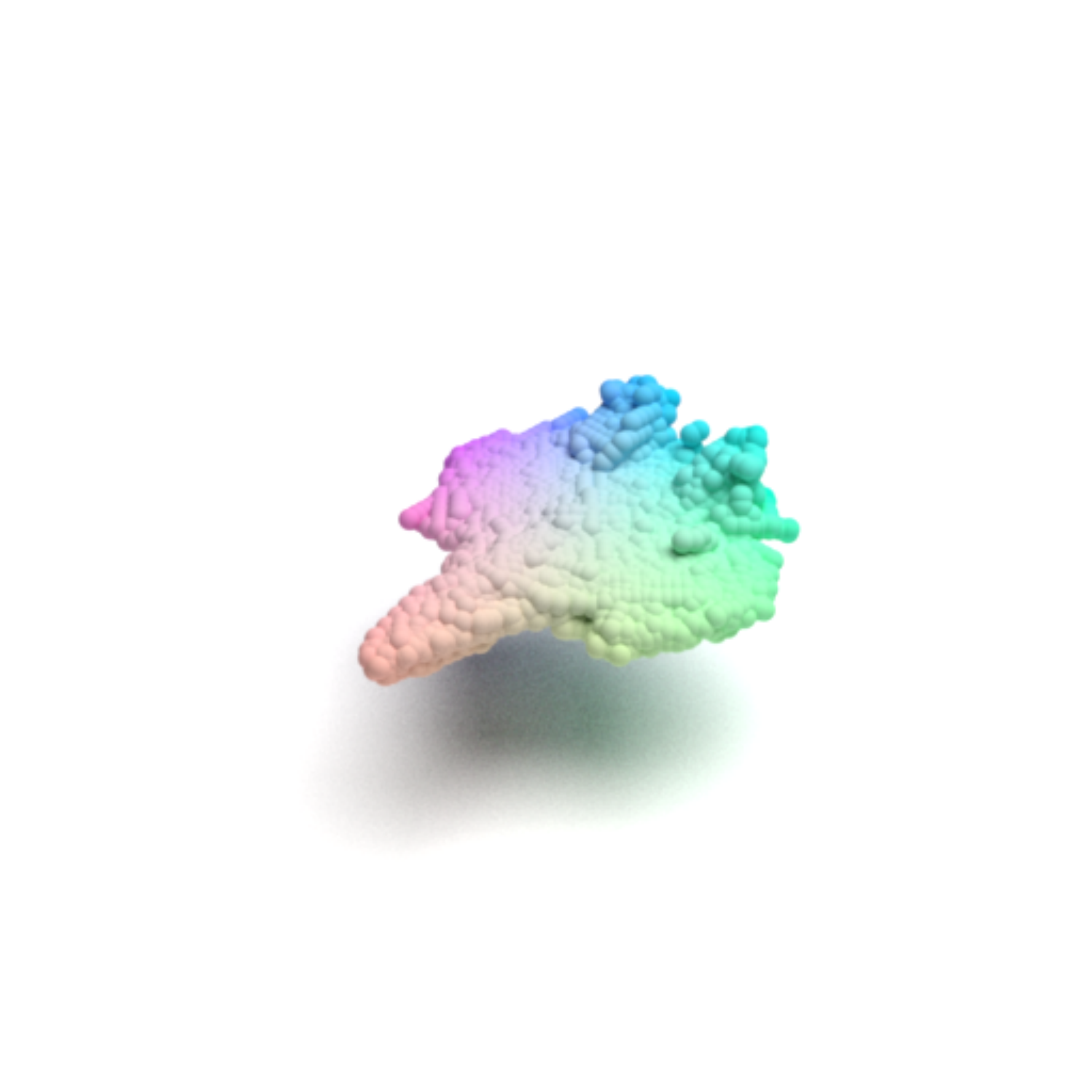}
\includegraphics[clip,trim=3cm 3cm 3cm 3cm, width=0.095\textwidth]{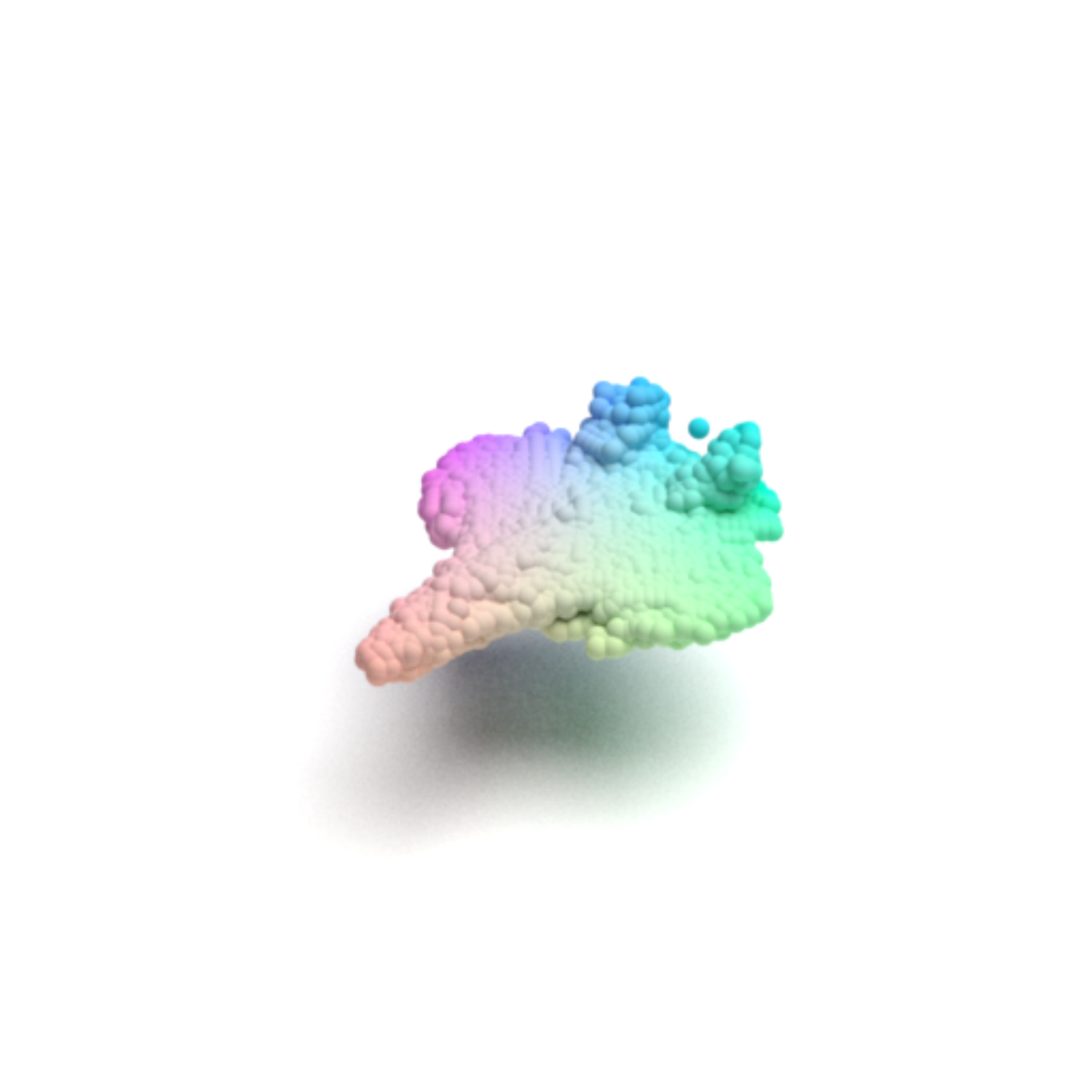}
\includegraphics[clip,trim=3cm 3cm 3cm 3cm, width=0.095\textwidth]{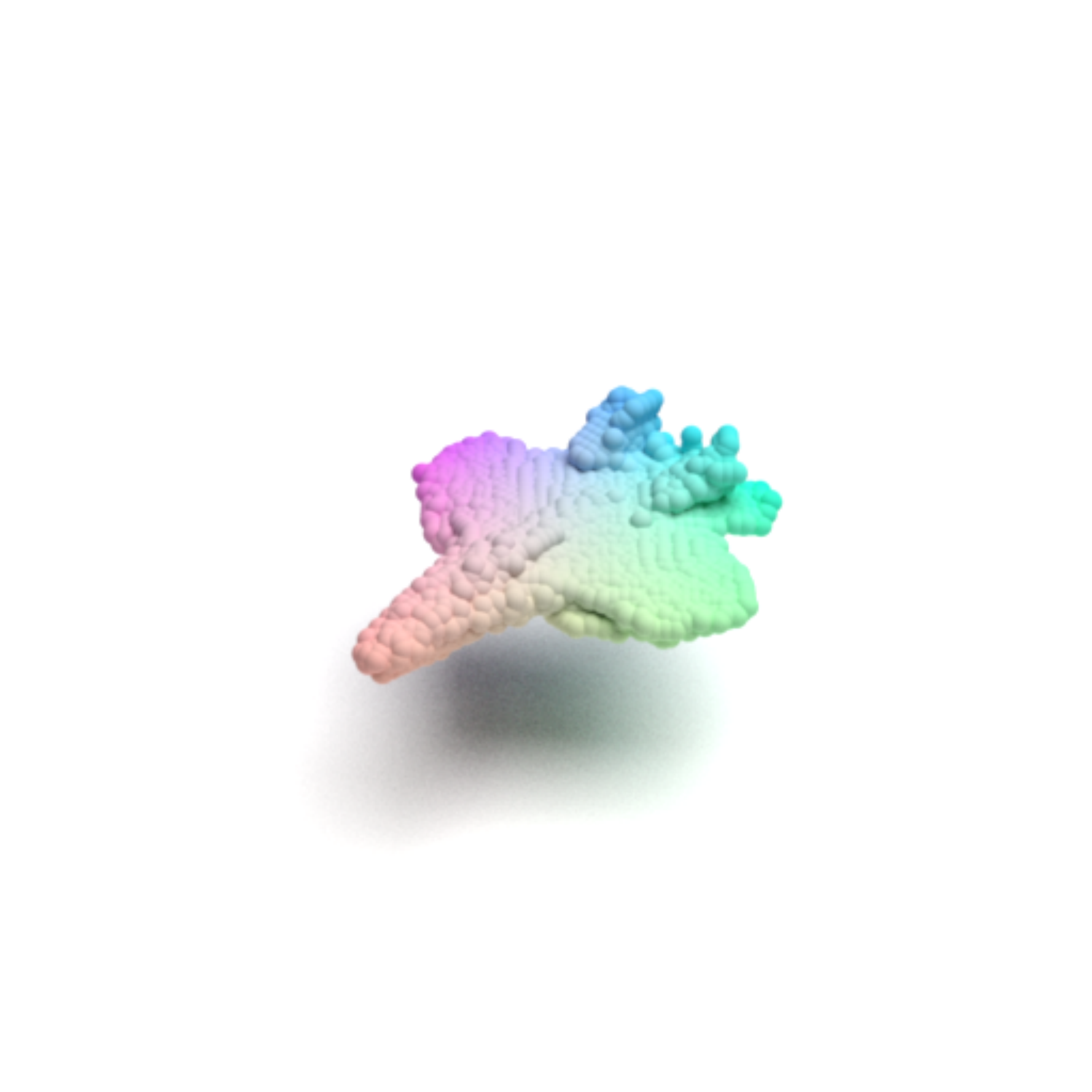}

\includegraphics[clip,trim=3cm 3cm 3cm 3cm, width=0.095\textwidth]{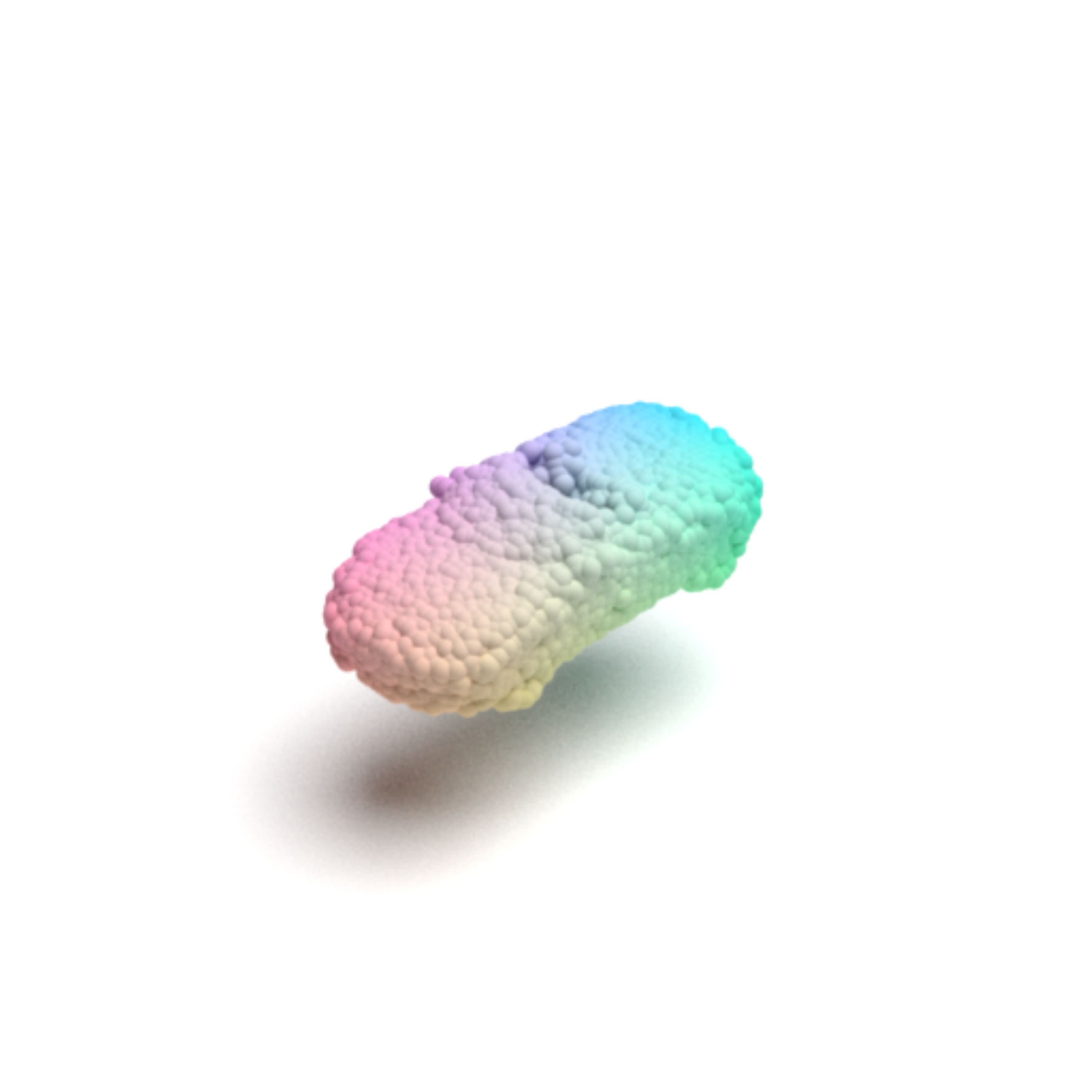}
\includegraphics[clip,trim=3cm 3cm 3cm 3cm, width=0.095\textwidth]{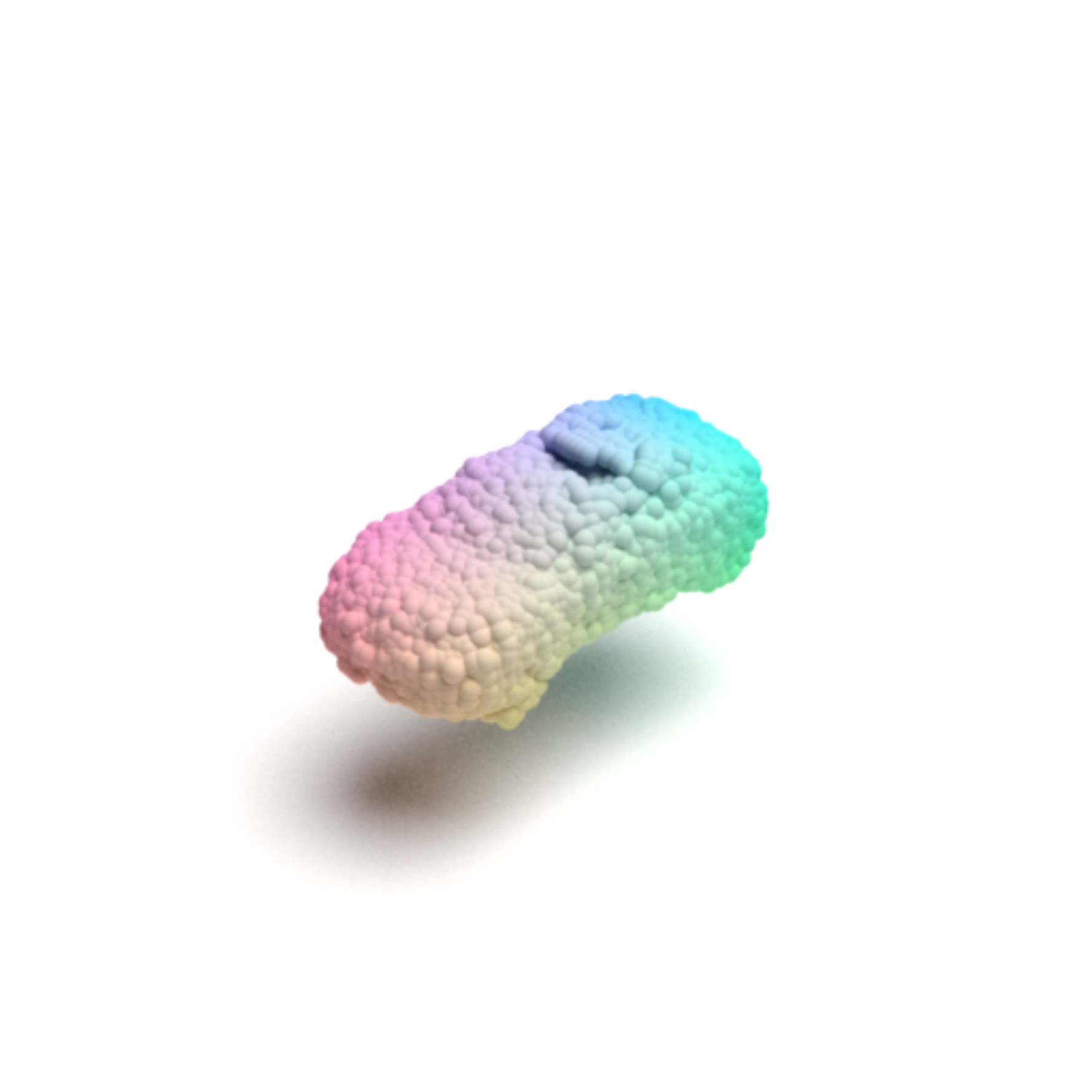}
\includegraphics[clip,trim=3cm 3cm 3cm 3cm, width=0.095\textwidth]{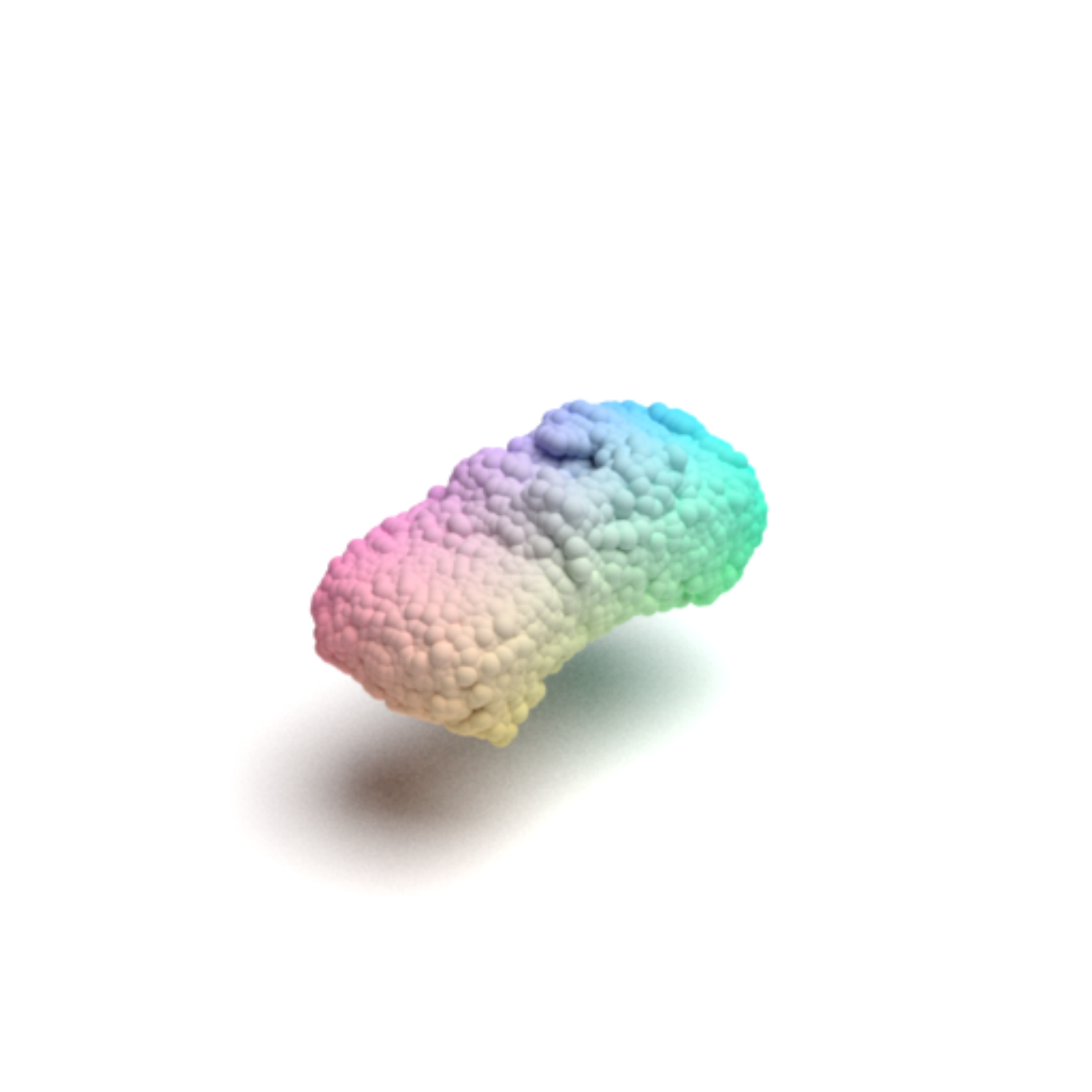}
\includegraphics[clip,trim=3cm 3cm 3cm 3cm, width=0.095\textwidth]{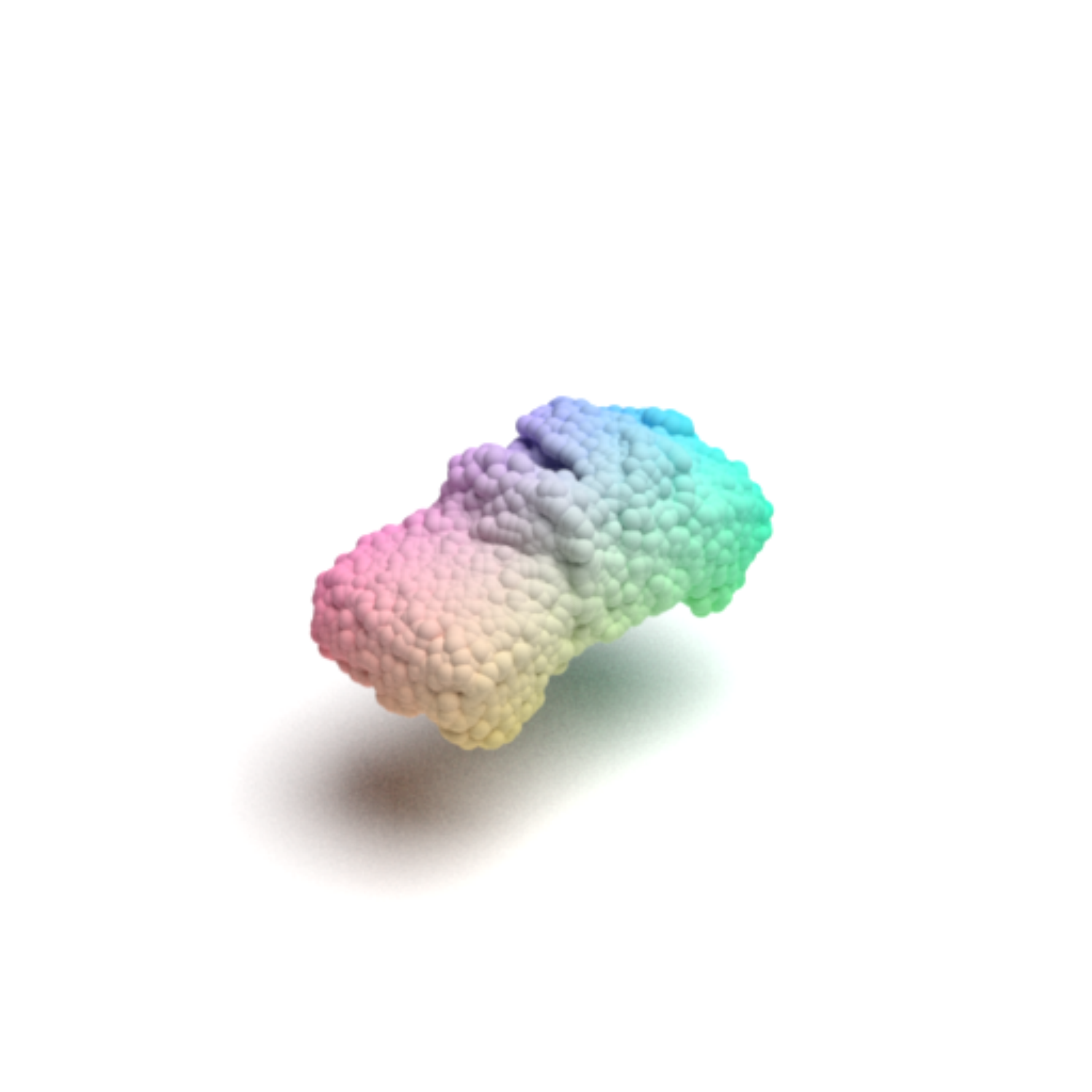}
\includegraphics[clip,trim=3cm 3cm 3cm 3cm, width=0.095\textwidth]{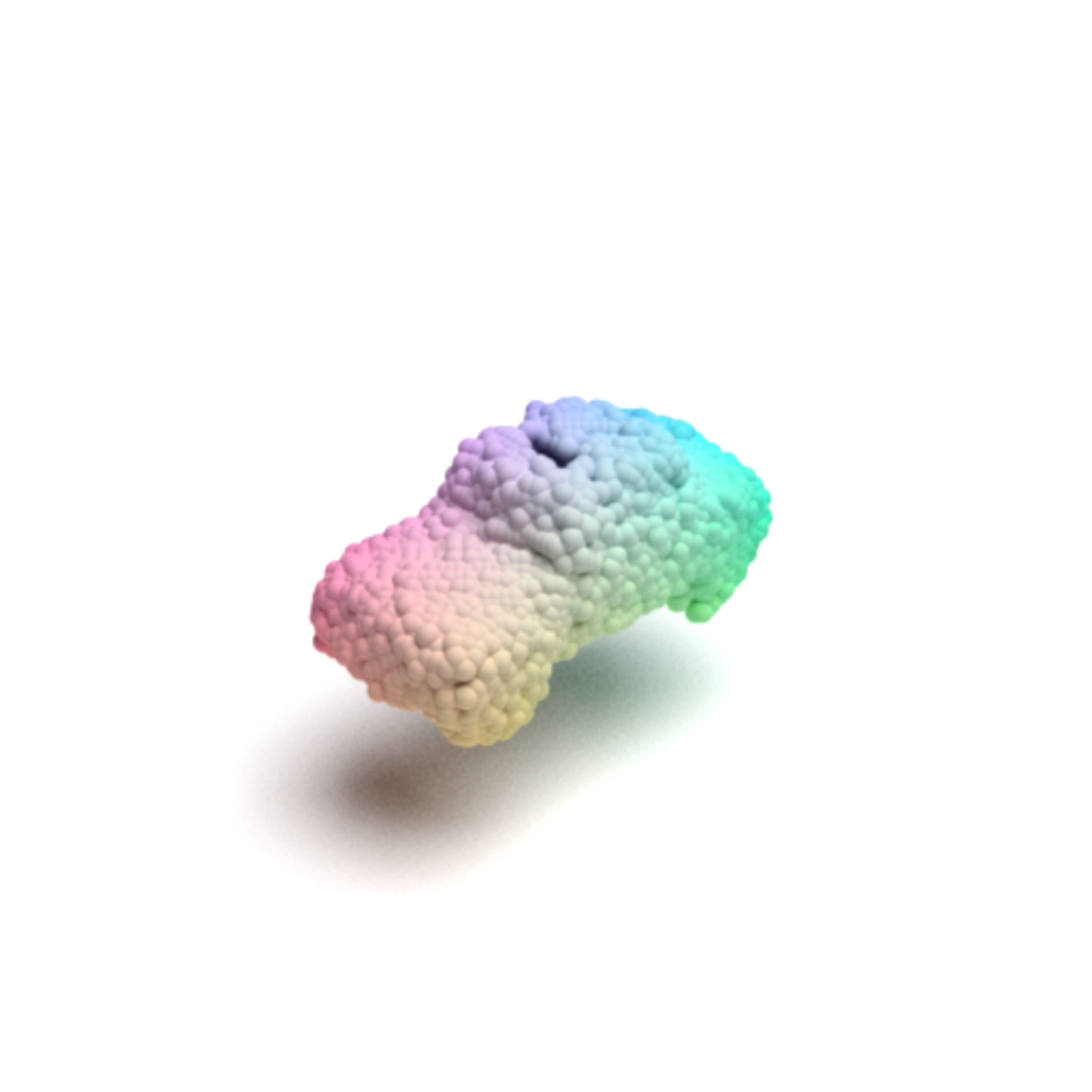}
\includegraphics[clip,trim=3cm 3cm 3cm 3cm, width=0.095\textwidth]{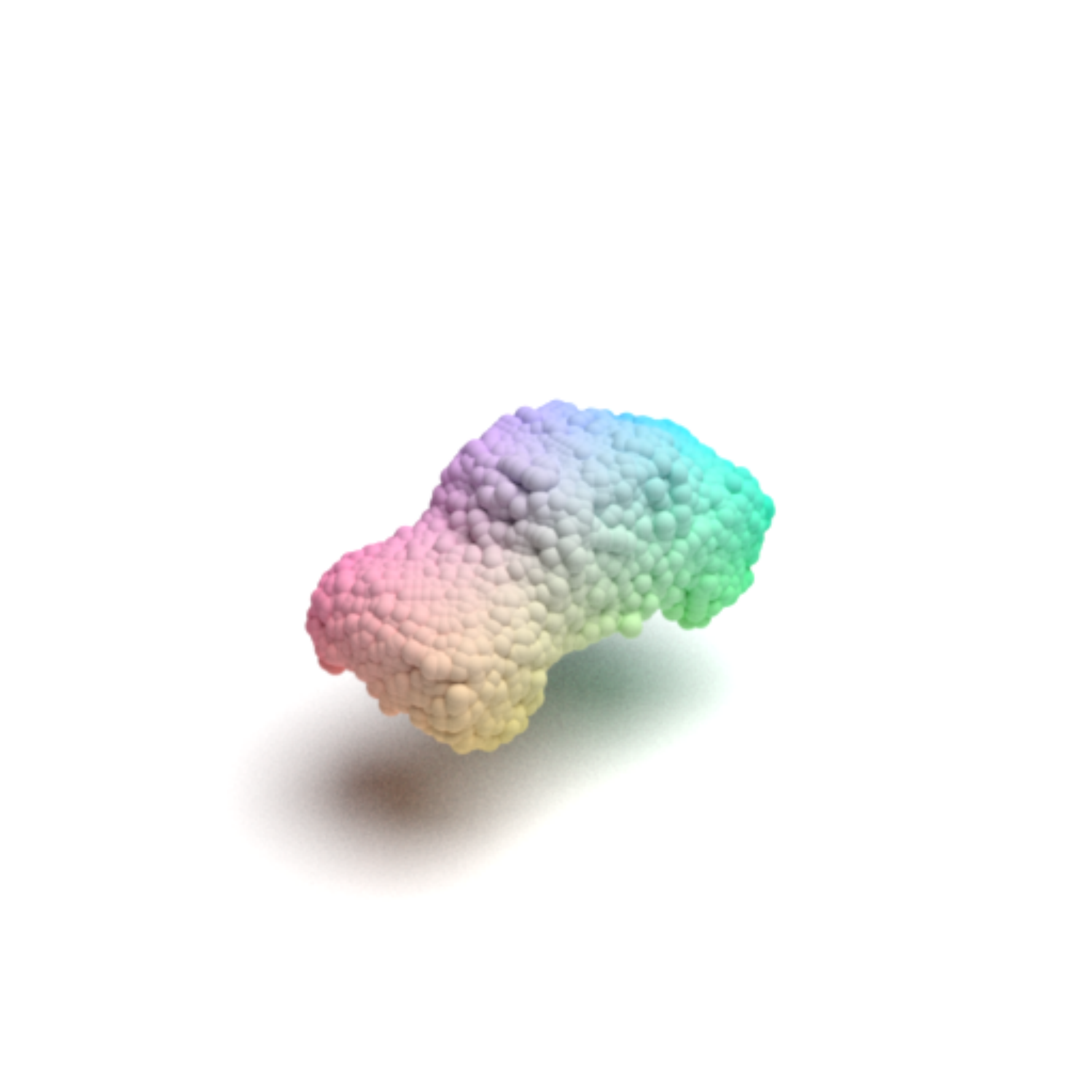}
\includegraphics[clip,trim=3cm 3cm 3cm 3cm, width=0.095\textwidth]{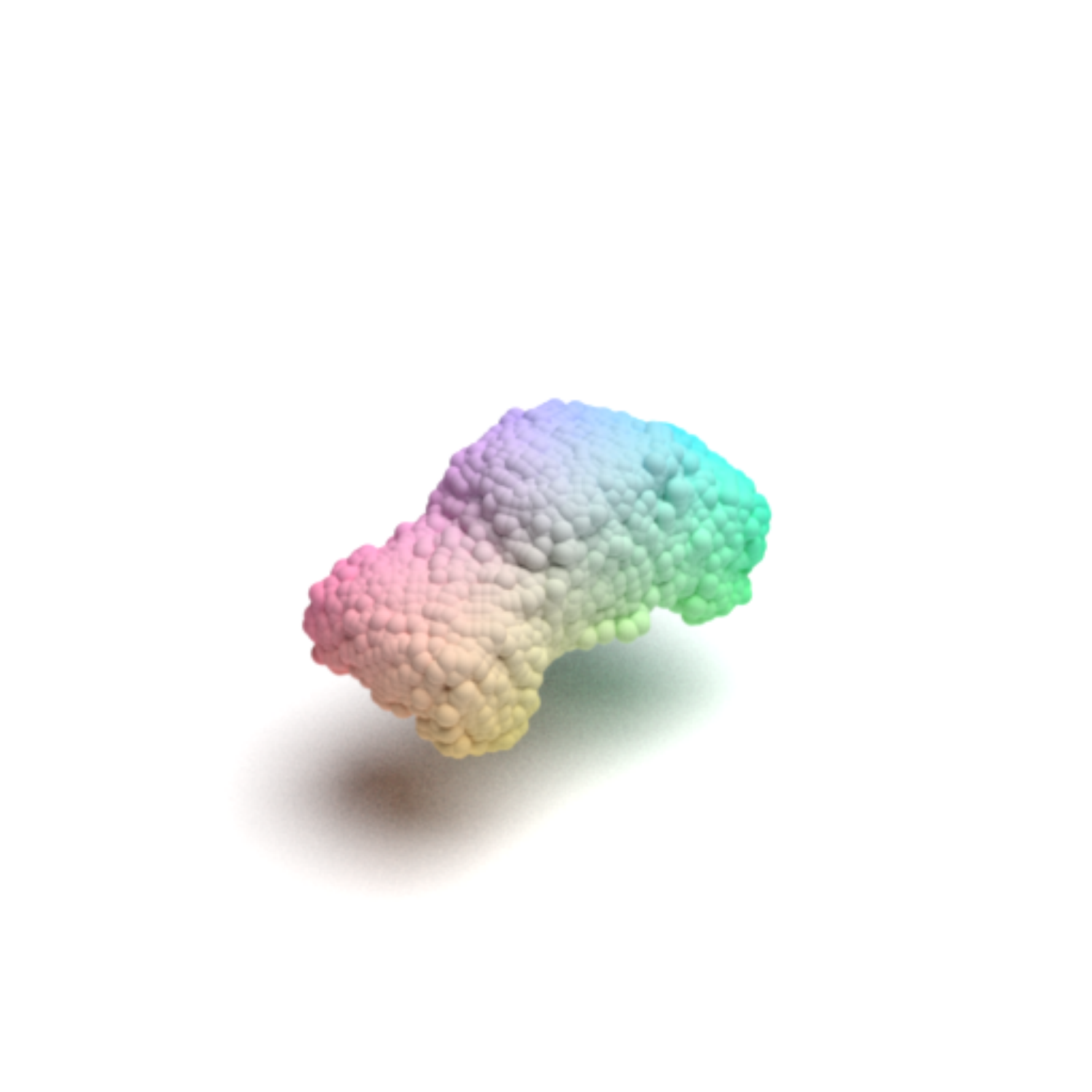}
\includegraphics[clip,trim=3cm 3cm 3cm 3cm, width=0.095\textwidth]{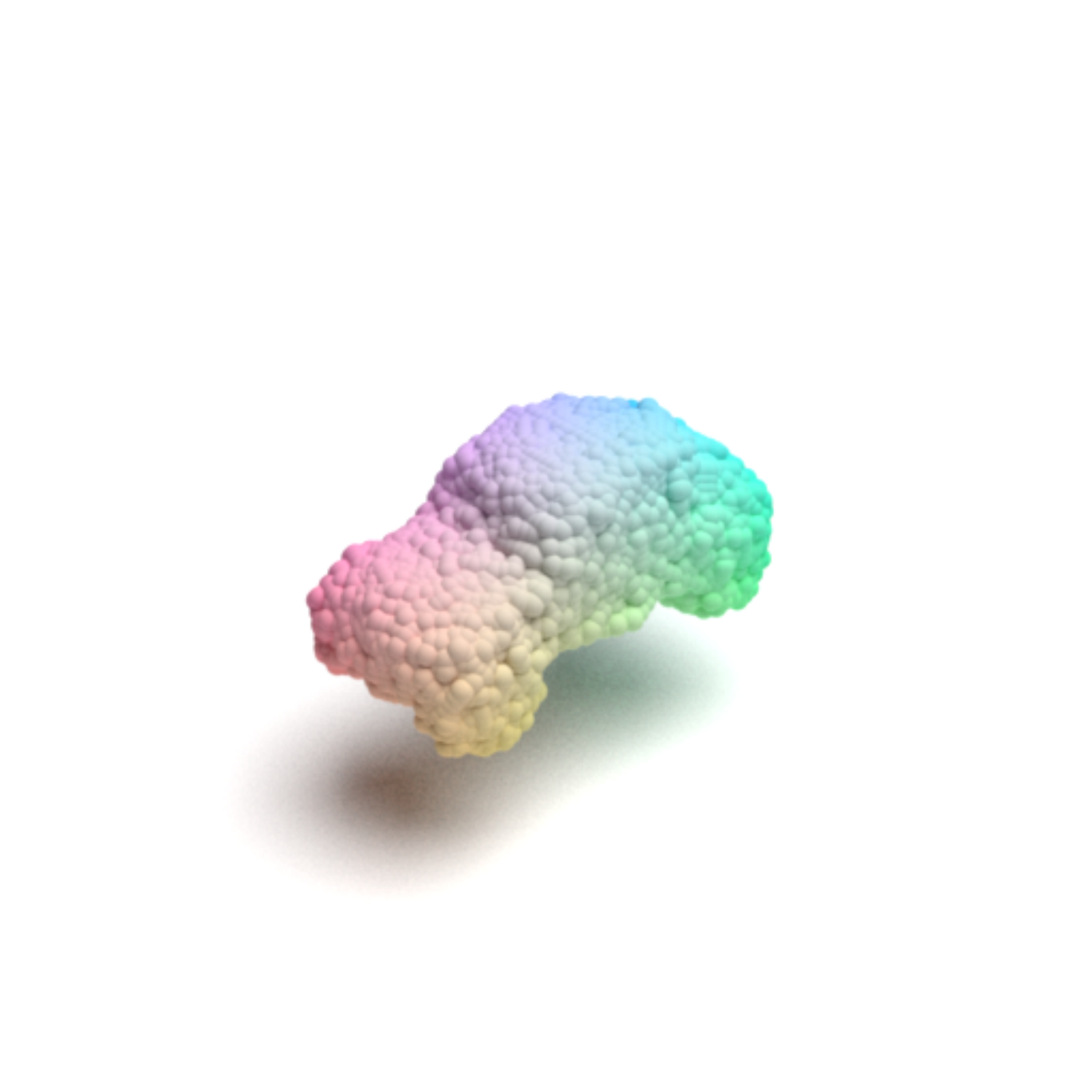}
\includegraphics[clip,trim=3cm 3cm 3cm 3cm, width=0.095\textwidth]{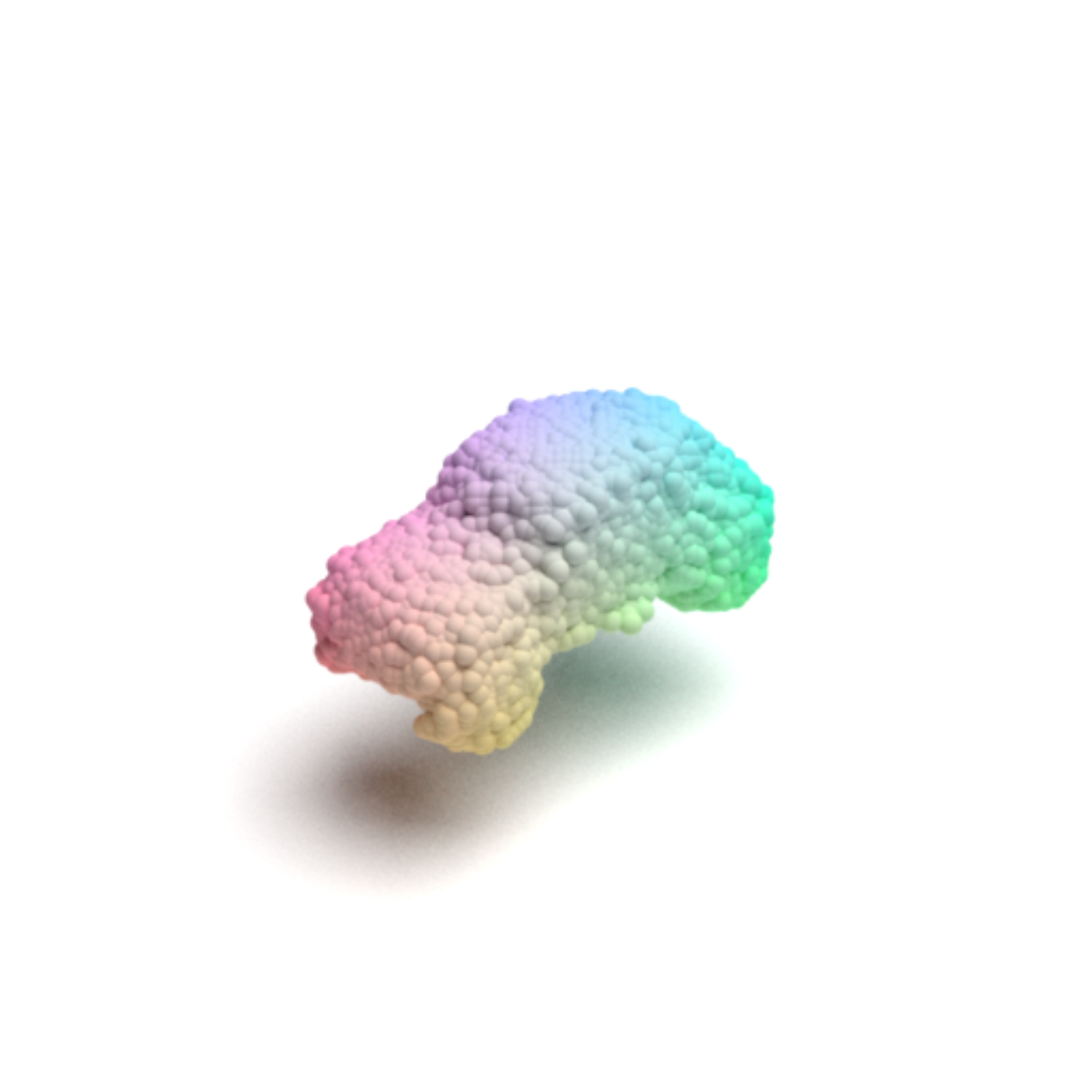}
\includegraphics[clip,trim=3cm 3cm 3cm 3cm, width=0.095\textwidth]{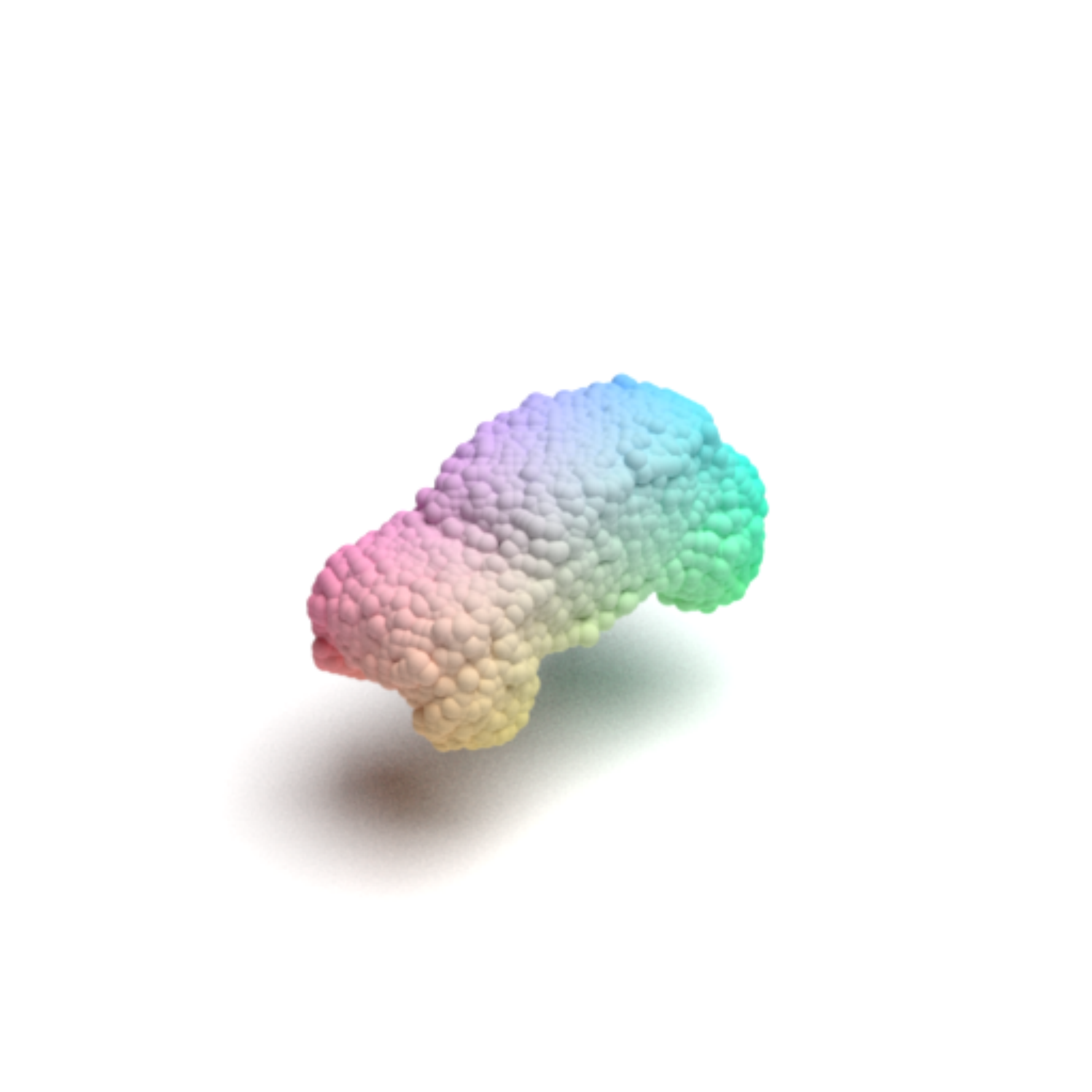}

\includegraphics[clip,trim=3cm 3cm 3cm 3cm, width=0.095\textwidth]{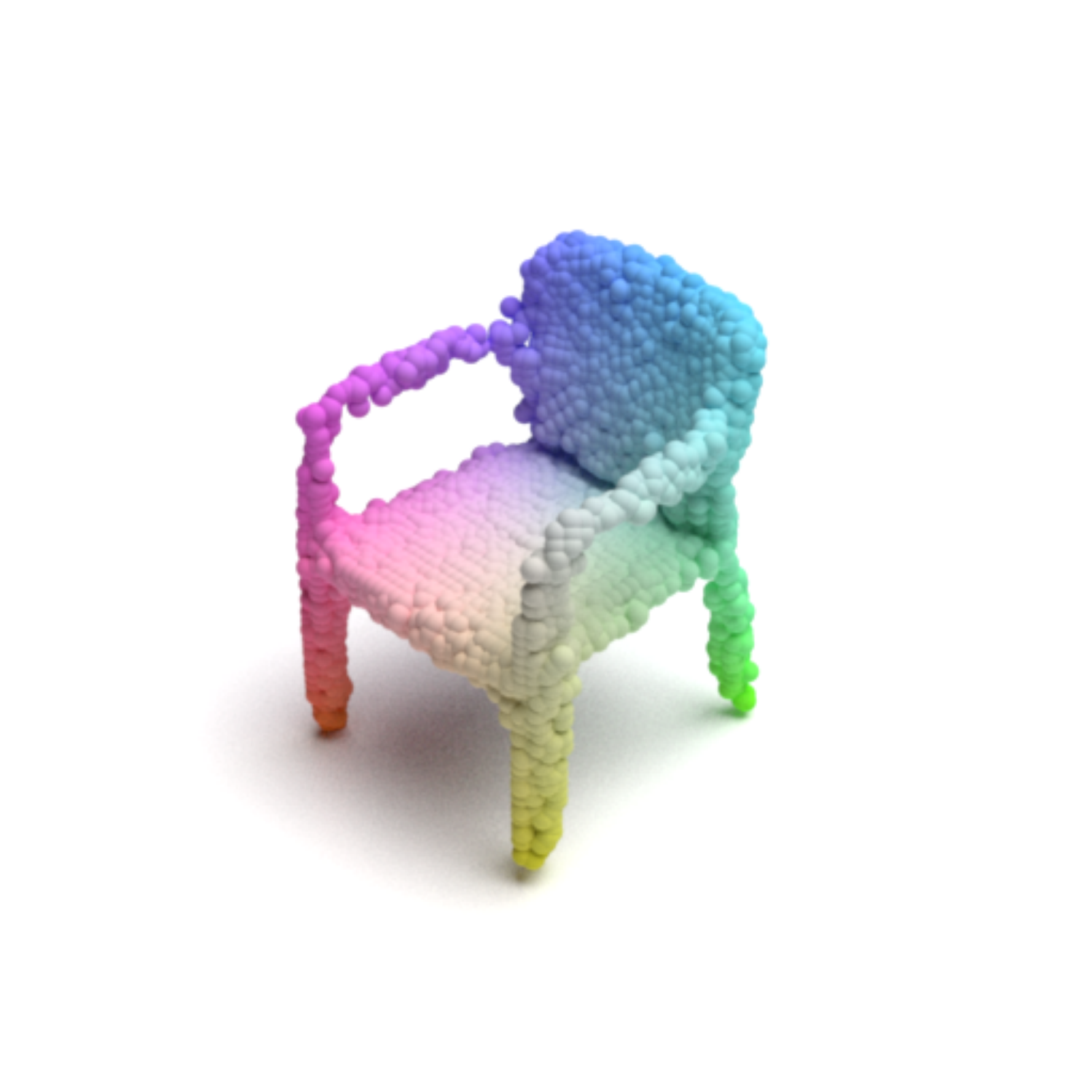}
\includegraphics[clip,trim=3cm 3cm 3cm 3cm, width=0.095\textwidth]{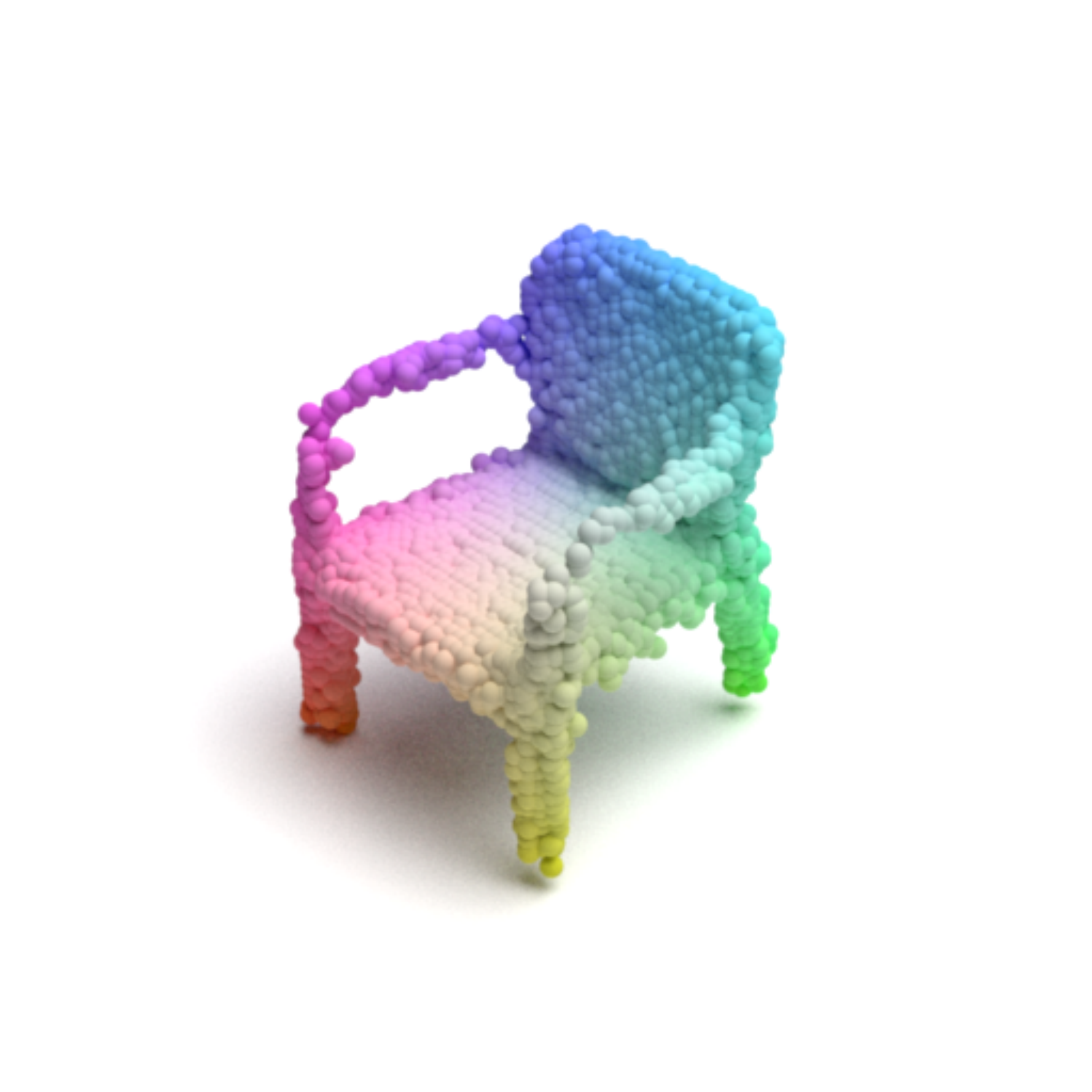}
\includegraphics[clip,trim=3cm 3cm 3cm 3cm, width=0.095\textwidth]{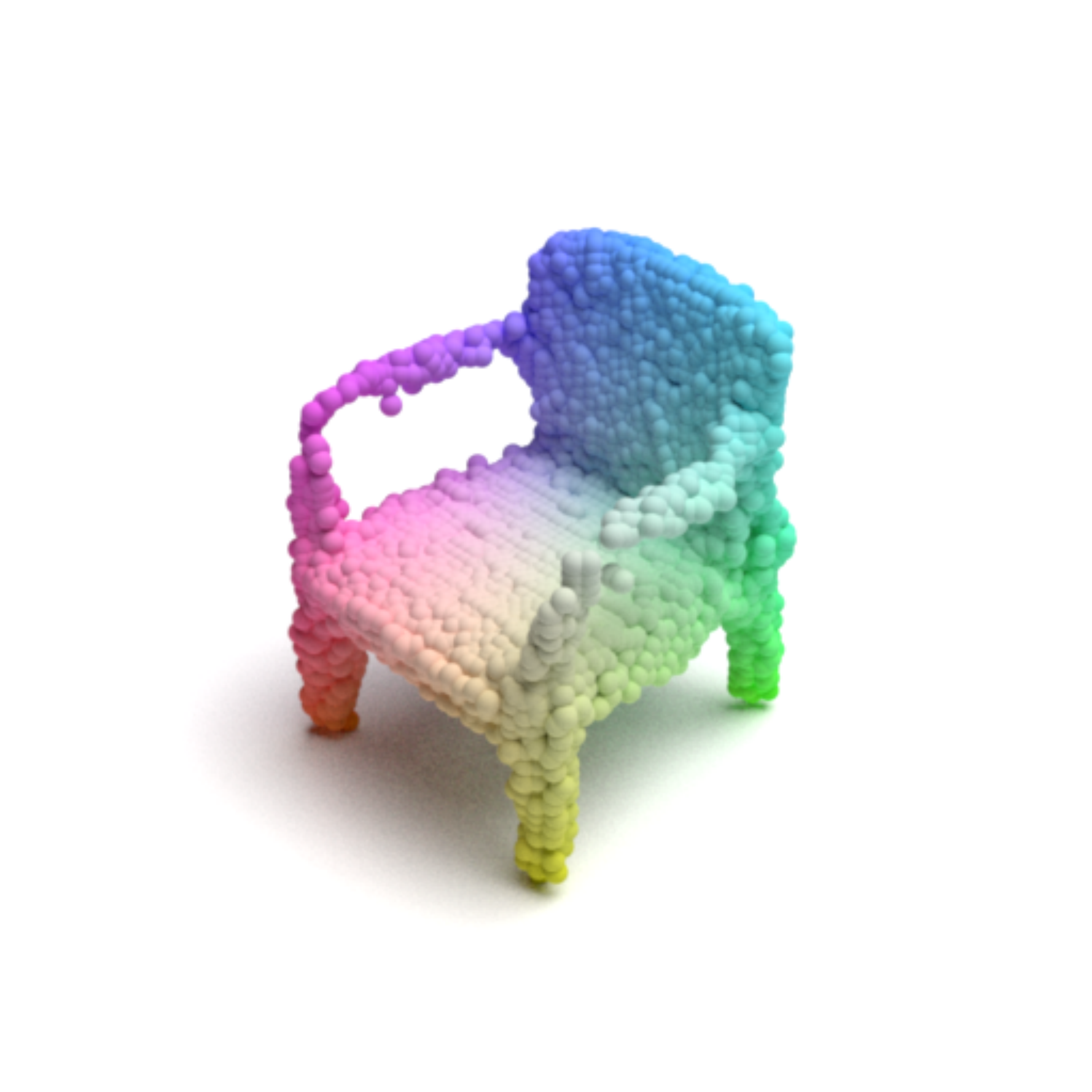}
\includegraphics[clip,trim=3cm 3cm 3cm 3cm, width=0.095\textwidth]{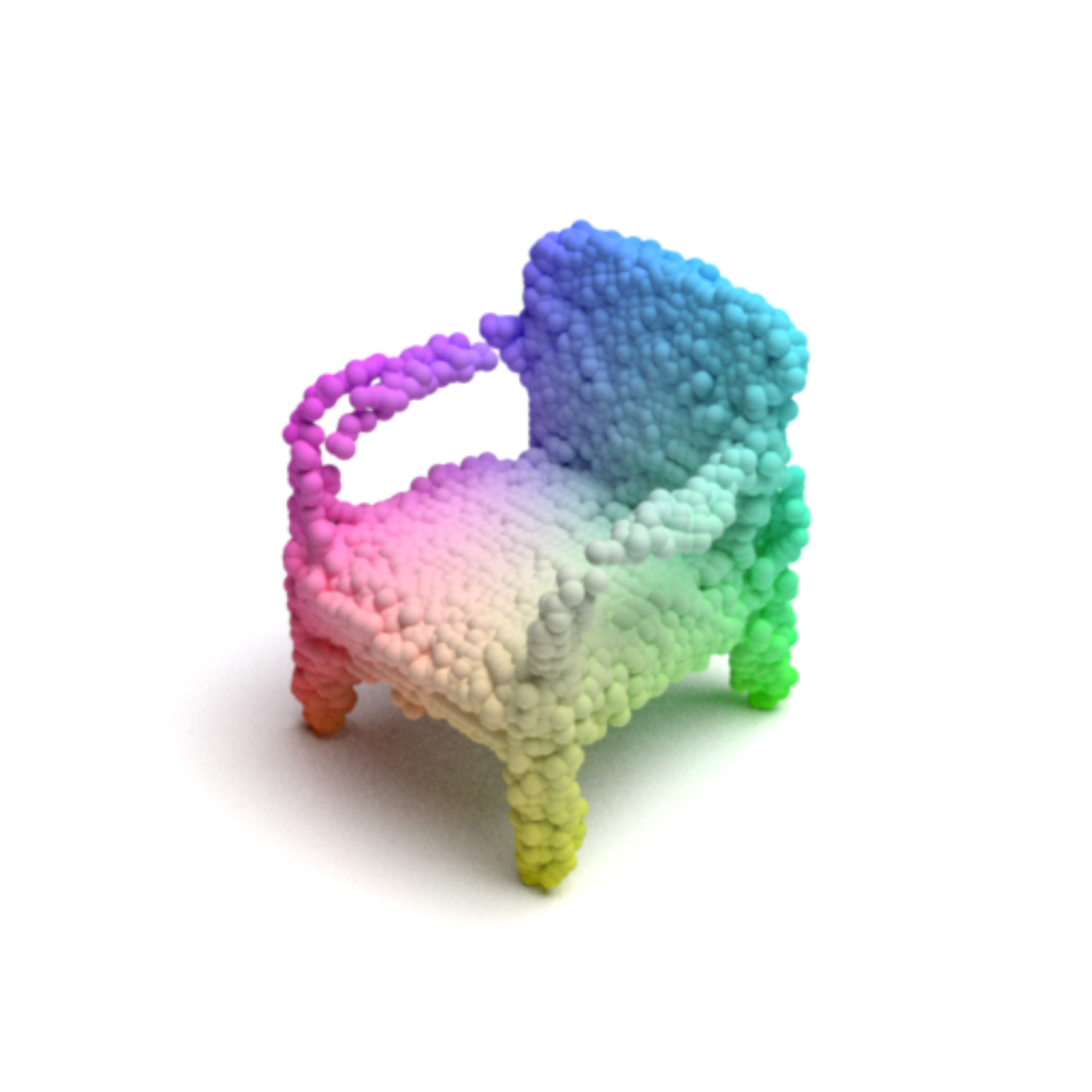}
\includegraphics[clip,trim=3cm 3cm 3cm 3cm, width=0.095\textwidth]{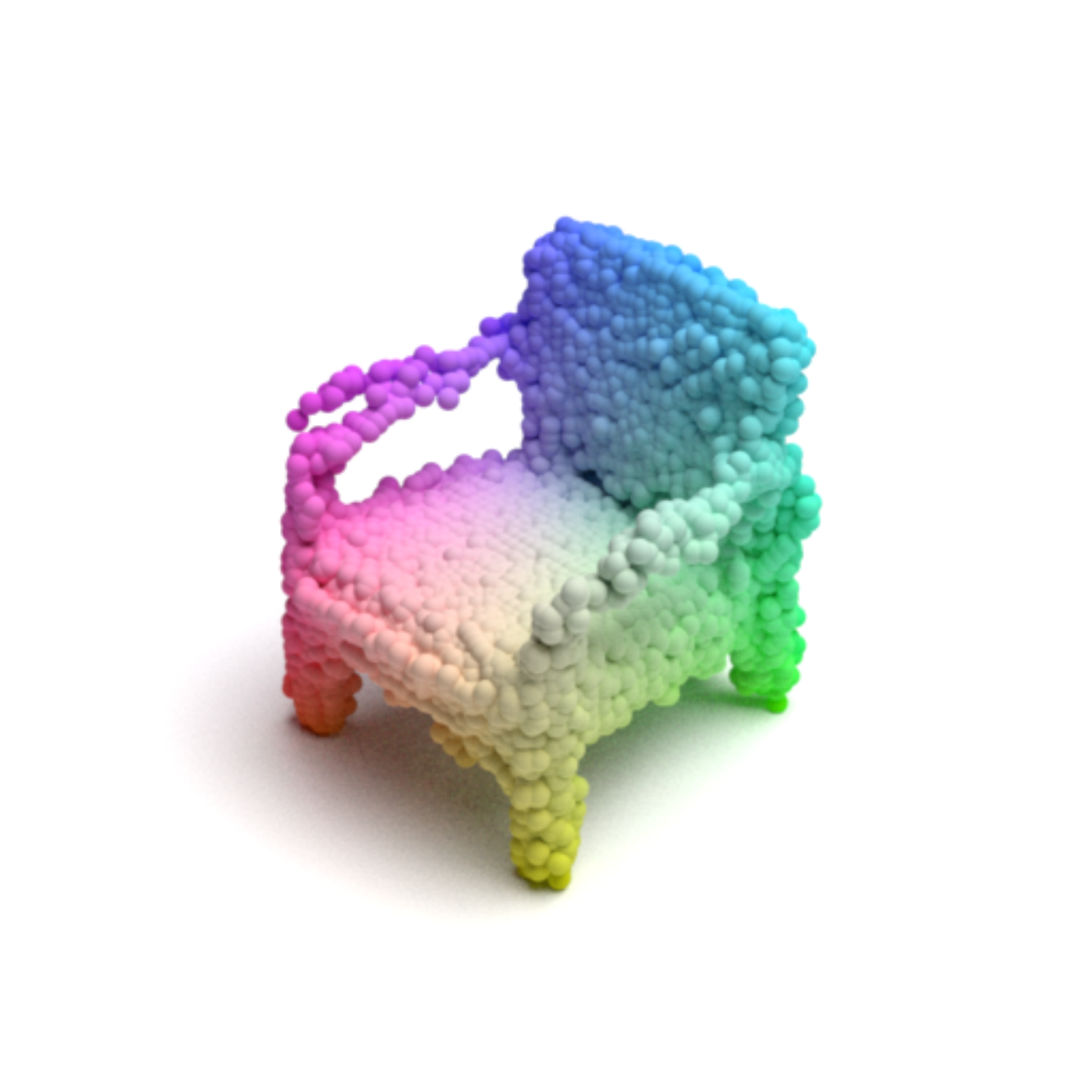}
\includegraphics[clip,trim=3cm 3cm 3cm 3cm, width=0.095\textwidth]{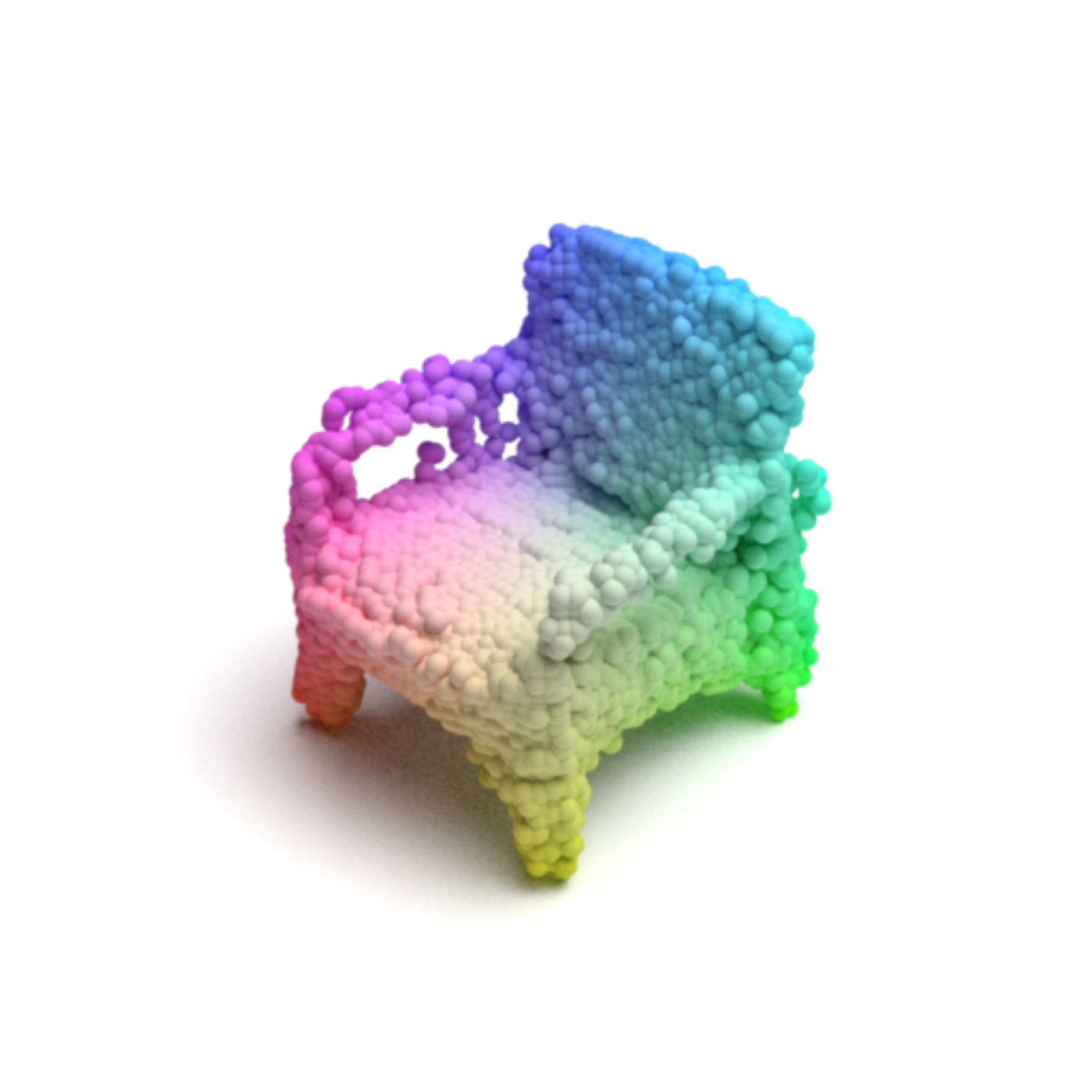}
\includegraphics[clip,trim=3cm 3cm 3cm 3cm, width=0.095\textwidth]{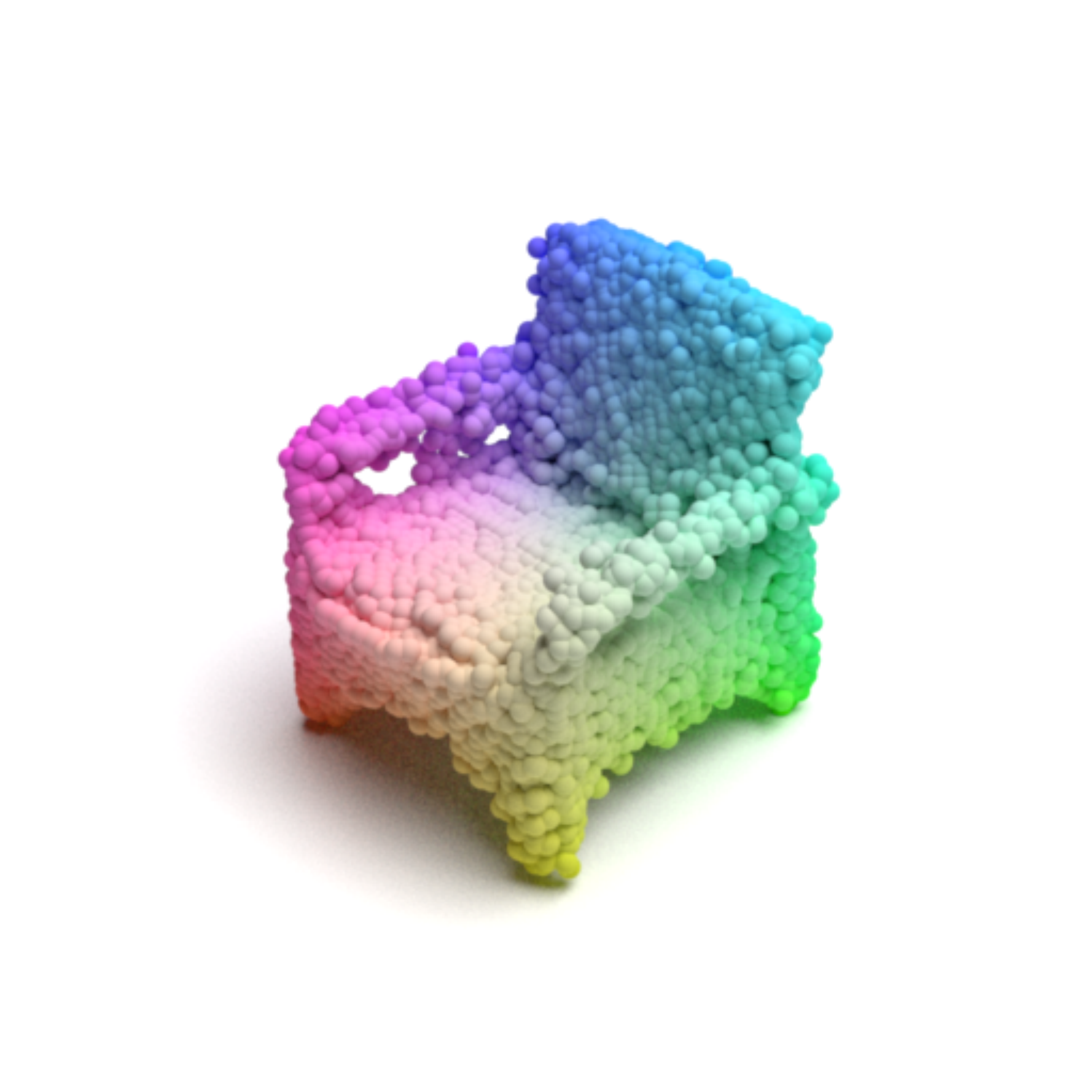}
\includegraphics[clip,trim=3cm 3cm 3cm 3cm, width=0.095\textwidth]{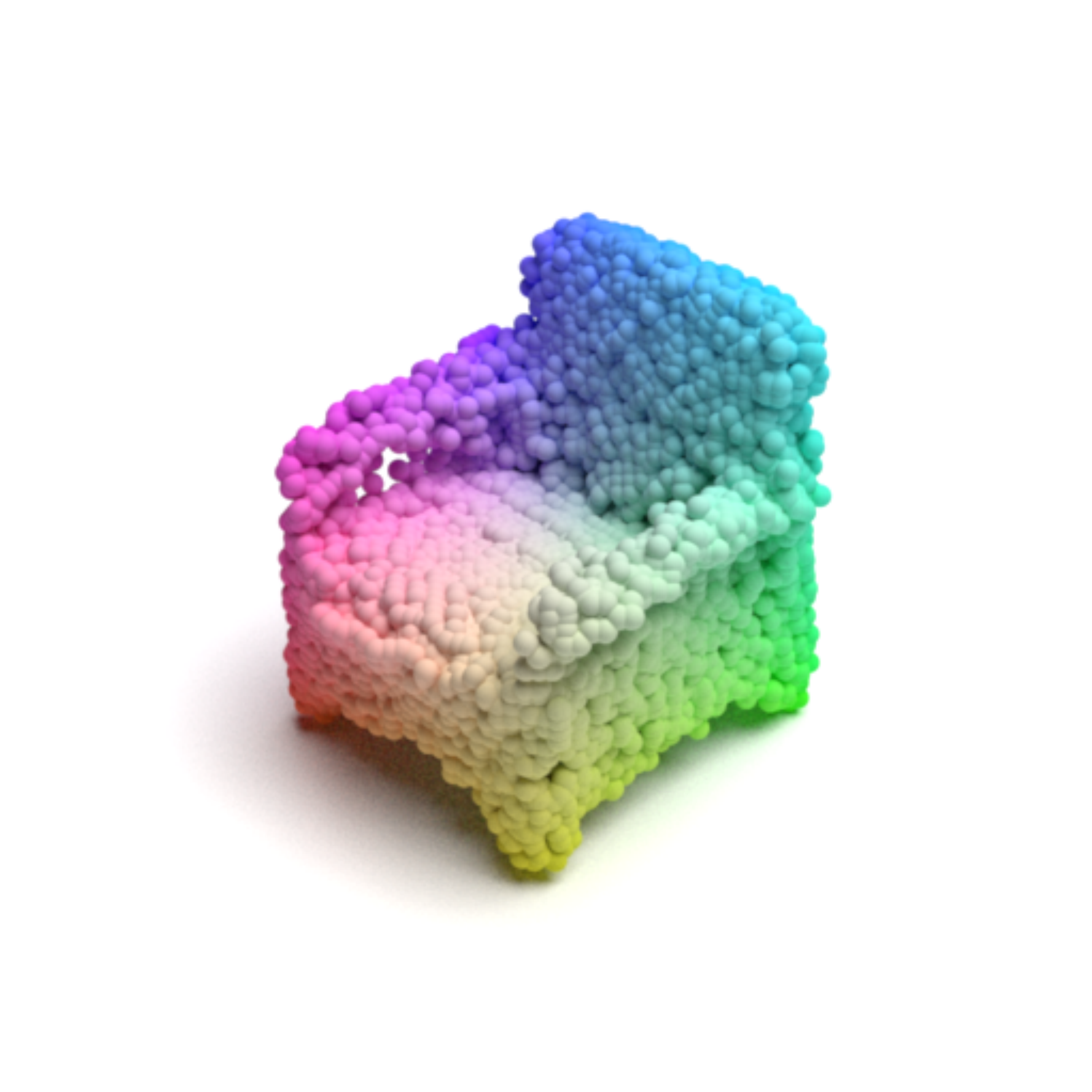}
\includegraphics[clip,trim=3cm 3cm 3cm 3cm, width=0.095\textwidth]{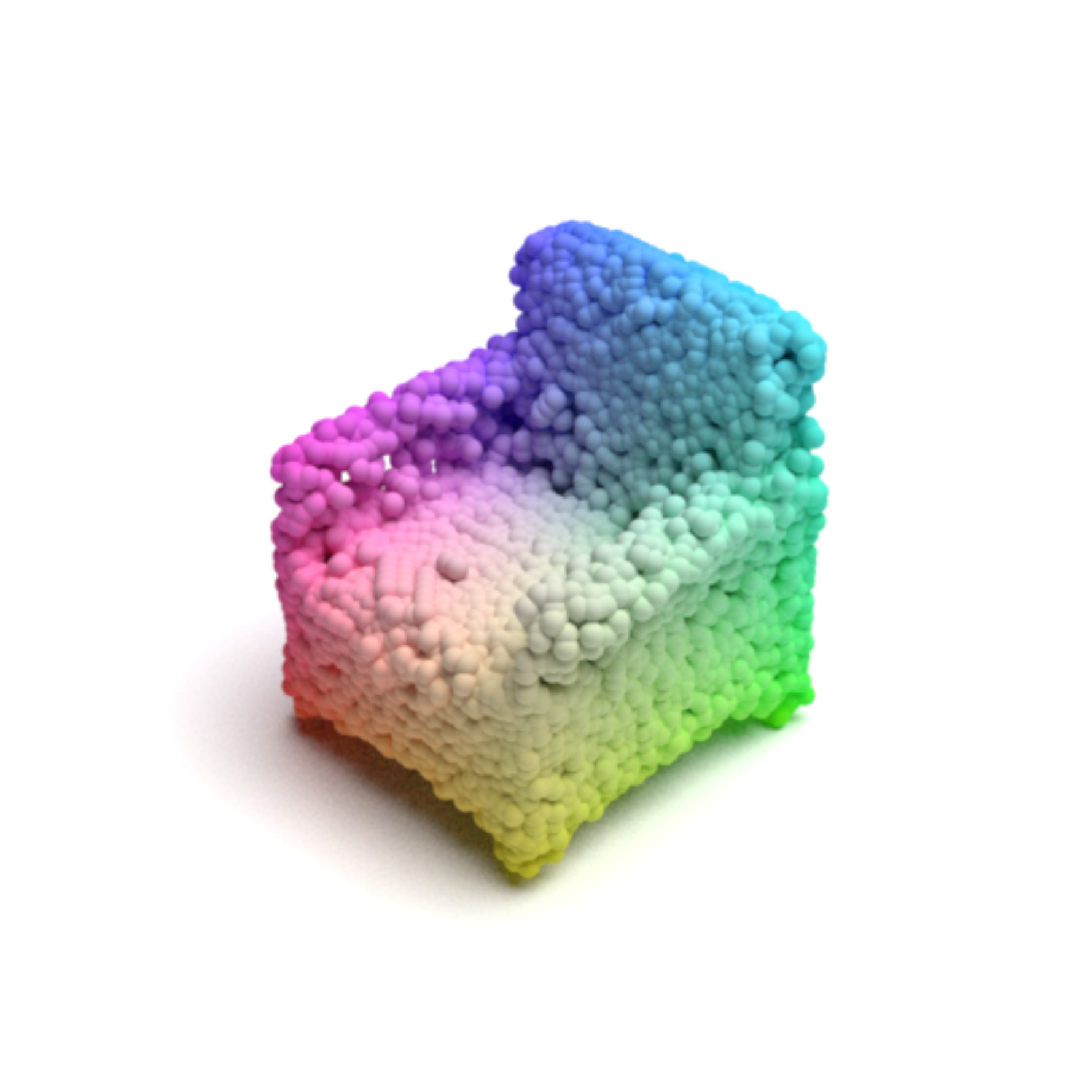}
\includegraphics[clip,trim=3cm 3cm 3cm 3cm, width=0.095\textwidth]{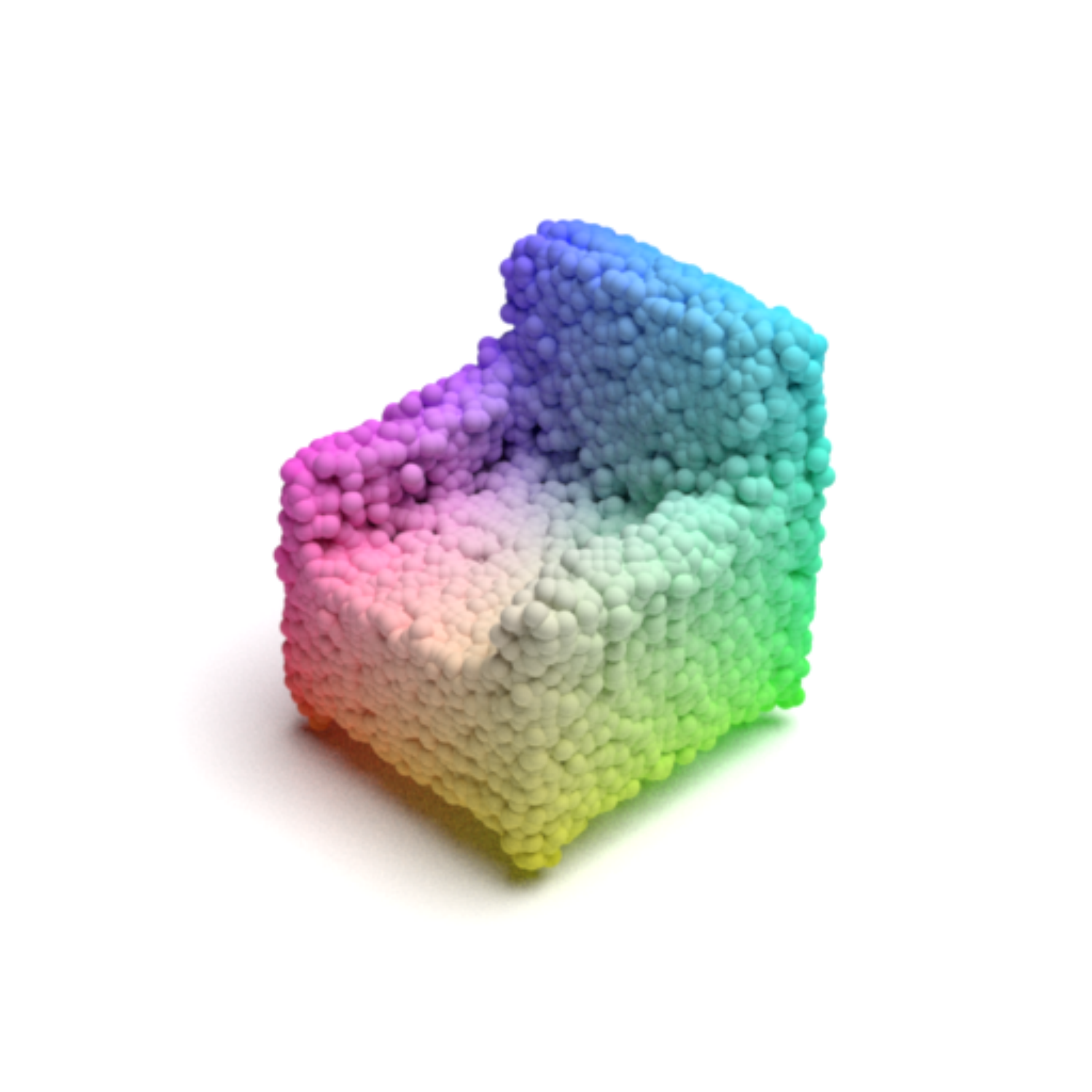}

\includegraphics[clip,trim=3cm 3cm 3cm 3cm, width=0.095\textwidth]{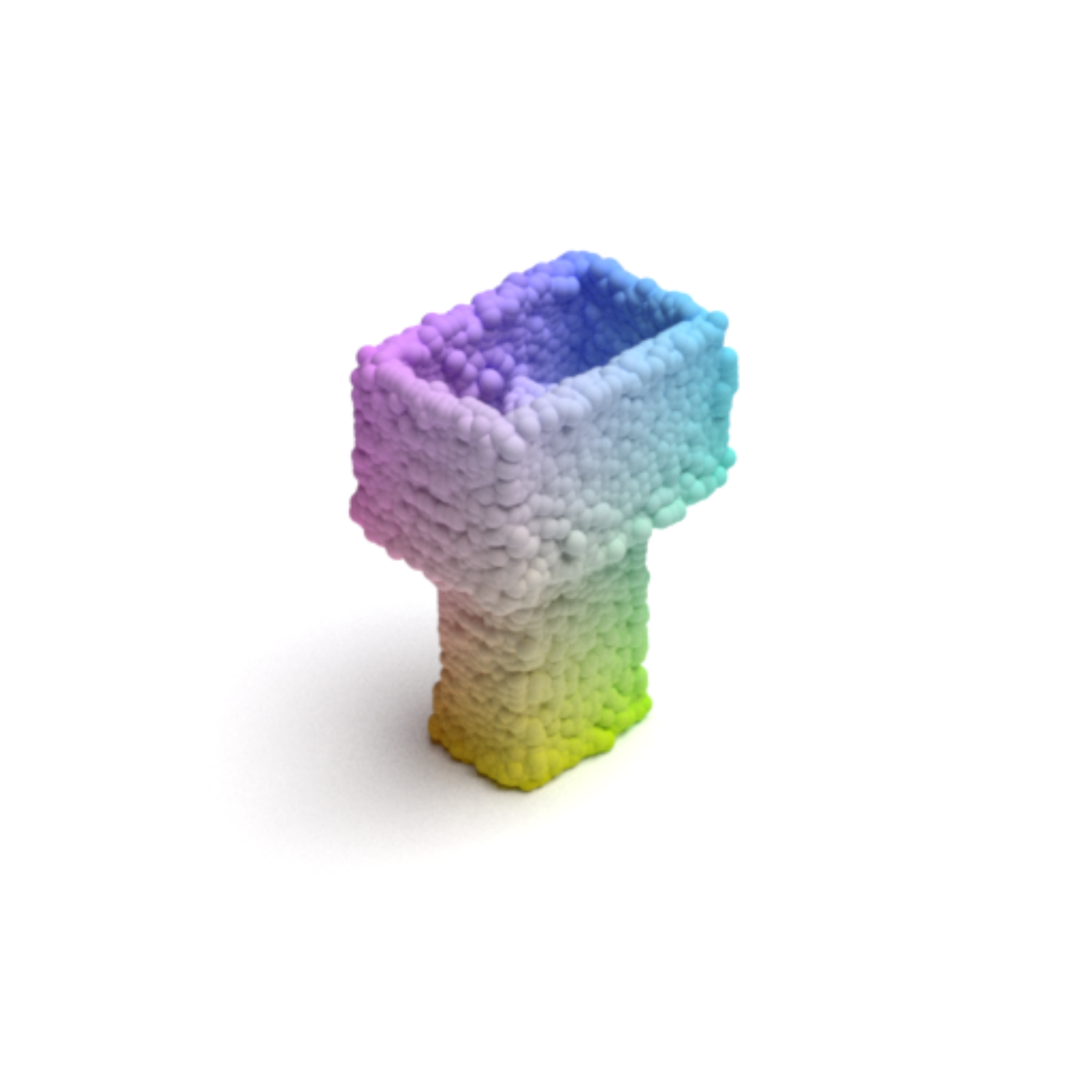}
\includegraphics[clip,trim=3cm 3cm 3cm 3cm, width=0.095\textwidth]{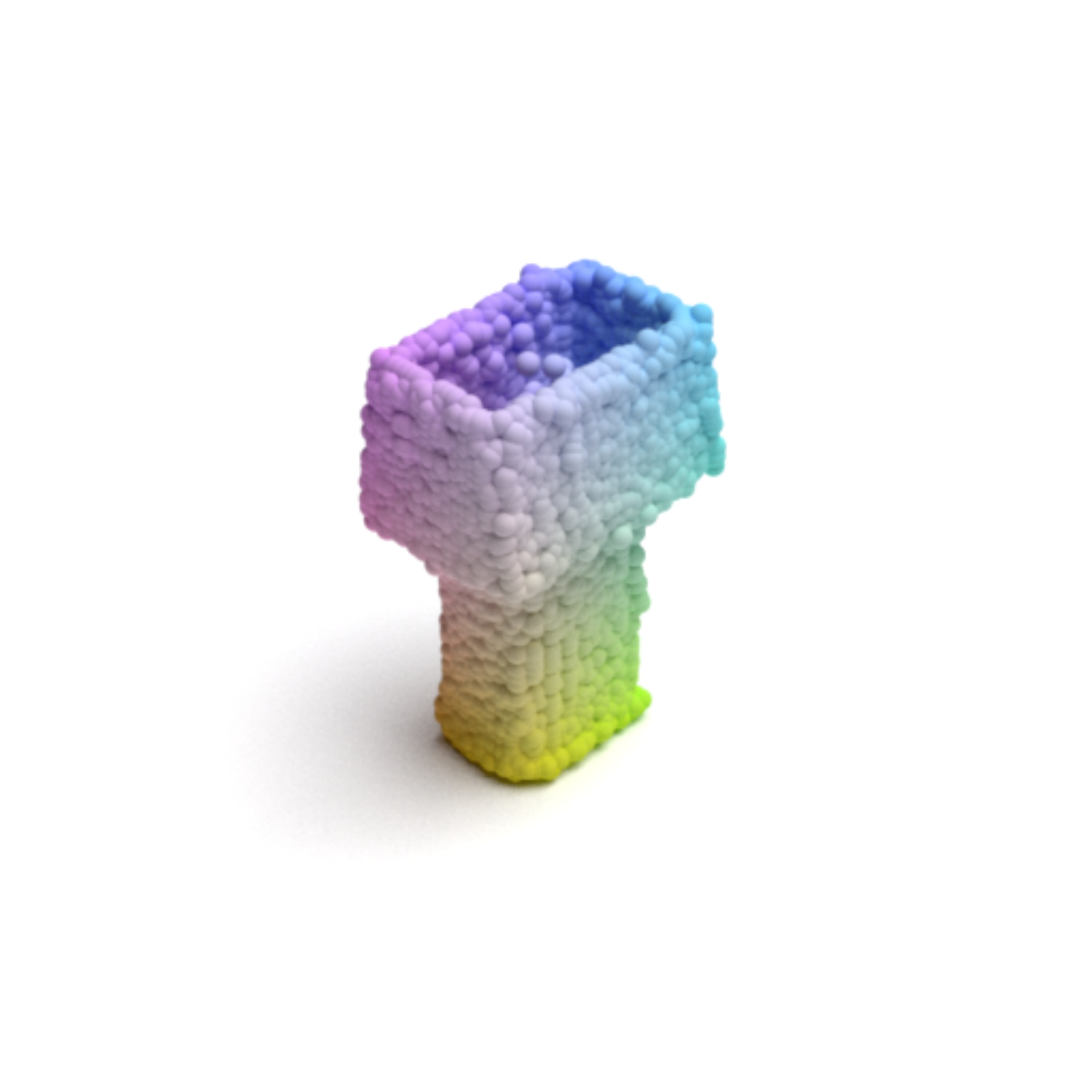}
\includegraphics[clip,trim=3cm 3cm 3cm 3cm, width=0.095\textwidth]{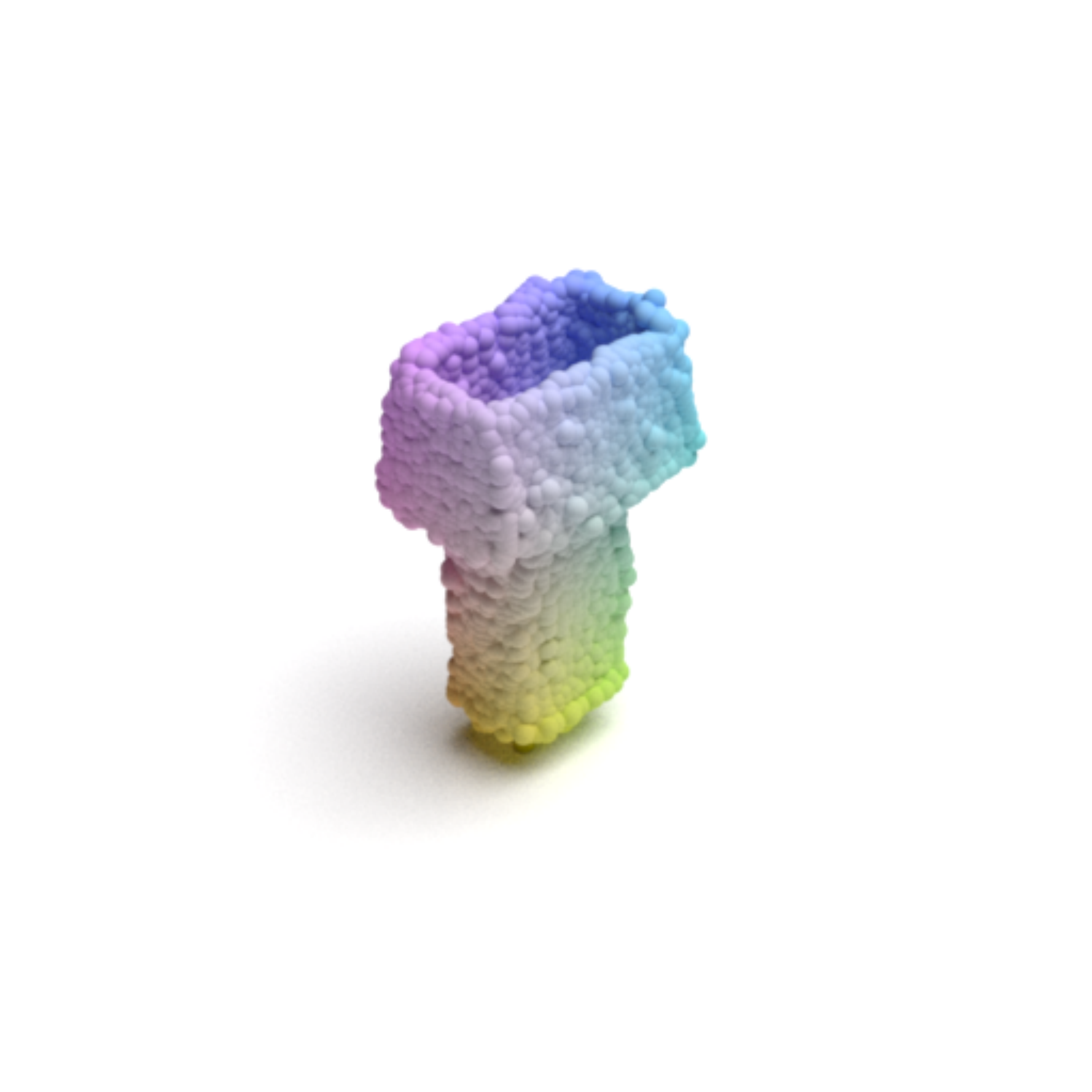}
\includegraphics[clip,trim=3cm 3cm 3cm 3cm, width=0.095\textwidth]{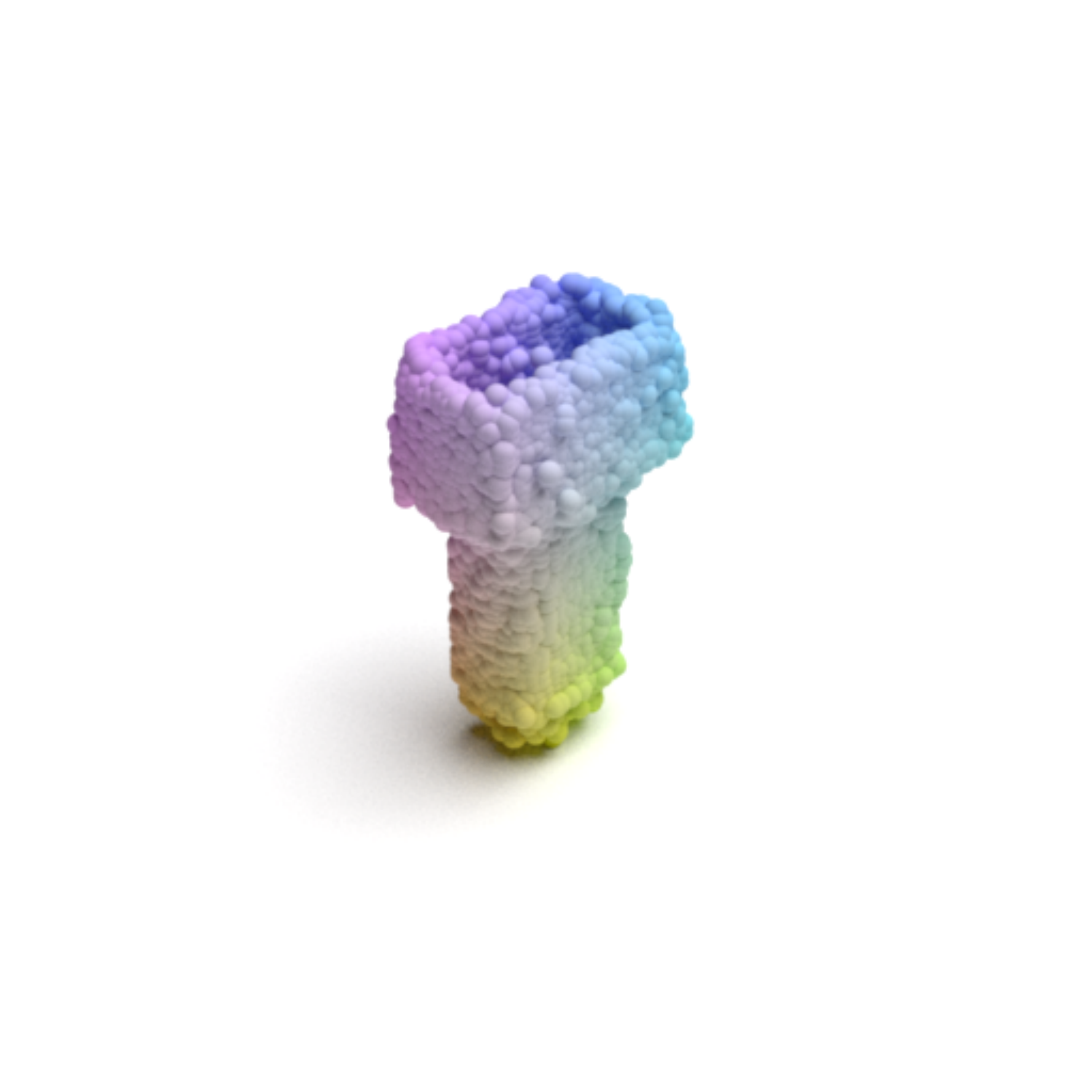}
\includegraphics[clip,trim=3cm 3cm 3cm 3cm, width=0.095\textwidth]{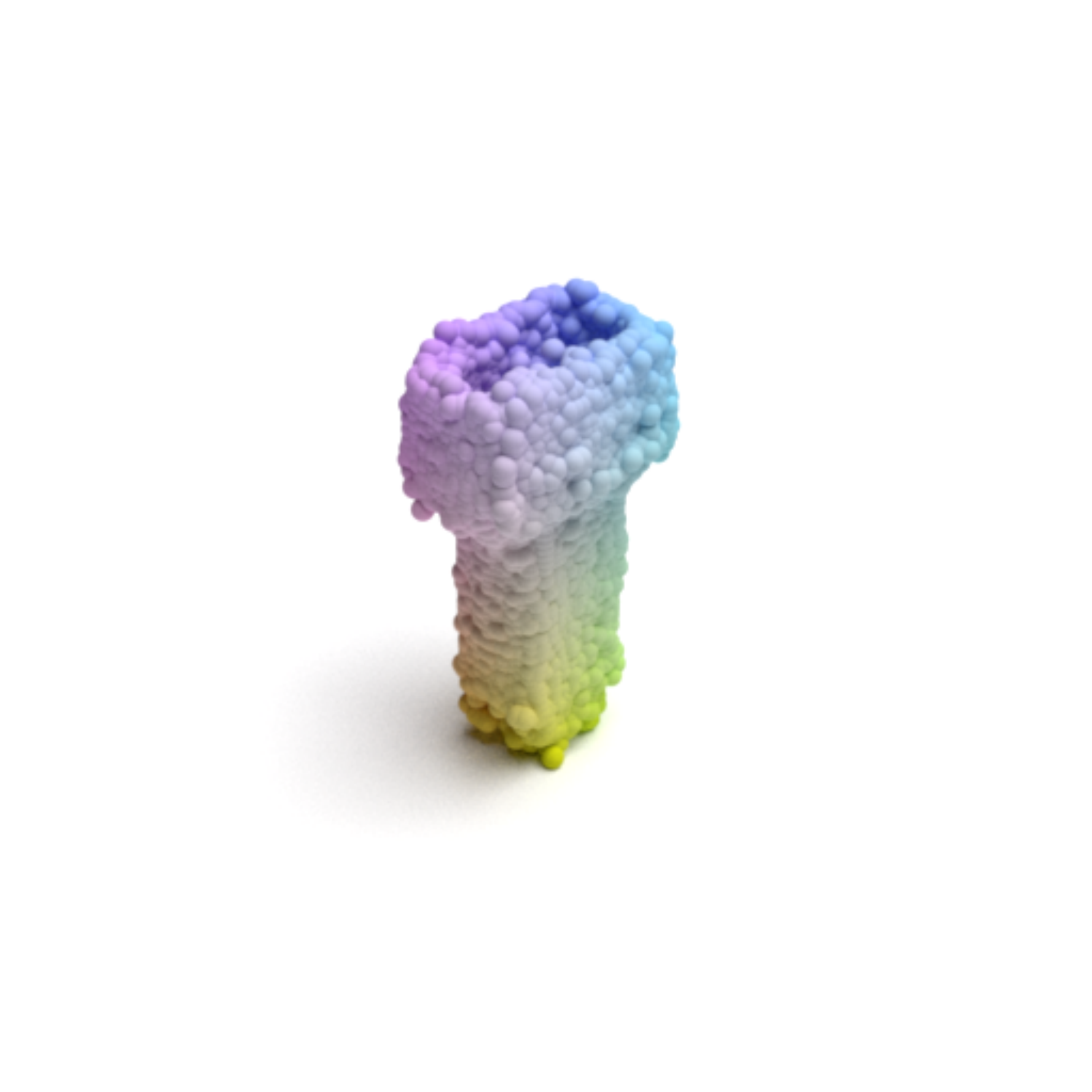}
\includegraphics[clip,trim=3cm 3cm 3cm 3cm, width=0.095\textwidth]{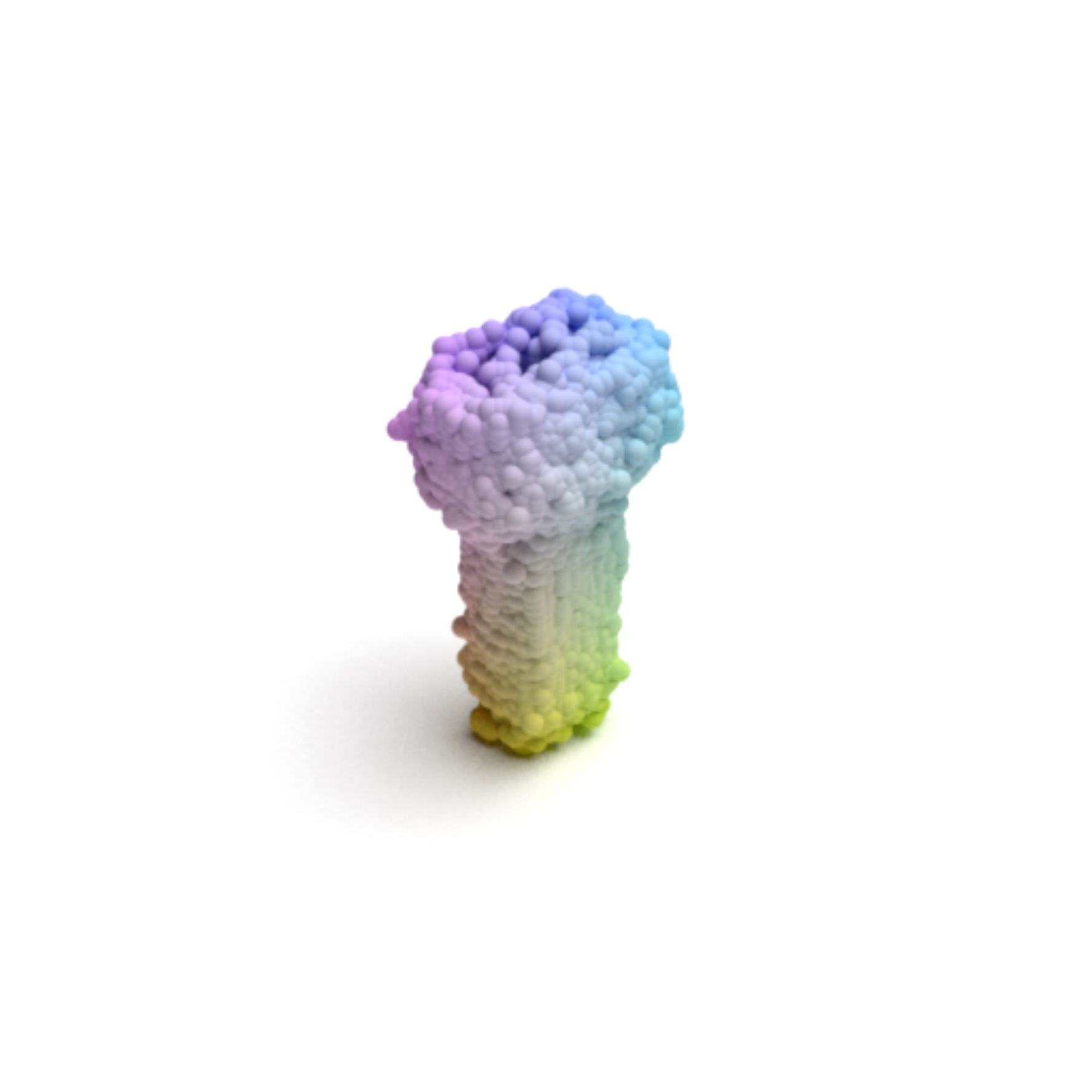}
\includegraphics[clip,trim=3cm 3cm 3cm 3cm, width=0.095\textwidth]{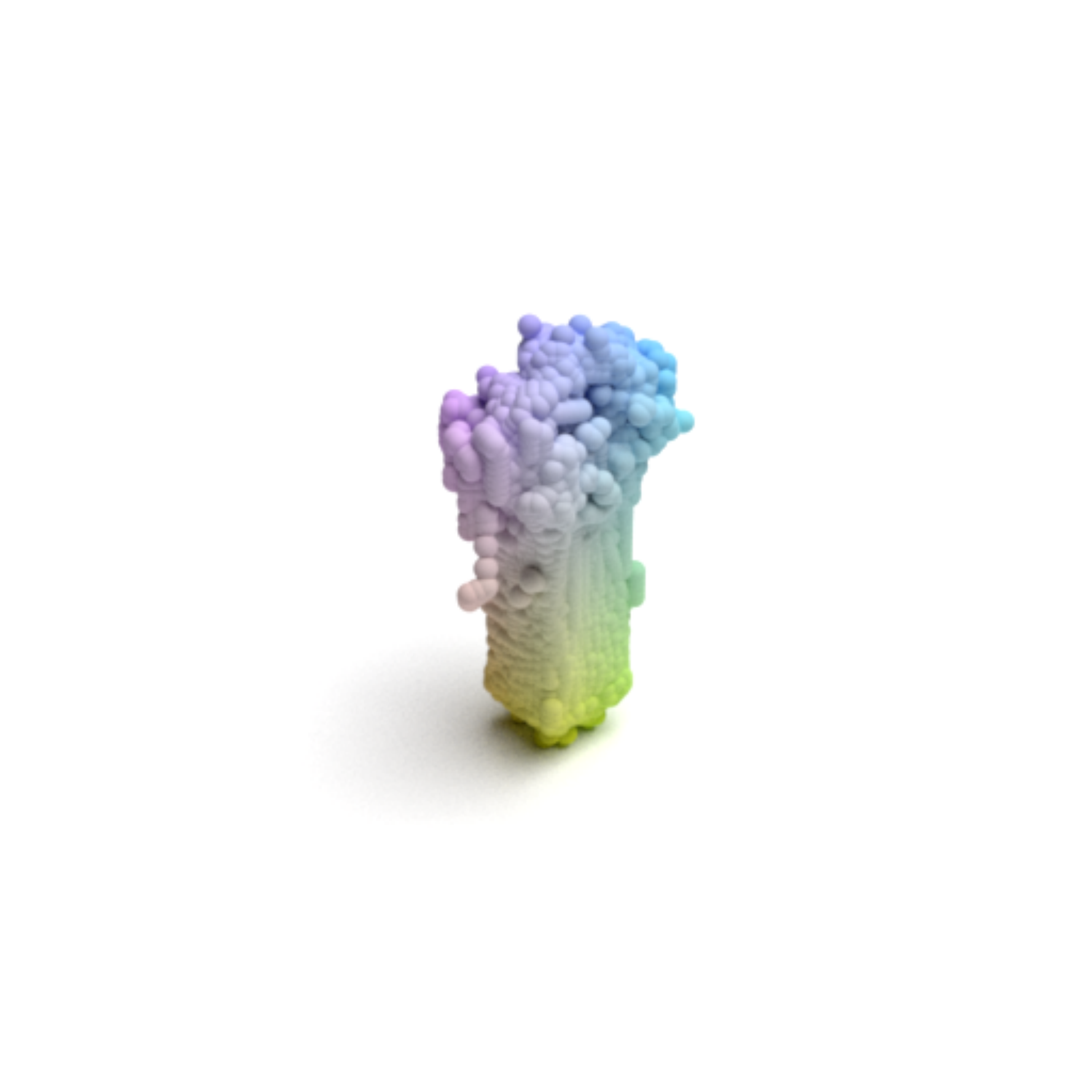}
\includegraphics[clip,trim=3cm 3cm 3cm 3cm, width=0.095\textwidth]{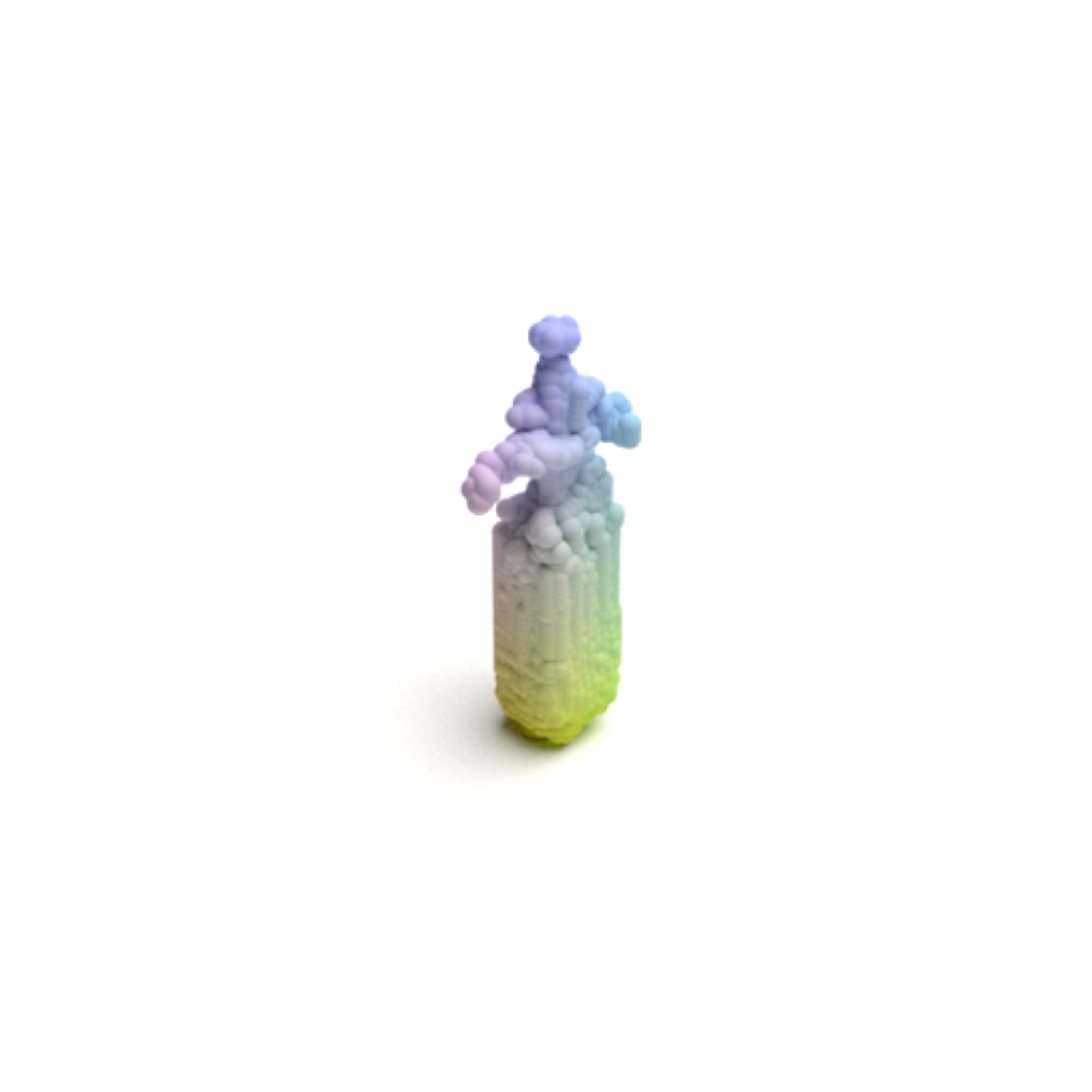}
\includegraphics[clip,trim=3cm 3cm 3cm 3cm, width=0.095\textwidth]{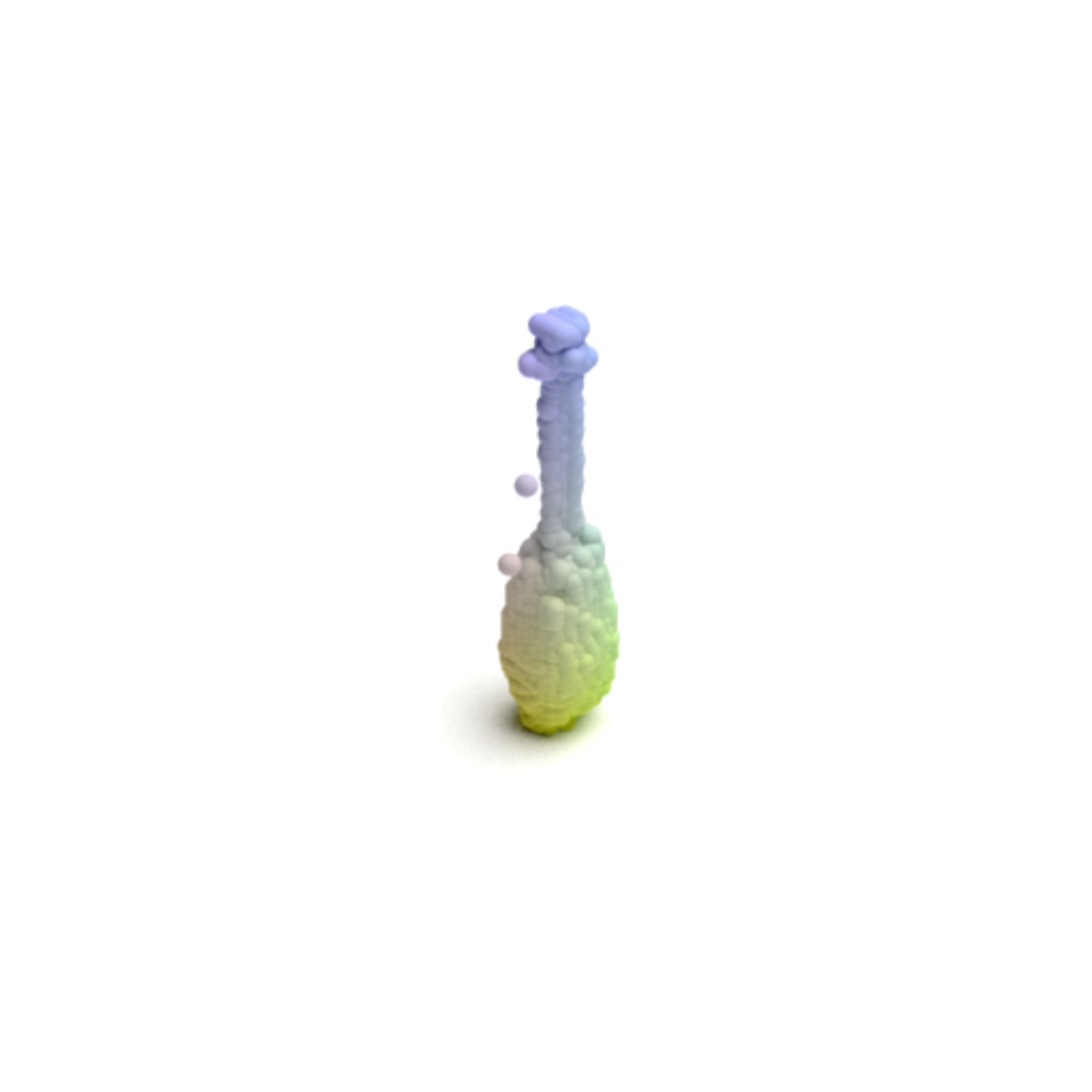}
\includegraphics[clip,trim=3cm 3cm 3cm 3cm, width=0.095\textwidth]{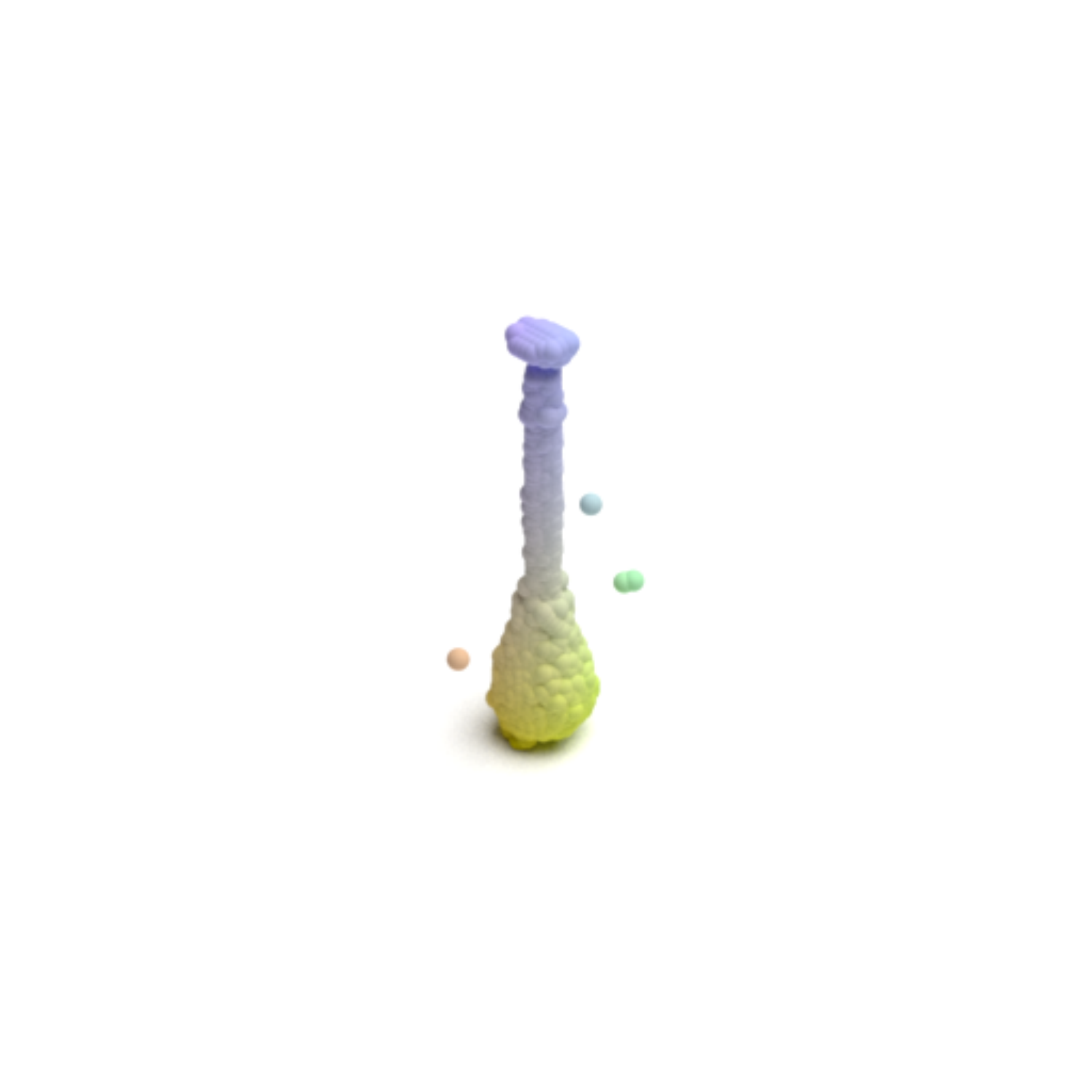}

\includegraphics[clip,trim=3cm 3cm 3cm 3cm, width=0.095\textwidth]{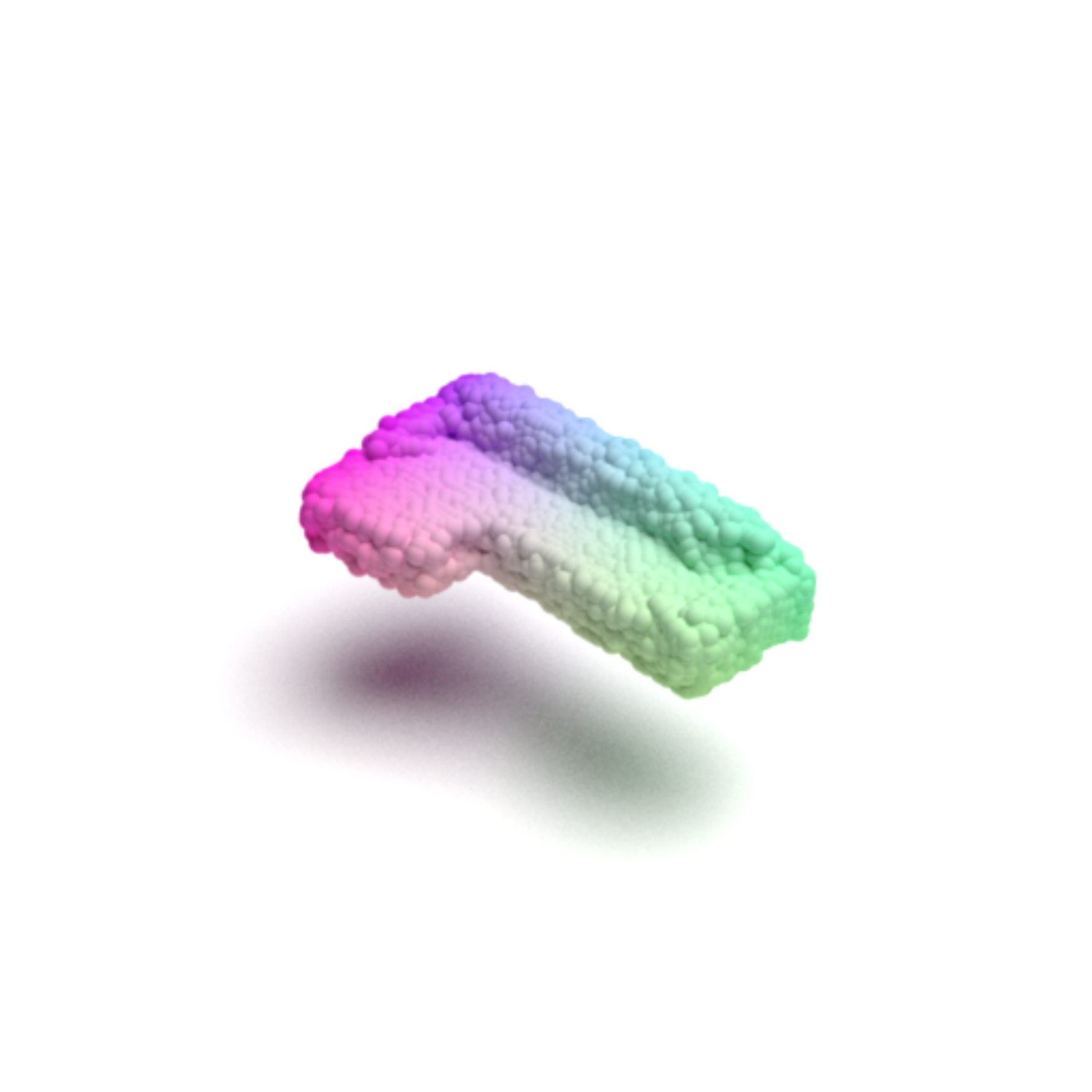}
\includegraphics[clip,trim=3cm 3cm 3cm 3cm, width=0.095\textwidth]{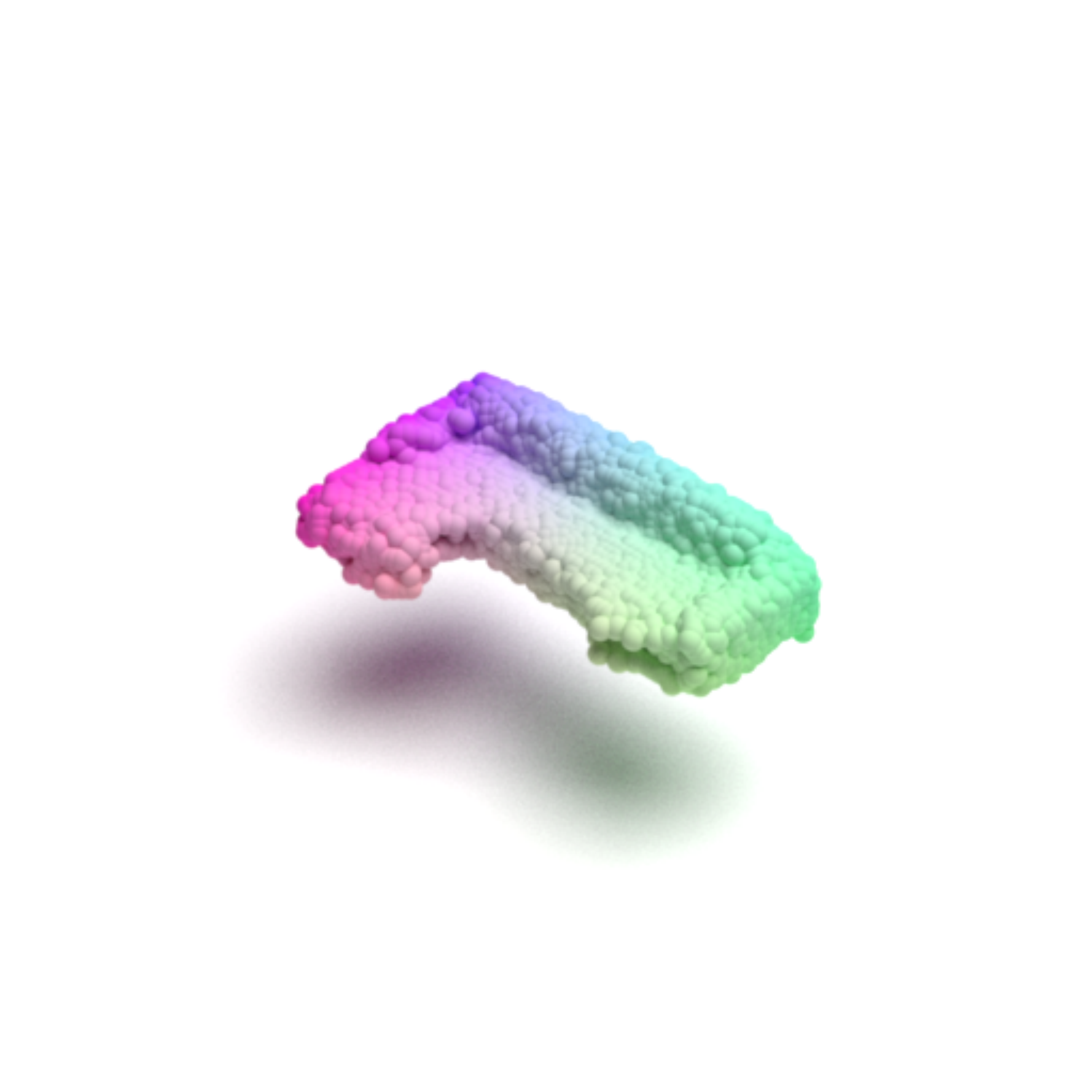}
\includegraphics[clip,trim=3cm 3cm 3cm 3cm, width=0.095\textwidth]{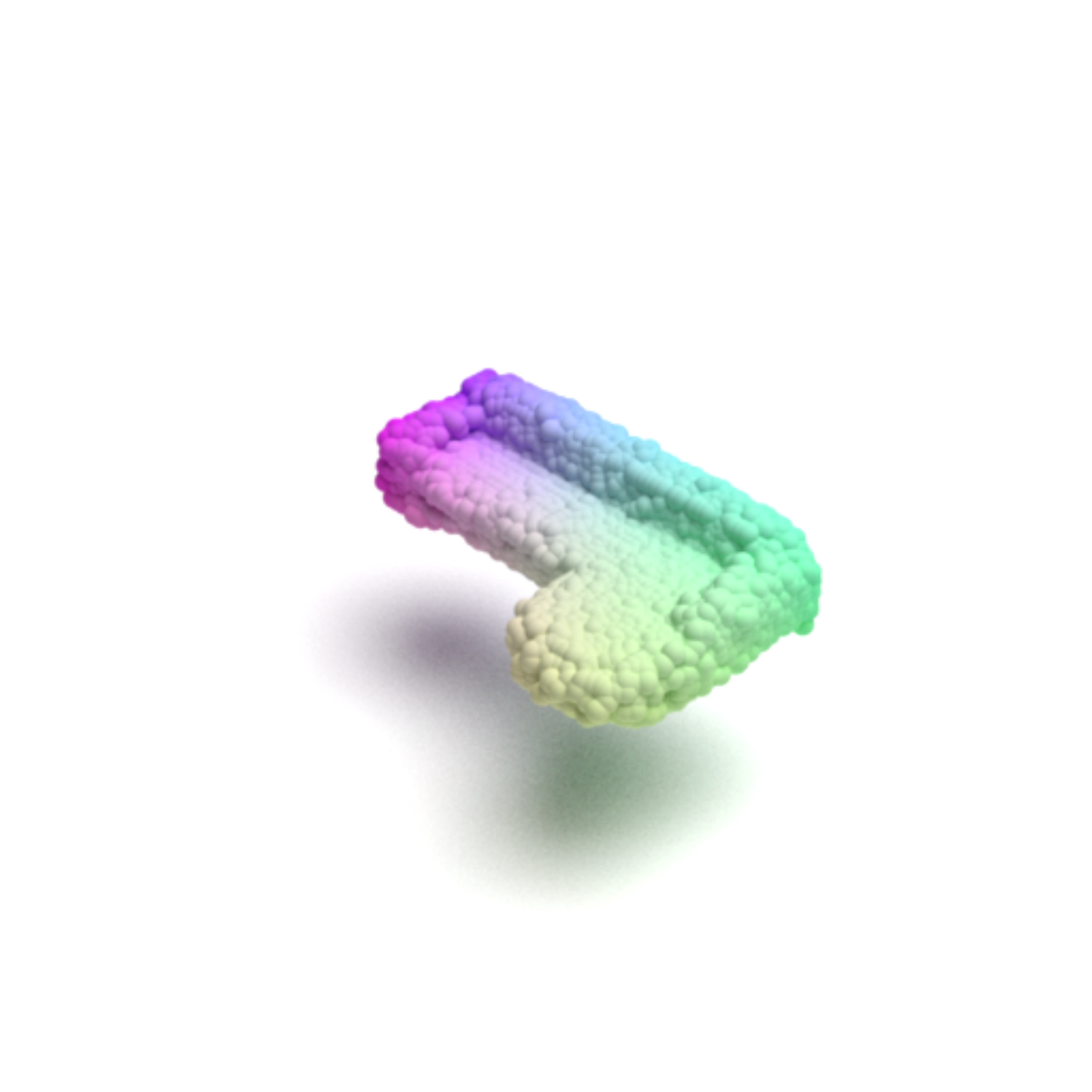}
\includegraphics[clip,trim=3cm 3cm 3cm 3cm, width=0.095\textwidth]{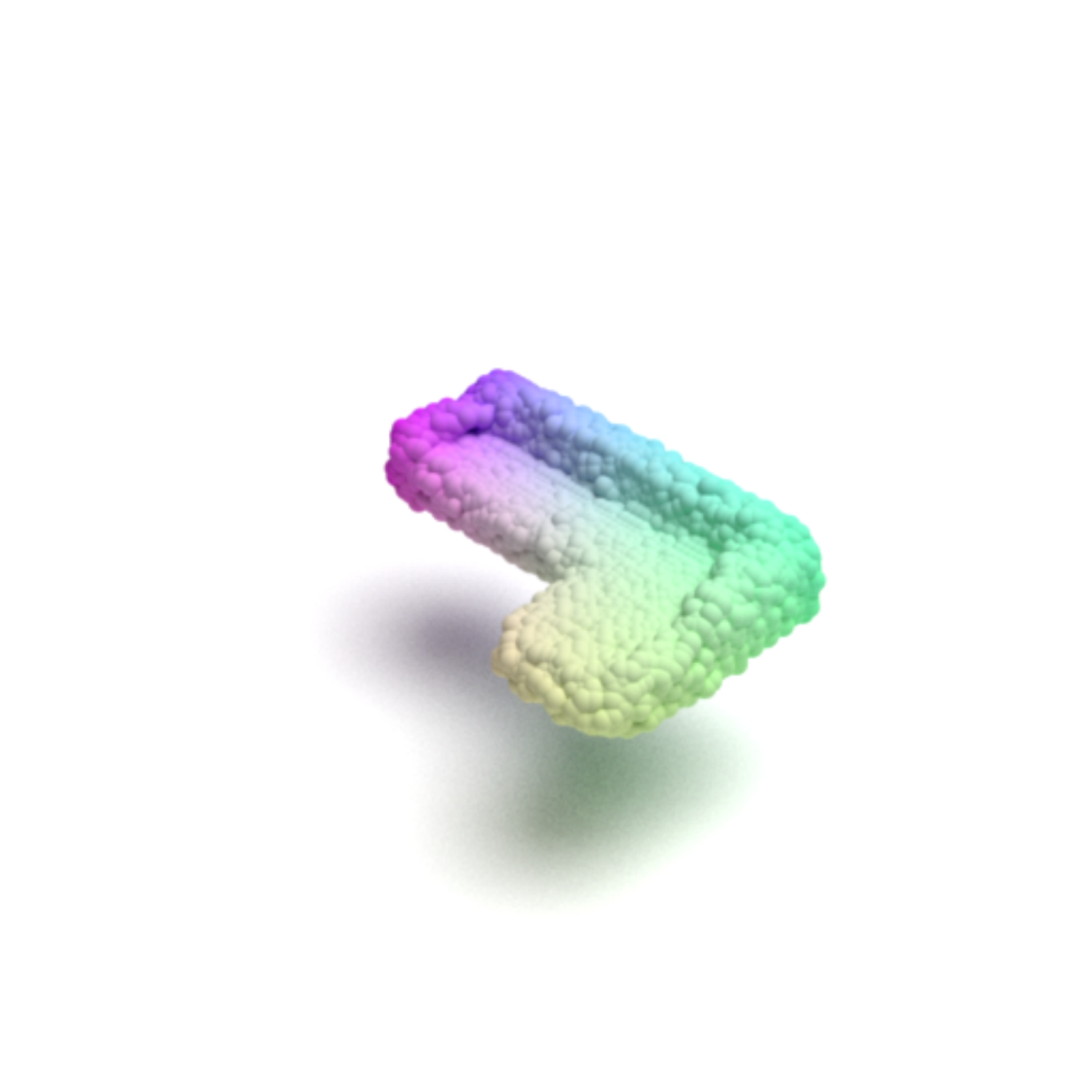}
\includegraphics[clip,trim=3cm 3cm 3cm 3cm, width=0.095\textwidth]{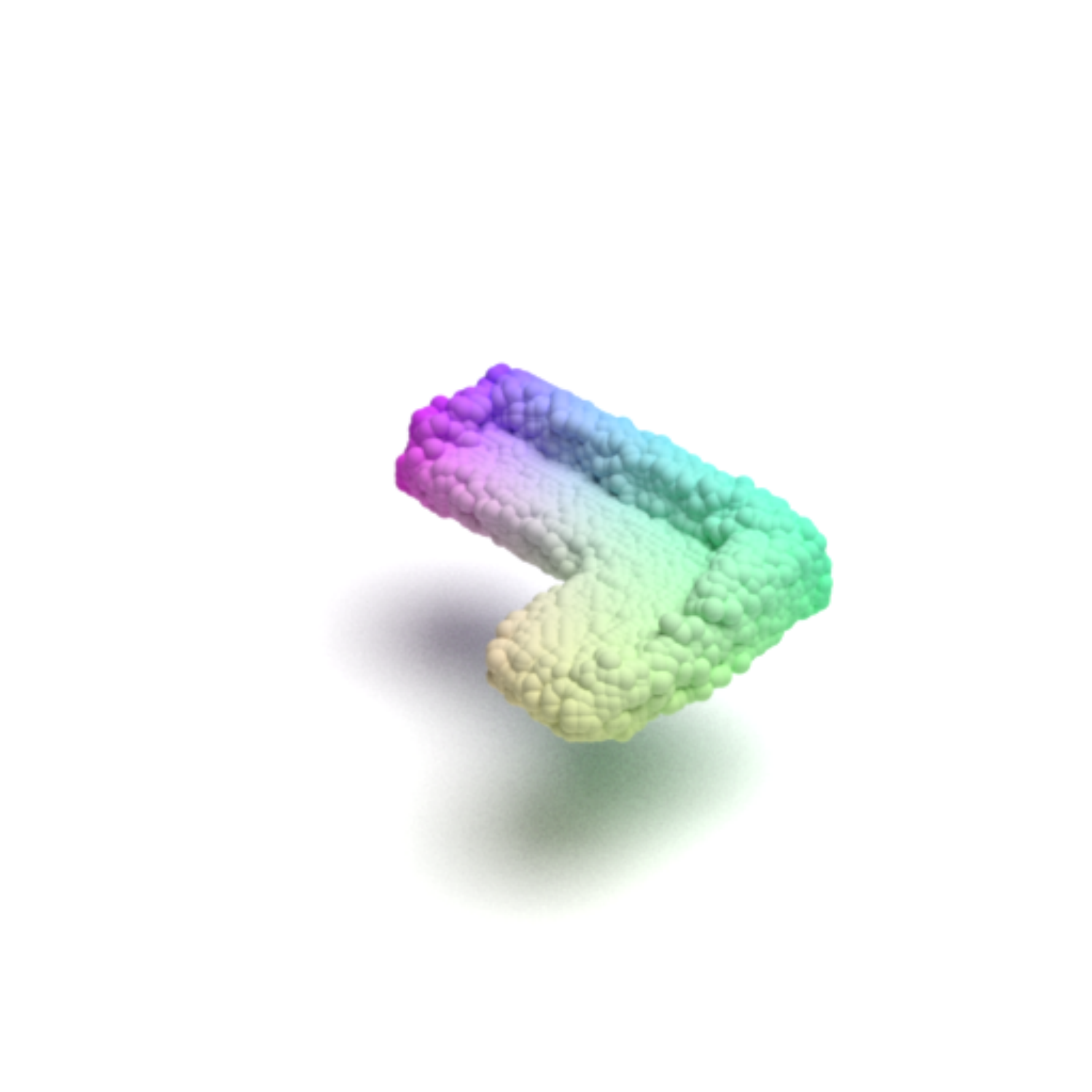}
\includegraphics[clip,trim=3cm 3cm 3cm 3cm, width=0.095\textwidth]{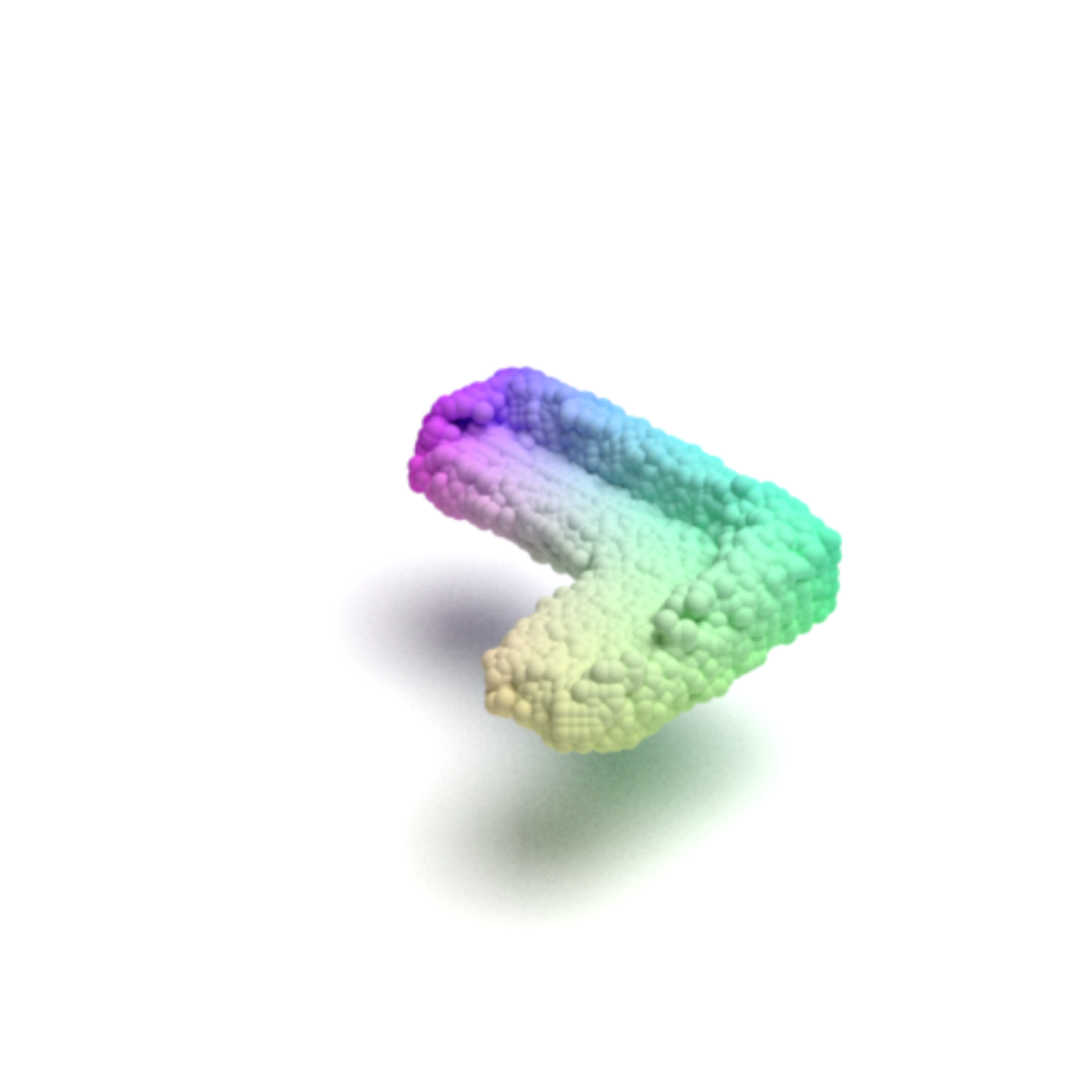}
\includegraphics[clip,trim=3cm 3cm 3cm 3cm, width=0.095\textwidth]{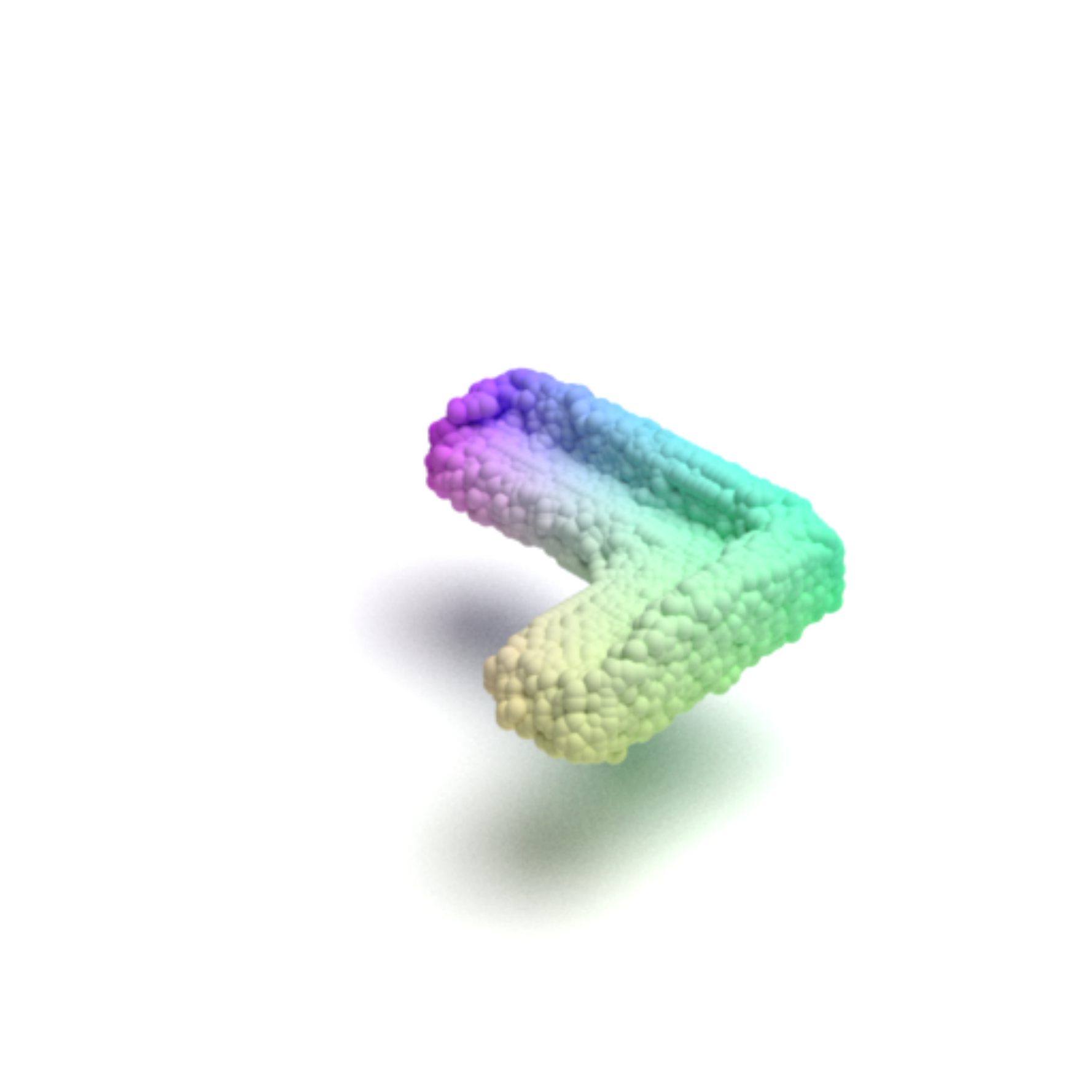}
\includegraphics[clip,trim=3cm 3cm 3cm 3cm, width=0.095\textwidth]{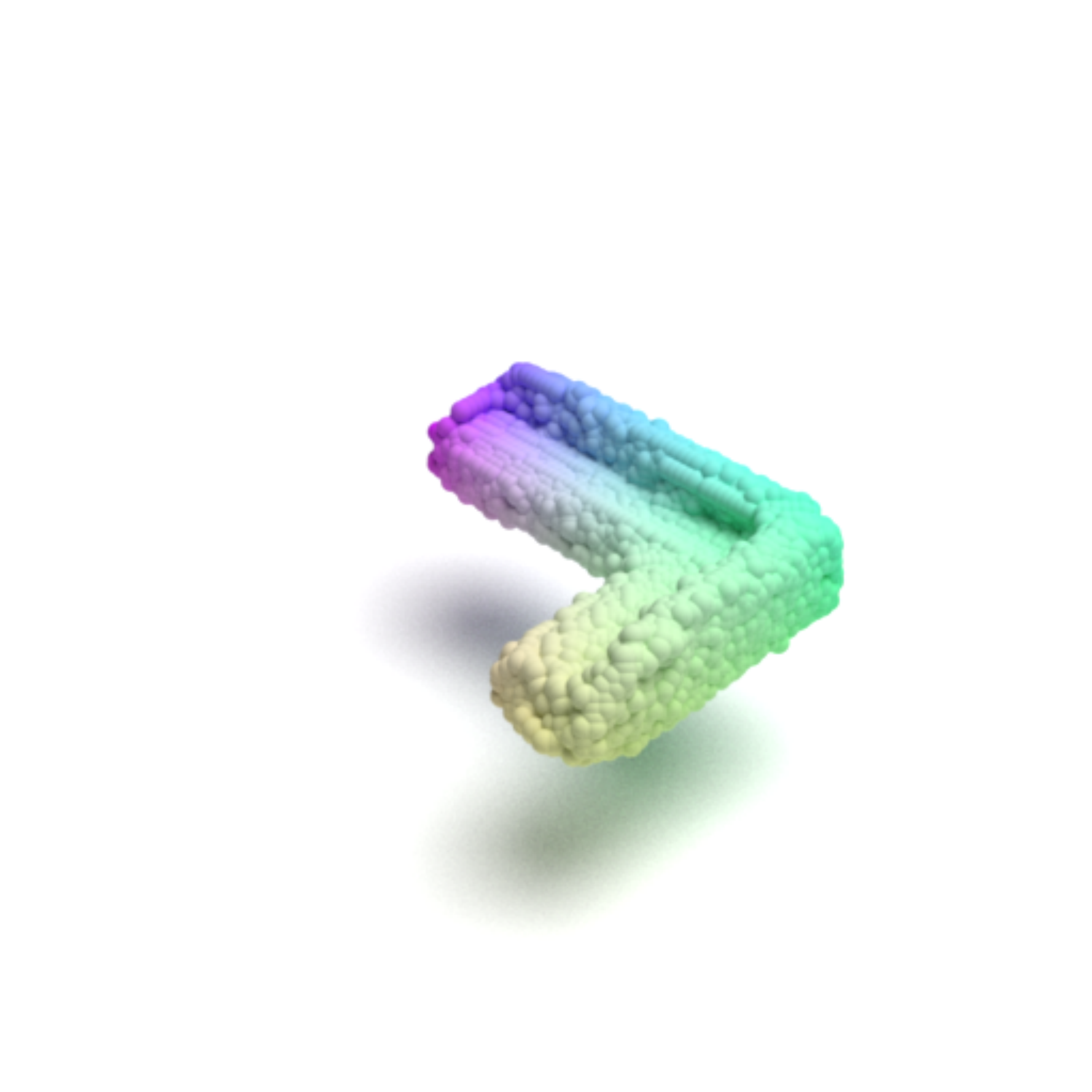}
\includegraphics[clip,trim=3cm 3cm 3cm 3cm, width=0.095\textwidth]{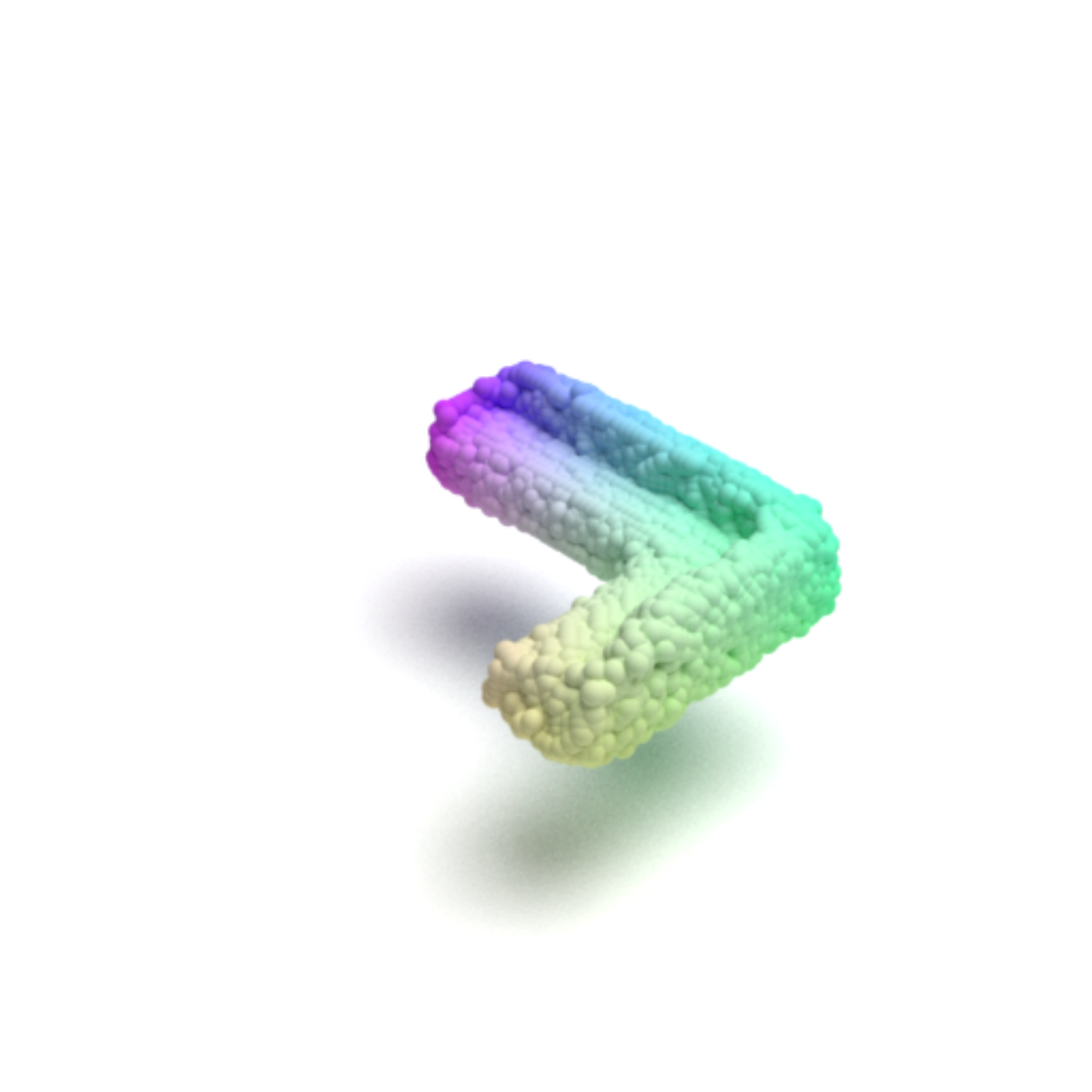}
\includegraphics[clip,trim=3cm 3cm 3cm 3cm, width=0.095\textwidth]{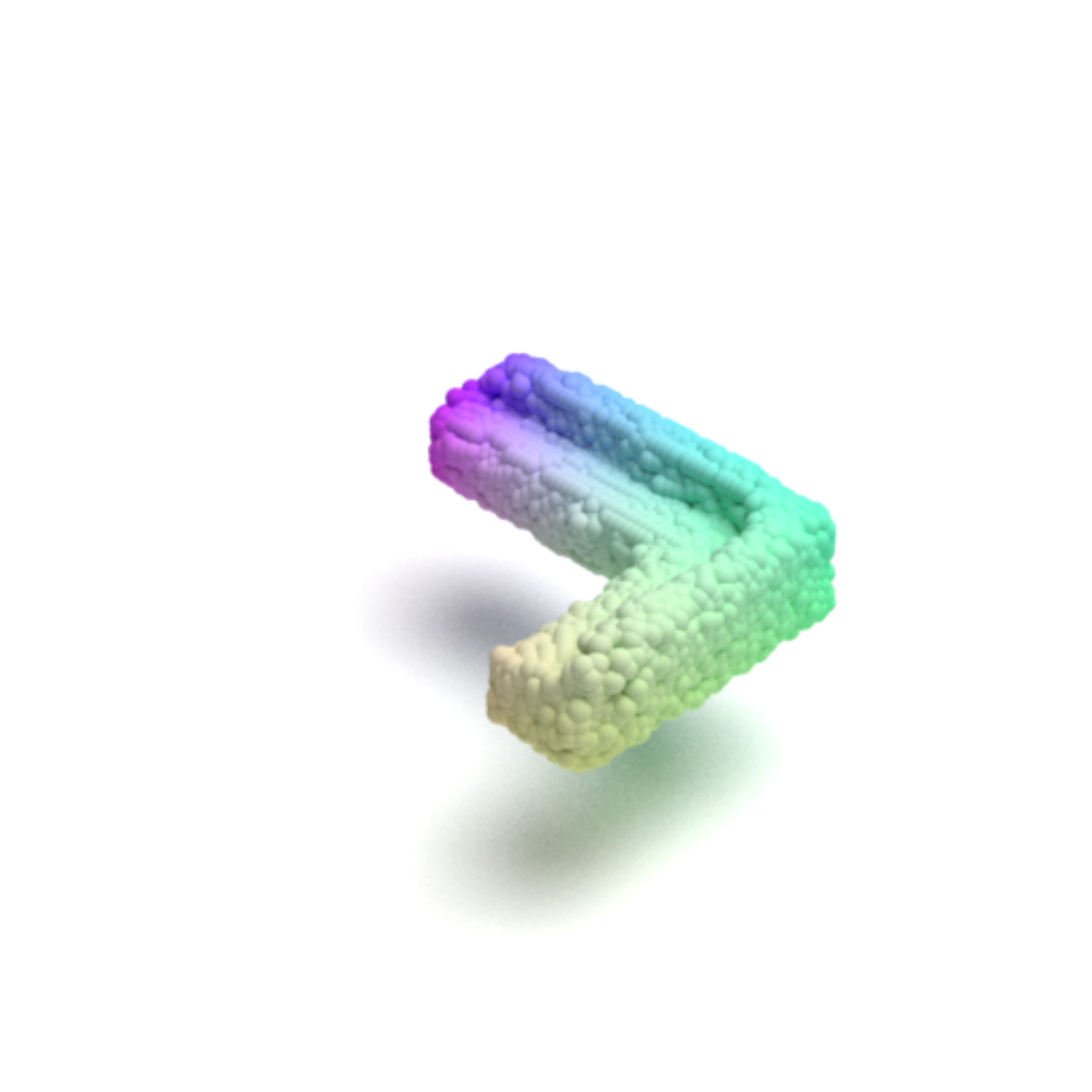}

\includegraphics[clip,trim=3cm 3cm 3cm 3cm, width=0.095\textwidth]{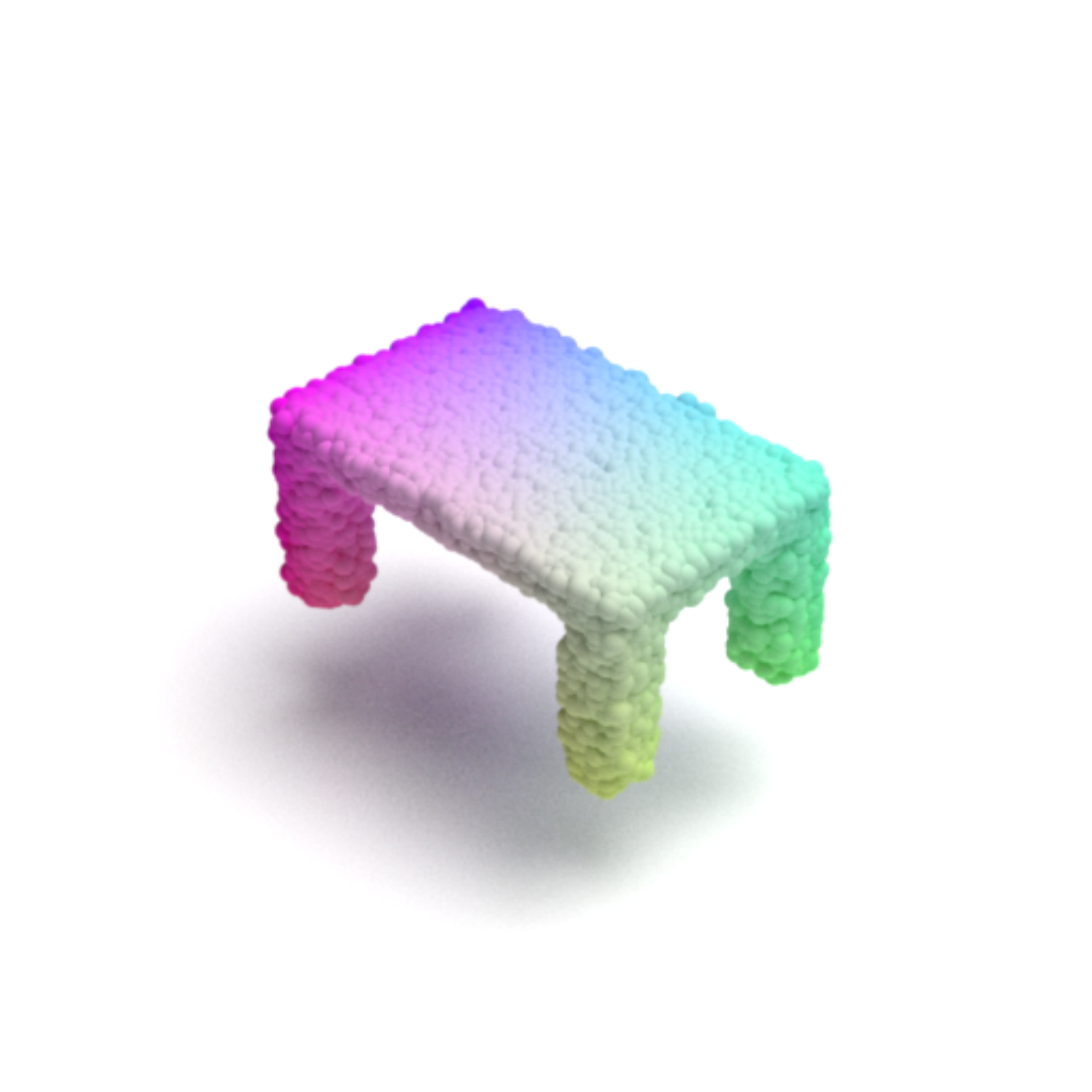}
\includegraphics[clip,trim=3cm 3cm 3cm 3cm, width=0.095\textwidth]{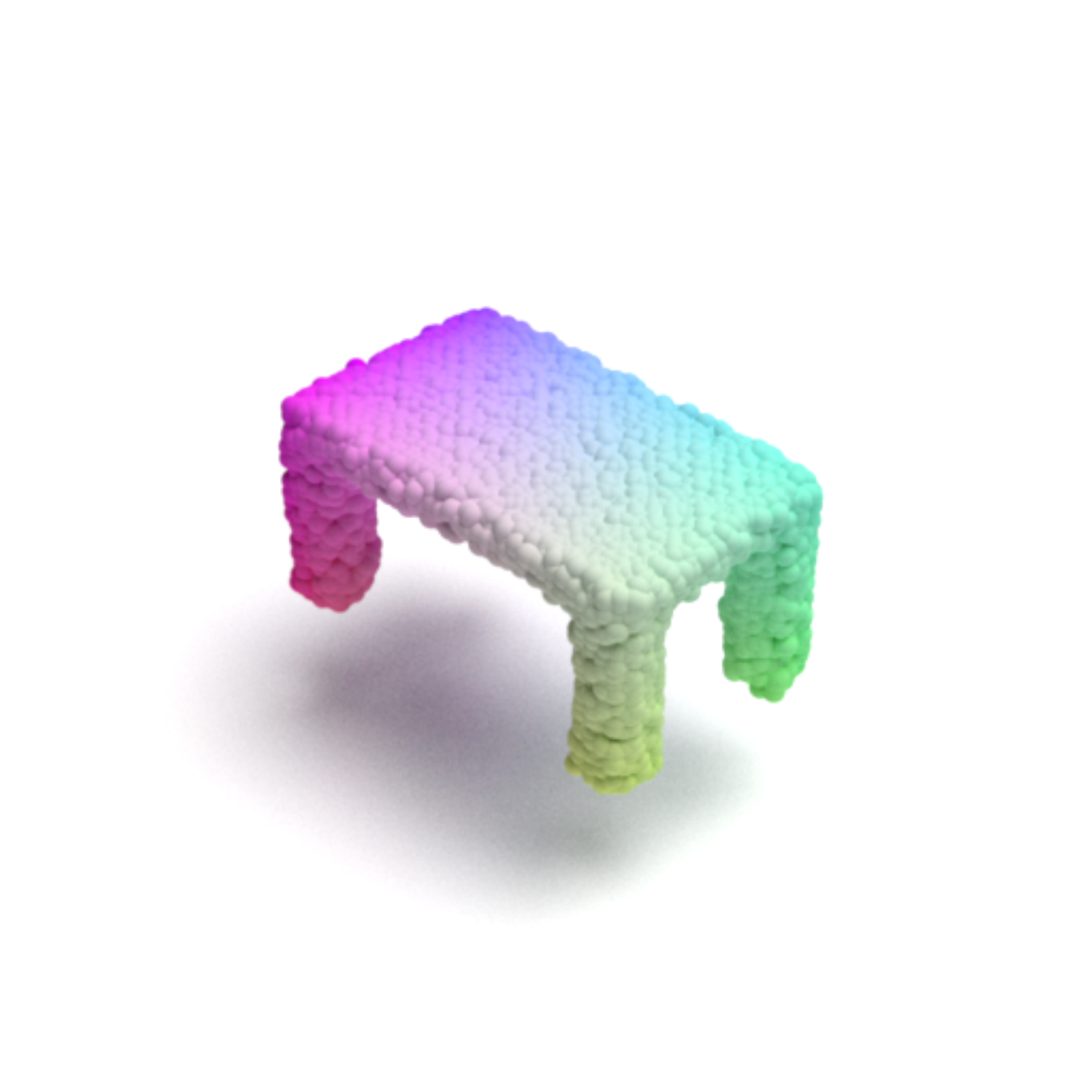}
\includegraphics[clip,trim=3cm 3cm 3cm 3cm, width=0.095\textwidth]{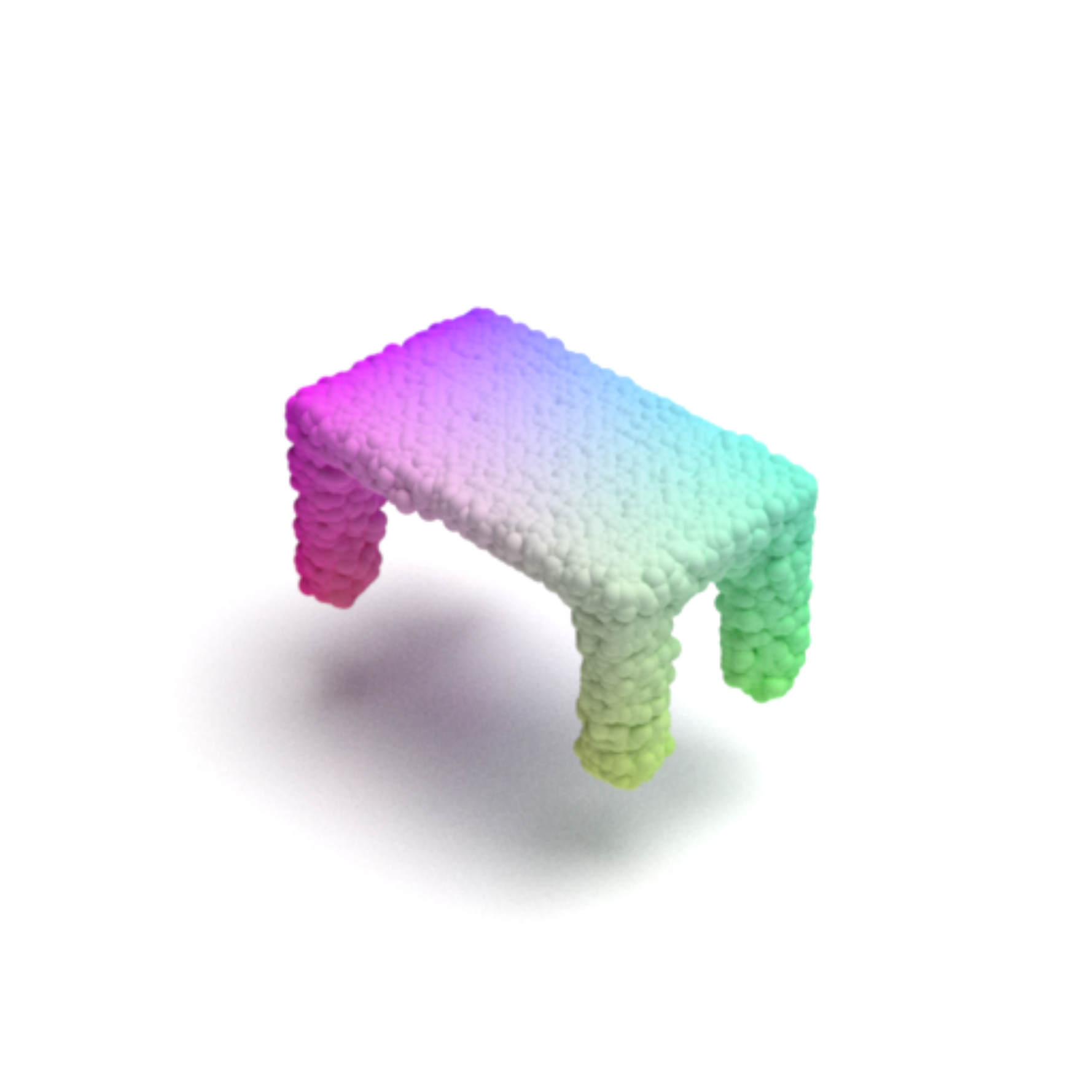}
\includegraphics[clip,trim=3cm 3cm 3cm 3cm, width=0.095\textwidth]{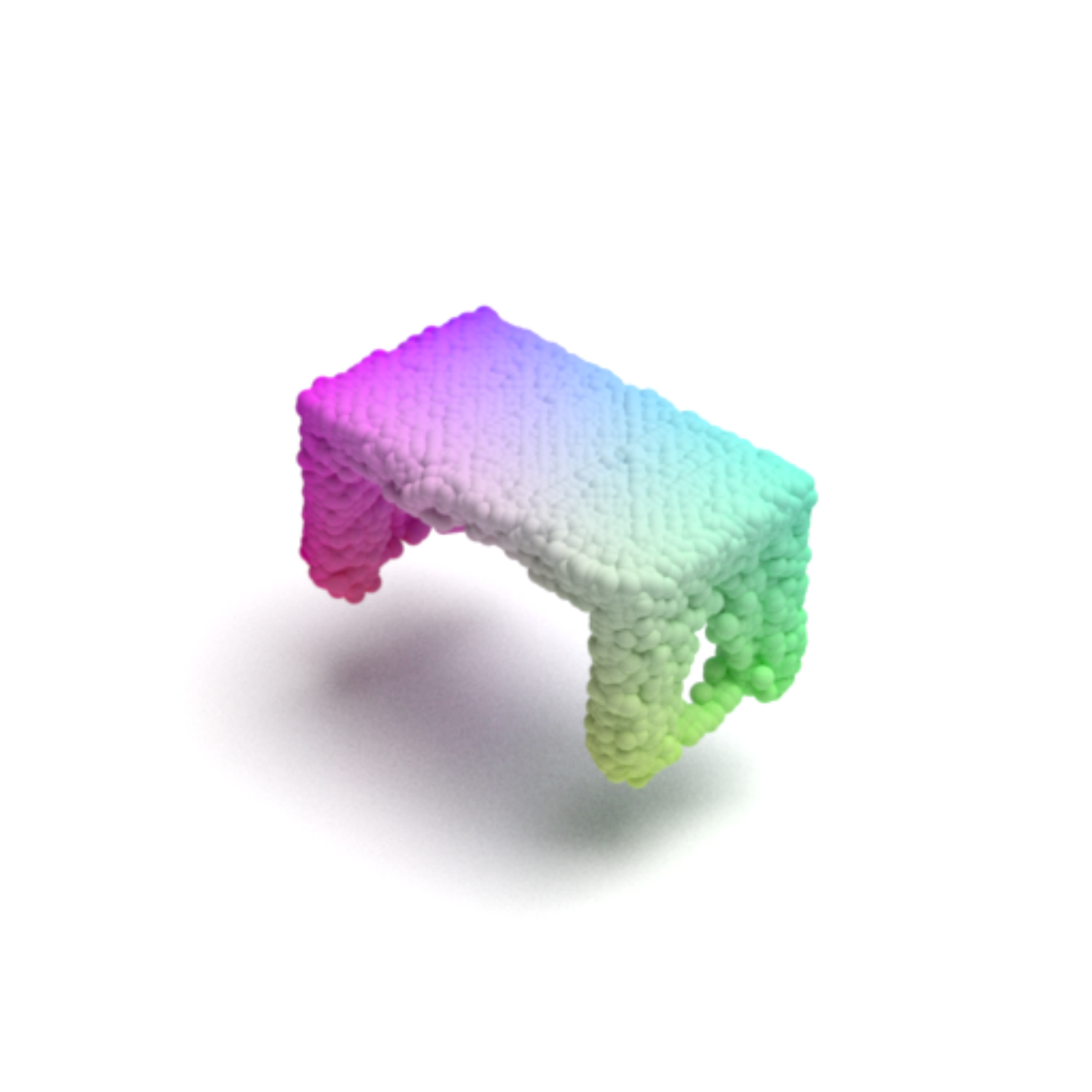}
\includegraphics[clip,trim=3cm 3cm 3cm 3cm, width=0.095\textwidth]{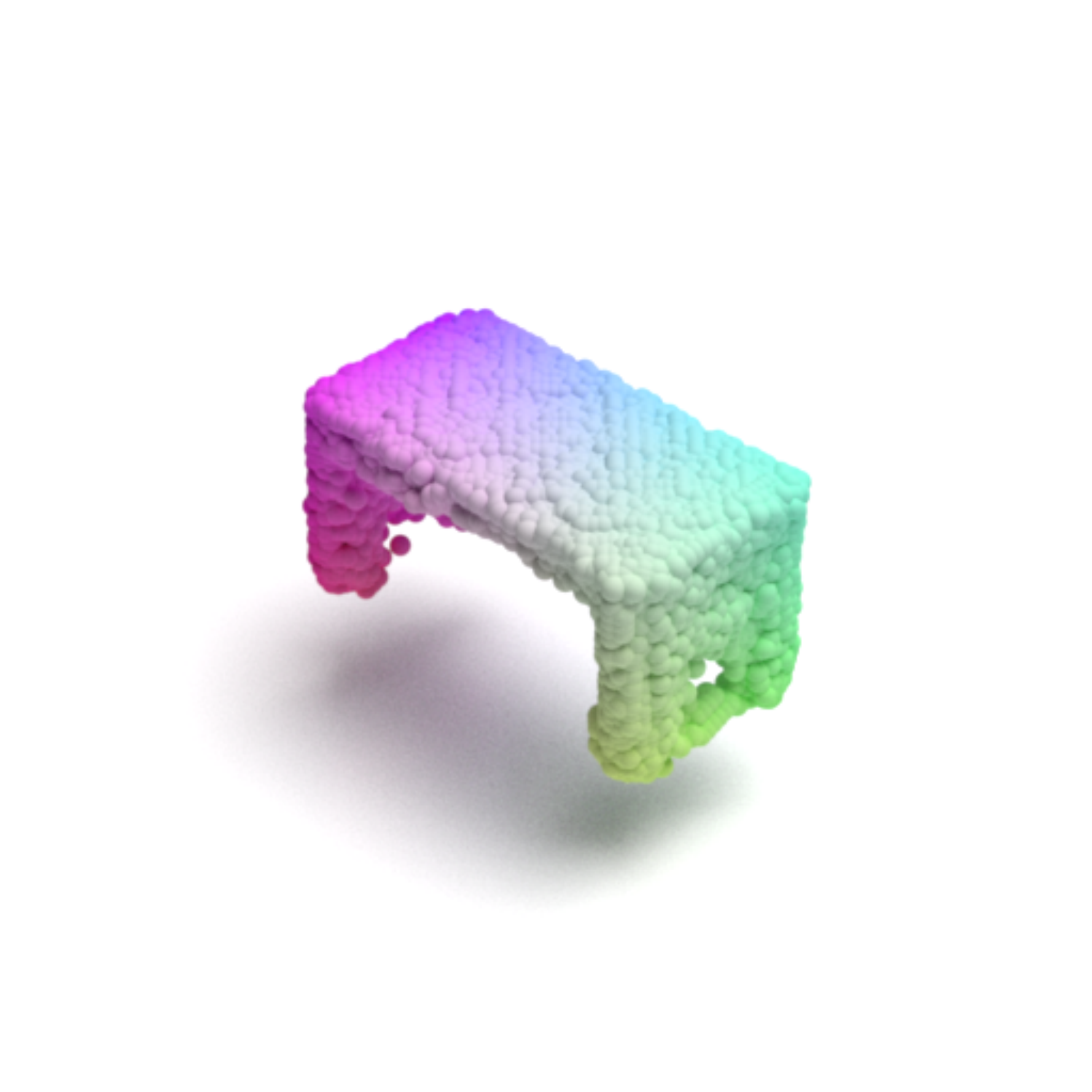}
\includegraphics[clip,trim=3cm 3cm 3cm 3cm, width=0.095\textwidth]{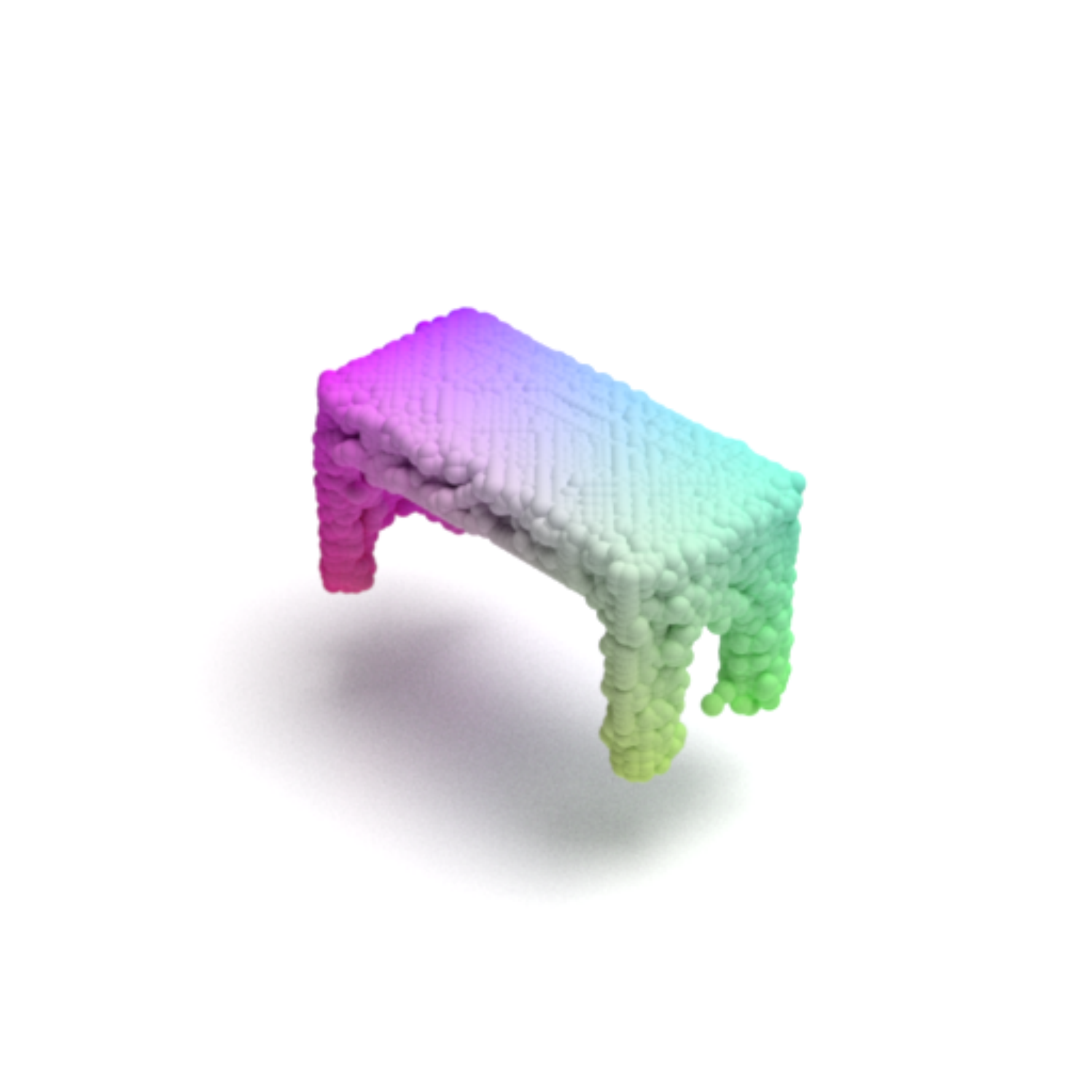}
\includegraphics[clip,trim=3cm 3cm 3cm 3cm, width=0.095\textwidth]{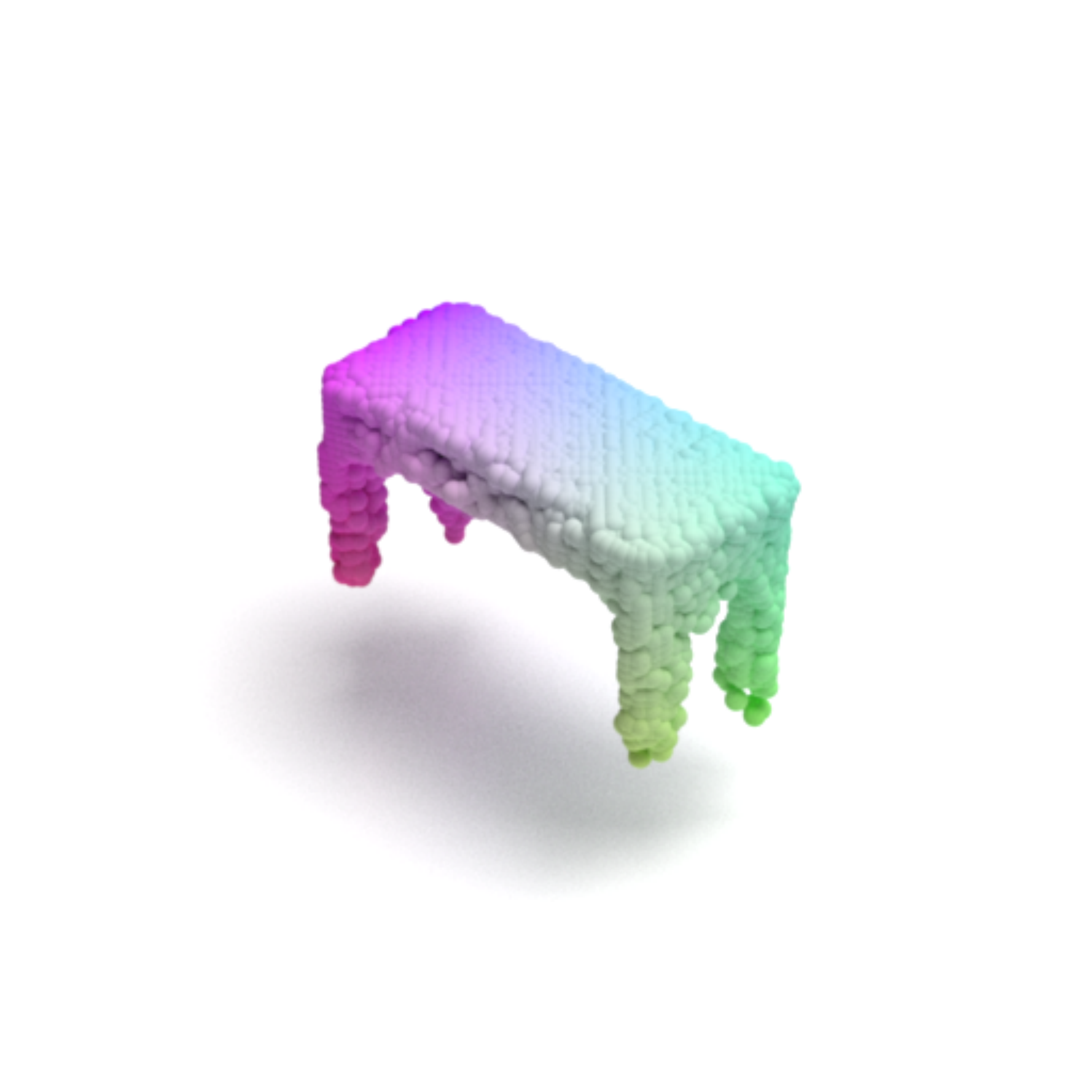}
\includegraphics[clip,trim=3cm 3cm 3cm 3cm, width=0.095\textwidth]{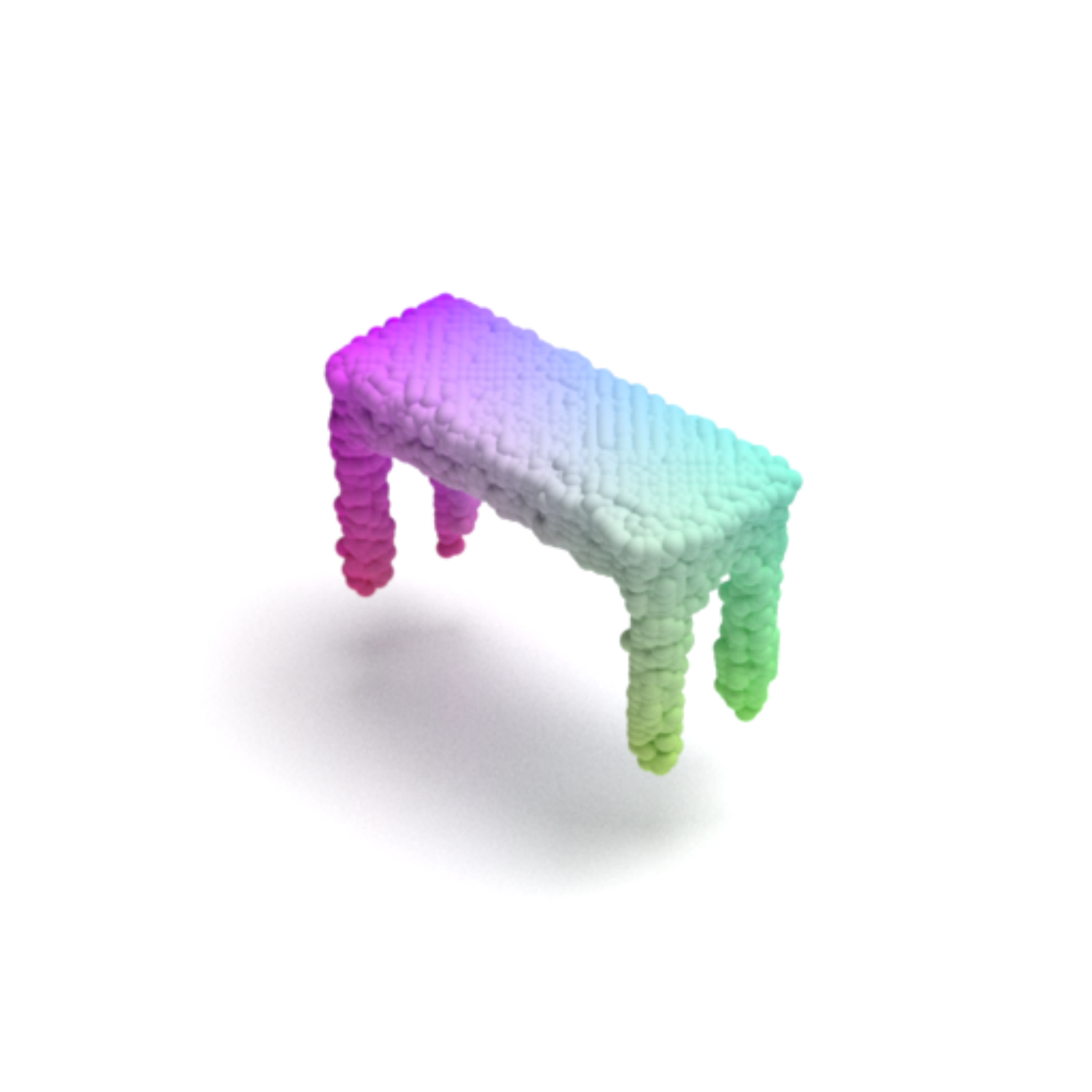}
\includegraphics[clip,trim=3cm 3cm 3cm 3cm, width=0.095\textwidth]{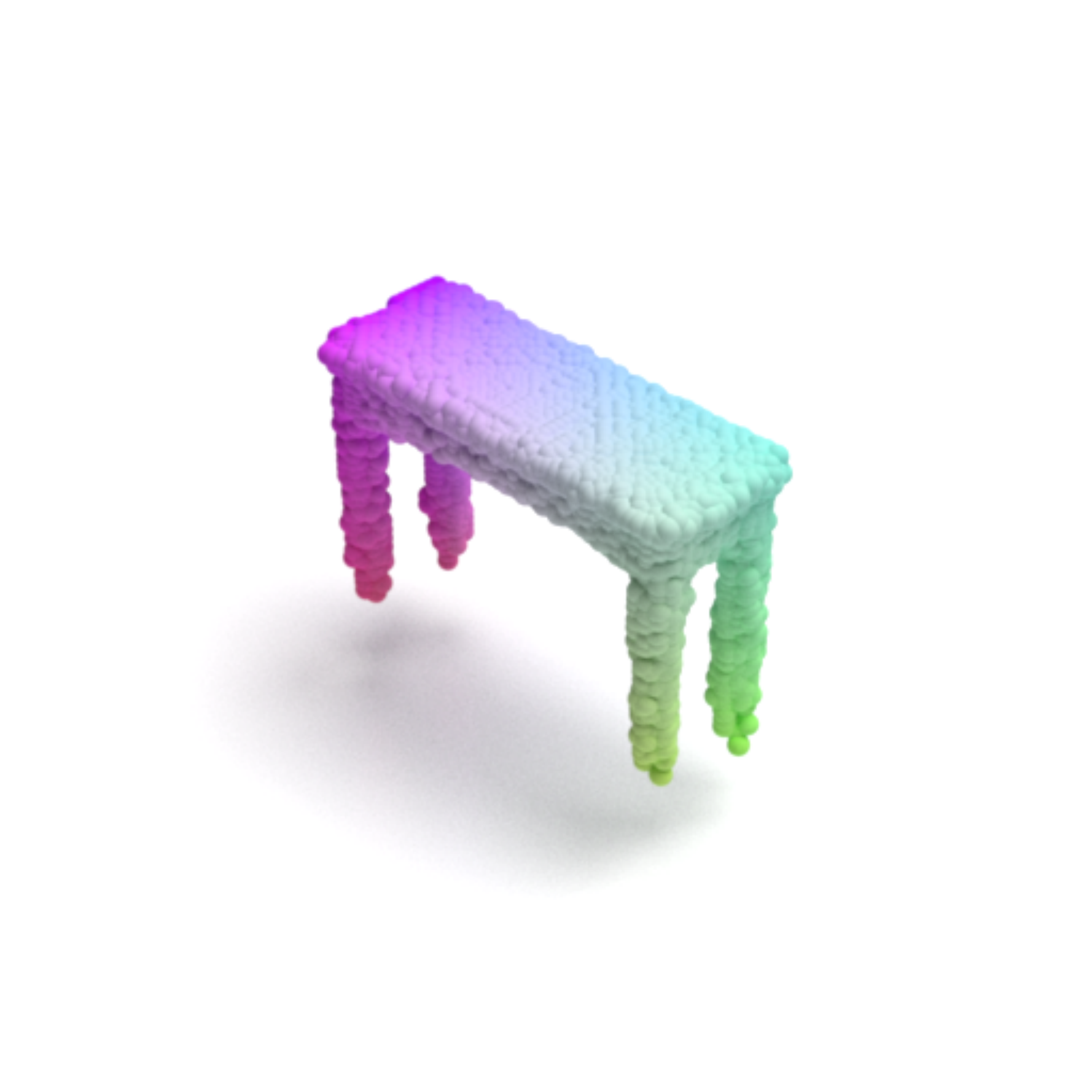}
\includegraphics[clip,trim=3cm 3cm 3cm 3cm, width=0.095\textwidth]{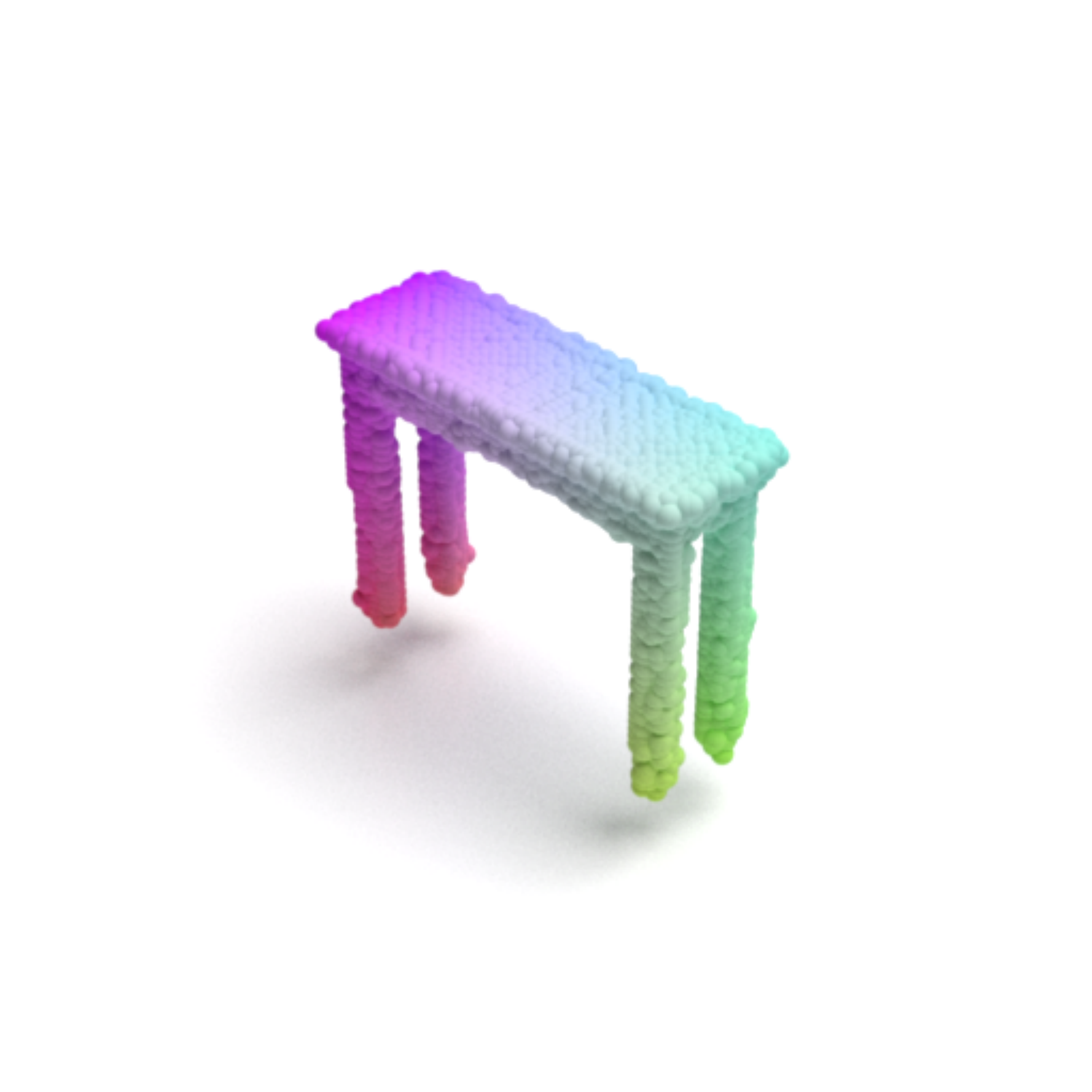}

\captionof{figure}{\textbf{Shape Interpolation.} The latent representations learned by our model are smooth. In this figure, we interpolate between two test objects (left-most and right-most column) by generating the intermediate 3D objects from the spherical interpolation of the latent codes.}
\label{fig:interpolation}
\end{figure*}

\begin{figure}[t]
    \centering
    \includegraphics[width=\linewidth]{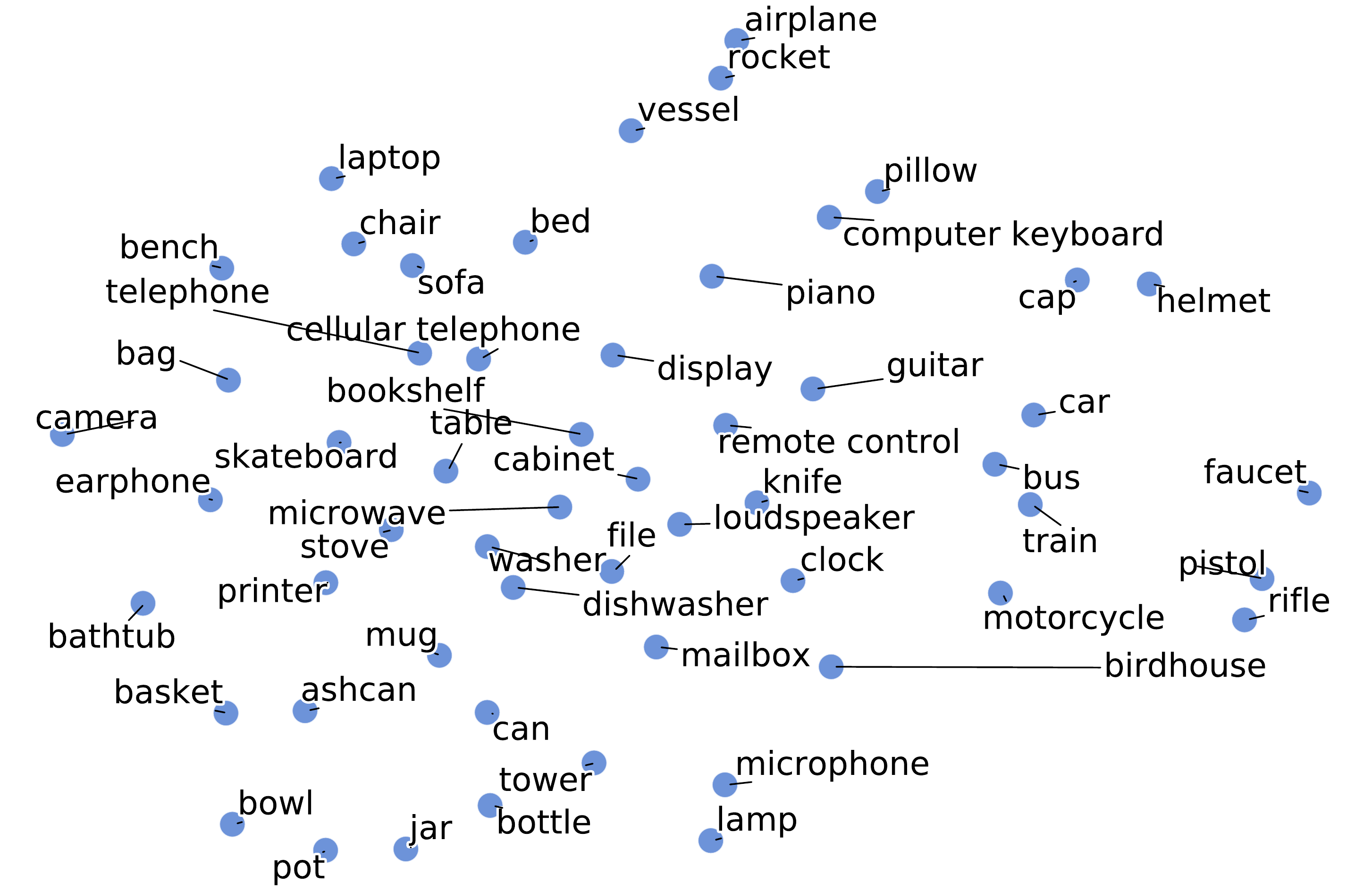}
    \caption{Word embeddings learned by our model for various class objects. We see that visually similar categories cluster together, even if they are not semantically similar. E.g., airplane, rocket and vessel are together, mailbox and microwave are also together.}
    \label{fig:tsne}
    \vspace{-0.2in}
\end{figure}

\medskip
\noindent\textbf{3D shape interpolation.}
We choose spherical over linear interpolation for analyzing the latent space as recommended by~\cite{karras2020analyzing}. Figure~\ref{fig:interpolation} shows samples generated by interpolating between two random objects picked from the test data.
\medskip
\medskip
\noindent\textbf{Learned class embeddings}
Our class conditional model learns embeddings for each of the class. Figure~\ref{fig:tsne} shows a TSNE~\cite{maaten2008visualizing} plot of the embeddings in two dimensions. We can immediately observe that object classes that look visually similar appear closer in the embedding space.

\begin{figure}[!h]
\centering

\raisebox{.2\height}{\rotatebox{90}{\small{Samples}}}
\includegraphics[clip,trim=3cm 3cm 3cm 3cm, width=0.085\textwidth]{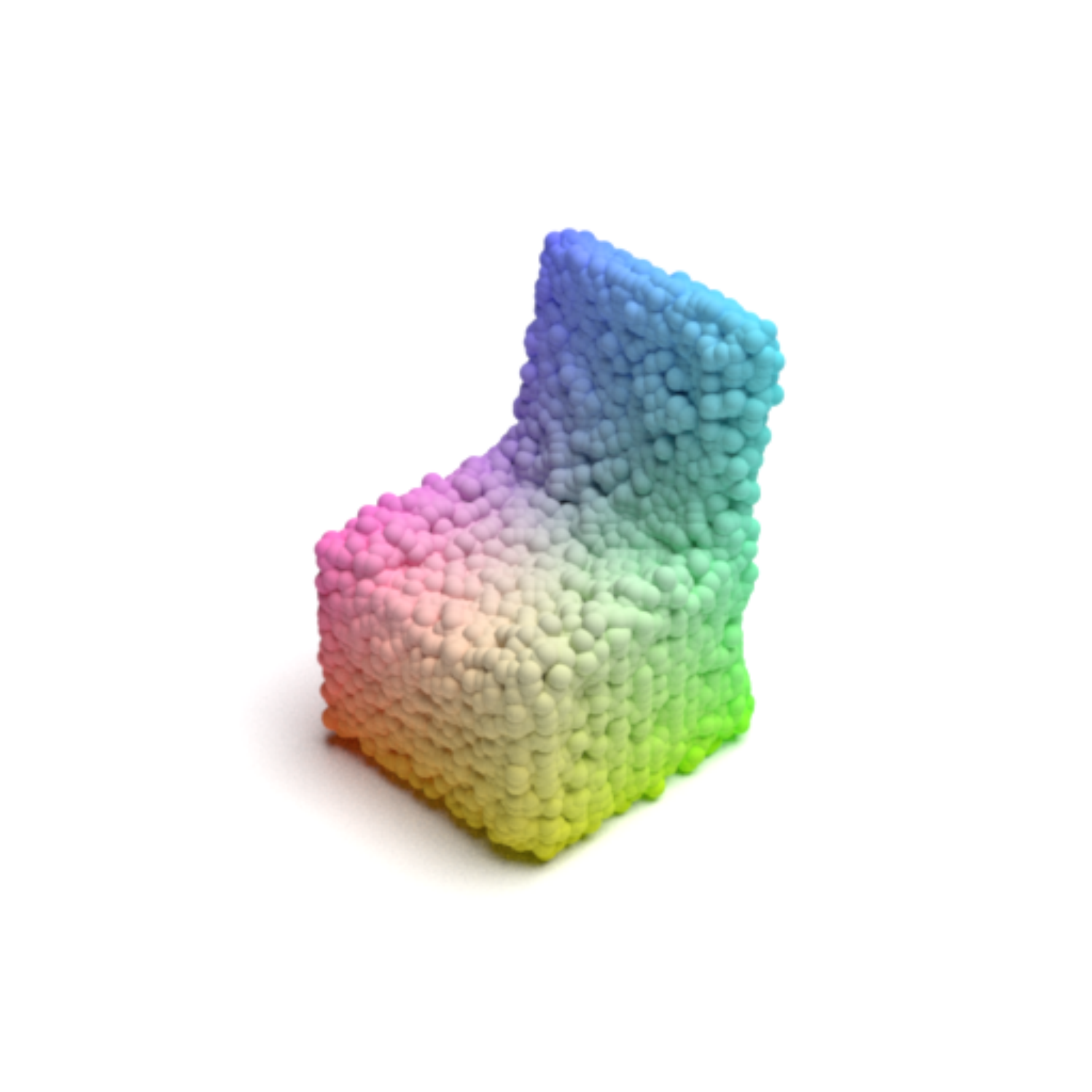}
\includegraphics[clip,trim=3cm 3cm 3cm 3cm, width=0.085\textwidth]{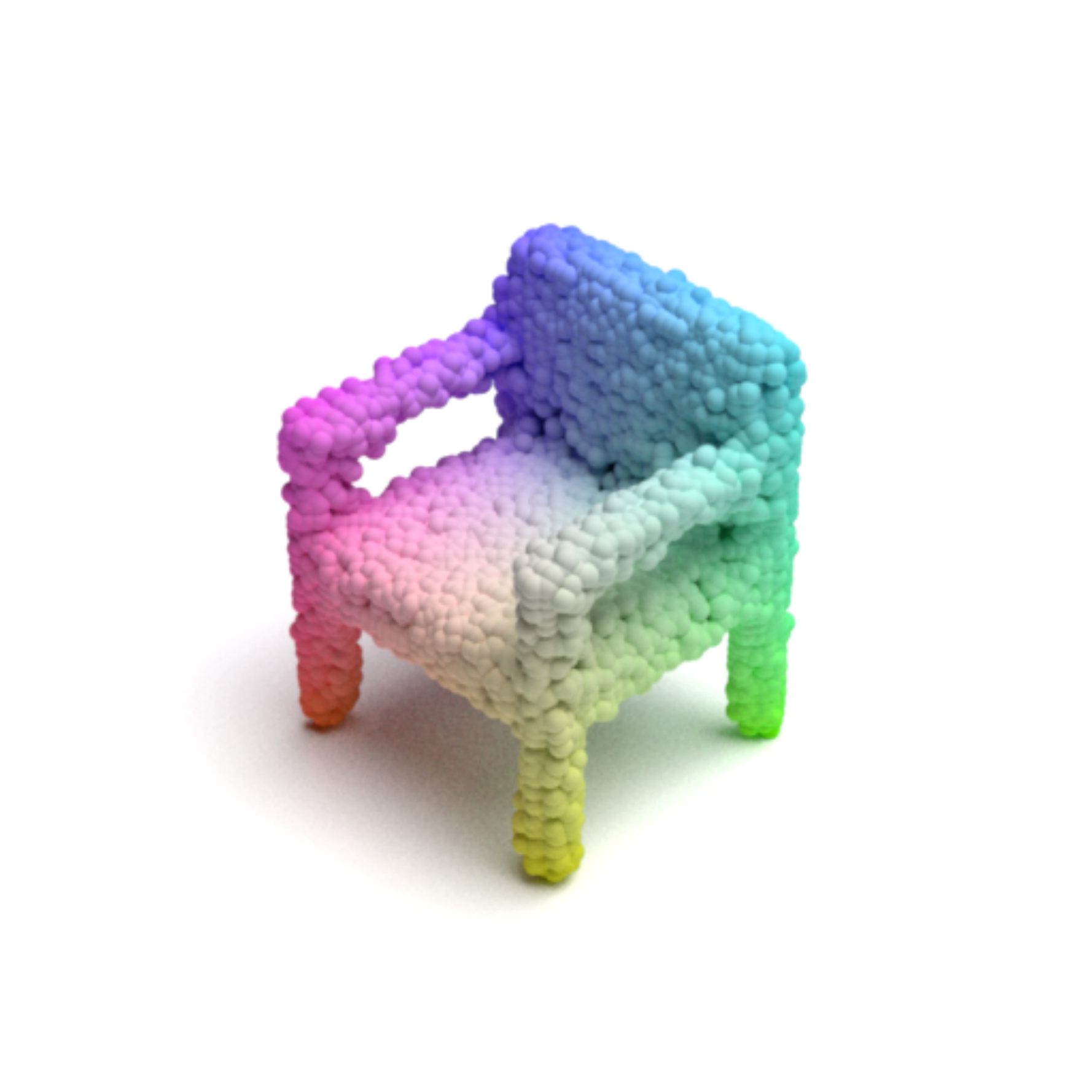}
\includegraphics[clip,trim=3cm 3cm 3cm 3cm, width=0.085\textwidth]{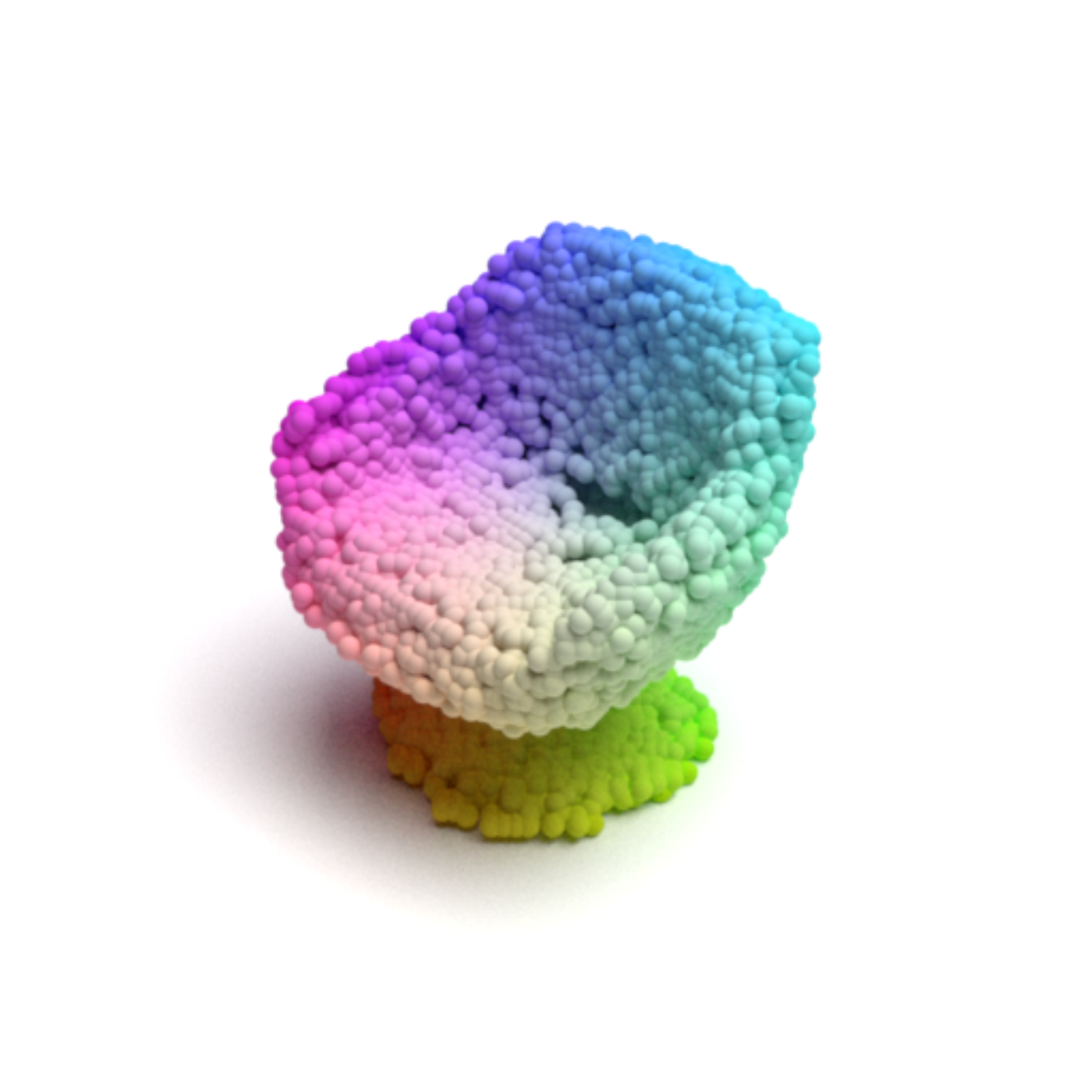}
\includegraphics[clip,trim=3cm 3cm 3cm 3cm, width=0.085\textwidth]{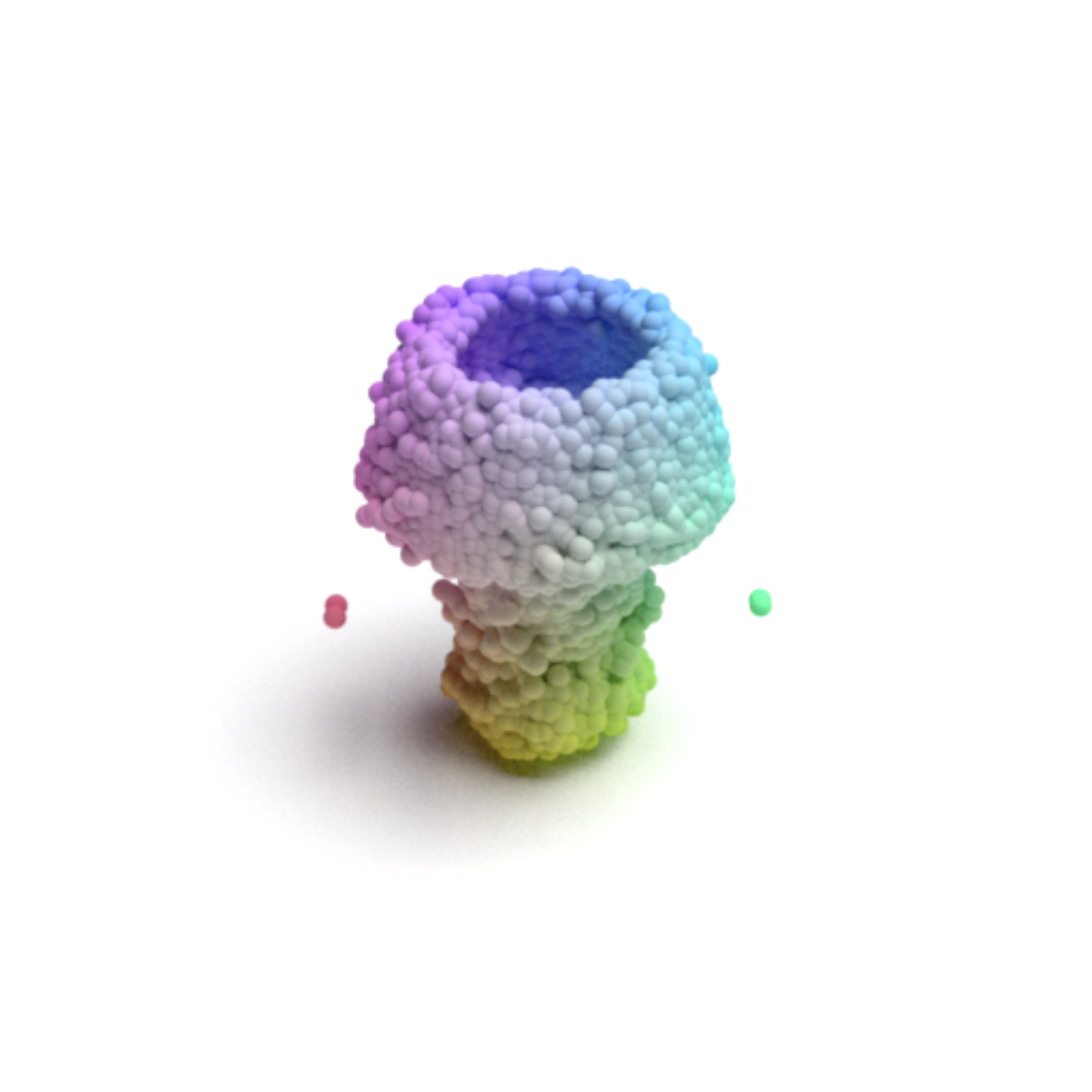}
\includegraphics[clip,trim=3cm 3cm 3cm 3cm, width=0.085\textwidth]{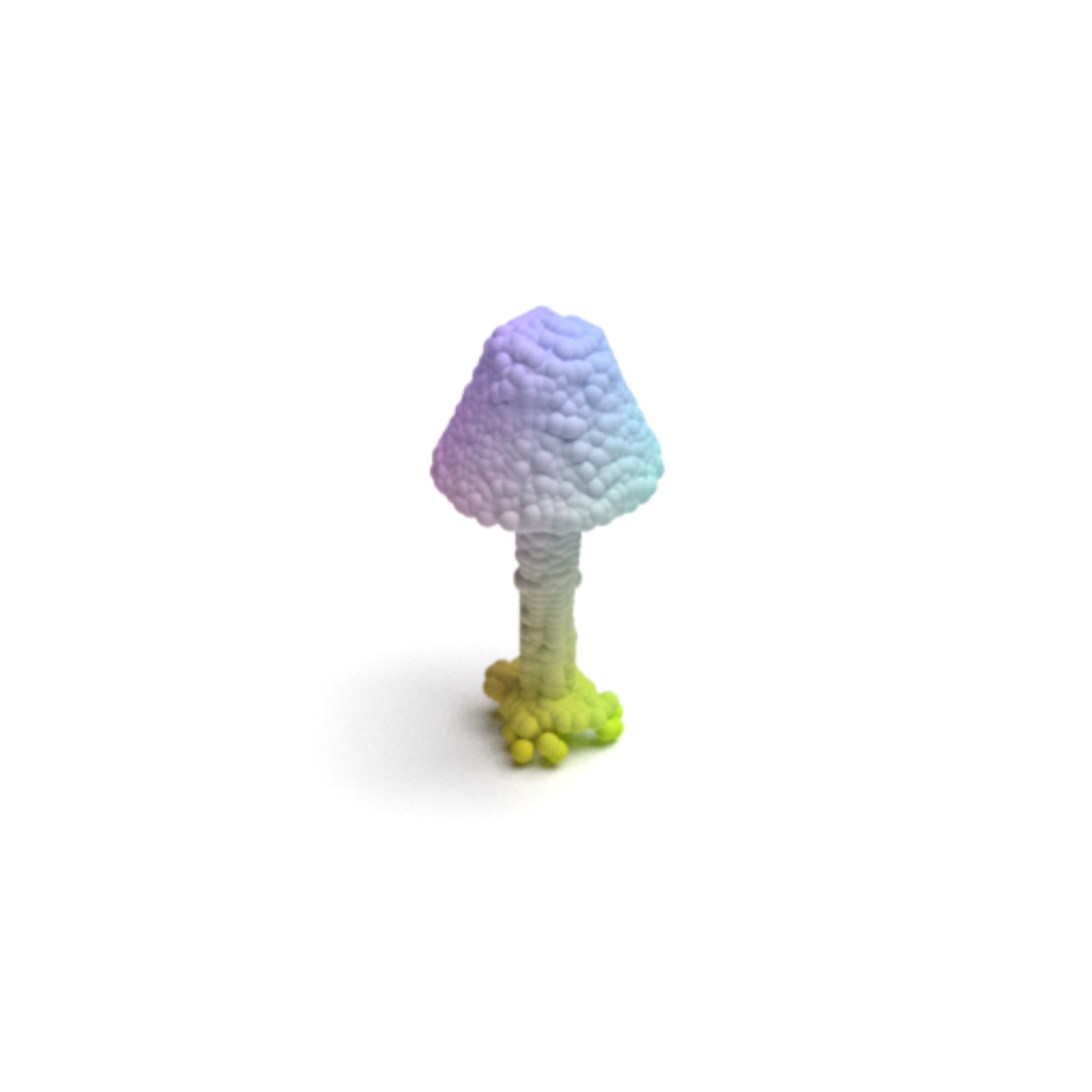}
\raisebox{0.5\height}{\centering{\rotatebox{90}{\small{NN1}}}}
\includegraphics[clip,trim=3cm 3cm 3cm 3cm, width=0.085\textwidth]{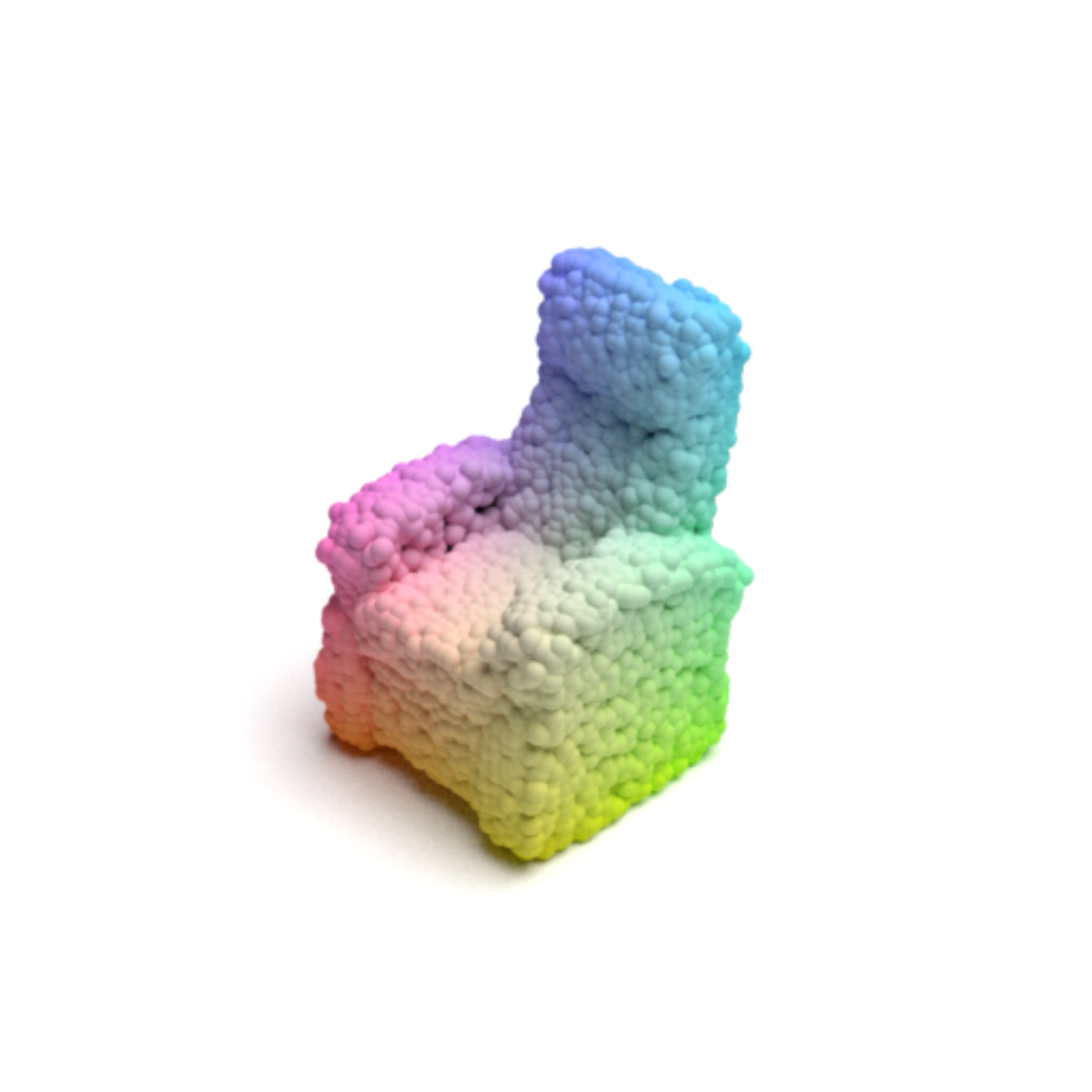}
\includegraphics[clip,trim=3cm 3cm 3cm 3cm, width=0.085\textwidth]{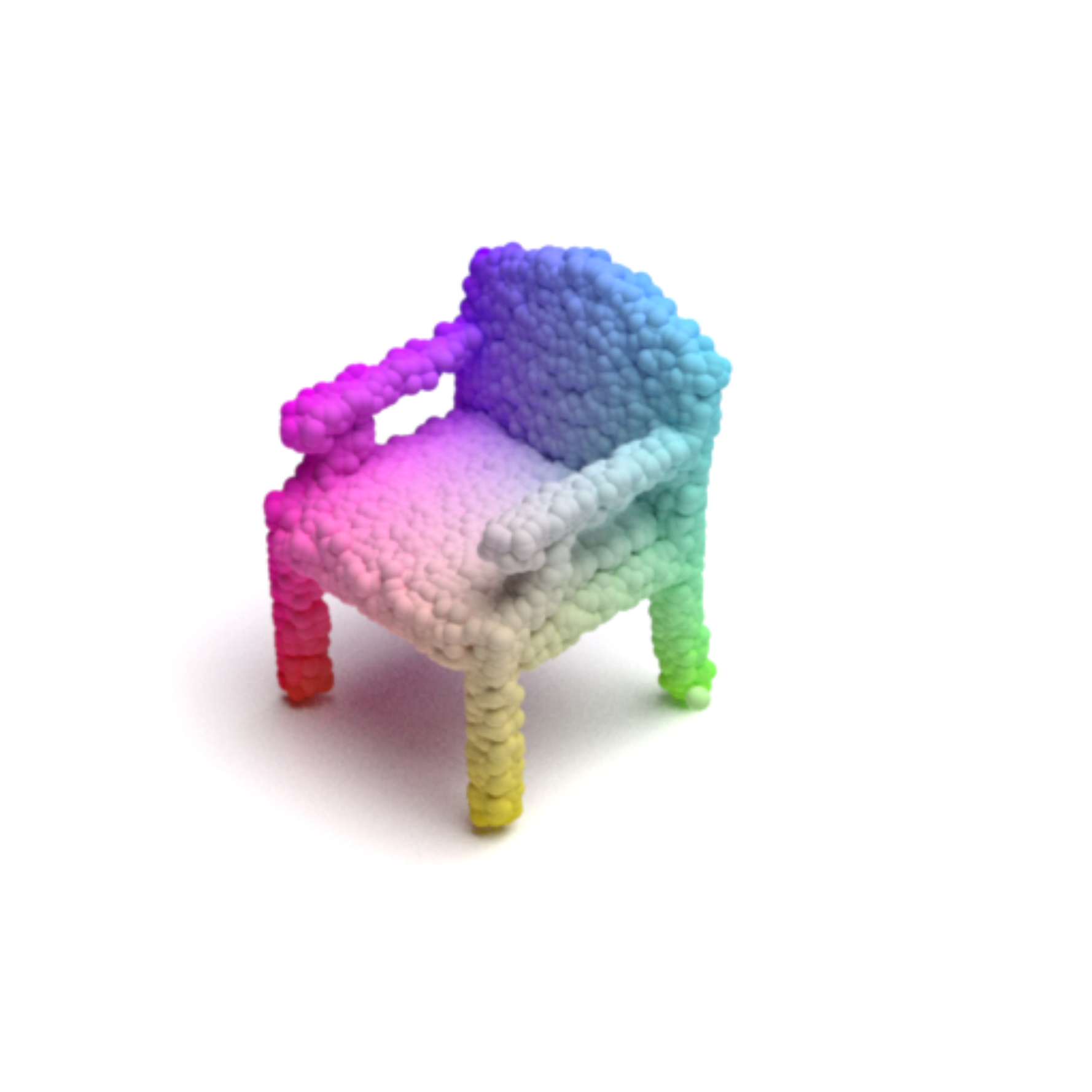}
\includegraphics[clip,trim=3cm 3cm 3cm 3cm, width=0.085\textwidth]{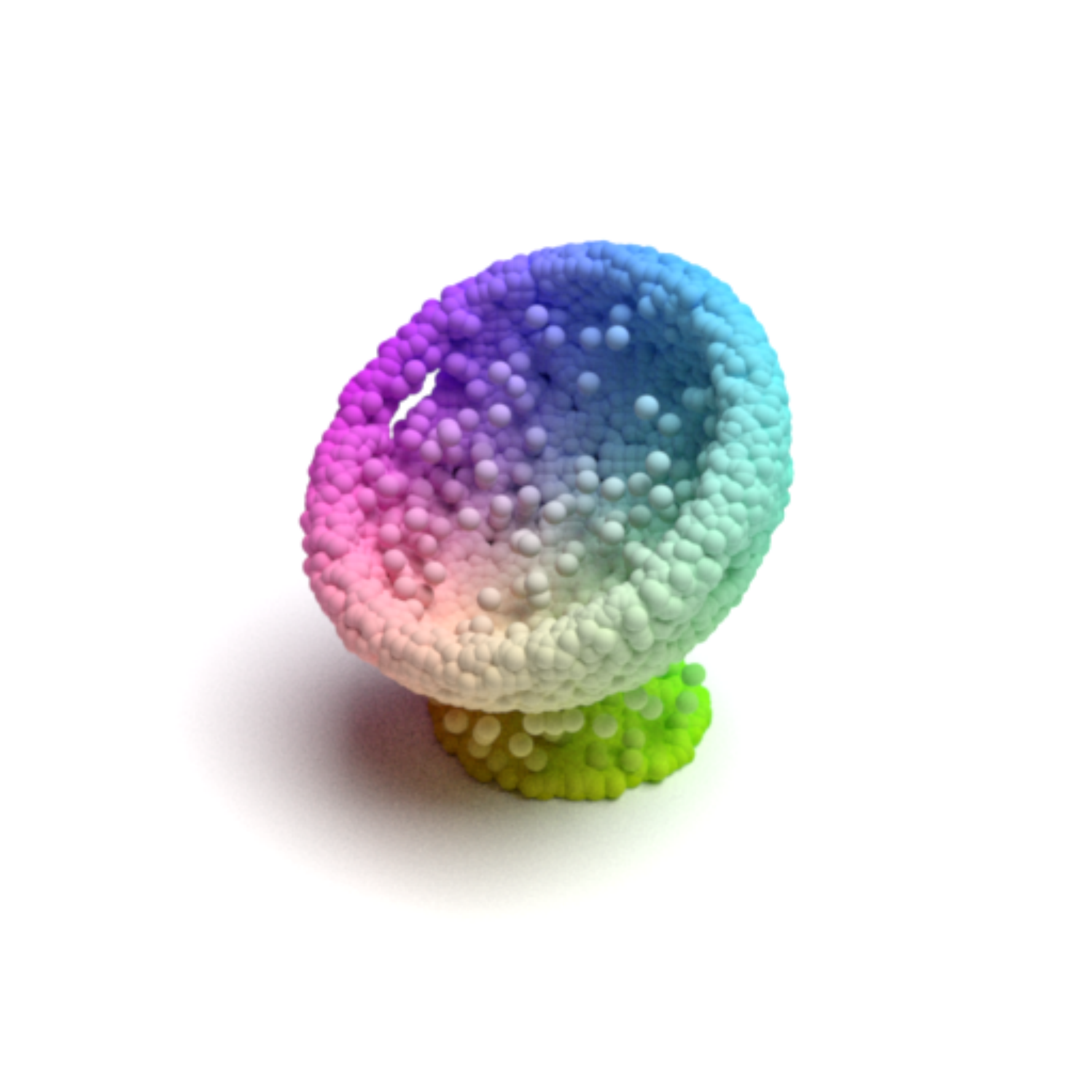}
\includegraphics[clip,trim=3cm 3cm 3cm 3cm, width=0.085\textwidth]{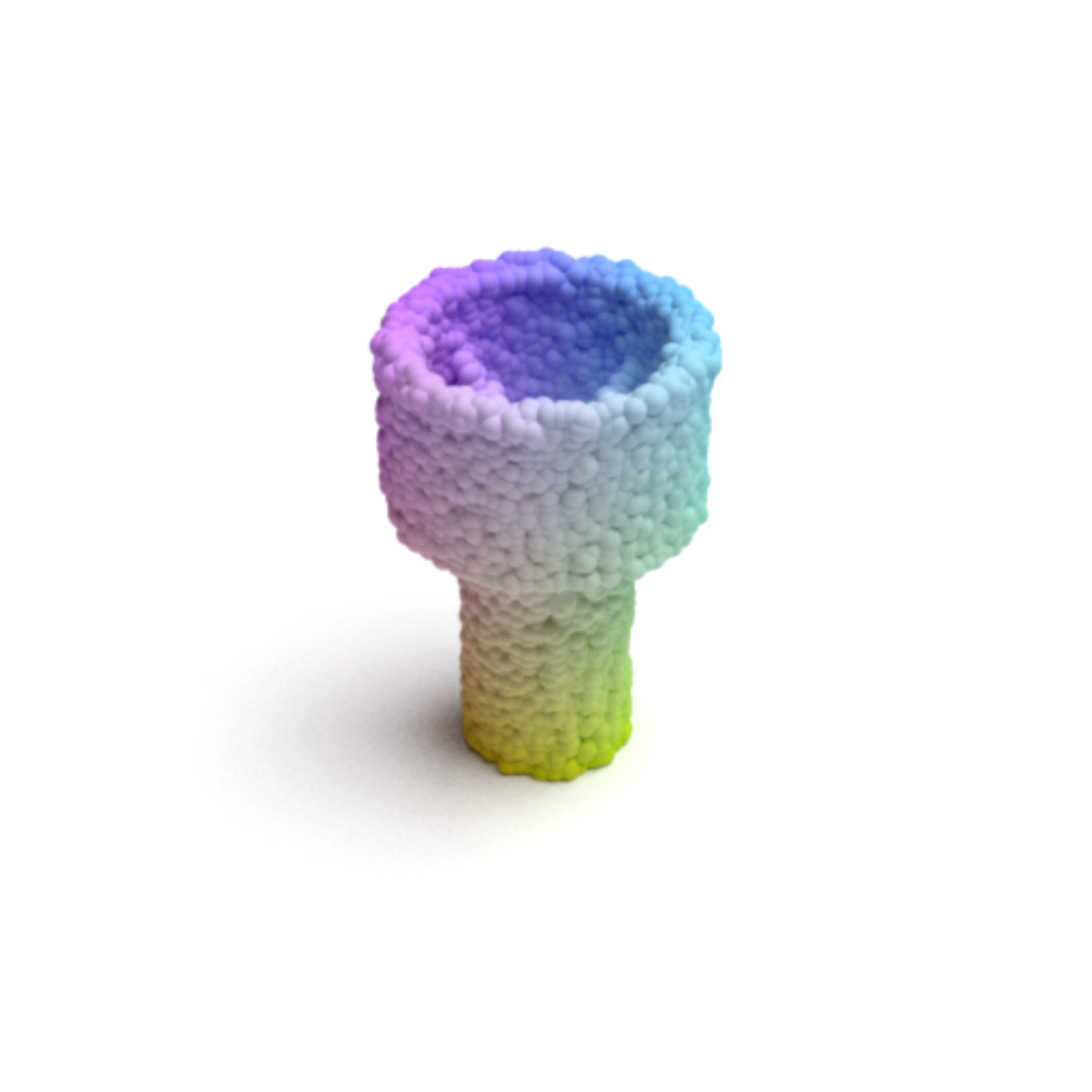}
\includegraphics[clip,trim=3cm 3cm 3cm 3cm, width=0.085\textwidth]{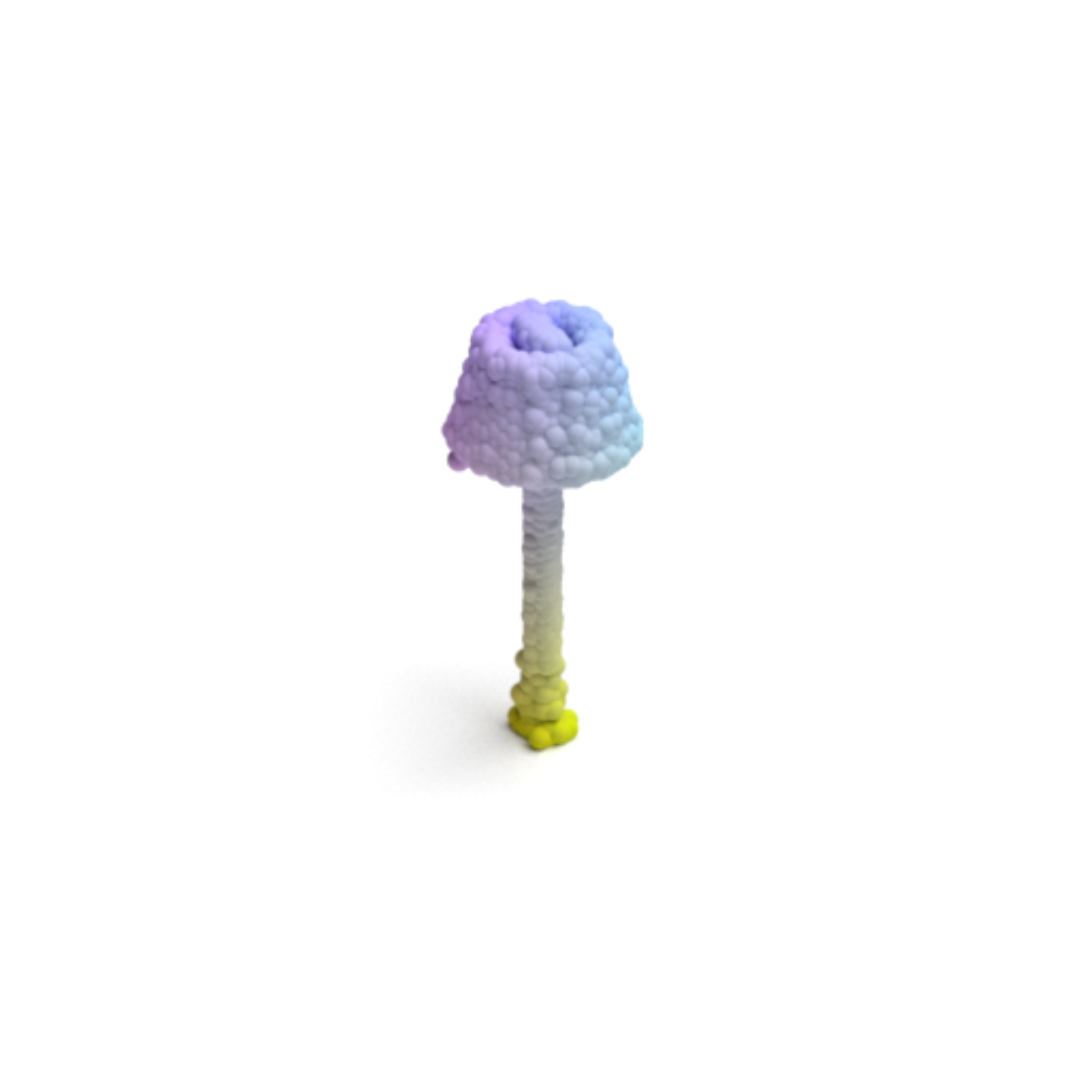}
\raisebox{0.5\height}{\centering{\rotatebox{90}{\small{NN2}}}}
\includegraphics[clip,trim=3cm 3cm 3cm 3cm, width=0.085\textwidth]{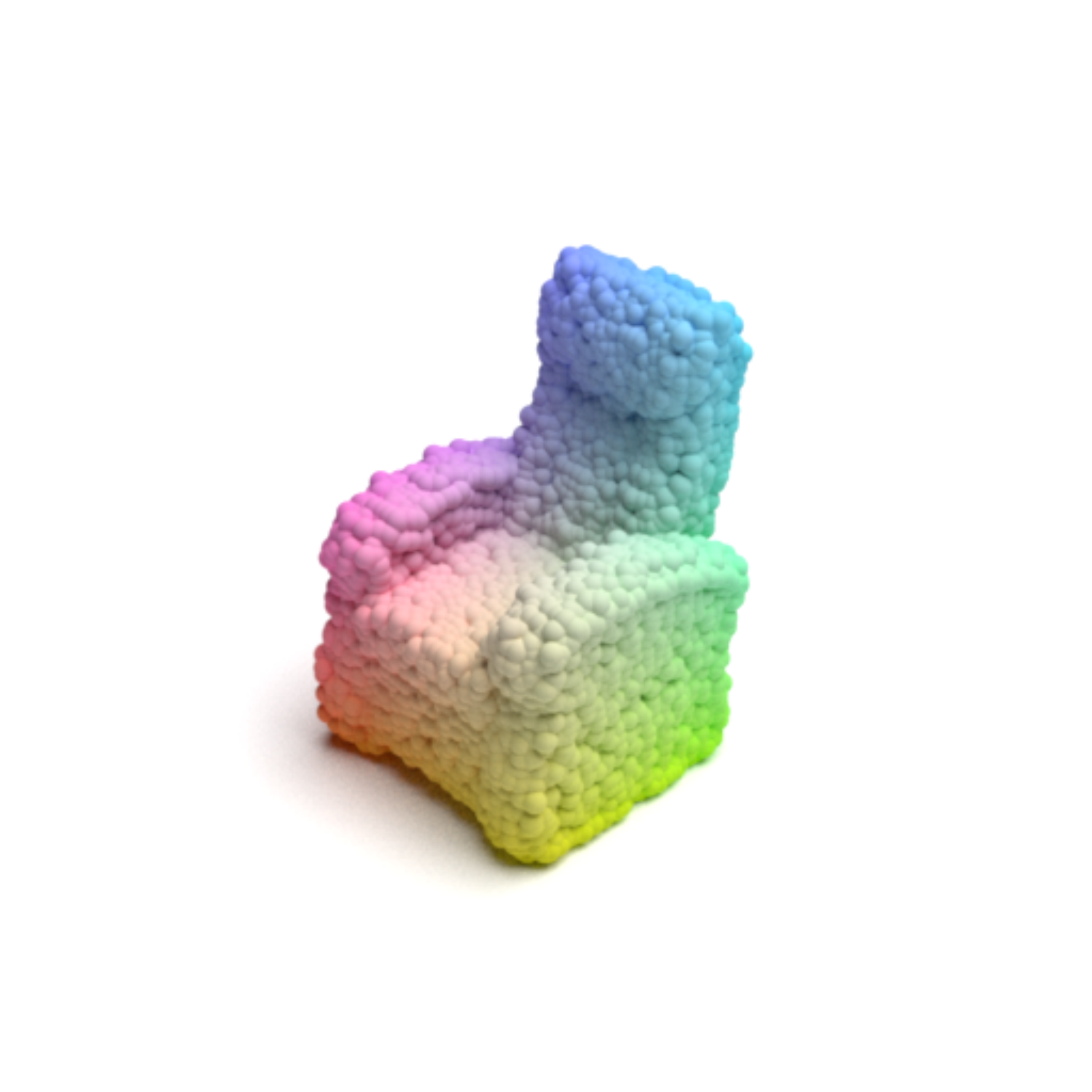}
\includegraphics[clip,trim=3cm 3cm 3cm 3cm, width=0.085\textwidth]{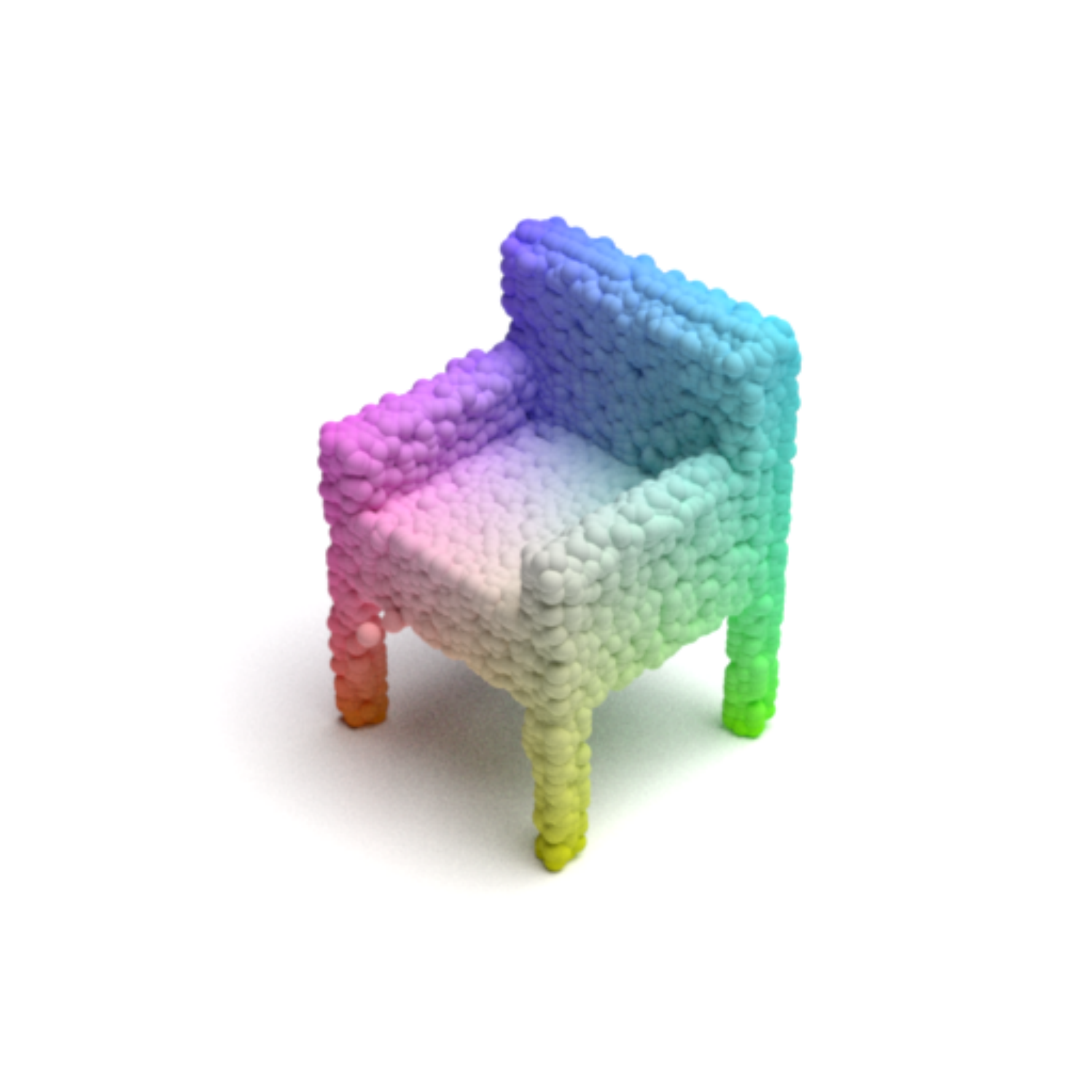}
\includegraphics[clip,trim=3cm 3cm 3cm 3cm, width=0.085\textwidth]{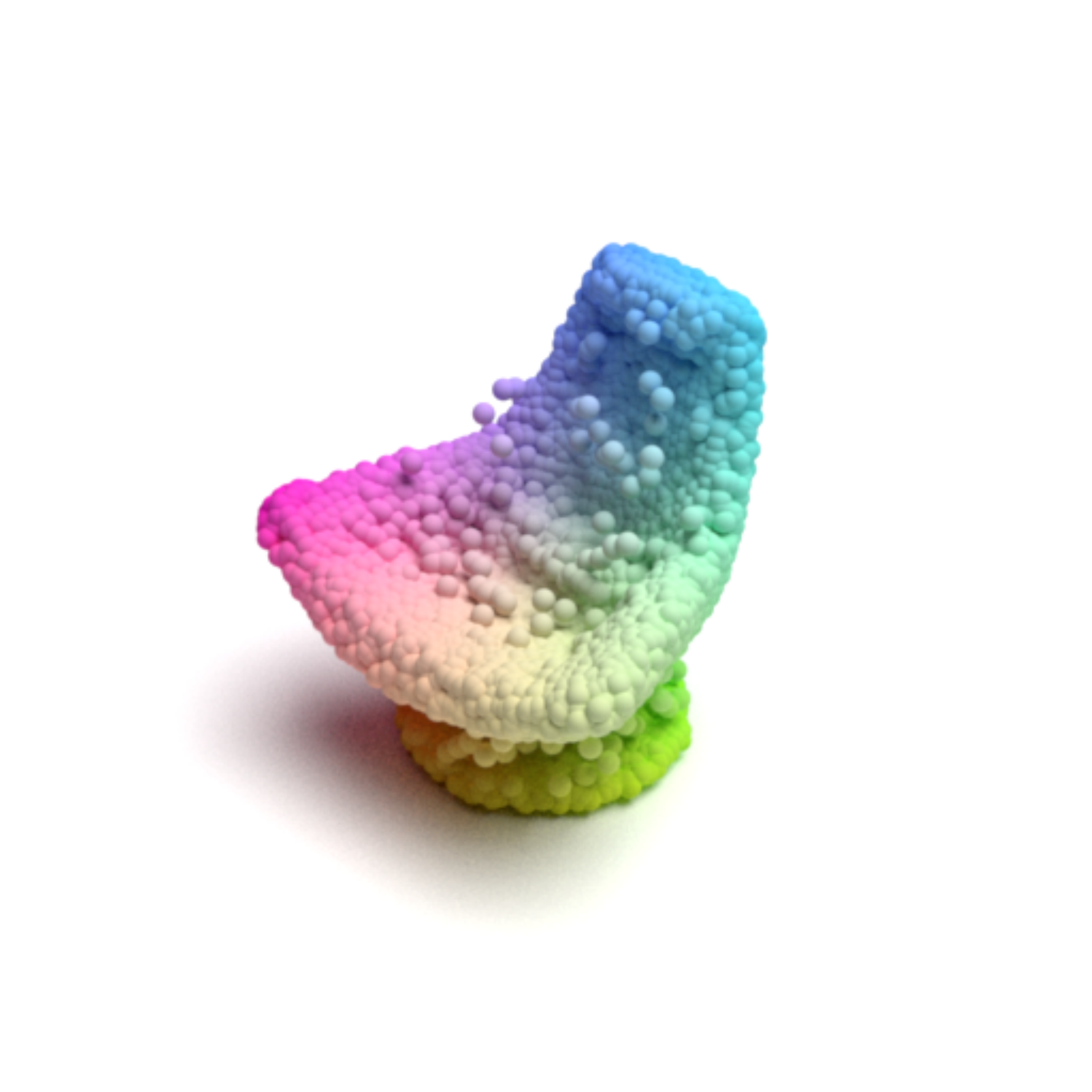}
\includegraphics[clip,trim=3cm 3cm 3cm 3cm, width=0.085\textwidth]{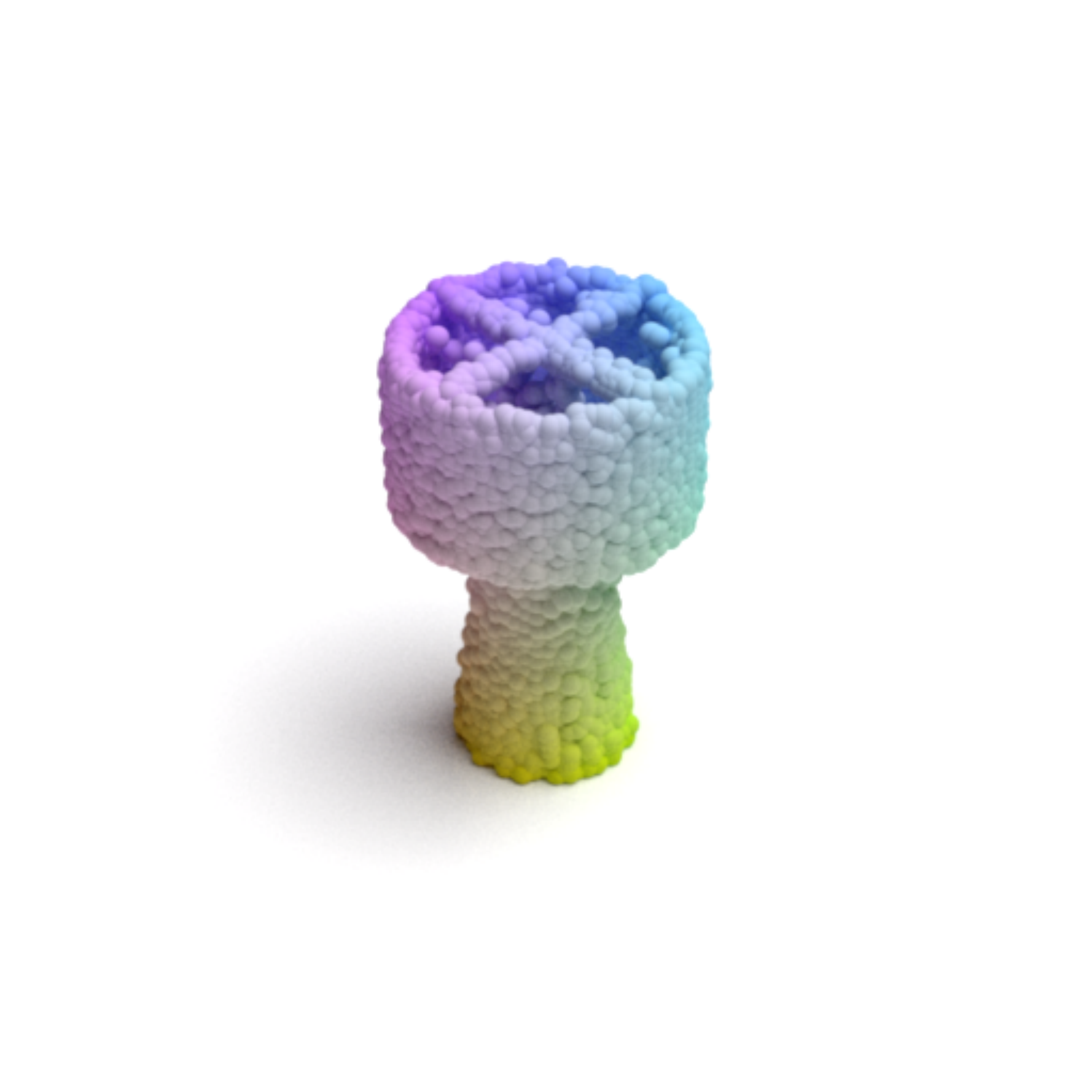}
\includegraphics[clip,trim=3cm 3cm 3cm 3cm, width=0.085\textwidth]{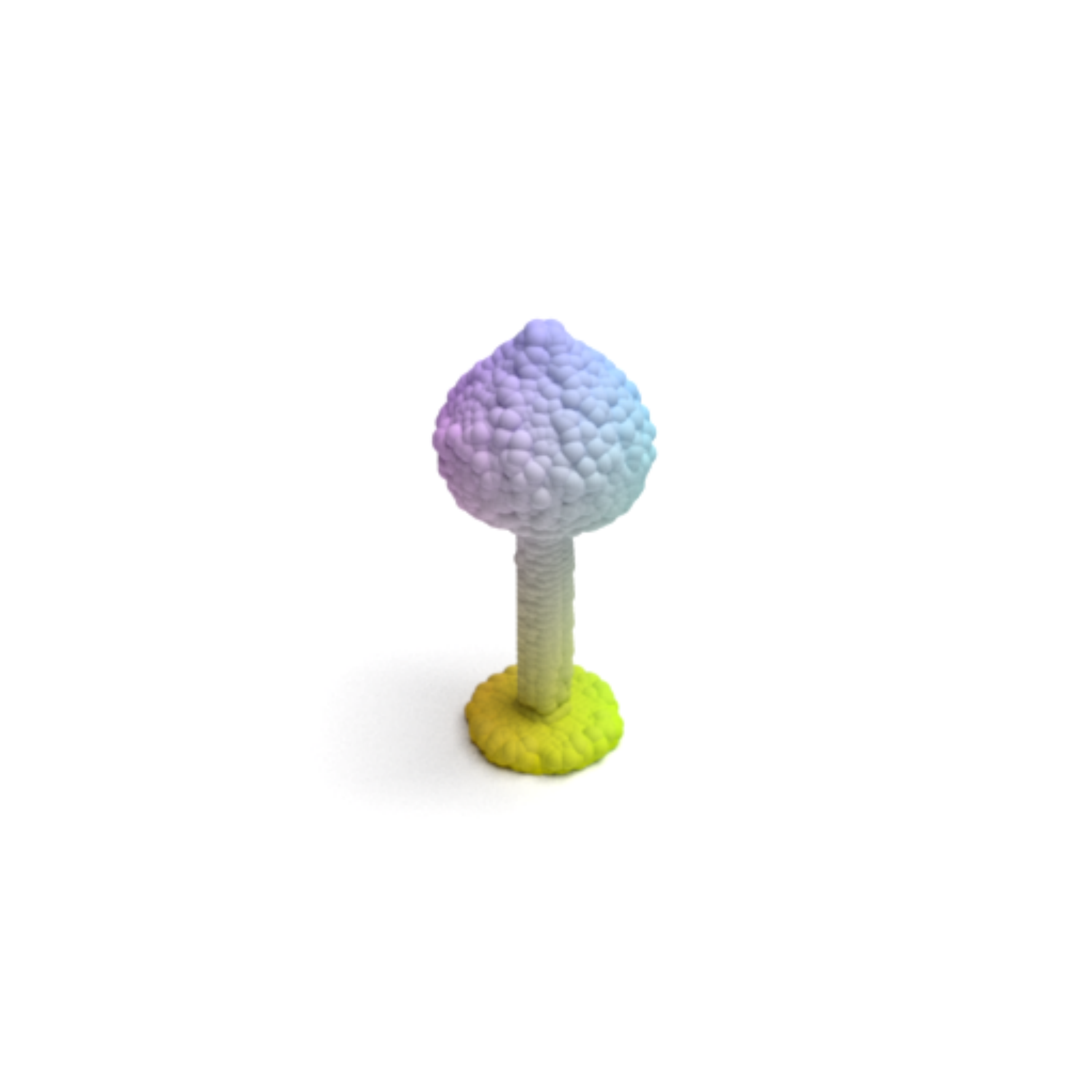}

\captionof{figure}{Two NN to generated 3D models: our model produces 3D models that differ from the two closest train samples.}
\label{fig:shape_synthesis}
\vspace{-0.25in}
\end{figure}

\medskip\noindent\textbf{3D Shape Synthesis.}
We train an IMLE-based model on the shape identities obtained from the trained identity encoder. Once the model is trained, we use it to sample the shape identities of new 3D objects. The viewpoint generator maps the shape identities to the depth map of the object as viewed from the given viewpoint. Figure~\ref{fig:shape_synthesis} shows a few generated 3D objects for two categories, along with the nearest neighbors for each generated sample. We observe that the generated samples are diverse and differ from their closest neighbors in the training set. %

\section{Conclusion and Limitations}
\label{sec:conculsion}

To summarize, we provide a simple framework to encode 3D shapes and generate 3D consistent depth maps for objects. Our approach is fairly generic and can be extended to work with a variety of image-based neural network techniques. For single view reconstruction, our method outperforms existing approaches that use different representations for 3D objects such as meshes, point clouds, implicit functions, and depth maps. That said, our method also faces the same limitations as faced by image-based architectures. It is data-hungry and requires multiple viewpoints generated from a large number of training objects in canonical pose to generalize well. We address this shortcoming by training a single class conditional model on all training objects, but generalizing to work well on a new category with few samples remains an open problem. Another limitation is that it can only model surfaces. More architectural changes may be required for the model to adapt to the 3D solid volumes, textures, etc., which we will explore in the future.


{\small
\bibliographystyle{ieee}
\bibliography{egbib}
}

\clearpage
\appendix

\section{Implementation Details}
\label{sec:implementation}
Here we provide various implementation details of our model. Although our architecture is class-conditional (i.e., can take object category as an input to generate category specific shapes), most of the baselines we have used in our paper are \emph{not}. We found extending existing methods to be class-conditional to be a non-trivial exercise, and hence, trained separate models for separate categories for our method for quantitative evaluations.

We trained our models for 100,000 iterations with batch size of 24 and $V_{in}=V_{out}=2$ for the results reported in the paper. Section~\ref{sec:batch} has more discussion regarding the choice of these hyperparameters. We also use Adam~\cite{kingma2014adam} optimizer with hyperparameters $\beta_1=0, \beta_2=0.99$ and learning rate $4\times10^{-3}$. We use a single 24 GB GPU (NVIDIA P6000) in all our experiments.

For our generative model, we train an IMLE-based model~\cite{li2019implicit, hoshen2019non} for $100$ epochs with a batch size of $128$. A simple 2-layer fully-connected neural network with each 128 units is used to map the $64$-dimensional noise vectors to $512$-dimensional shape identity vectors. We use Adam~\cite{kingma2014adam} optimizer with learning rate $10^{-3}$ and $\beta_1 = 0.5$ and $\beta_2 = 0.999$ to learn the network weights.

\subsection{Identity Encoder}
\label{sec:encoder}
Table~\ref{tab:encoder} shows the detailed architecture of identity encoder. We follow an architecture similar to StyleGANv2 discriminator, although any architecture can be easily plugged-in. Our encoder takes in a single depth map input, discretized in 256 bins, and applies $1\times1$ convolutions in its first layer. It follows up with 4 residual blocks each consisting of 3 convolutional layers each downsizing the feature maps by a factory of 2. The residual connections have been shown helpful in design of various classifiers (unlike the generator network, where till now skip connections have shown limited benefits). The final block in the encoder consists of two fully-connected layers and it generates a 512 dimension identity embedding of the input depth map.

\begin{table}[h]
    \centering
    \caption{Encoder architecture - All `convolutional' layers are represented as $\text{Conv2d}(\text{C}_\text{in}, \text{C}_\text{out}, \text{kernel}\_\text{size}, \text{stride}, \text{padding})$. All `fully-connected' layers are represented as $\text{Linear}(\text{C}_\text{in}, \text{C}_\text{out})$. LeakyReLU for Leaky Rectified Linear Units with a negative slope of $0.2$.  `SkipConv2d' layers de\text{No}tes applying `Conv2d' to the output of the previous block and adding  it to the output of previous layer.}
     \begin{tabular}{@{}l|c@{}}
    \toprule
    Block & Layers\\
    \midrule
    conv & $\begin{array} {ccc}  \text{Conv2d}(256, 256, 1, 1, 0) \\ \text{LeakyReLU} \end{array}$ \\
    
    \midrule
    ResBlock & $\begin{array} {ccc}  \text{Conv2d}(256, 256, 3, 1, 1) \\ \text{LeakyReLU} \\ \text{Conv2d}(256, 512, 3, 2, 0) \\ \text{LeakyReLU} \\  \text{SkipConv2d}(256, 512, 1, 2, 0) \end{array}$ \\
    
    \midrule
    ResBlock & $\begin{array} {ccc}  \text{Conv2d}(512, 512, 3, 1, 1) \\ \text{LeakyReLU} \\ \text{Conv2d}(512, 512, 3, 2, 0) \\ \text{LeakyReLU} \\  \text{SkipConv2d}(512, 512, 1, 2, 0) \end{array}$ \\
    
    \midrule
    ResBlock & $\begin{array} {ccc}  \text{Conv2d}(512, 512, 3, 1, 1) \\ \text{LeakyReLU} \\ \text{Conv2d}(512, 512, 3, 2, 0) \\ \text{LeakyReLU} \\  \text{SkipConv2d}(512, 512, 1, 2, 0) \end{array}$ \\
    
    \midrule
    ResBlock & $\begin{array} {ccc}  \text{Conv2d}(512, 512, 3, 1, 1) \\ \text{LeakyReLU} \\ \text{Conv2d}(512, 512, 3, 2, 0) \\ \text{LeakyReLU} \\  \text{SkipConv2d}(512, 512, 1, 2, 0) \end{array}$ \\
    
    \midrule
    final & $\begin{array} {ccc}  \text{Conv2d}(513, 512, 3, 1, 1) \\ \text{LeakyReLU} \\ \text{Linear(8192, 512)} \\ \text{Linear(512, 512)} \end{array}$ \\
    
    \bottomrule
    \end{tabular}
    \label{tab:encoder}
\end{table}

\subsection{Viewpoint Generator}
\label{sec:decoder}
Table~\ref{tab:decoder} shows the detailed architecture of viewpoint generator. The generator is conditional on class and viewpoint embeddings. Each of the embeddings consists of a $256\times4\times4$ feature map. These feature maps, stacked together, act as a $512\times4\times4$ dimensional input to the generator. From here the feature maps pass through a series of modulated convolutional layers. Each modulated convolution scales the image according to the style generated by the `Style block'. Input to this style block is the embedding generated by encoder. A sequence of fully-connected layers apply an affine transform on this embedding to generate the style. All upsampling layers are simple bilinear upsampling transform.

\begin{table}[t]
    \centering
    \caption{Generator architecture. Following the architecture similar to~\cite{karras2018style}, we first apply a sequence of fully connected layers representing an affine transform of encoder output to a style vector. Instead of using a constant learned input, we use a class and viewpoint conditional embedding in the beginning of generator. These are followed by a series of Modulated Convolution layers, each represented in the table as ModConv($\text{C}_\text{in}$, $\text{C}_\text{out}$, kernel, upsample, downsample). A modulated convolution layer scales the input feature maps of the layers with incoming style vectors. Refer to Table~\ref{tab:encoder} for rest of the notations.}
     \begin{tabular}{@{}l|c@{}}
    \toprule
    Block & Layers\\
    \midrule
    Style & $\begin{array} {ccc}  \text{Pixel\text{No}rm}() \\ \text{Linear}(512, 512) \\  \text{Linear}(512, 512) \\  \text{Linear}(512, 512) \\ \text{Linear}(512, 512) \\  \text{Linear}(512, 512) \\  \text{Linear}(512, 512) \\ \text{Linear}(512, 512) \\ \text{Linear}(512, 512) \end{array}$ \\
    
    \midrule
    Viewpoint Embedding & Embedding(20, 4096) \\
    
    \midrule
    Class Embedding & Embedding(55, 4096) \\
    
    \midrule
    conv1 & $\begin{array} {ccc}  \text{ModConv}(512, 512, 3, \text{No}, \text{No}) \\ \text{LeakyReLU} \end{array}$ \\
    
    \midrule
    to\_rgb1 & $\begin{array} {ccc}  \text{ModConv}(512, 256, 1, \text{No}, \text{No}) \end{array}$ \\

    \midrule
    StyledConv & $\begin{array} {ccc}  \text{ModConv}(512, 512, 3, \text{Yes}, \text{No}) \\ \text{LeakyReLU} \\  \text{ModConv}(512, 512, 3, \text{No}, \text{No}) \\ \text{LeakyReLU} \\ \text{ModConv}(512, 512, 3, \text{Yes}, \text{No}) \\ \text{LeakyReLU} \\ \text{ModConv}(512, 512, 3, \text{No}, \text{No}) \\ \text{LeakyReLU} \\  \text{ModConv}(512, 512, 3, \text{Yes}, \text{No}) \\ \text{LeakyReLU} \\ \text{ModConv}(512, 512, 3, \text{No}, \text{No}) \\ \text{LeakyReLU} \\  \text{ModConv}(512, 256, 3, \text{Yes}, \text{No}) \\ \text{LeakyReLU} \\ \text{ModConv}(256, 256, 3, \text{No}, \text{No}) \\ \text{LeakyReLU} \end{array}$ \\
    
    \midrule
    to\_rgbs & $\begin{array} {ccc}  \text{UpSample()} \\ \text{ModConv}(512, 256, 1, \text{No}, \text{No}) \\ \text{UpSample()} \\ \text{ModConv}(512, 256, 1, \text{No}, \text{No}) \\ \text{UpSample()} \\ \text{ModConv}(512, 256, 1, \text{No}, \text{No}) \\ \text{UpSample()} \\ \text{ModConv}(256, 256, 1, \text{No}, \text{No}) \end{array}$ \\
    \bottomrule
    \end{tabular}
    \label{tab:decoder}
\end{table}

\subsection{On Image to Image Translation}
Generating a new view-point from an existing view-point of an image is under-constrained problem. Recent Image-to-Image translation advances~\cite{isola2017image,wang2018high,park2019semantic,liu2019learning} have shown impressive results in transforming images, particularly in case of domain transfer. Richardson et al.~\cite{richardson2020surprising} indicate that much of improvements in image translation domain can be attributed to the use of encoder-decoder bottleneck with large and wide spatial dimensions. While these bottlenecks allow network to perform impressively in domain transfer tasks, it introduces a strong locality bias. This means the network fails to learn simple non-local transformations such as rotating the face along camera axis.

In our work, we chose to have a bottleneck with spatial dimension of $1\times1$ to remove this locality bias, while at the same time generating output depthmaps of high quality.

\subsection{Choice for Generative model}
Implicit Maximum Likelihood Estimation~\cite{li2019implicit} is a recently introduced technique to learn the weights of an implicit generative model without an adversarial loss. IMLE overcomes the three main challenges with Generative Adversarial Networks (GANs) - mode collapse, vanishing gradients, and training instability~\cite{li2019implicit, hoshen2019non}. We found that the generative model trained using IMLE gave superior results, both qualitative and quantitative, compared to other generative modeling techniques such as Variational AutoEncoder (VAE) and GAN.

\section{Ablation studies}
In this section, we discuss various ablation experiments we conducted to analyse different components of our model. For all ablation experiments, we vary only the mentioned hyperparameters while keeping the rest constant. We train the model for only 75000 iterations (approximately 15 epochs for the chair class) for faster experimentation.

\paragraph{Discretizing Depth.}
We first want to see how varying the resolution for depth affects the performance of our model. Table~\ref{tab:ablation_resolution_recon} and~\ref{tab:ablation_resolution_svr} shows the reconstruction from all view-points and single view-point respectively, for different resolutions of the model. At first glance, it seems surprising that lower resolution seem to perform better in terms of reconstruction metrics. However, we attribute this finding to the fact that higher resolution models contain more parameter and need longer training to reach their best performance.

\begin{table}[h]
    \centering
    \caption{Reconstruction on test data (chair) with varying depth resolution}
    \resizebox{\linewidth}{!}{
    \begin{tabular}{@{}lccc@{}}
    \toprule
    precision ($2^\text{x}$) & CD (mean) & CD (median) & EMD \\
    \midrule
    5 & 0.62 & 0.34 & 0.0632 \\
    6 & 0.65 & 0.33 & 0.0676 \\
    7 & 0.73 & 0.34 & 0.0732 \\
    8 & 0.87 & 0.38 & 0.0853 \\
    \bottomrule
    \end{tabular}
    }
    \label{tab:ablation_resolution_recon}
\end{table}

\begin{table}[h]
    \centering
    \caption{Single View Reconstruction (chair) with varying depth resolution}
    \resizebox{\linewidth}{!}{
     \begin{tabular}{@{}lccc@{}}
    \toprule    
    precision ($2^\text{x}$) & CD (mean) & CD (median) & EMD \\
    \midrule
    5 & 0.76 & 0.45 & 0.0737 \\
    6 & 0.94 & 0.49 & 0.0791 \\
    7 & 1.03 & 0.40 & 0.0860 \\
    8 & 1.25 & 0.56 & 0.0989 \\
    \bottomrule
    \end{tabular}
    }
    \label{tab:ablation_resolution_svr}
\end{table}

\paragraph{Varying $V_{in}$ and $V_{out}$.}
\label{sec:batch}

Next we analyze our choice of  $V_{in}$ and $V_{out}$ for all the experiments we performed. Having smaller $V_{in}$ allows for having larger batch sizes, and hence, model to see more variety of shapes in a mini-batch. However, having larger $V_{in}$ allows for model to learn better encodings from various viewpoints of same shape. We found that $V_{in}=V_{out}=2$ as a good balance between quantity and quality of mini-batch. Table~\ref{tab:ablation_batch_recon} and~\ref{tab:ablation_batch_svr} demonstrate quantitatively the impact of varying $V_{in}$ and $V_{out}$ on reconstruction metrics.

\begin{table}[h]
    \centering
    \caption{Reconstruction on test data (chair) with varying $V_{in}$}
    \resizebox{\linewidth}{!}{
     \begin{tabular}{@{}llccc@{}}
    \toprule
    Batch Size & $V_{in}$ & CD (mean) & CD (median) & EMD \\
    \midrule
    48  & 1 & 0.98 & 0.42 & 0.0925 \\
    24  & 2 & 0.87 & 0.38 & 0.0853 \\
    16  & 3 & 0.91 & 0.45 & 0.0908 \\
    12  & 4 & 0.96 & 0.43 & 0.0855 \\
    \bottomrule
    \end{tabular}
    }
    \label{tab:ablation_batch_recon}
\end{table}

\begin{table}[h]
    \centering
    \caption{Single View Reconstruction (chair) with varying $V_{in}$}
    \resizebox{\linewidth}{!}{
     \begin{tabular}{@{}llccc@{}}
    \toprule
    Batch Size & $V_{in}$ & CD (mean) & CD (median) & EMD \\
    \midrule
    48  & 1 & 1.25 & 0.55 & 0.0988 \\
    24  & 2 & 1.25 & 0.56 & 0.0989 \\
    16  & 3 & 1.45 & 0.72 & 0.1079 \\
    12  & 4 & 1.46 & 0.70 & 0.1036 \\
    \bottomrule
    \end{tabular}
    }
    \label{tab:ablation_batch_svr}
\end{table}

\paragraph{Orthographic vs Perspective.}

We further study the impact of perspective and orthographic depth map outputs. We observed that model learns faster and shows better performance with orthographic projection outputs. We hypothesize that orthographic depth outputs contain more points to learn from within a single image. Table~\ref{tab:ablation_projection_recon} and~\ref{tab:ablation_projection_svr} demonstrate the differences between the two projections.

\begin{table}[h]
    \centering
    \caption{Reconstruction on test data (chair) with varying projection}
    \resizebox{\linewidth}{!}{
     \begin{tabular}{@{}lccc@{}}
    \toprule
   Projection & CD (mean) & CD (median) & EMD \\
    \midrule
    Orthographic  & 0.87 & 0.38 & 0.0853 \\
    Perspective   & 1.1 & 0.72 & 0.0886 \\
    \bottomrule
    \end{tabular}
    }
    \label{tab:ablation_projection_recon}
\end{table}

\begin{table}[h]
    \centering
    \caption{Single View Reconstruction (chair) with varying projection}
    \resizebox{\linewidth}{!}{
     \begin{tabular}{@{}lccc@{}}
    \toprule
   Projection & CD (mean) & CD (median) & EMD \\
    \midrule
    Orthographic  & 0.87 & 0.38 & 0.0853 \\
    Perspective   & 1.7 & 0.97 & 0.1019 \\
    \bottomrule
    \end{tabular}
    }
    \label{tab:ablation_projection_svr}
\end{table}

\section{Generative Modeling}

\begin{figure*}
\centering

\includegraphics[clip,trim=3cm 20cm 3cm 20cm, width=0.85\textwidth]{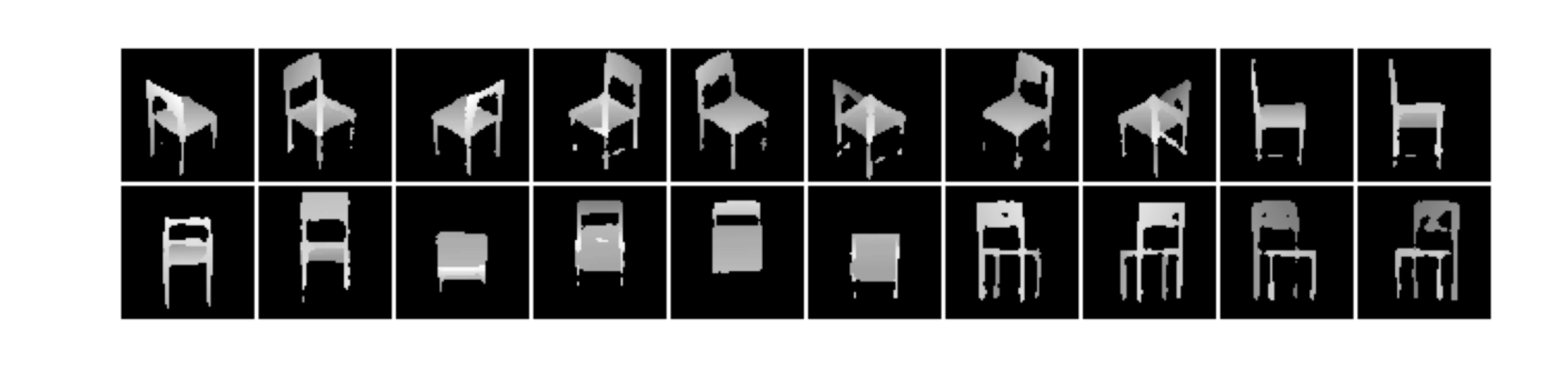}
\raisebox{0.5\height}{\centering{\includegraphics[clip,trim=3cm 3cm 3cm 3cm, width=0.1\textwidth]{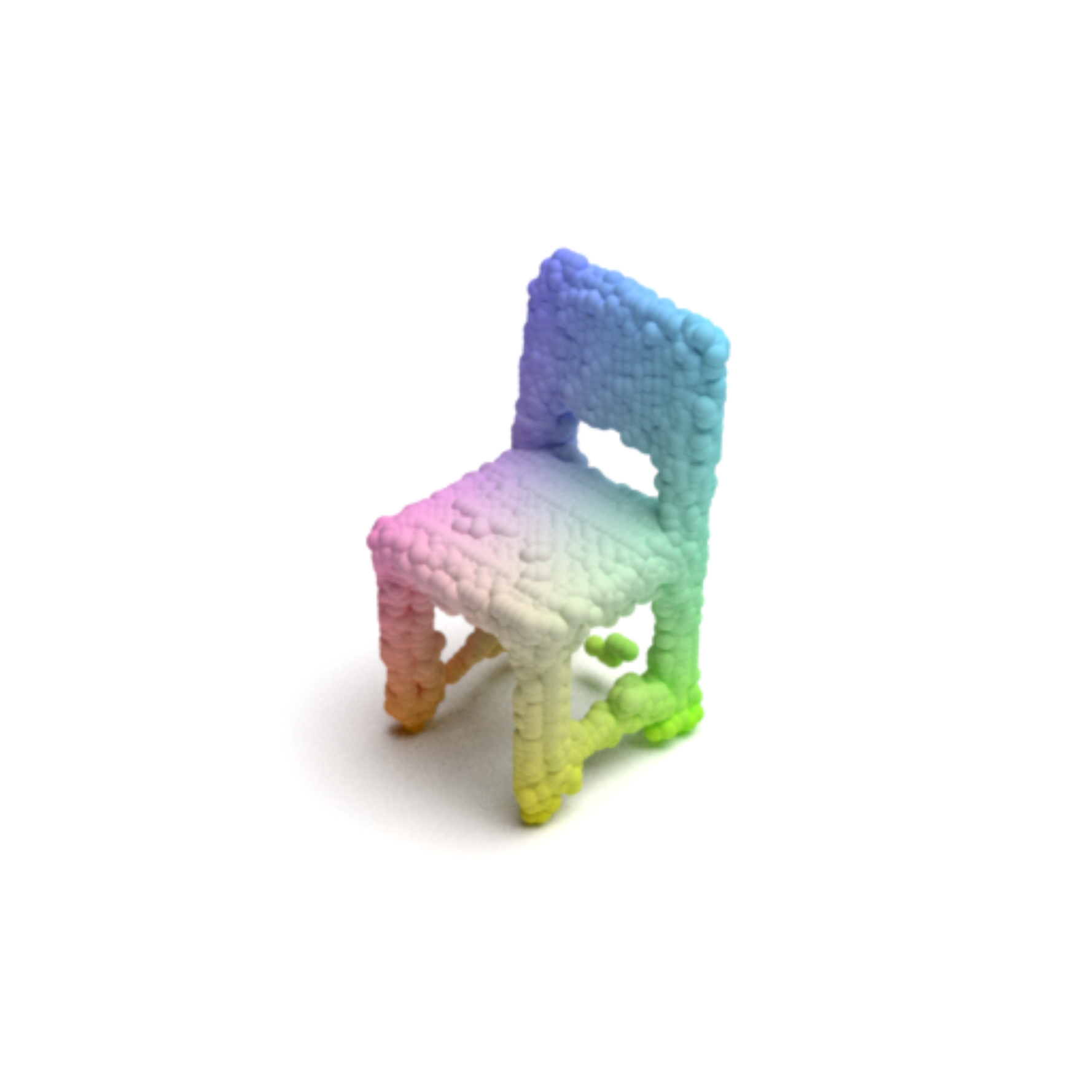}}}
\includegraphics[clip,trim=3cm 20cm 3cm 20cm, width=0.85\textwidth]{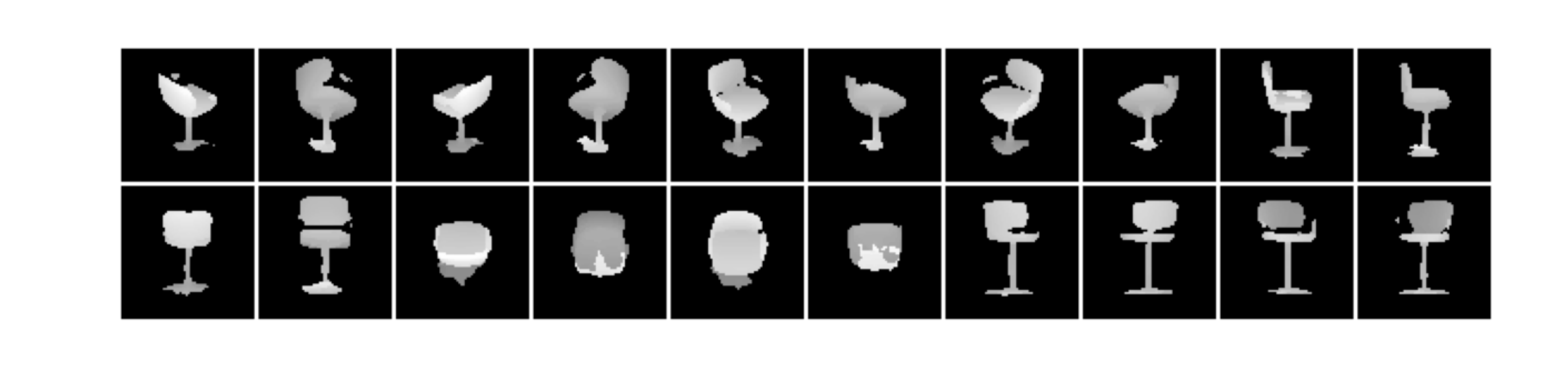}
\raisebox{0.5\height}{\centering{\includegraphics[clip,trim=3cm 3cm 3cm 3cm, width=0.1\textwidth]{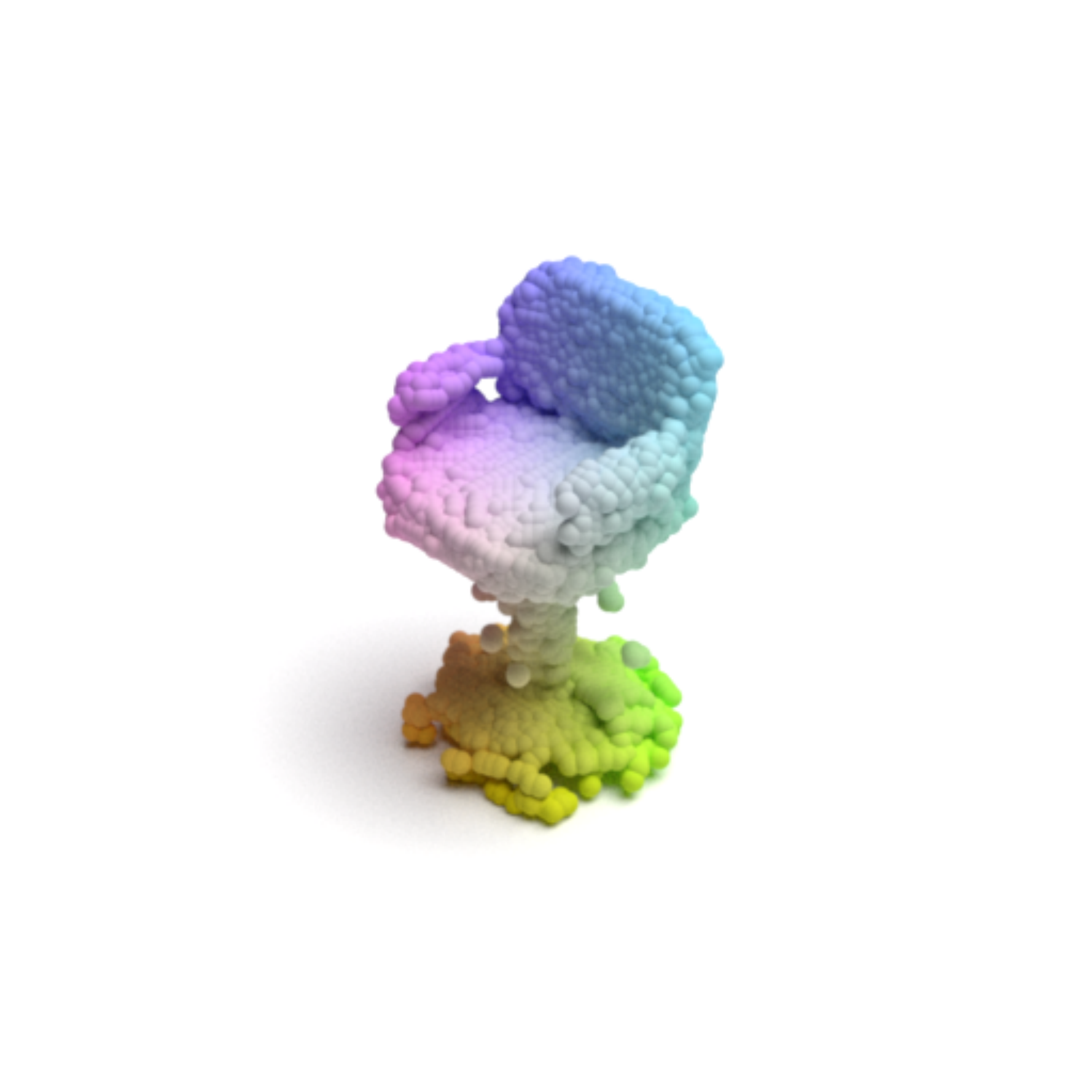}}}
\includegraphics[clip,trim=3cm 20cm 3cm 20cm, width=0.85\textwidth]{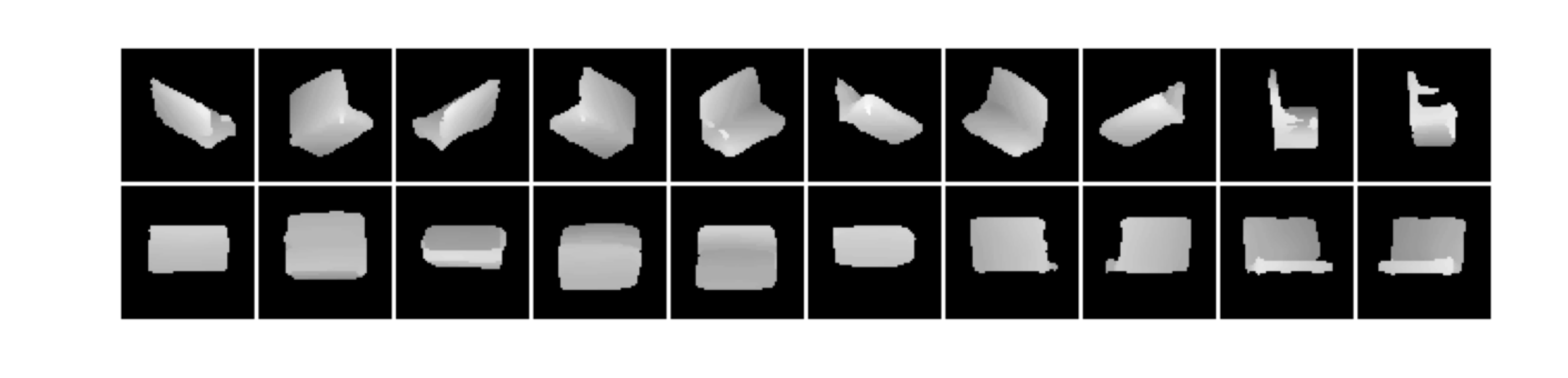}
\raisebox{0.5\height}{\centering{\includegraphics[clip,trim=3cm 3cm 3cm 3cm, width=0.1\textwidth]{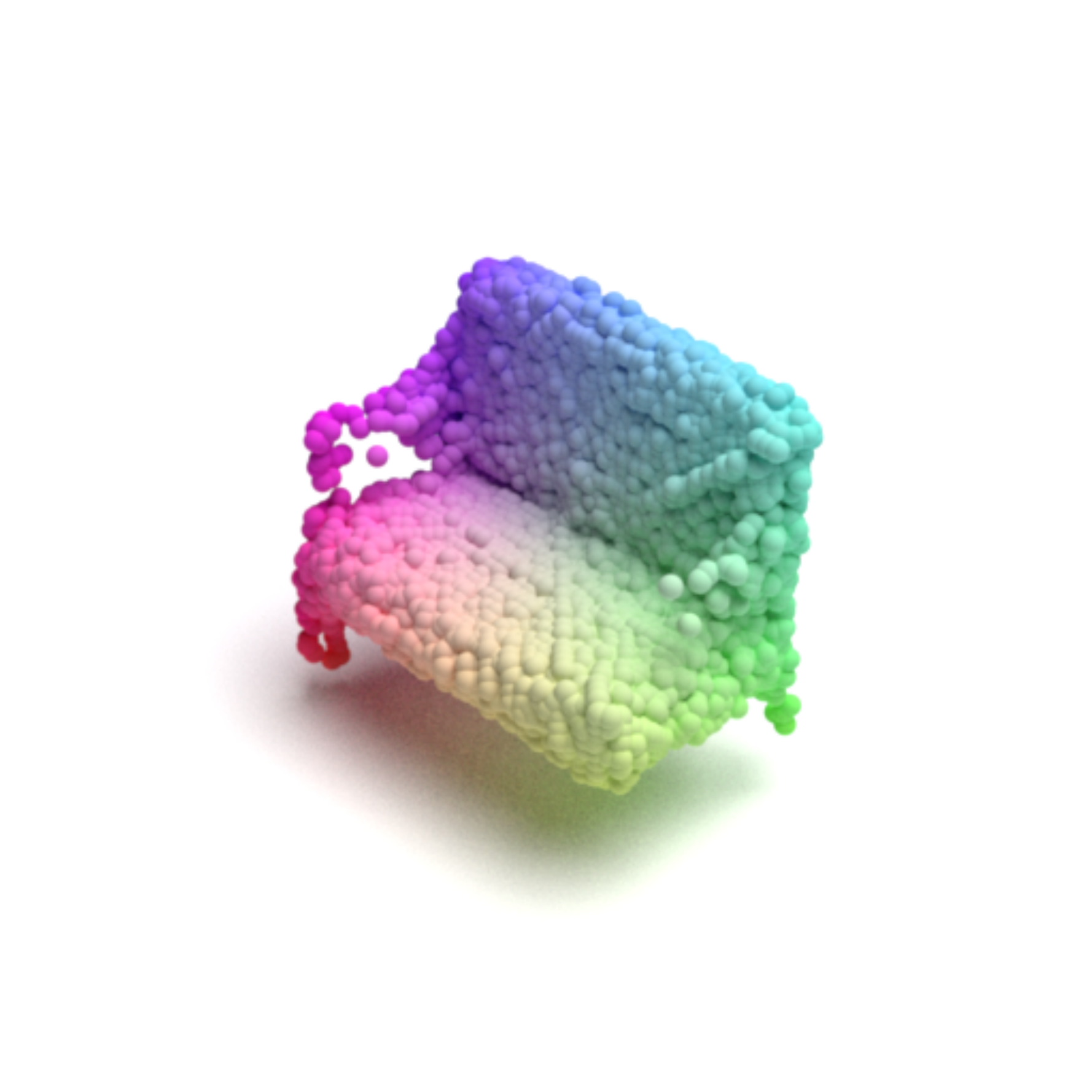}}}
\includegraphics[clip,trim=3cm 20cm 3cm 20cm, width=0.85\textwidth]{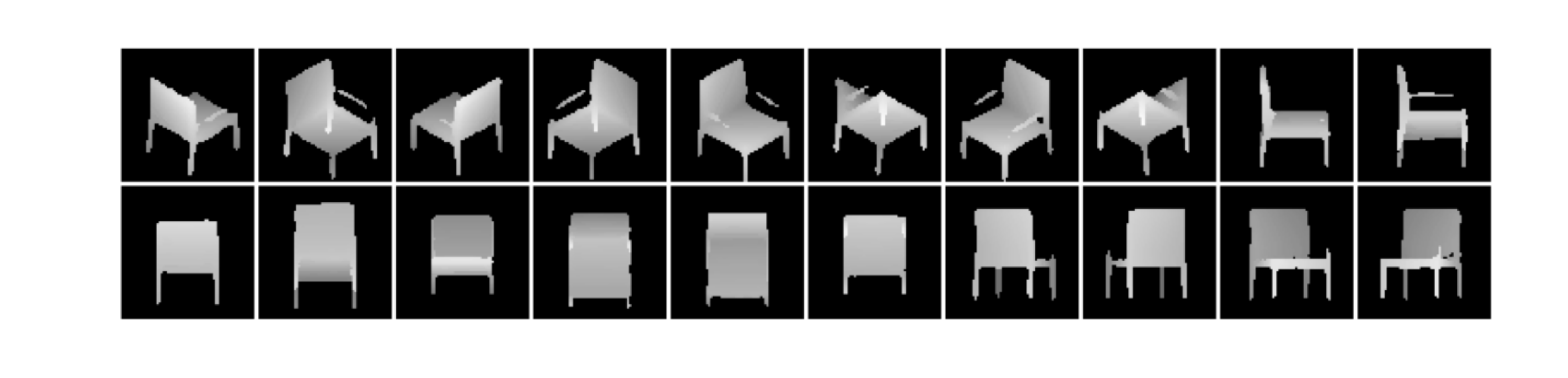}
\raisebox{0.5\height}{\centering{\includegraphics[clip,trim=3cm 3cm 3cm 3cm, width=0.1\textwidth]{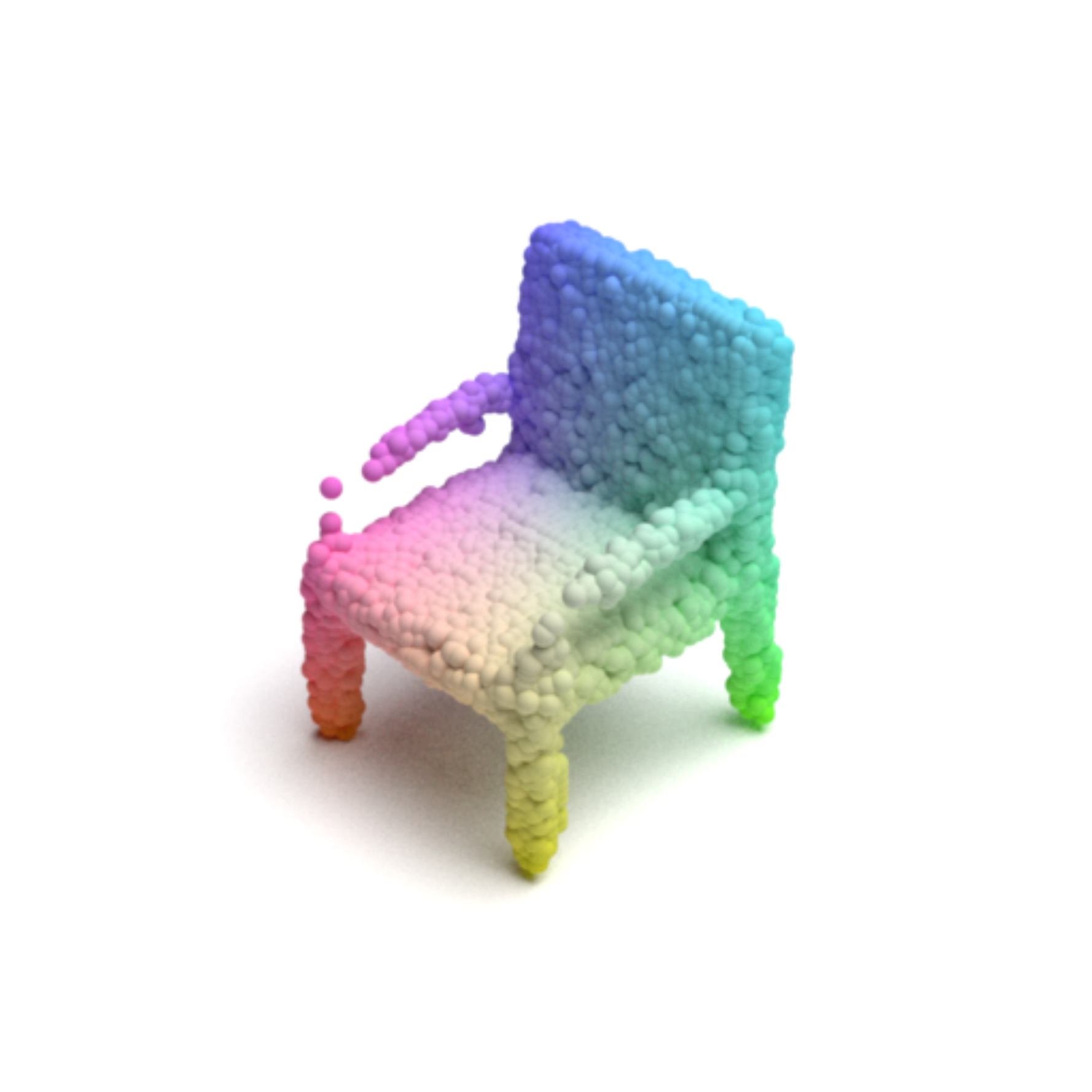}}}
\includegraphics[clip,trim=3cm 20cm 3cm 20cm, width=0.85\textwidth]{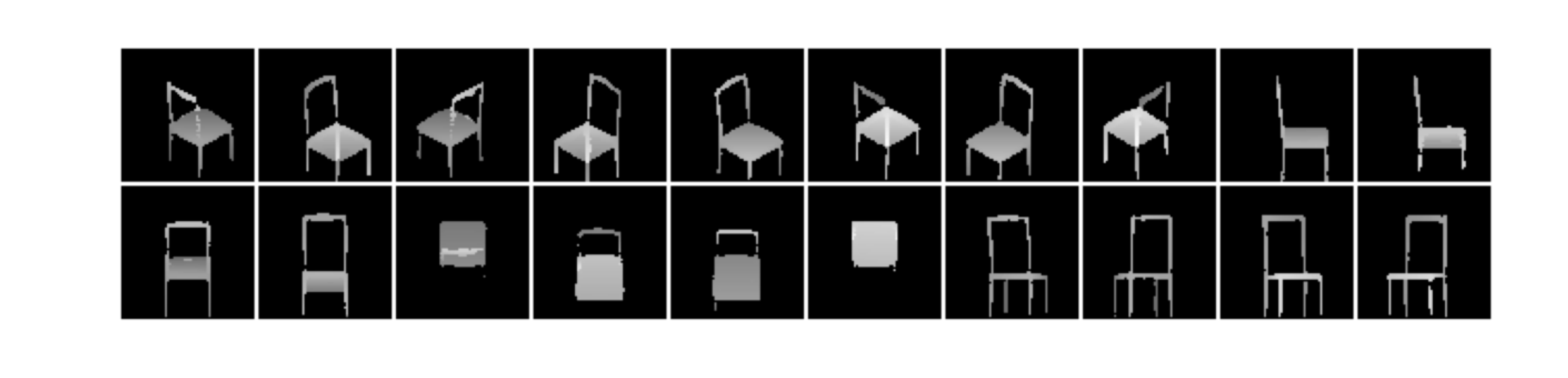}
\raisebox{0.5\height}{\centering{\includegraphics[clip,trim=3cm 3cm 3cm 3cm, width=0.1\textwidth]{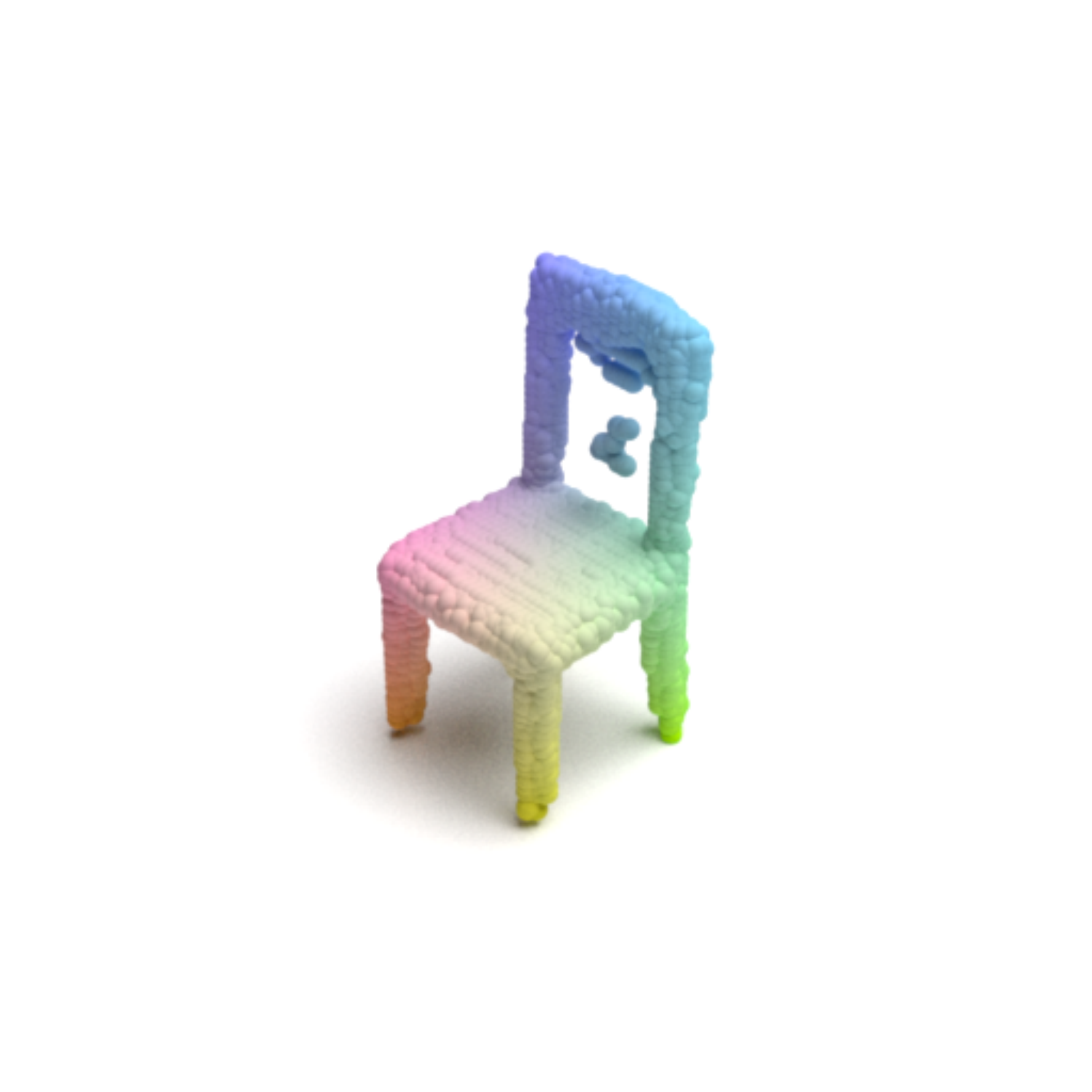}}}

\captionof{figure}{\textbf{Synthesized chair objects} Each row shows the depth maps of the object from 20 views and corresponding 3D shape}
\label{fig:gen_dm_chair}
\end{figure*}

\begin{figure*}
\centering

\includegraphics[clip,trim=3cm 20cm 3cm 20cm, width=0.85\textwidth]{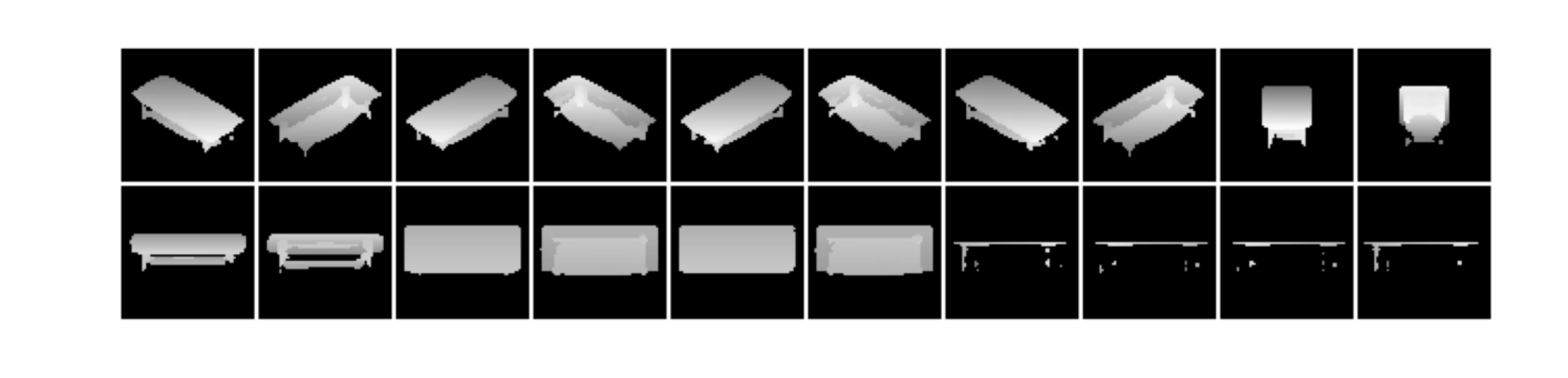}
\raisebox{0.5\height}{\centering{\includegraphics[clip,trim=3cm 3cm 3cm 3cm, width=0.1\textwidth]{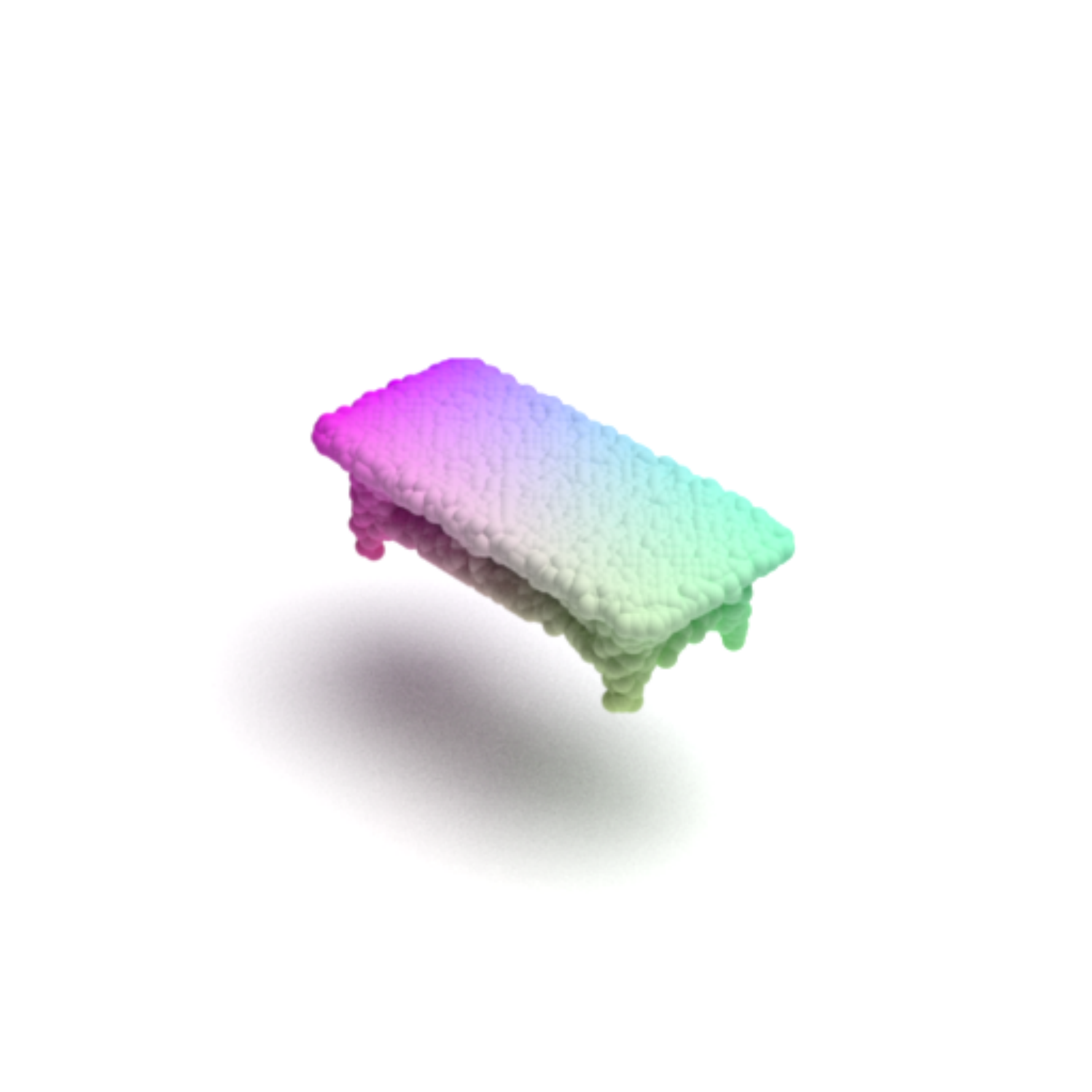}}}
\includegraphics[clip,trim=3cm 20cm 3cm 20cm, width=0.85\textwidth]{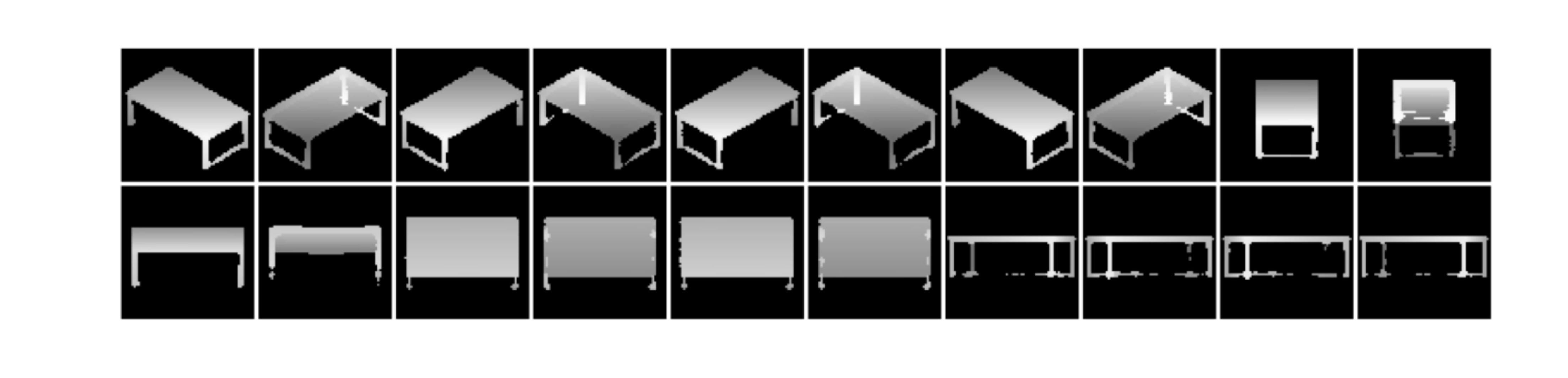}
\raisebox{0.5\height}{\centering{\includegraphics[clip,trim=3cm 3cm 3cm 3cm, width=0.1\textwidth]{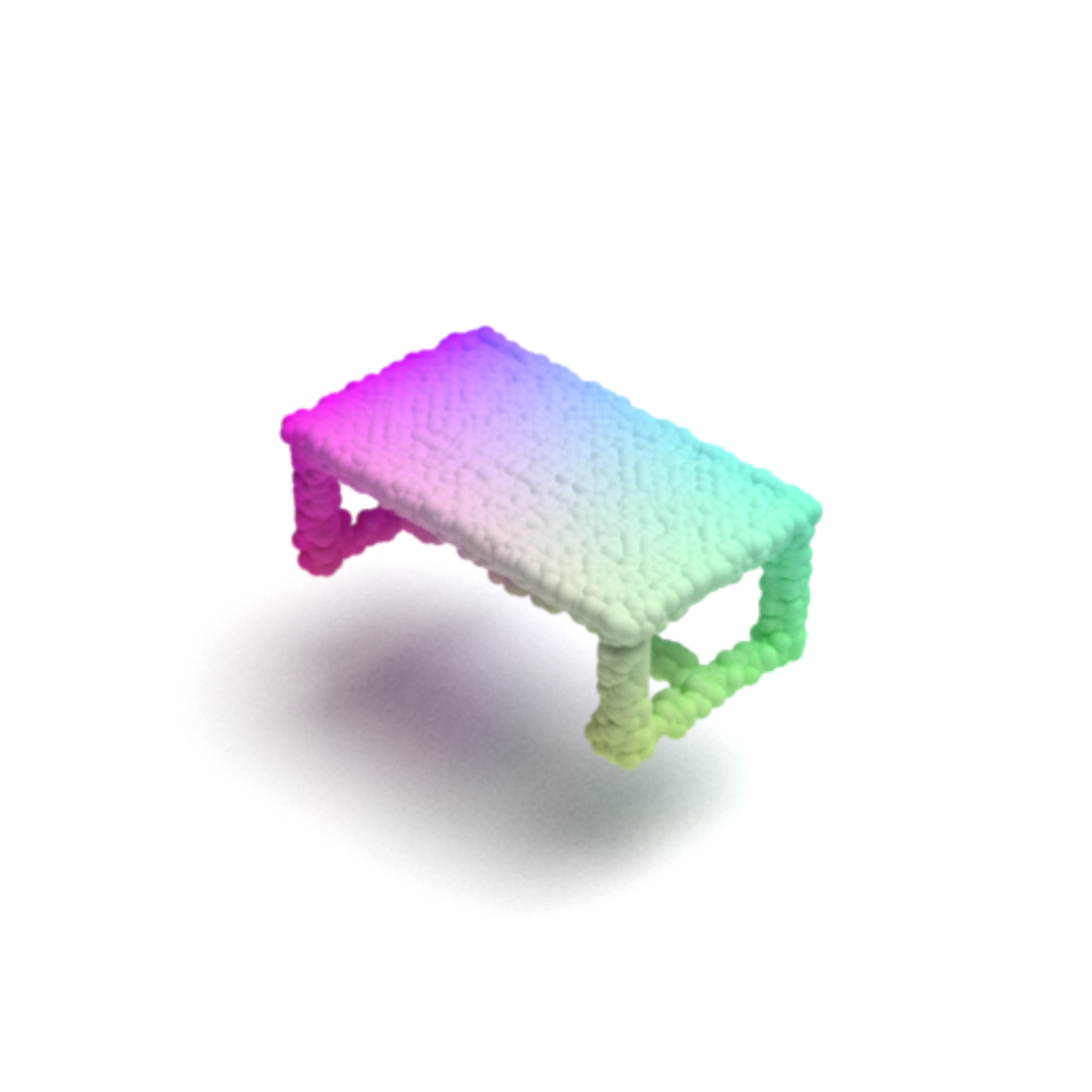}}}
\includegraphics[clip,trim=3cm 20cm 3cm 20cm, width=0.85\textwidth]{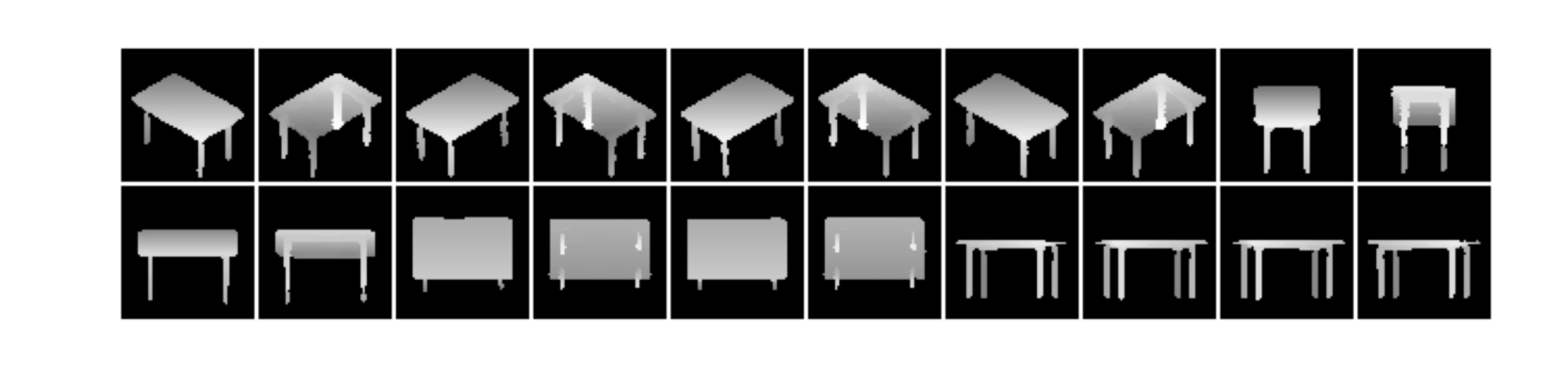}
\raisebox{0.5\height}{\centering{\includegraphics[clip,trim=3cm 3cm 3cm 3cm, width=0.1\textwidth]{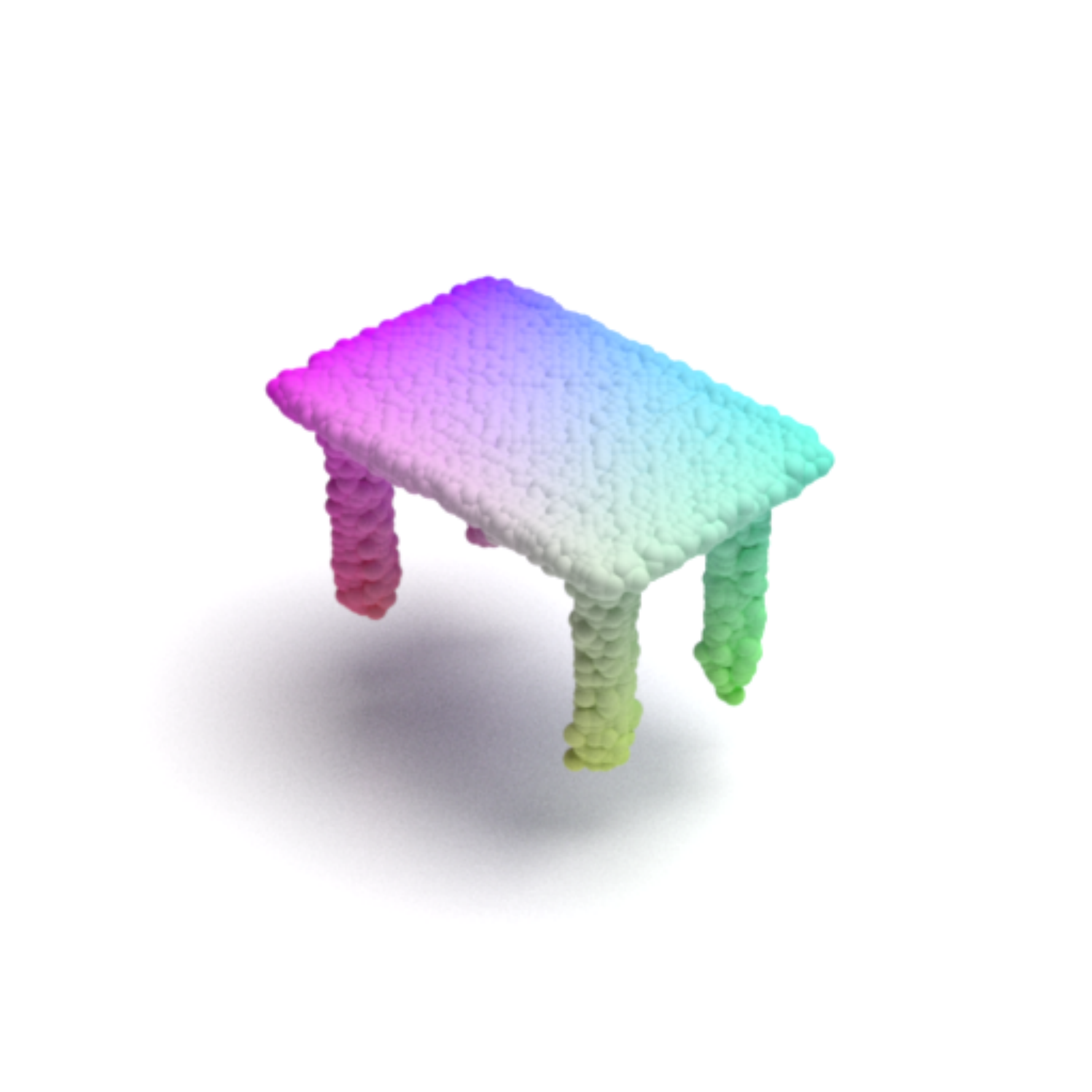}}}
\includegraphics[clip,trim=3cm 20cm 3cm 20cm, width=0.85\textwidth]{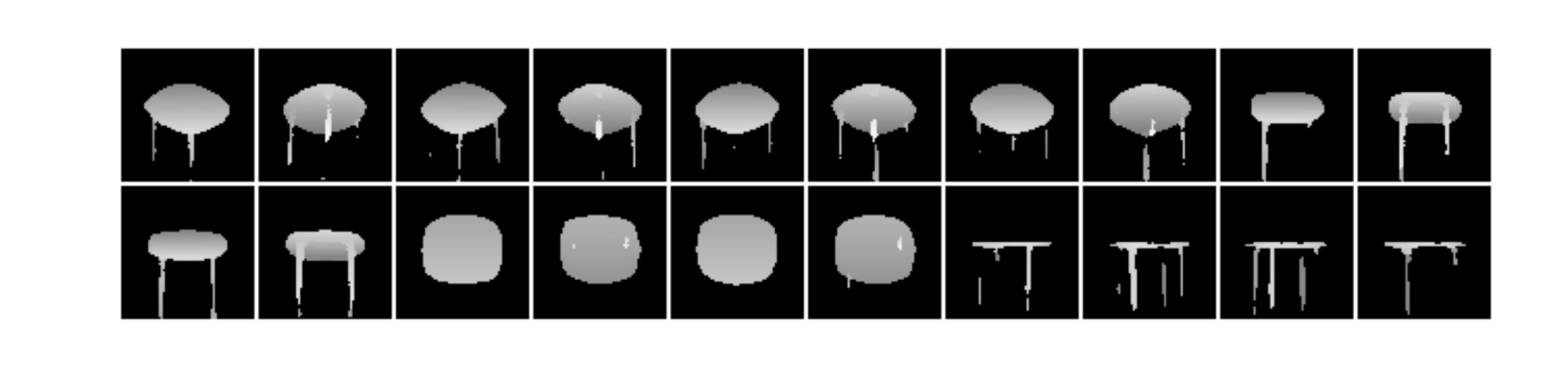}
\raisebox{0.5\height}{\centering{\includegraphics[clip,trim=3cm 3cm 3cm 3cm, width=0.1\textwidth]{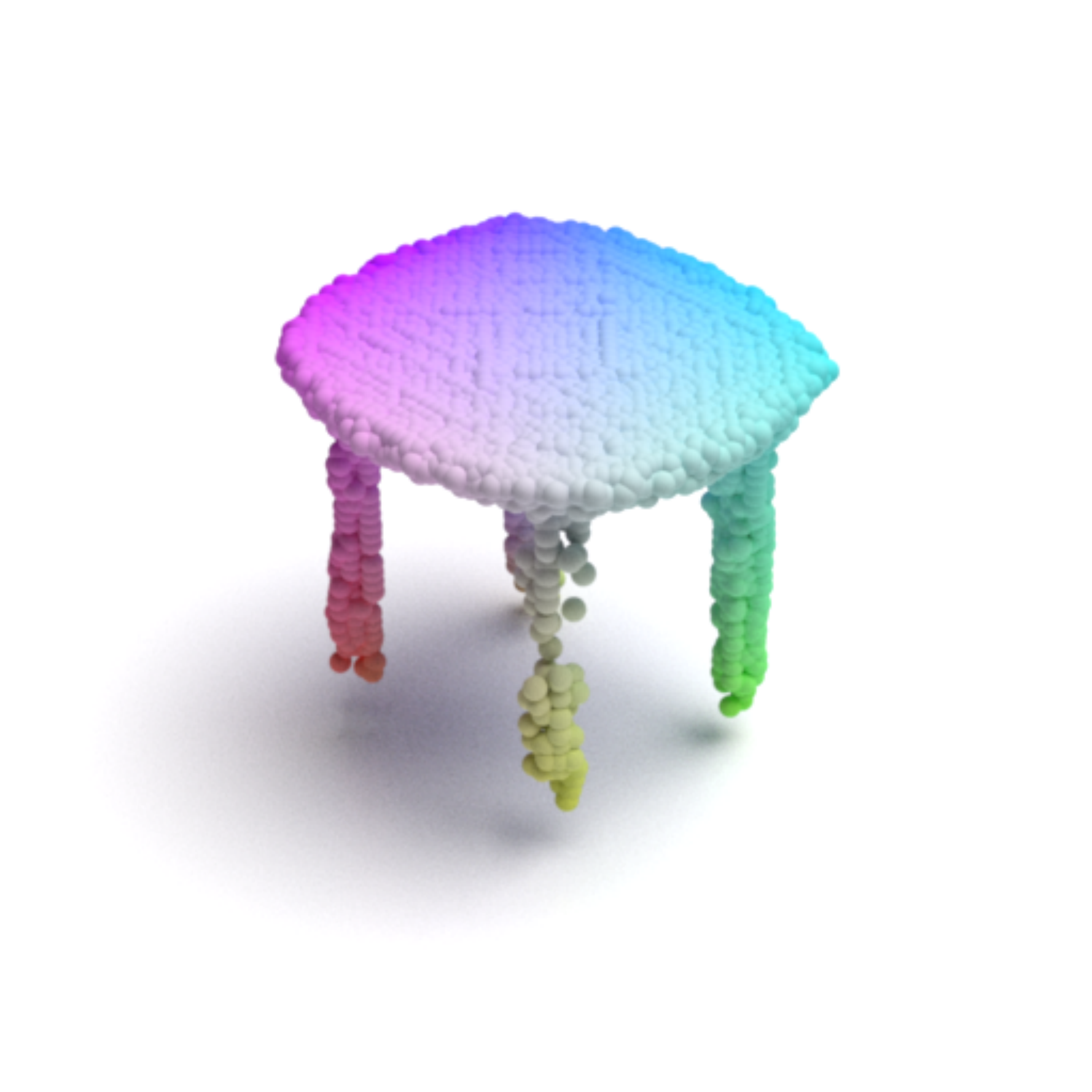}}}

\captionof{figure}{\textbf{Synthesized table objects} Each row shows the depth maps of the object from 20 views and corresponding 3D shape}
\label{fig:gen_dm_table}
\end{figure*}

\begin{figure*}
\centering
\hspace*{-0.5cm}%

\includegraphics[clip,trim=3cm 20cm 3cm 20cm, width=0.85\textwidth]{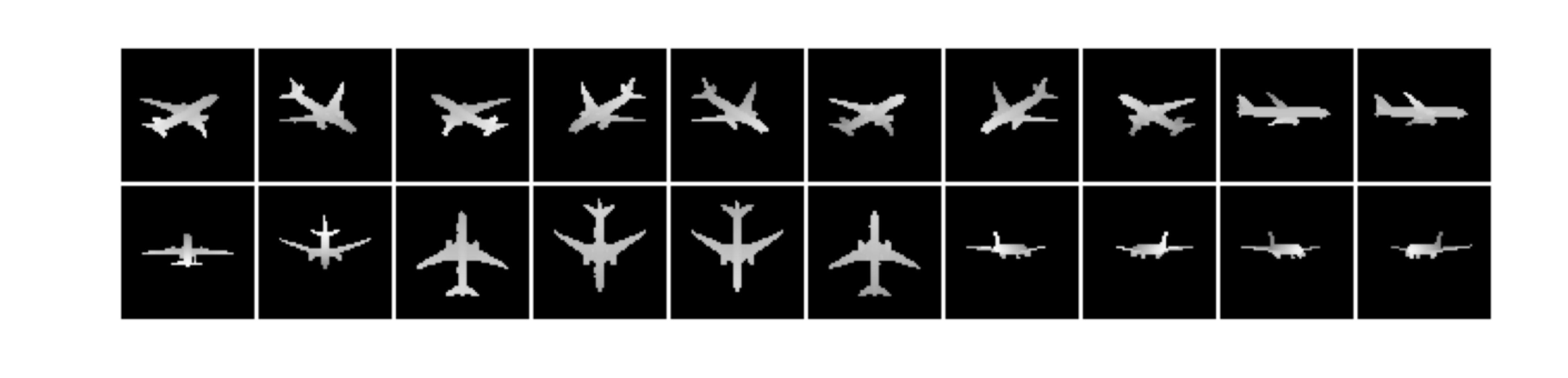}
\raisebox{0.5\height}{\centering{\includegraphics[clip,trim=3cm 3cm 3cm 3cm, width=0.1\textwidth]{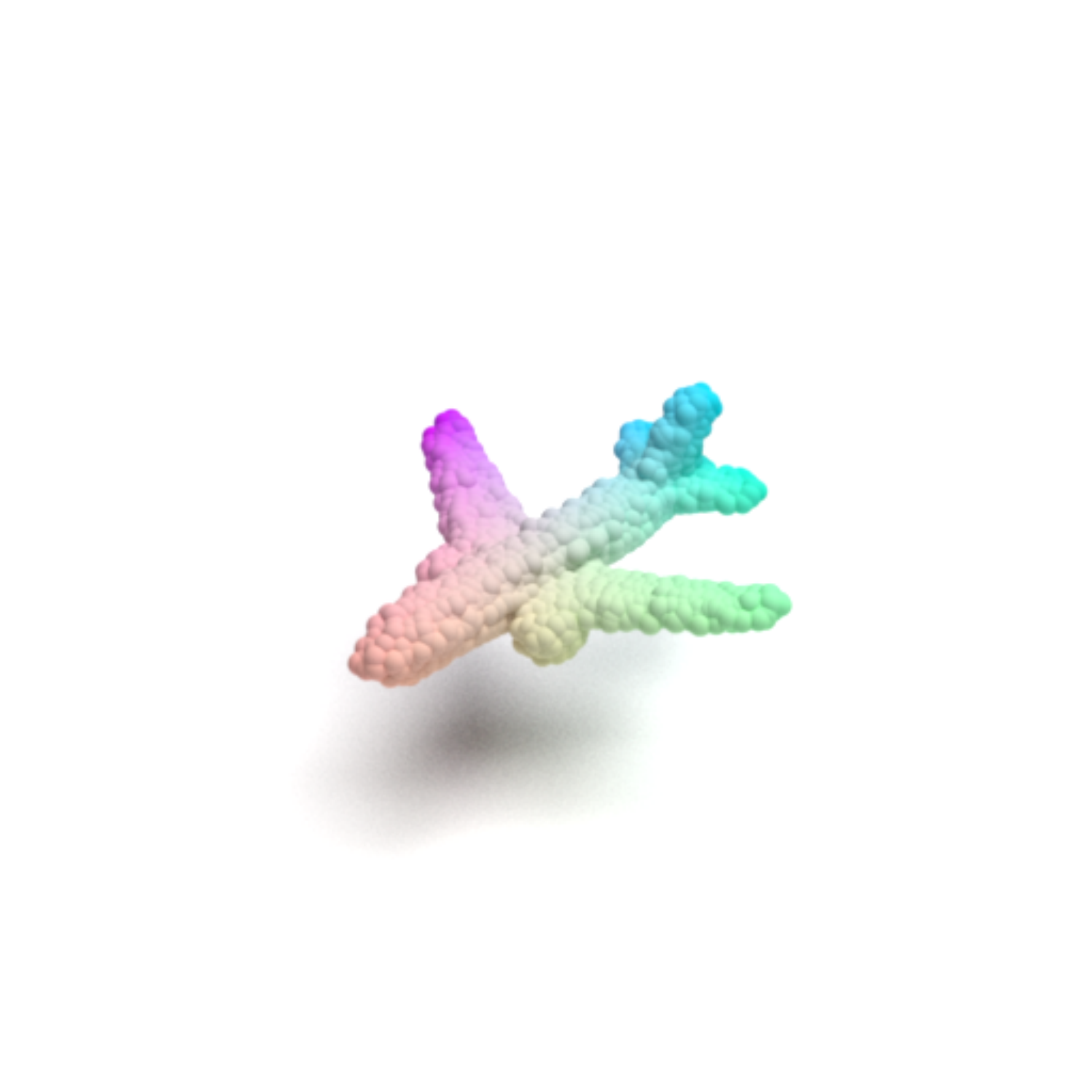}}}
\includegraphics[clip,trim=3cm 20cm 3cm 20cm, width=0.85\textwidth]{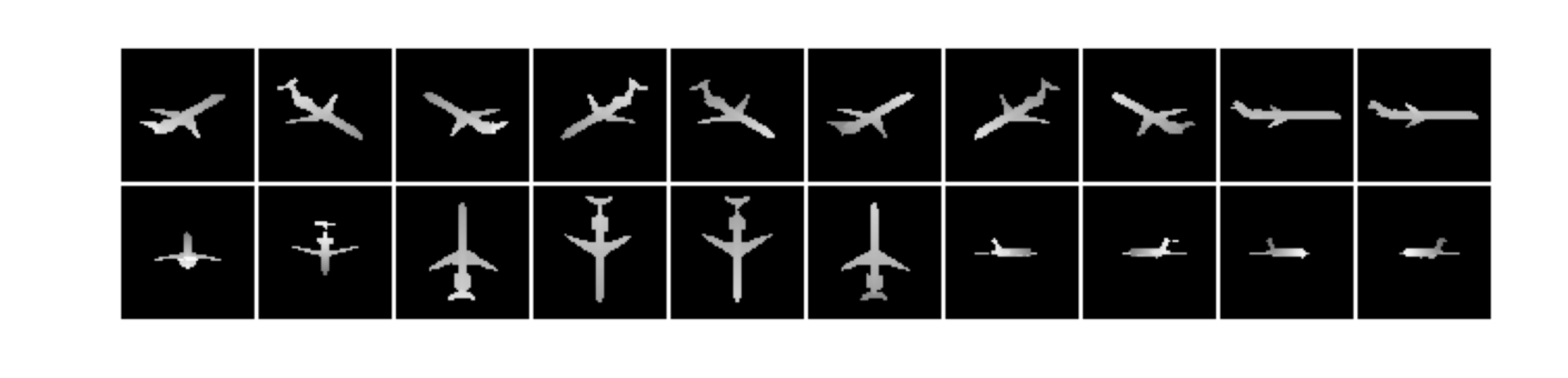}
\raisebox{0.5\height}{\centering{\includegraphics[clip,trim=3cm 3cm 3cm 3cm, width=0.1\textwidth]{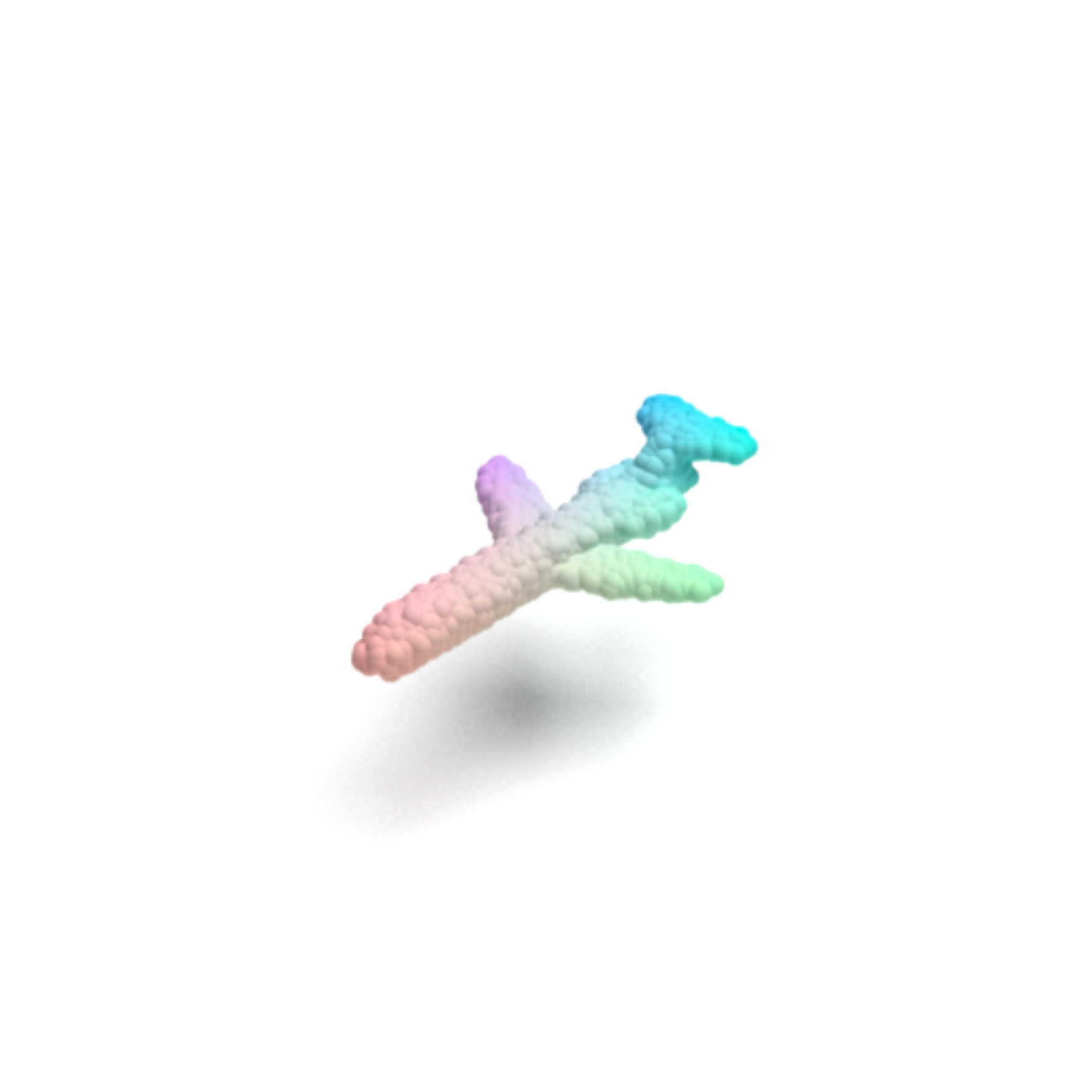}}}
\includegraphics[clip,trim=3cm 20cm 3cm 20cm, width=0.85\textwidth]{{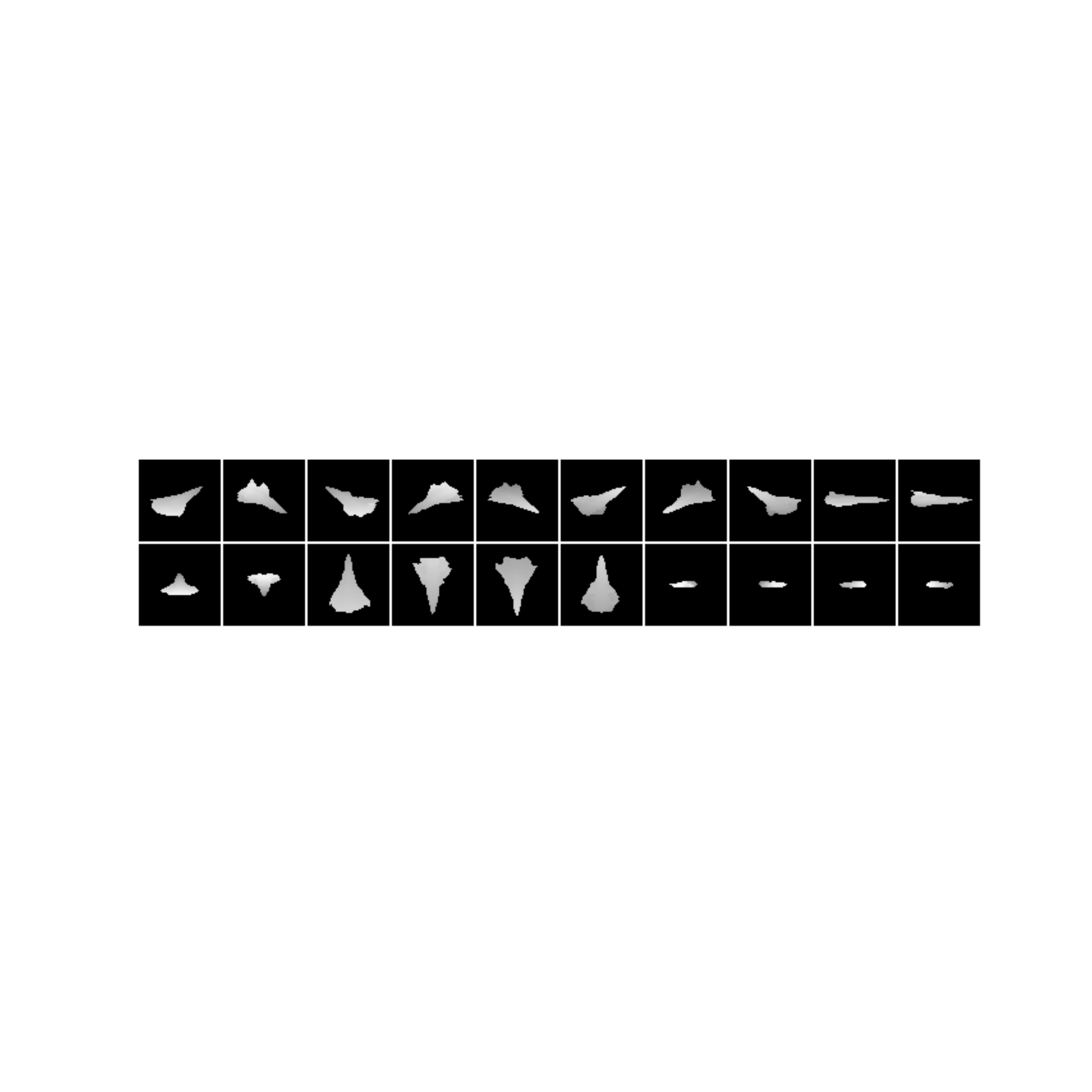}}
\raisebox{0.5\height}{\centering{\includegraphics[clip,trim=3cm 3cm 3cm 3cm, width=0.1\textwidth]{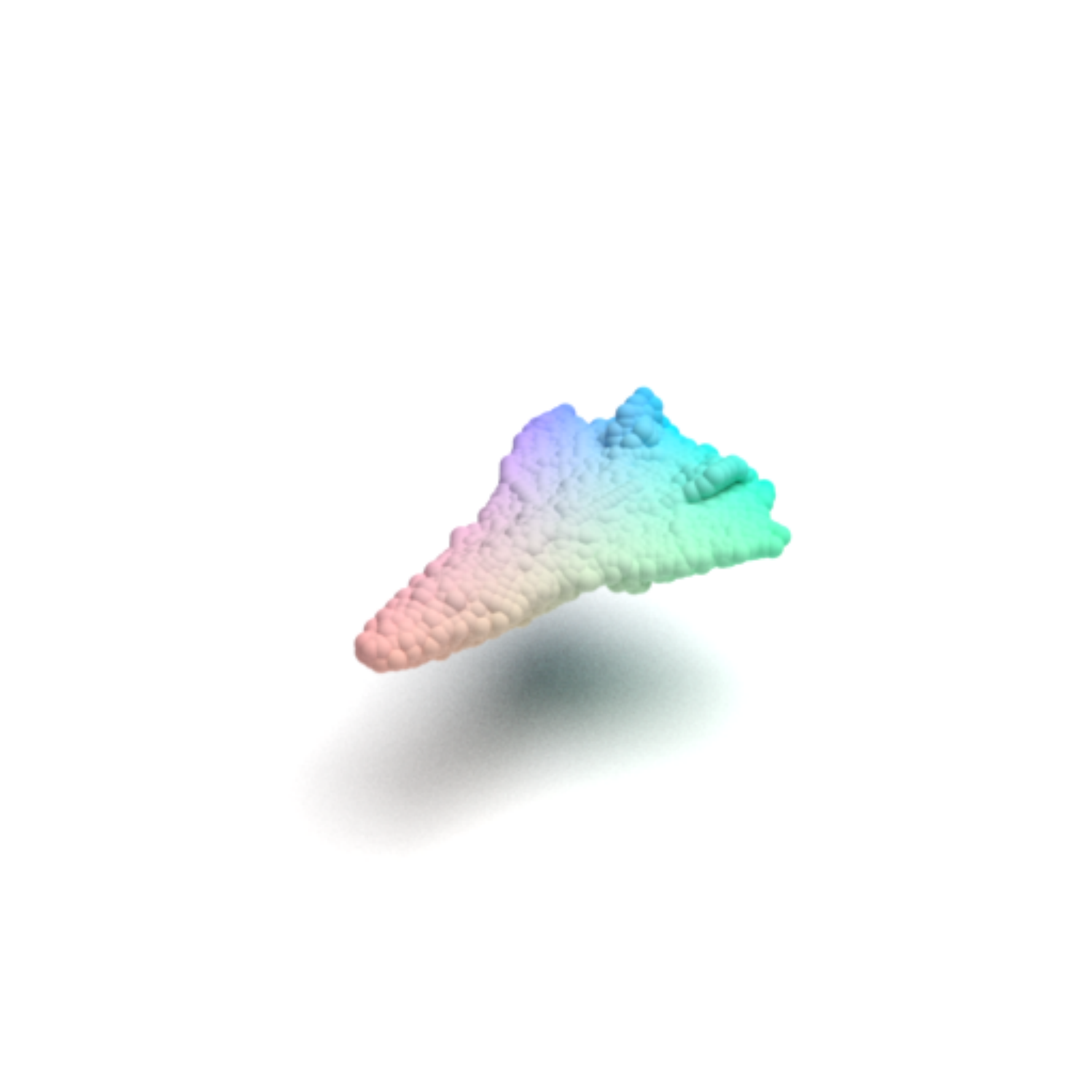}}}
\includegraphics[clip,trim=3cm 20cm 3cm 20cm, width=0.85\textwidth]{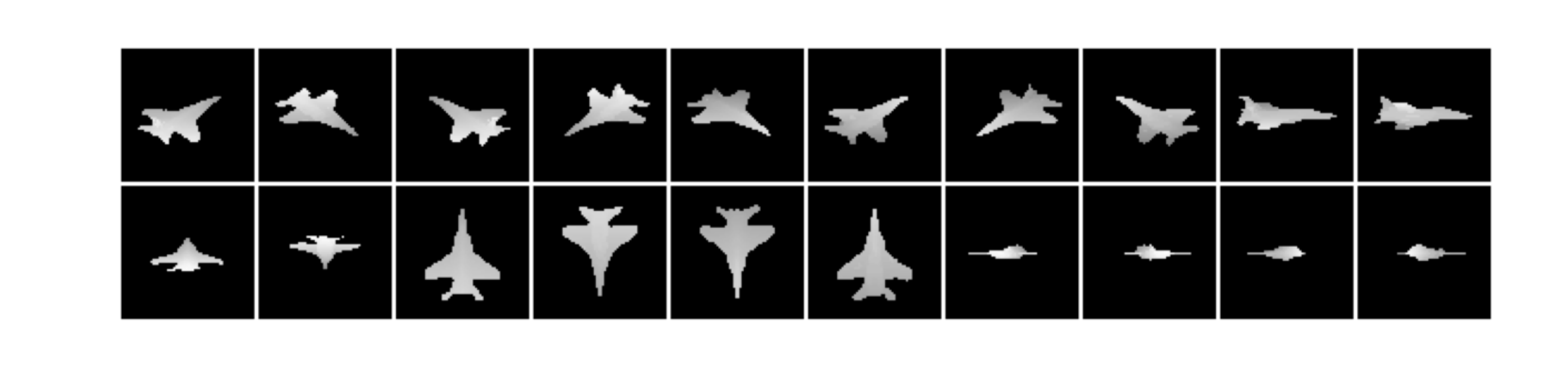}
\raisebox{0.5\height}{\centering{\includegraphics[clip,trim=3cm 3cm 3cm 3cm, width=0.1\textwidth]{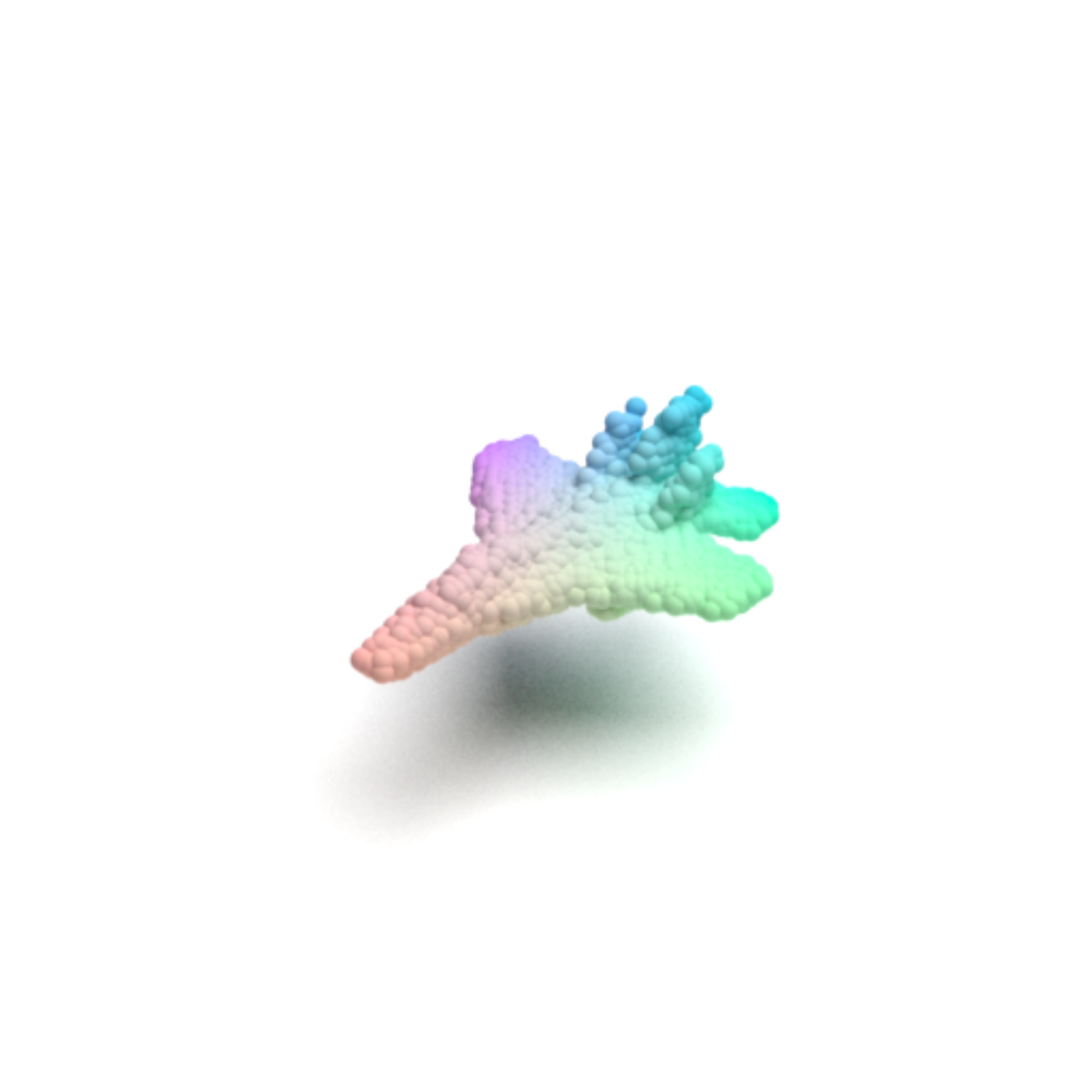}}}

\captionof{figure}{\textbf{Synthesized airplane objects} Each row shows the depth maps of the object from 20 views and corresponding 3D shape}
\label{fig:gen_dm_airplane}
\end{figure*}

\begin{figure*}
\centering
\hspace*{-0.5cm}%

\includegraphics[clip,trim=3cm 20cm 3cm 20cm, width=0.85\textwidth]{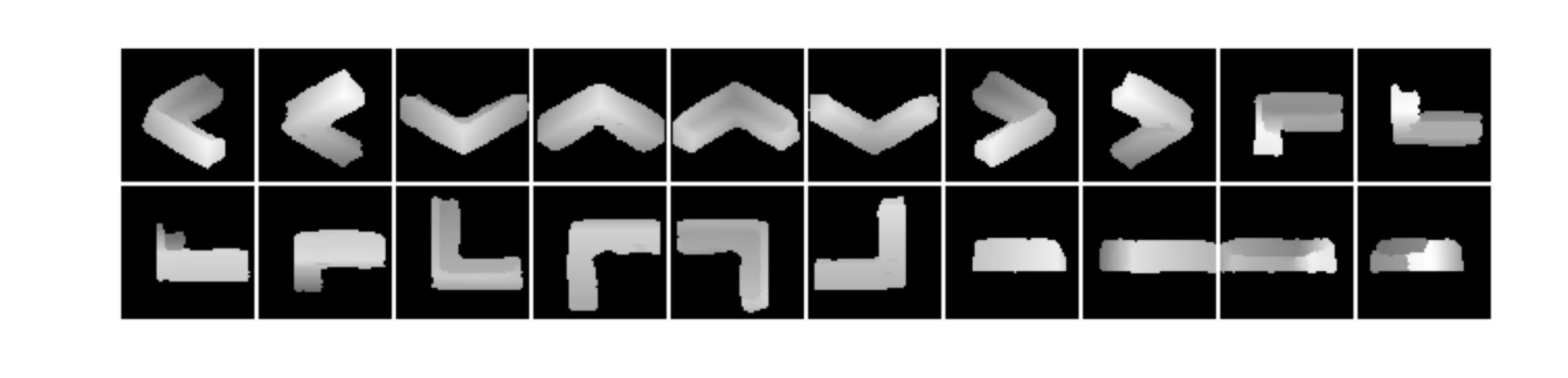}
\raisebox{0.5\height}{\centering{\includegraphics[clip,trim=3cm 3cm 3cm 3cm, width=0.1\textwidth]{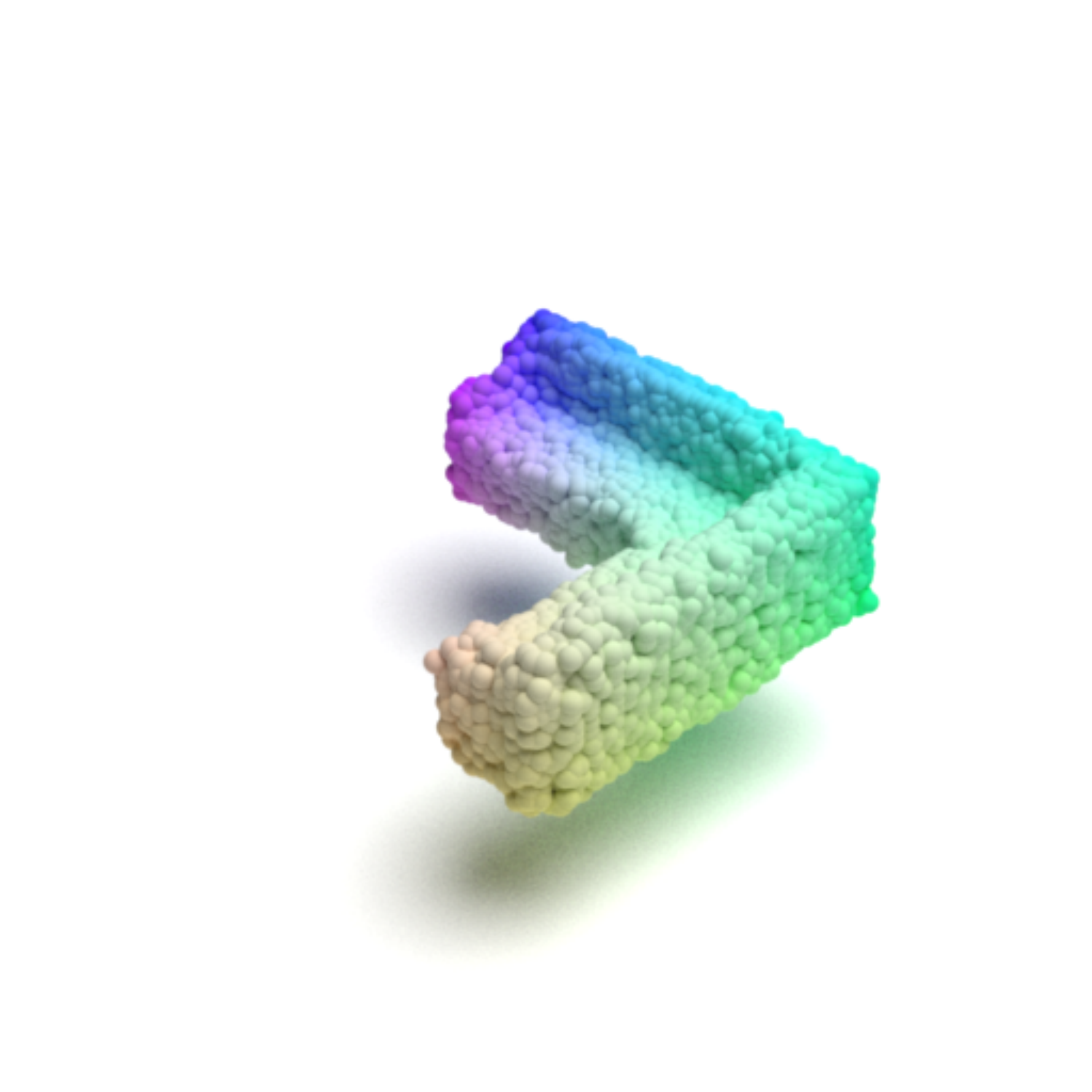}}}
\includegraphics[clip,trim=3cm 20cm 3cm 20cm, width=0.85\textwidth]{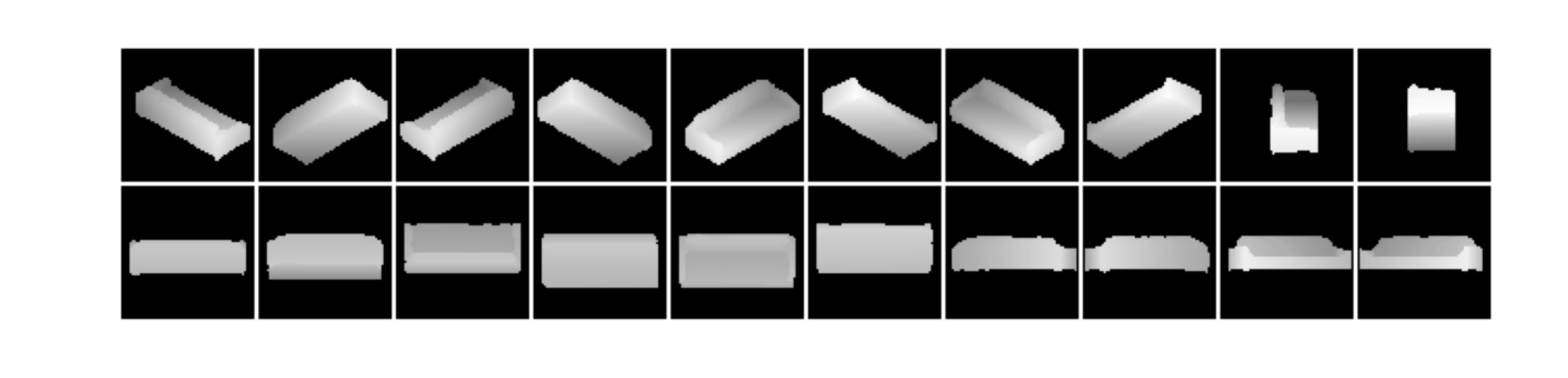}
\raisebox{0.5\height}{\centering{\includegraphics[clip,trim=3cm 3cm 3cm 3cm, width=0.1\textwidth]{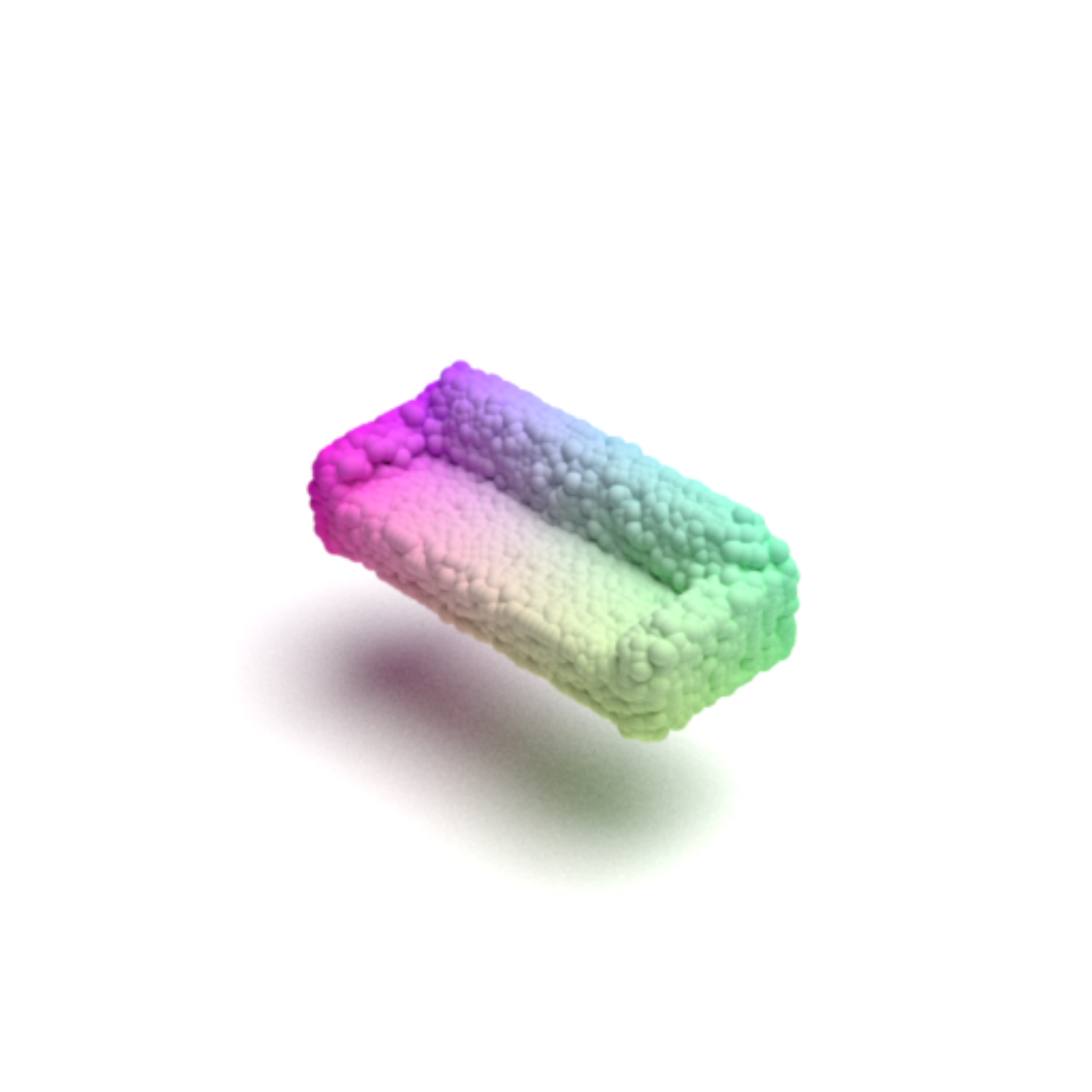}}}
\includegraphics[clip,trim=3cm 20cm 3cm 20cm, width=0.85\textwidth]{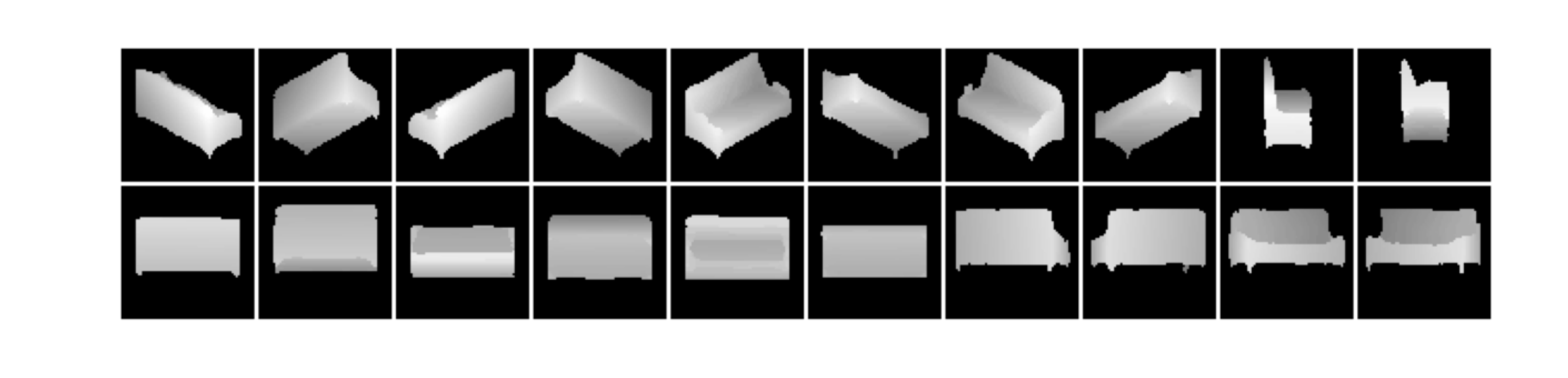}
\raisebox{0.5\height}{\centering{\includegraphics[clip,trim=3cm 3cm 3cm 3cm, width=0.1\textwidth]{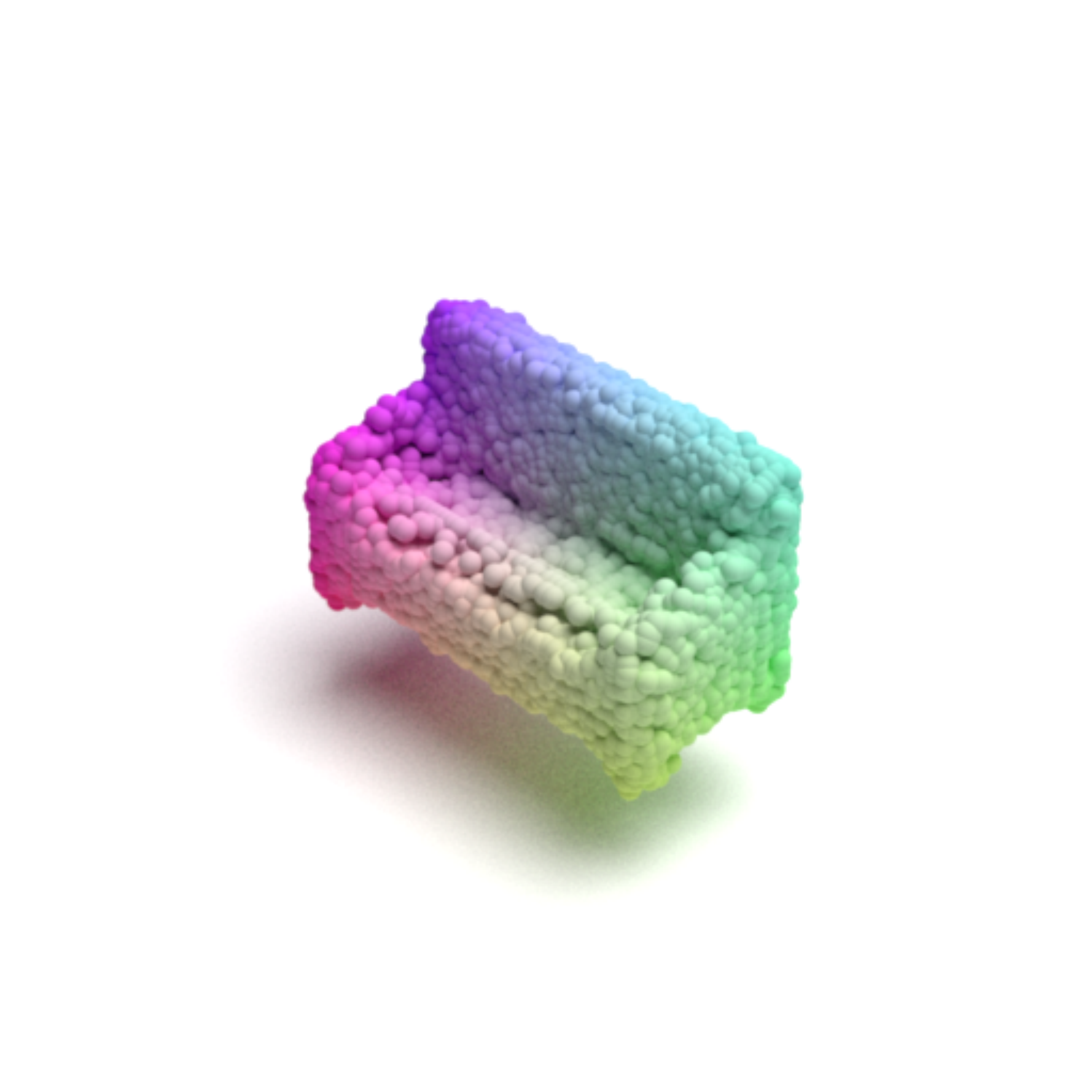}}}
\includegraphics[clip,trim=3cm 20cm 3cm 20cm, width=0.85\textwidth]{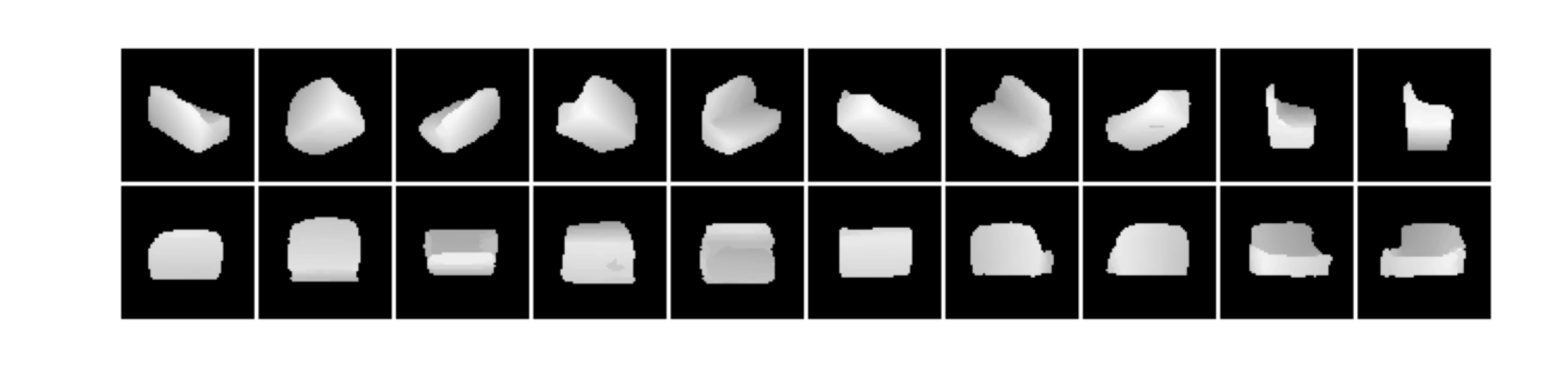}
\raisebox{0.5\height}{\centering{\includegraphics[clip,trim=3cm 3cm 3cm 3cm, width=0.1\textwidth]{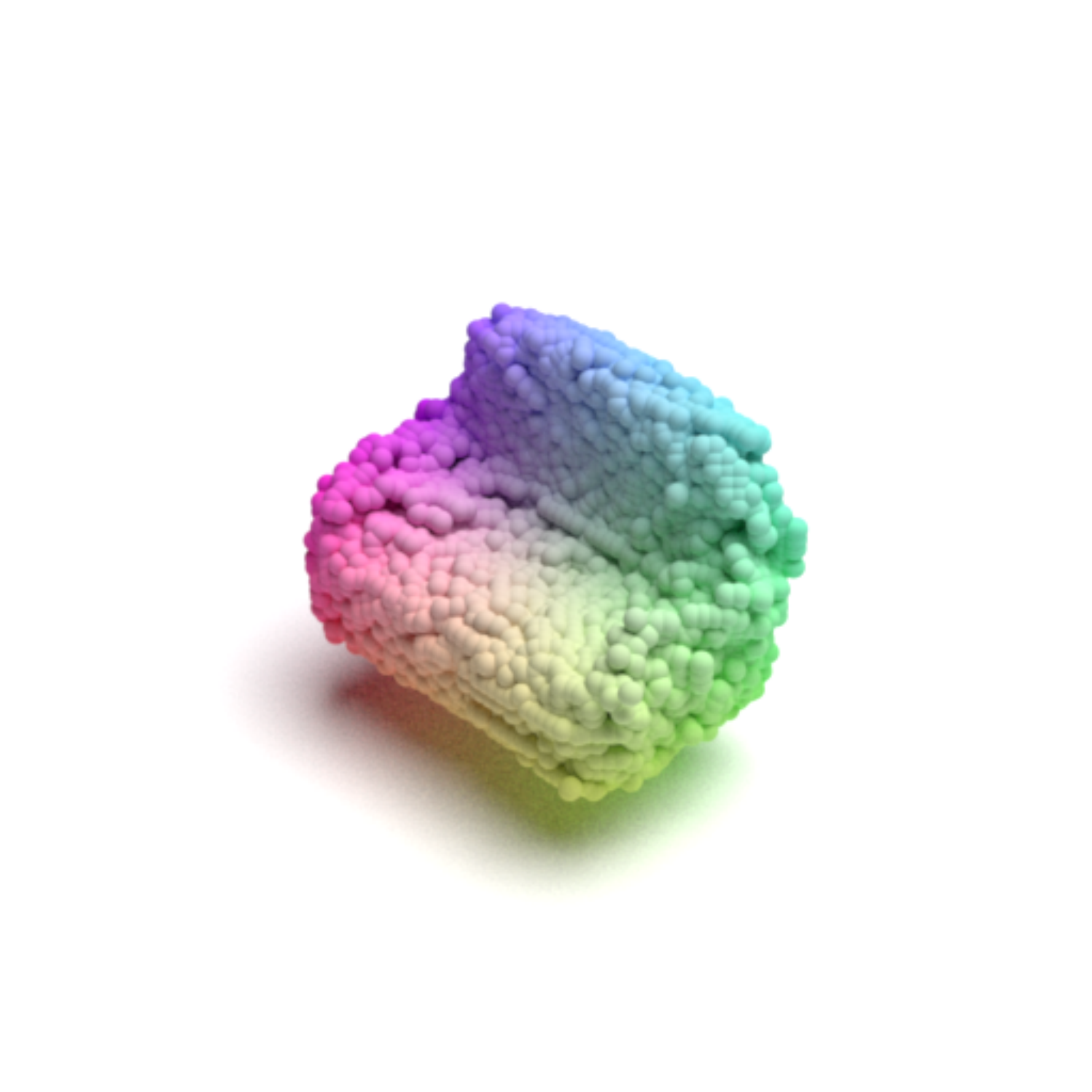}}}

\captionof{figure}{\textbf{Synthesized sofa objects} Each row shows the depth maps of the object from 20 views and corresponding 3D shape}
\label{fig:gen_dm_sofa}
\end{figure*}

\begin{figure*}
\centering
\hspace*{-0.5cm}%

\includegraphics[clip,trim=3cm 20cm 3cm 20cm, width=0.85\textwidth]{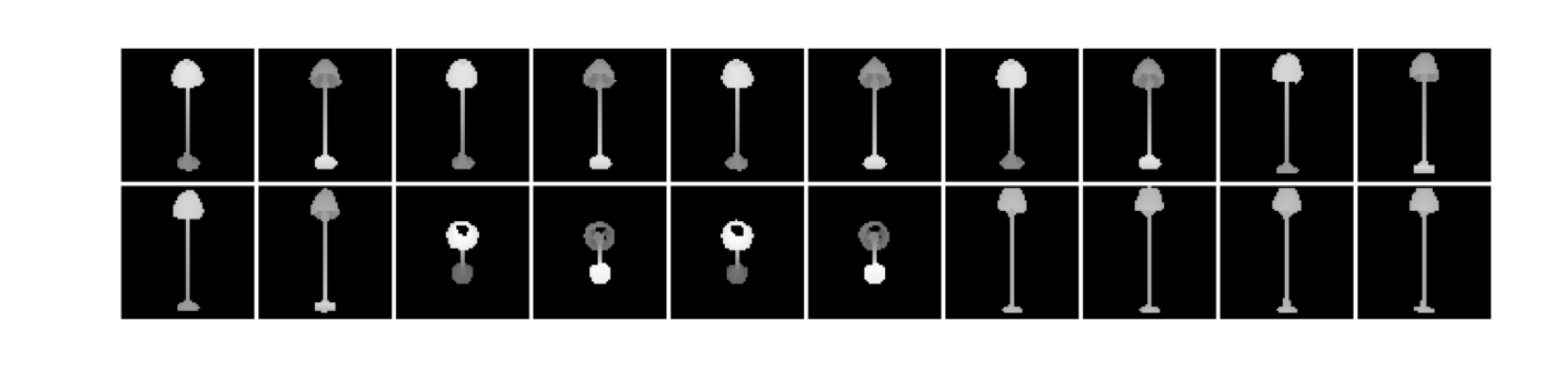}
\raisebox{0.5\height}{\centering{\includegraphics[clip,trim=3cm 3cm 3cm 3cm, width=0.1\textwidth]{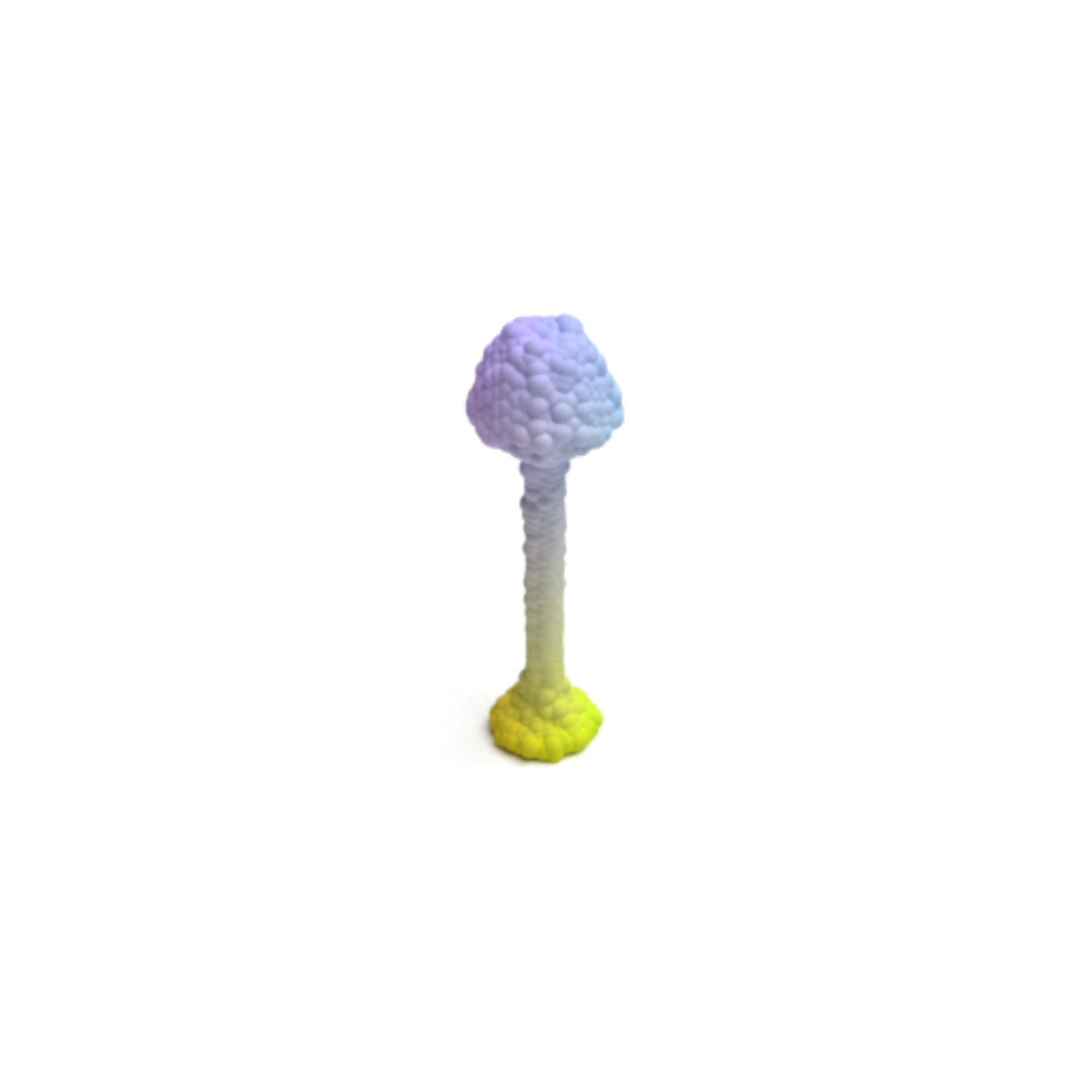}}}
\includegraphics[clip,trim=3cm 20cm 3cm 20cm, width=0.85\textwidth]{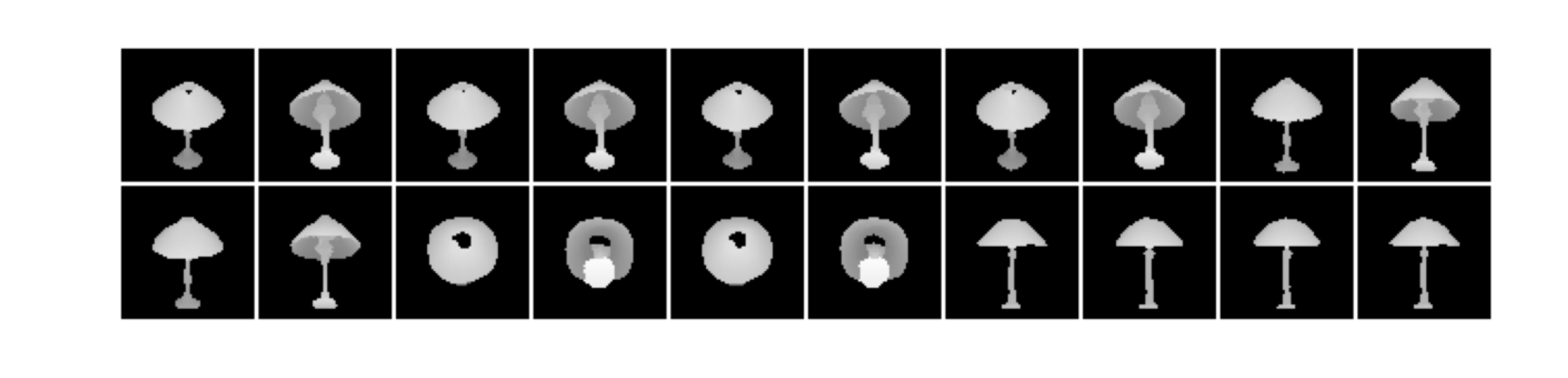}
\raisebox{0.5\height}{\centering{\includegraphics[clip,trim=3cm 3cm 3cm 3cm, width=0.1\textwidth]{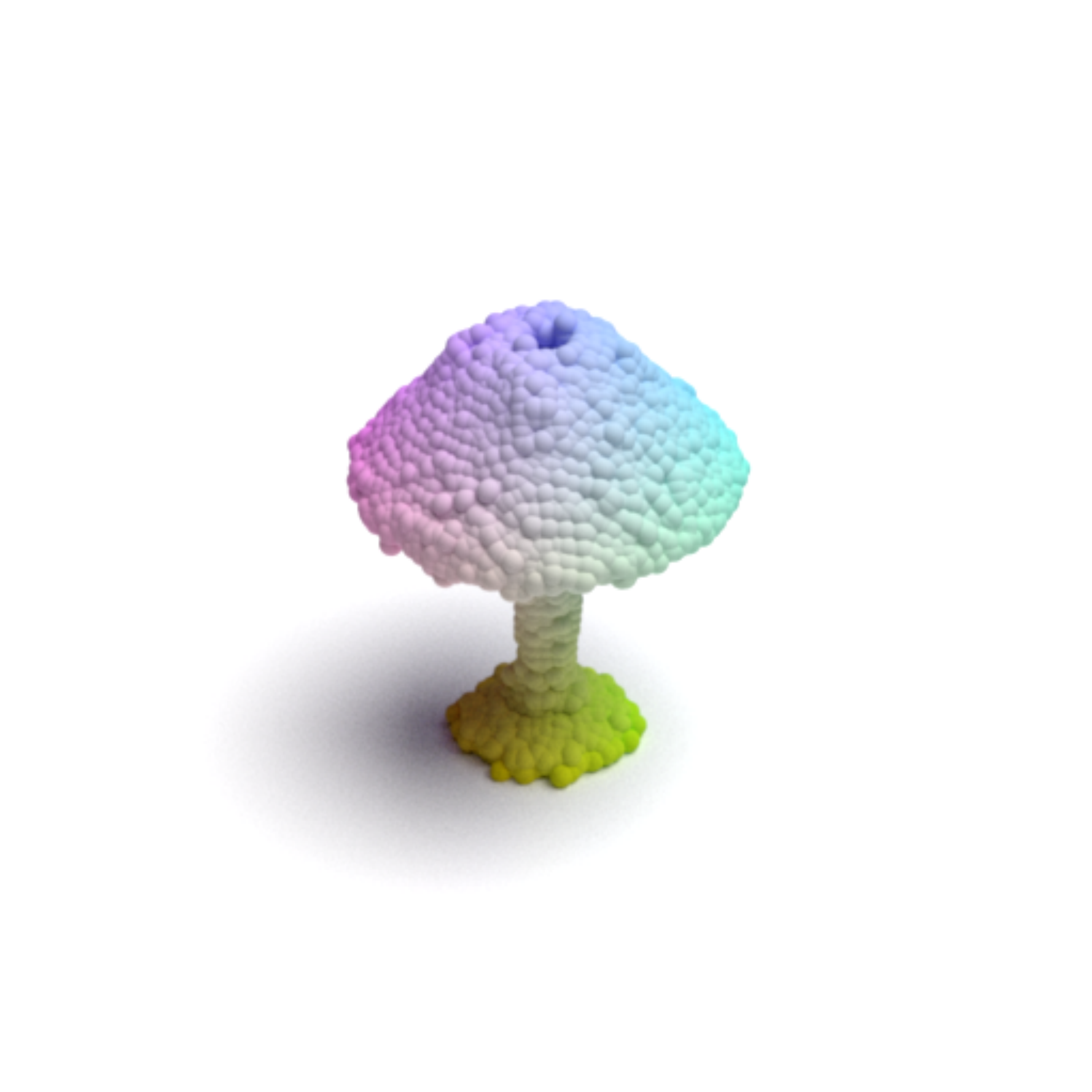}}}
\includegraphics[clip,trim=3cm 20cm 3cm 20cm, width=0.85\textwidth]{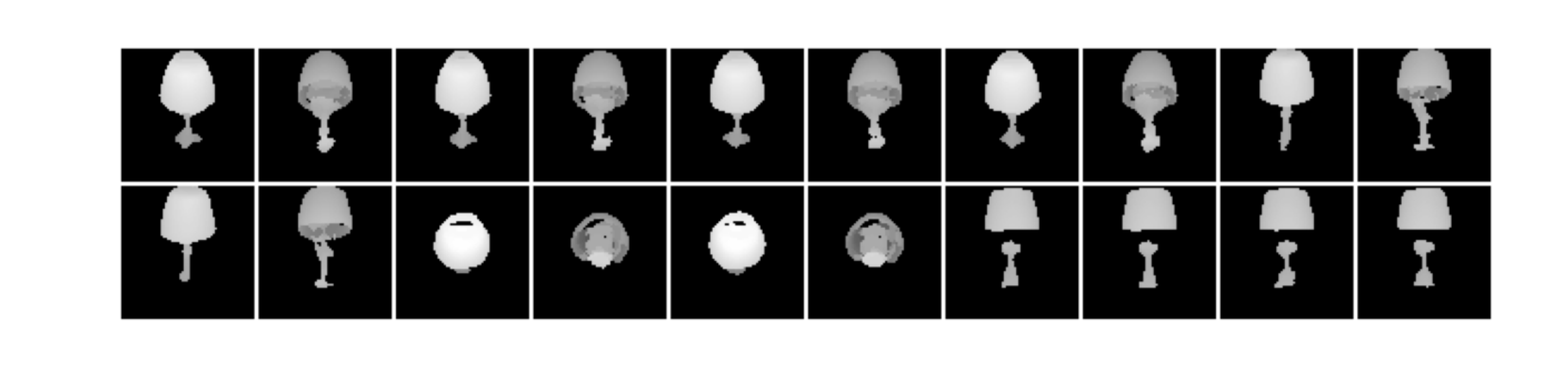}
\raisebox{0.5\height}{\centering{\includegraphics[clip,trim=3cm 3cm 3cm 3cm, width=0.1\textwidth]{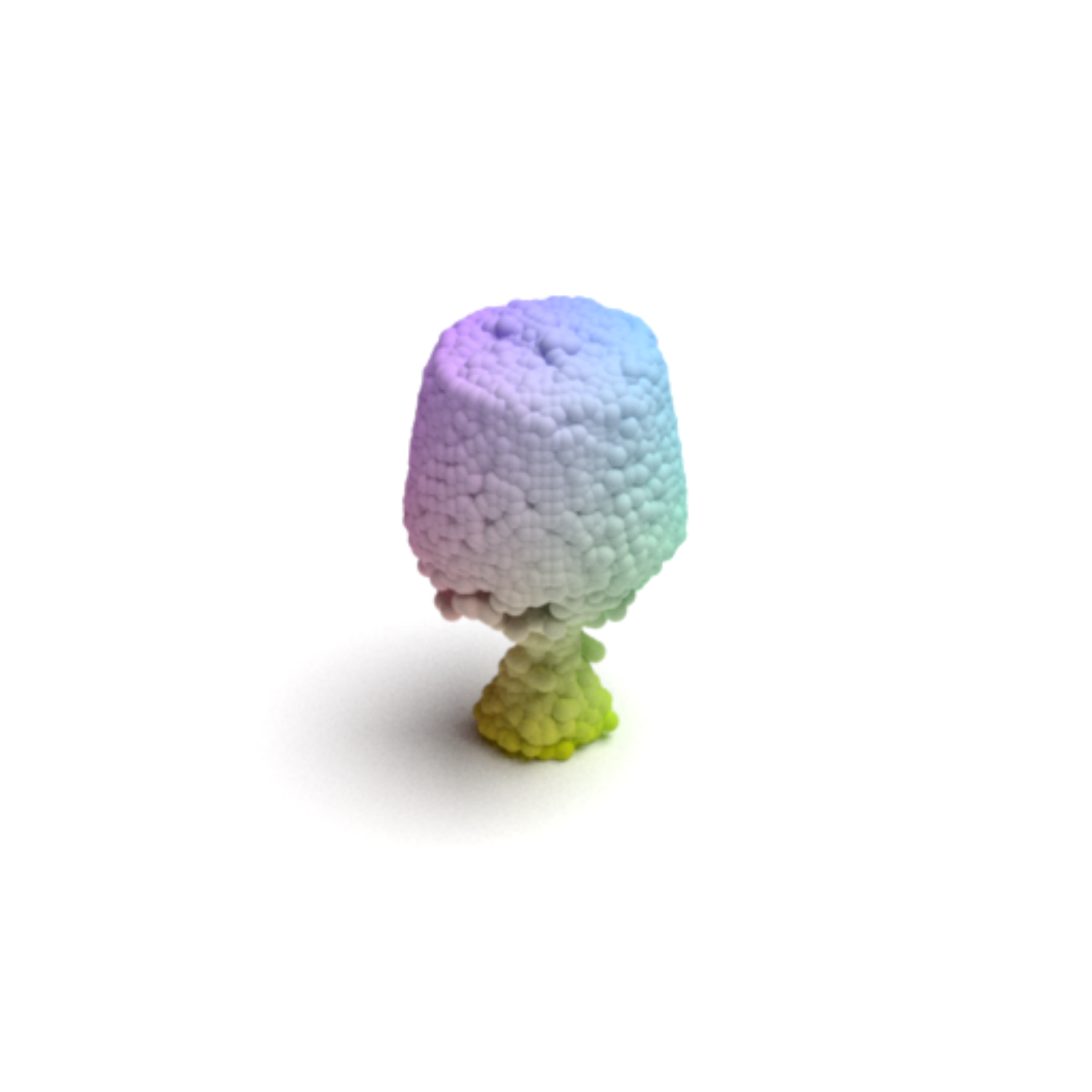}}}
\includegraphics[clip,trim=3cm 20cm 3cm 20cm, width=0.85\textwidth]{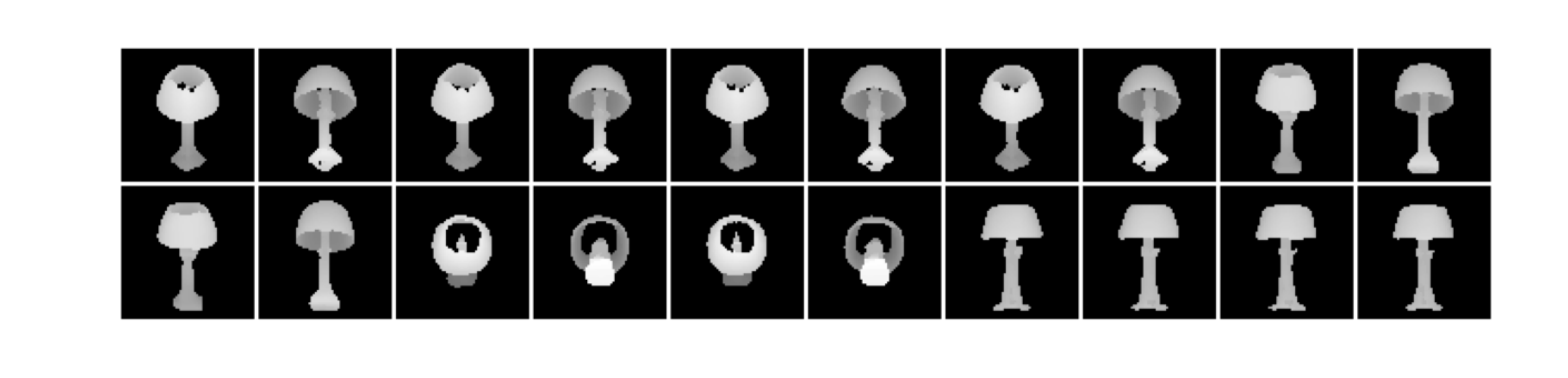}
\raisebox{0.5\height}{\centering{\includegraphics[clip,trim=3cm 3cm 3cm 3cm, width=0.1\textwidth]{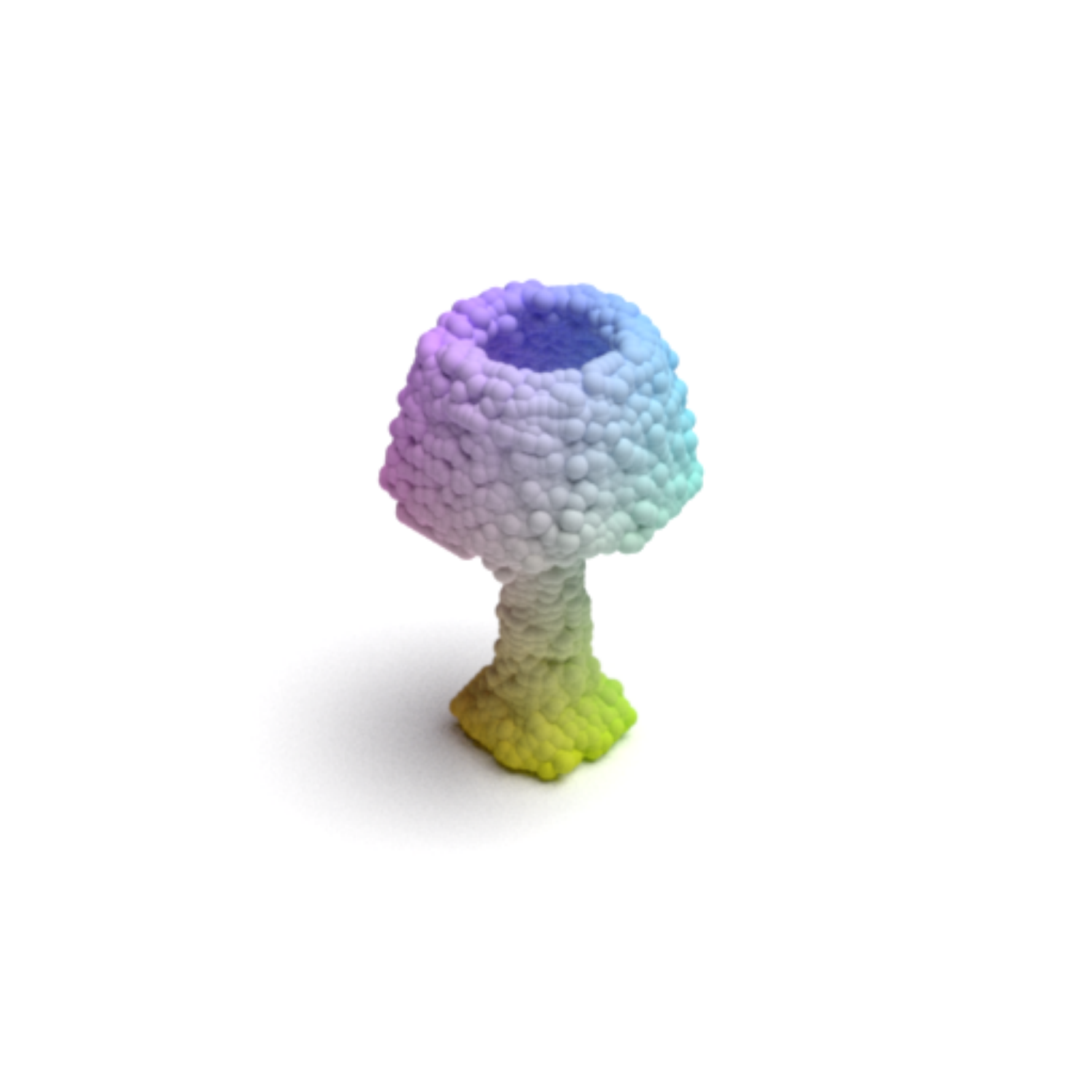}}}

\captionof{figure}{\textbf{Synthesized lamp objects} Each row shows the depth maps of the object from 20 views and corresponding 3D shape}
\label{fig:gen_dm_lamp}
\end{figure*}

\begin{figure*}
\centering
\hspace*{-0.5cm}%

\includegraphics[clip,trim=3cm 20cm 3cm 20cm, width=0.85\textwidth]{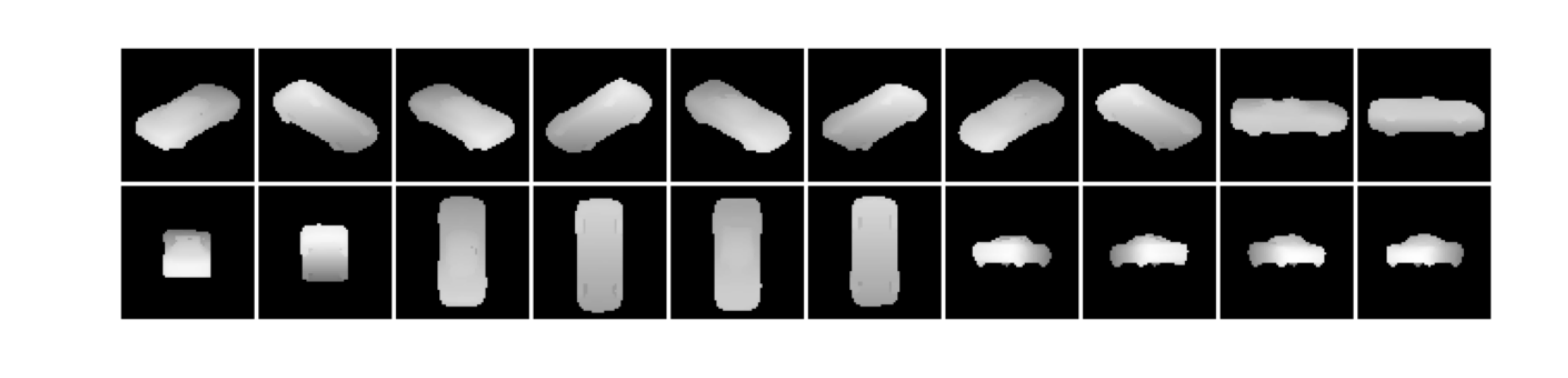}
\raisebox{0.5\height}{\centering{\includegraphics[clip,trim=3cm 3cm 3cm 3cm, width=0.1\textwidth]{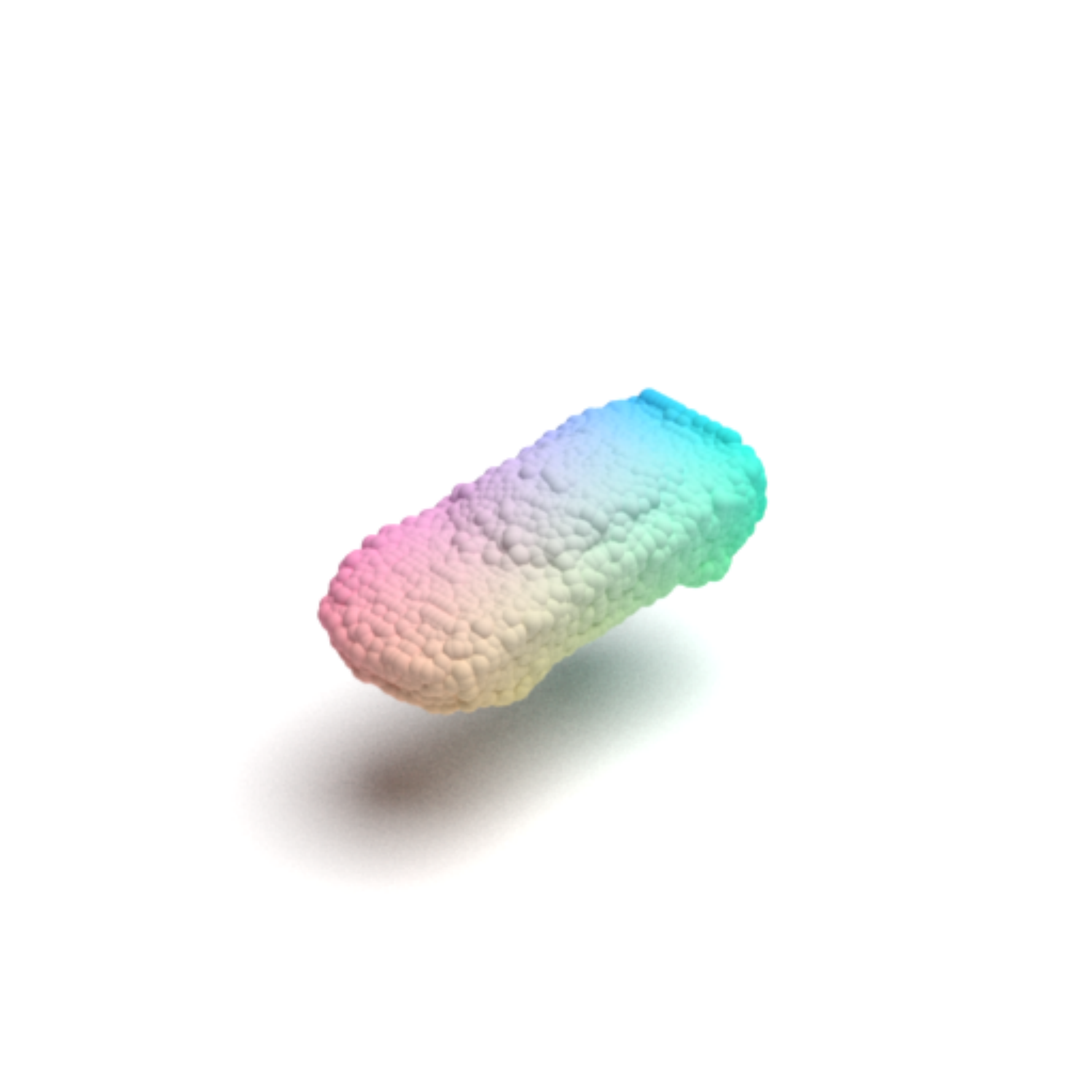}}}
\includegraphics[clip,trim=3cm 20cm 3cm 20cm, width=0.85\textwidth]{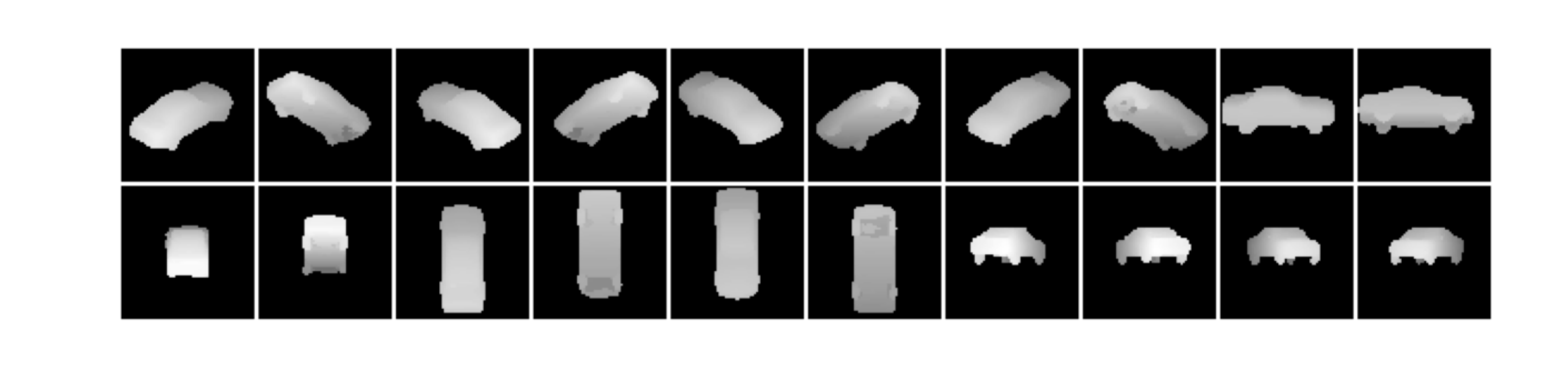}
\raisebox{0.5\height}{\centering{\includegraphics[clip,trim=3cm 3cm 3cm 3cm, width=0.1\textwidth]{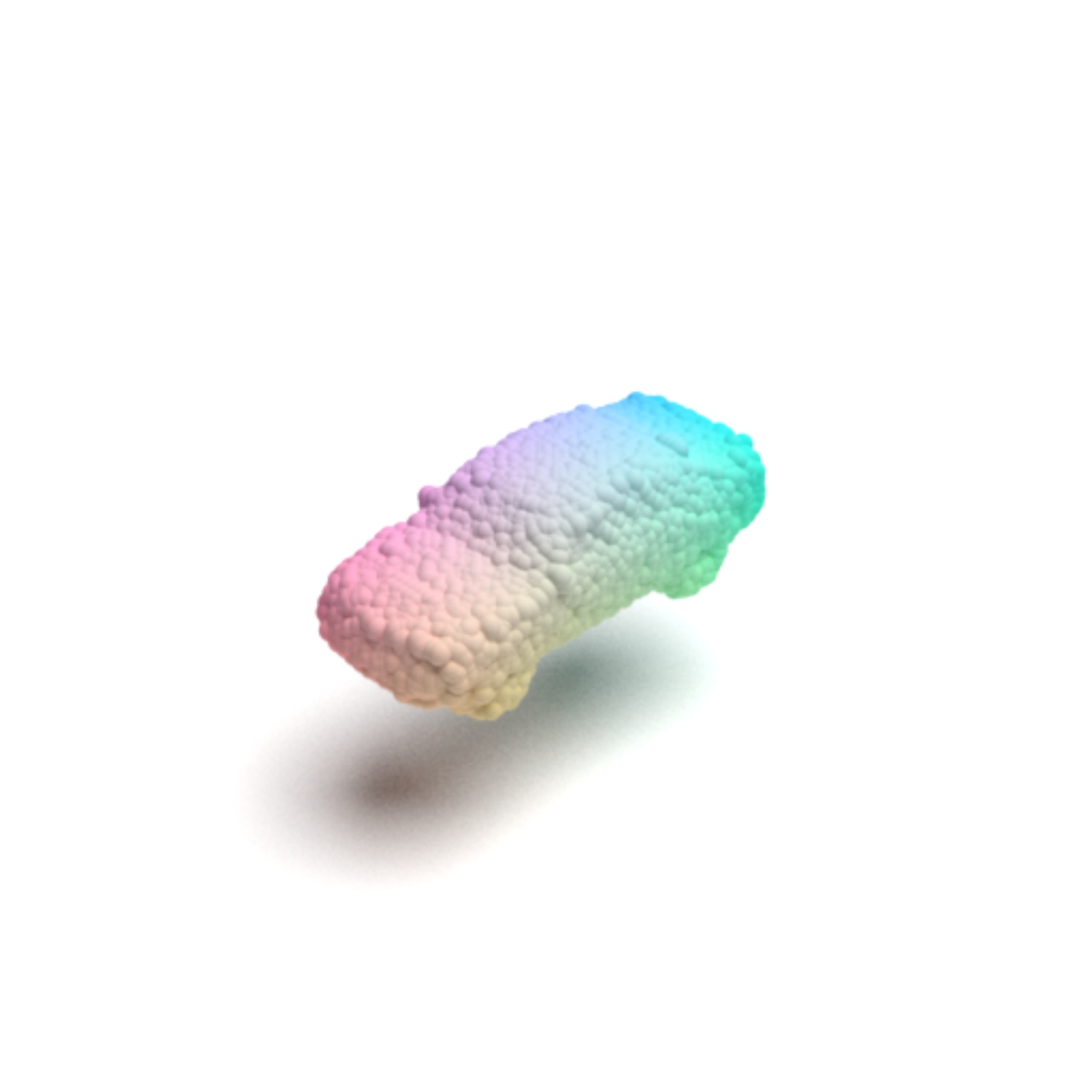}}}
\includegraphics[clip,trim=3cm 20cm 3cm 20cm, width=0.85\textwidth]{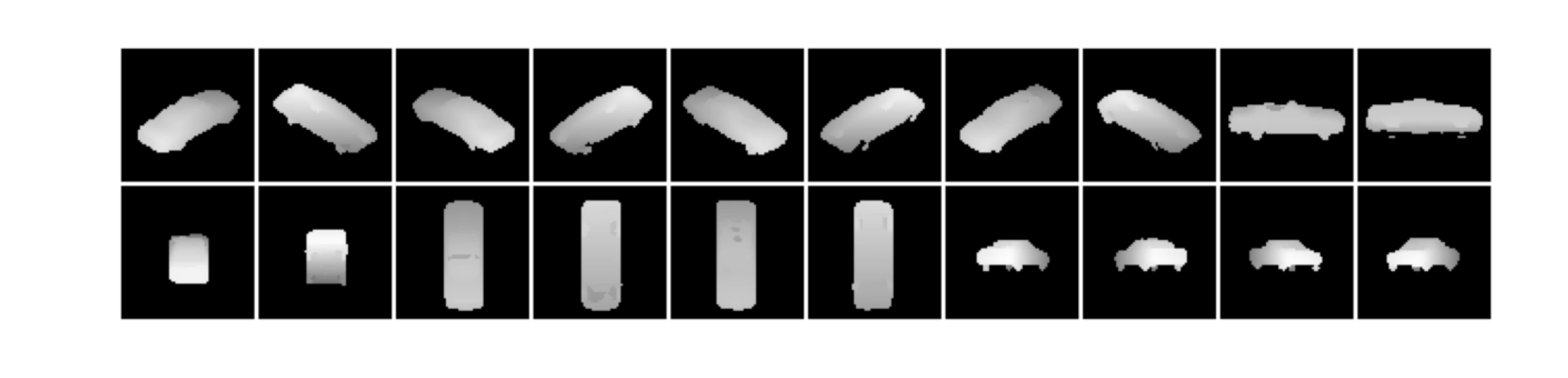}
\raisebox{0.5\height}{\centering{\includegraphics[clip,trim=3cm 3cm 3cm 3cm, width=0.1\textwidth]{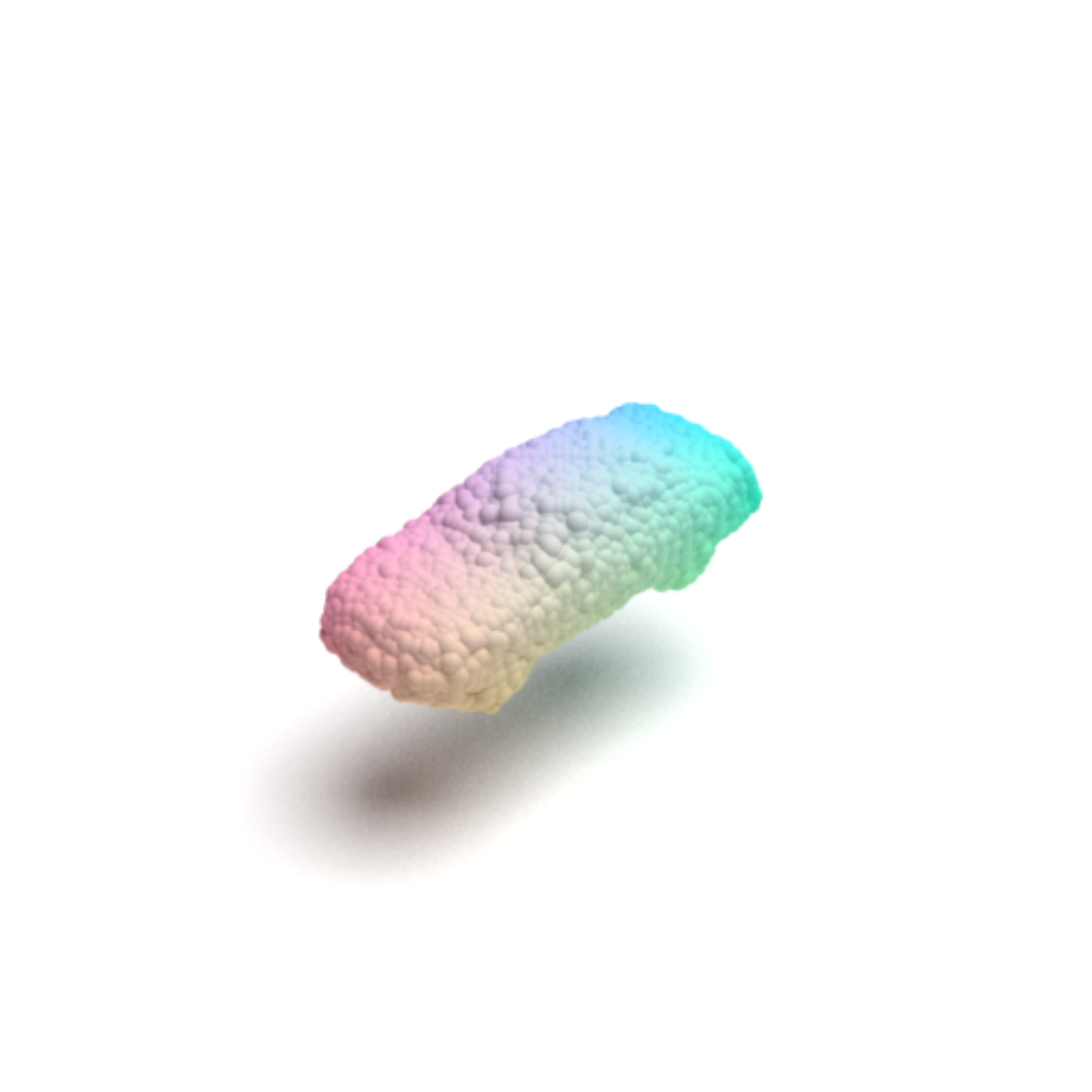}}}
\includegraphics[clip,trim=3cm 20cm 3cm 20cm, width=0.85\textwidth]{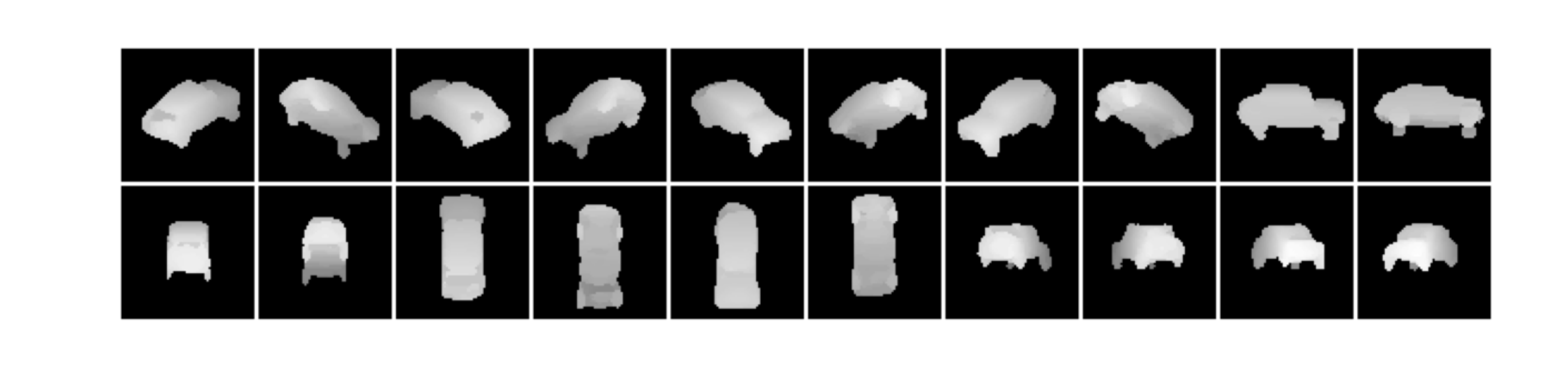}
\raisebox{0.5\height}{\centering{\includegraphics[clip,trim=3cm 3cm 3cm 3cm, width=0.1\textwidth]{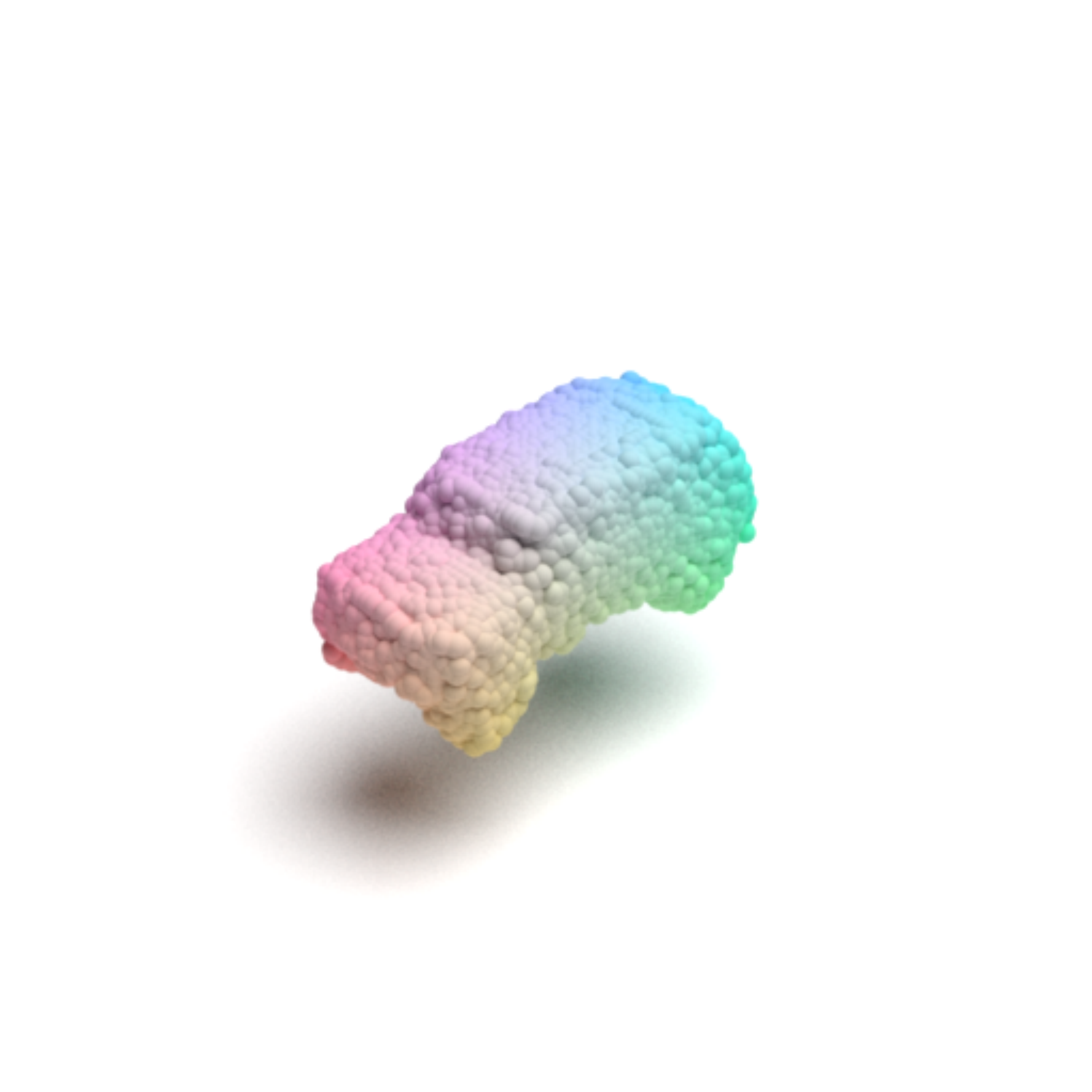}}}

\captionof{figure}{\textbf{Synthesized car objects} Each row shows the depth maps of the object from 20 views and corresponding 3D shape}
\label{fig:gen_dm_car}
\end{figure*}

Figures~\ref{fig:gen_dm_chair},~\ref{fig:gen_dm_table},~\ref{fig:gen_dm_airplane},~\ref{fig:gen_dm_sofa},~\ref{fig:gen_dm_lamp},~\ref{fig:gen_dm_car} show few generated depth maps as well as the reconstructed point clouds from different categories using IMLE. Our model can generate diverse and sharp 3D consistent depth maps for various categories.

\clearpage

\end{document}